\RequirePackage{silence} % :-\
\documentclass[ twoside,openright,titlepage,numbers=noenddot,headinclude,%1headlines,% letterpaper a4paper
                footinclude=true,cleardoublepage=empty,abstractoff, % <--- obsolete, remove (todo)
                BCOR=5mm,paper=a4,fontsize=11pt,%11pt,a4paper,%
                ngerman,american,%
                ]{scrreprt}
                
\usepackage{ragged2e} % fo
\usepackage{strivedi}
\usepackage{xcolor}

\usepackage{framed}
\newenvironment{aside}[1]{%
	\definecolor{shadecolor}{gray}{0.9}%
	\begin{shaded}{\color{Maroon}\noindent\textsc{#1}}\\%
	}{%
	\end{shaded}%
}

\PassOptionsToPackage{utf8}{inputenc}
  \usepackage{inputenc}

\PassOptionsToPackage{
  tocaligned=false, % the left column of the toc will be aligned (no indentation)
  dottedtoc=false,  % page numbers in ToC flushed right
  parts=true,       % use part division
  eulerchapternumbers=true, % use AMS Euler for chapter font (otherwise Palatino)
  linedheaders=false,       % chaper headers will have line above and beneath
  floatperchapter=true,     % numbering per chapter for all floats (i.e., Figure 1.1)
  listings=true,    % load listings package and setup LoL
  subfig=true,      % setup for preloaded subfig package
  eulermath=false,  % use awesome Euler fonts for mathematical formulae (only with pdfLaTeX)
  beramono=true,    % toggle a nice monospaced font (w/ bold)
  minionpro=false   % setup for minion pro font; use minion pro small caps as well (only with pdfLaTeX)
}{classicthesis}

\newcommand{\myTitle}{Discriminative Learning of Similarity and Group Equivariant Representations\xspace}
\newcommand{\mySubtitle}{PhD Thesis\xspace}

\newcommand{\myName}{Shubhendu Trivedi\xspace}

\newcommand{\myFaculty}{Put data here\xspace}

\newcommand{\myUni}{Toyota Technological Institute at Chicago\xspace}

\newcommand{\myTime}{August 2018\xspace}
\newcommand{\myVersion}{version 4.4}

\newcounter{dummy} % necessary for correct hyperlinks (to index, bib, etc.)
\providecommand{\mLyX}{L\kern-.1667em\lower.25em\hbox{Y}\kern-.125emX\@}

\usepackage{babel}

\usepackage{csquotes}
\PassOptionsToPackage{
  backend=bibtex8,bibencoding=ascii,%
  language=auto,%
  style=numeric-comp,%
  sorting=nyt, % name, year, title
  maxbibnames=10, % default: 3, et al.
  natbib=true % natbib compatibility mode (\citep and \citet still work)
}{biblatex}
\usepackage{biblatex}

\PassOptionsToPackage{fleqn}{amsmath}       % math environments and more by the AMS
\usepackage{amsmath}

\PassOptionsToPackage{T1}{fontenc} % T2A for cyrillics
  \usepackage{fontenc}
\usepackage{textcomp} % fix warning with missing font shapes
\usepackage{scrhack} % fix warnings when using KOMA with listings package
\usepackage{xspace} % to get the spacing after macros right
\usepackage{mparhack} % get marginpar right
\PassOptionsToPackage{printonlyused,smaller}{acronym}
  \usepackage{acronym} % nice macros for handling all acronyms in the thesis
\usepackage{tabularx} % better tables
\usepackage{caption}
\usepackage{subfig}
\usepackage{listings}
\PassOptionsToPackage{hyperfootnotes=false,pdfpagelabels}{hyperref}
  \usepackage{hyperref}  % backref linktocpage pagebackref
  \usepackage{graphicx}

\hypersetup{%
  colorlinks=true, linktocpage=true, pdfstartpage=3, pdfstartview=FitV,%
  breaklinks=true, pdfpagemode=UseNone, pageanchor=true, pdfpagemode=UseOutlines,%
  plainpages=false, bookmarksnumbered, bookmarksopen=true, bookmarksopenlevel=1,%
  hypertexnames=true, pdfhighlight=/O,%nesting=true,%frenchlinks,%
  urlcolor=webbrown, linkcolor=RoyalBlue, citecolor=webgreen, %pagecolor=RoyalBlue,%
  pdftitle={\myTitle},%
  pdfauthor={\textcopyright\ \myName, \myUni, \myFaculty},%
  pdfsubject={},%
  pdfkeywords={},%
  pdfcreator={pdfLaTeX},%
  pdfproducer={LaTeX with hyperref and classicthesis}%
}

\makeatletter
\@ifpackageloaded{babel}%
  {%
    \addto\extrasamerican{%
    }%
    \addto\extrasngerman{%
    }%
    }{\relax}
\makeatother

\usepackage{classicthesis}
\begin{document}
\frenchspacing
\raggedbottom
\selectlanguage{american} % american ngerman
%\renewcommand*{\bibname}{new name}
%\setbibpreamble{}
\pagenumbering{roman}
\pagestyle{plain}
%********************************************************************
% Frontmatter
%*******************************************************
%*******************************************************
% Little Dirty Titlepage
%*******************************************************
\thispagestyle{empty}
%\pdfbookmark[1]{Titel}{title}
%*******************************************************
\begin{center}
    \spacedlowsmallcaps{Shubhendu Trivedi} \\ \medskip

    \begingroup
        \color{Maroon}\spacedallcaps{Discriminative Learning of Similarity and Group Equivariant Representations}
    \endgroup
\end{center}

%*******************************************************
% Titlepage
%*******************************************************
\begin{titlepage}
    % if you want the titlepage to be centered, uncomment and fine-tune the line below (KOMA classes environment)
    \begin{addmargin}[-1cm]{-1cm}
    \begin{center}
        \large

        \hfill

        \begingroup
            \color{Maroon}\spacedallcaps{Discriminative Learning of Similarity and Group Equivariant Representations} \\ \bigskip
        \endgroup

        \spacedlowsmallcaps{Shubhendu Trivedi}

        \vfill

        \mySubtitle \\ \medskip
        %\myDegree \\
        %\myDepartment \\
        %\myFaculty \\
        %\myUni \\ \bigskip

        \myTime \\[100pt]

		---\; Dissertation Committee\;---\\[100pt]
		
		Dr. Kevin Gimpel \\ {\small Toyota Technological Institute at Chicago} \\[20pt]
		Dr. Risi Kondor \\ {\small The University of Chicago} \\ [20pt]
		Dr. Brian D. Nord \\ {\small Fermilab \& The University of Chicago} \\ [20pt]
		Dr. Gregory Shakhnarovich \\[3pt] {\small \textbf{(Thesis Advisor)} \\[3pt] Toyota Technological Institute at Chicago}

    \end{center}
  \end{addmargin}
\end{titlepage}

\thispagestyle{empty}

\hfill

\vfill

\noindent \textit{Discriminative Learning of Similarity and Group Equivariant Representations}, \\ \copyright  \hspace{0.5mm} Shubhendu Trivedi, \myTime %\myDegree,
%\textcopyright\ \myTime

%\bigskip
%
%\noindent\spacedlowsmallcaps{Supervisors}: \\
%\myProf \\
%\myOtherProf \\
%\mySupervisor
%
%\medskip
%
%\noindent\spacedlowsmallcaps{Location}: \\
%\myLocation
%
%\medskip
%
%\noindent\spacedlowsmallcaps{Time Frame}: \\
%\myTime

    \begin{addmargin}[-1cm]{-1cm}
	\begin{center}		
		\color{Maroon}\spacedallcaps{Discriminative Learning of Similarity and Group Equivariant Representations}\\[20pt] \normalsize \color{black}
		A thesis presented\\[2pt]
		by\\[5pt]
		\color{Maroon}\spacedallcaps{Shubhendu Trivedi}\\[10pt] \normalsize \color{black}
		in partial fulfillment of the requirements for the degree of\\[4pt]
		Doctor of Philosophy in Computer Science.\\[4pt]
		Toyota Technological Institute at Chicago\\[4pt]
		Chicago, Illinois\\[4pt]
		August, 2018\\[30pt]

---\;Thesis Committee\;---\\[30pt]
\begin{table}[h]
	\begin{tabular}{l l l}
		Dr. Kevin Gimpel & &\\
		\rule{4.8cm}{1pt}\quad\quad\quad & \rule{5cm}{1pt}\;\;\; & \rule{3cm}{1pt}\\[2pt]
		Committee member & Signature & Date\\[25pt]
		
		Dr. Risi Kondor & &\\
		\rule{4.8cm}{1pt}\quad\quad\quad & \rule{5cm}{1pt}\;\;\; & \rule{3cm}{1pt}\\[2pt]
		Committee member & Signature & Date\\[25pt]
		
		Dr. Brian D. Nord & &\\
		\rule{4.8cm}{1pt}\quad\quad\quad & \rule{5cm}{1pt}\;\;\; & \rule{3cm}{1pt}\\[2pt]
		Committee member & Signature & Date\\[25pt]
		
		Dr. Gregory Shakhnarovich & &\\
		\rule{4.8cm}{1pt}\quad\quad\quad & \rule{5cm}{1pt}\;\;\; & \rule{3cm}{1pt}\\[2pt]
		Thesis/Research Advisor & Signature & Date\\[25pt]
		
		Dr. Avrim Blum & &\\
		\rule{4.8cm}{1pt}\quad\quad\quad & \rule{5cm}{1pt}\;\;\; & \rule{3cm}{1pt}\\[2pt]
		Chief Academic Officer & Signature & Date
		
	\end{tabular}
\end{table}

\end{center}		
\end{addmargin}

\begin{addmargin}[-1cm]{-1cm}
\begin{center}		
\includegraphics{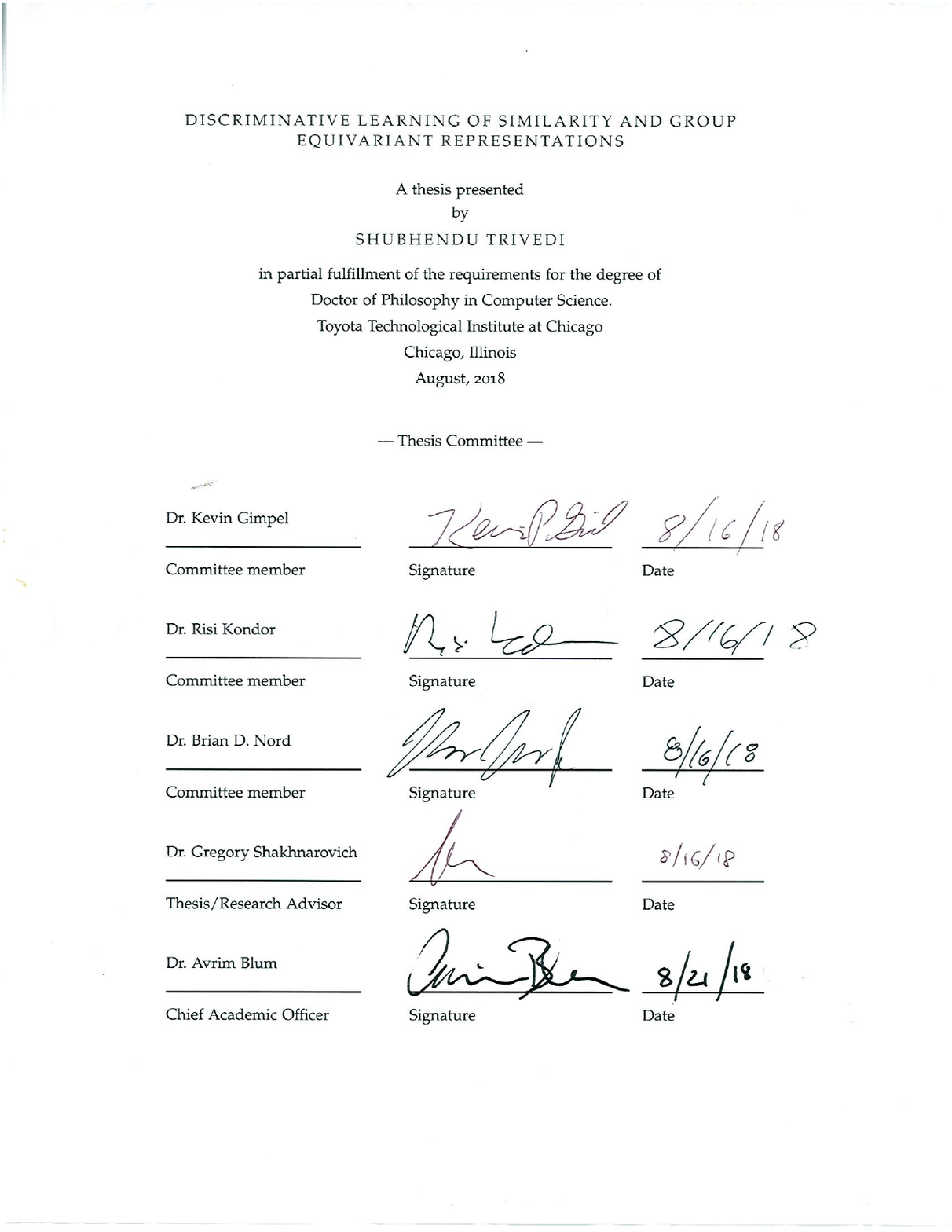}
\end{center}		
\end{addmargin}

%*******************************************************
% Dedication
%*******************************************************
\thispagestyle{empty}
%\phantomsection
\refstepcounter{dummy}
\pdfbookmark[1]{Dedication}{Dedication}

\vspace*{7cm}
\begin{center}
	
%	This is the dedication.
	
\begin{flushleft}

\vspace{1cm}
\emph{... for (and in veneration of) my loving parents:} \emph{Smt. Jyotsna Trivedi and Shri M. L. Trivedi.}

\vspace{1cm}

\emph{... to the memory of my grandfather: Shri R. C. Trivedi.}

\vspace{1cm}
\emph{... for one of my dearest friends:
Babar Majeed Saggu.}

\vspace{1cm}
\emph{... and finally to:
	M.}
\end{flushleft}
	
\end{center}
\include{FrontBackmatter/Epigraph}
%%%\cleardoublepage\include{FrontBackmatter/Foreword}
%*******************************************************
% Abstract
%*******************************************************
%\renewcommand{\abstractname}{Abstract}
\refstepcounter{dummy}
\pdfbookmark[1]{BTL}{BTL}
\begingroup
\let\clearpage\relax
\let\cleardoublepage\relax
\let\cleardoublepage\relax

\chapter*{Franz Kafka: Before The Law}
Before the law sits a gatekeeper. To this gatekeeper comes a man from the country who asks to gain entry into the law. But the gatekeeper says that he cannot grant him entry at the moment. The man thinks about it and then asks if he will be allowed to come in later on. “It is possible,” says the gatekeeper, “but not now.” At the moment the gate to the law stands open, as always, and the gatekeeper walks to the side, so the man bends over in order to see through the gate into the inside. When the gatekeeper notices that, he laughs and says: “If it tempts you so much, try it in spite of my prohibition. But take note: I am powerful. And I am only the most lowly gatekeeper. But from room to room stand gatekeepers, each more powerful than the other. I can’t endure even one glimpse of the third.” The man from the country has not expected such difficulties: the law should always be accessible for everyone, he thinks, but as he now looks more closely at the gatekeeper in his fur coat, at his large pointed nose and his long, thin, black Tartar’s beard, he decides that it would be better to wait until he gets permission to go inside. The gatekeeper gives him a stool and allows him to sit down at the side in front of the gate. There he sits for days and years. He makes many attempts to be let in, and he wears the gatekeeper out with his requests. The gatekeeper often interrogates him briefly, questioning him about his homeland and many other things, but they are indifferent questions, the kind great men put, and at the end he always tells him once more that he cannot let him inside yet. The man, who has equipped himself with many things for his journey, spends everything, no matter how valuable, to win over the gatekeeper. The latter takes it all but, as he does so, says, “I am taking this only so that you do not think you have failed to do anything.” During the many years the man observes the gatekeeper almost continuously. He forgets the other gatekeepers, and this one seems to him the only obstacle for entry into the law. He curses the unlucky circumstance, in the first years thoughtlessly and out loud, later, as he grows old, he still mumbles to himself. He becomes childish and, since in the long years studying the gatekeeper he has come to know the fleas in his fur collar, he even asks the fleas to help him persuade the gatekeeper. Finally his eyesight grows weak, and he does not know whether things are really darker around him or whether his eyes are merely deceiving him. But he recognizes now in the darkness an illumination which breaks inextinguishably out of the gateway to the law. Now he no longer has much time to live. Before his death he gathers in his head all his experiences of the entire time up into one question which he has not yet put to the gatekeeper. He waves to him, since he can no longer lift up his stiffening body. The gatekeeper has to bend way down to him, for the great difference has changed things to the disadvantage of the man. “What do you still want to know, then?” asks the gatekeeper. “You are insatiable.” “Everyone strives after the law,” says the man, “so how is that in these many years no one except me has requested entry?” The gatekeeper sees that the man is already dying and, in order to reach his diminishing sense of hearing, he shouts at him, “Here no one else can gain entry, since this entrance was assigned only to you. I’m going now to close it.”

\noindent
\footnotesize{\emph{[Trans. by Ian Johnston. Here, law might originate from the Hebrew word Torah, thus also having the meaning truth]}}
\endgroup

\vfill

\clearpage%*******************************************************
% Abstract
%*******************************************************
%\renewcommand{\abstractname}{Abstract}
\pdfbookmark[1]{Abstract}{Abstract}
\begingroup
\let\clearpage\relax
\let\cleardoublepage\relax
\let\cleardoublepage\relax

\chapter*{Abstract}
One of the most fundamental problems in machine learning is to compare examples: Given a pair of objects we want to return a value which indicates degree of (dis)similarity. Similarity is often task specific, and pre-defined distances can perform poorly, leading to work in metric learning. However, being able to learn a similarity-sensitive distance function also presupposes access to a rich, discriminative representation for the objects at hand.  In this dissertation we present contributions towards both ends. In the first part of the thesis, assuming good representations for the data, we present a formulation for metric learning that makes a more direct attempt to optimize for the k-NN accuracy as compared to prior work. Our approach considers the choice of k neighbors as a discrete valued latent variable, and casts the metric learning problem as a large margin structured prediction problem. We present experiments comparing to a suite of popular metric learning methods. We also present extensions of this formulation to metric learning for kNN regression, and discriminative learning of Hamming distance. In the second part, we consider a situation where we are on a limited computational budget i.e. optimizing over a space of possible metrics would be infeasible, but access to a label aware distance metric is still desirable. We present a simple, and computationally inexpensive approach for estimating a well motivated metric that relies only on gradient estimates, we also discuss theoretical as well as experimental results of using this approach in regression and multiclass settings. In the final part, we address representational issues, considering group equivariant neural networks (GCNNs). Equivariance to symmetry transformations is explicitly encoded in GCNNs; a classical CNN being the simplest example. Following recent work by Kondor et. al., we present a SO(3)-equivariant neural network architecture for spherical data, that operates entirely in Fourier space, while using tensor products and the Clebsch-Gordan decomposition as the only source of non-linearity. We report strong experimental results, and emphasize the wider applicability of our approach, in that it also provides a formalism for the design of fully Fourier neural networks that are equivariant to the action of any continuous compact group.

\vspace{2mm}
\noindent
\textbf{Thesis Advisor:} Gregory Shakhnarovich

\vspace{2mm}
\noindent
\textbf{Title: } Associate Professor
%\begin{center}
%\url{https://plg.uwaterloo.ca/~migod/research/beckOOPSLA.html}
%\end{center}

\endgroup

\vfill

%\clearpage\include{FrontBackmatter/Summary}
\clearpage%*******************************************************
% Publications
%*******************************************************
\pdfbookmark[1]{Publications}{publications}
\chapter*{Publications}%\graffito{This is just an early --~and currently ugly~-- test!}
The ideas in this thesis have appeared (or are about to appear) in the following publications, pre-prints and technical reports

\begin{enumerate}[{[}1{]}]
	\item Shubhendu Trivedi, David McAllester, and Gregory Shakhnarovich. "Discriminative Metric Learning by Neighborhood Gerrymandering." In: \emph{Advances in Neural Processing Systems}. 2014, pp. 3392--3400
	\item Shubhendu Trivedi, Jialei Wang, Samory Kpotufe, and Gregory Shakhnarovich.
	“A Consistent Estimator of the Expected Gradient Outerproduct.” In: \emph{Proceedings
	of the 30th International Conference on Uncertainty in Artificial Intelligence}. AUAI.
	2014, pp. 819–828.
	\item Shubhendu Trivedi and Jialei Wang. "The Expected Jacobian Outerproduct" Preprint. 2018.
	\item Risi Kondor, Shubhendu Trivedi, and Zhen Lin. "A Fully Fourier Space Spherical Convolutional Neural Network based on Clebsch-Gordan Transforms." \emph{Provisional US patent application}, 2018.
	\item Risi Kondor, Zhen Lin, and Shubhendu Trivedi. “Clebsch-Gordan Nets: a Fully
	Fourier Space Spherical Convolutional Neural Network.” arXiv:1806.09231, Pre-print, 2018.
\end{enumerate}

\vspace{2mm}
\noindent 
The following publications, pre-prints and technical reports that the dissertation author was also a contributor in (as a result of work initiated after January 2013), but are not part of this dissertation.

\begin{enumerate}[{[}1{]}]\addtocounter{enumi}{5}
	\item Fei Song, Shubhendu Trivedi, Yutao Wang, G\'abor N. S\'ark\"ozy, and Neil T. Heffernan. "Applying Clustering to the Problem of Predicting Retention within an ITS: Comparing Regularity Clustering with Traditional Methods." In: \emph{Proceedings of the 26th AAAI FLAIRS Conference}. 2013, pp. 527--532
	\item Risi Kondor, Truong Hy Song, Horace Pan, Brandon M. Anderson, and Shubhendu
	Trivedi. “Covariant compositional networks for learning graphs.” arXiv:1801.02144, Pre-print, 2018.
	\item Truong Son Hy, Shubhendu Trivedi, Horace Pan, Brandon M. Anderson, and
	Risi Kondor. “Predicting molecular properties with covariant compositional networks.”
	In: \emph{The Journal of Chemical Physics} 148.24 (2018), p. 241745.
	\item Risi Kondor and Shubhendu Trivedi. “On the Generalization of Equivariance
	and Convolution in Neural Networks to the Action of Compact Groups.” In:
	\emph{Proceedings of the 35th International Conference on Machine Learning}. PMLR, 2018, pp. 2747–2755.
	\item Rohit Nagpal and Shubhendu Trivedi. "A Module-Theoretic Perspective on Equivariant Steerable Convolutional Neural Networks", Pre-print, 2018. 
	\item Joao Caldeira, W. L. Kimmy Wu, Brian D. Nord, Camille Avestruz, Shubhendu Trivedi, and Kyle T. Story. "DeepCMB: Lensing Reconstruction of the Cosmic Microwave Background with Deep Neural Networks", Pre-print, 2018.
	\item Zhen Lin, Nick D. Huang, W. L. Kimmy Wu, Brian D. Nord, and Shubhendu Trivedi. "DeepCMB: Classification of Sunyaev-Zel'dovich Clusters in Millimeter Wave Maps using Deep Learning", Pre-print, 2018.	
	
\end{enumerate}

%\noindent Put your publications from the thesis here. The packages \texttt{multibib} or \texttt{bibtopic} etc. can be used to handle multiple different bibliographies in your document.

%\emph{Attention}: This requires a separate run of \texttt{bibtex} for your \texttt{refsection}, \eg, \texttt{ClassicThesis1-blx} for this file. You might also use \texttt{biber} as the backend for \texttt{biblatex}. See also \url{http://tex.stackexchange.com/questions/128196/problem-with-refsection}.

\clearpage%*******************************************************
% Abstract
%*******************************************************
%\renewcommand{\abstractname}{Abstract}
\pdfbookmark[1]{Credit}{Credit}
\begingroup
\let\clearpage\relax
\let\cleardoublepage\relax
\let\cleardoublepage\relax

\chapter*{Credit Assignment}

\begin{enumerate}
	\item[1] Work presented in chapter \ref{ch:mlng} was joint work with Gregory Shakhnarovich and David McAllester. G. Shakhnarovich was the primary contributor in an earlier iteration of the work presented. The latent structural SVM formulation was originally due to D. McAllester and G. Shakhnarovich. The dissertation author was the primary contributor in later iterations, and contributed ideas, proposed inference procedures, refinements, carried out experiments and contributed to the write-up. Some of the sections and figures in  chapter \ref{ch:mlng} are excerpted directly from the following report: Shubhendu Trivedi, David McAllester, and Gregory Shakhnarovich. "Discriminative Metric Learning by Neighborhood Gerrymandering." In: \emph{Advances in Neural Processing Systems}. 2014, pp. 3392--3400
	\item[2] Research presented in sections \ref{sec:asym} and \ref{sec:hamming} was joint work with Behnam Neyshabur and Gregory Shakhnarovich. The idea of using asymmetry is due to B. Neyshabur. The dissertation author was the primary contributor and contributed ideas, did the experimental evaluation as well as the complete write-up. 
	\item[3] Work presented in section \ref{sec:mlknnregression} was joint work with Gregory Shakhnarovich. The dissertation author was the primary contributor in all aspects of the presented work.
	\item[4] Work presented in chapter \ref{ch:EGOP} was joint work with Jialei Wang, Samory Kpotufe and Gregory Shakhnarovich. The dissertation author initiated the project with S. Kpotufe and G. Shakhnarovich. The idea of using the expected gradient outer product is due to G. Shakhnarovich. J. Wang and S. Kpotufe were the primary contributors in the theoretical analysis. The dissertation author was the primary contributor in the experimental evaluation, as well as contributed ideas for the theoretical analysis and did part of the write-up. Some of the text and figures in chapter \ref{ch:EGOP} are excerpted directly from the following report: Shubhendu Trivedi, Jialei Wang, Samory Kpotufe, and Gregory Shakhnarovich.	“A Consistent Estimator of the Expected Gradient Outerproduct.” In: \emph{Proceedings of the 30th International Conference on Uncertainty in Artificial Intelligence}. AUAI.	2014, pp. 819–828. 
	\item[5] Work on the expected Jacobian outer product presented in chapter \ref{ch:EJOP} was joint with Jialei Wang. The dissertation author was the primary contributor (jointly with J. Wang) in all aspects of the presented work and contributed to the theoretical analysis, did the experimental evaluation and did the complete write-up. The work also involved inputs by S. Kpotufe. 
	\item[6] Work presented in chapter \ref{ch:S2CNN} was joint work with Risi Kondor and Zhen Lin. The presented work is a direct consequence of a theoretical result (not part of the dissertation) that appeared in the following publication: Risi Kondor and Shubhendu Trivedi. “On the Generalization of Equivariance and Convolution in Neural Networks to the Action of Compact Groups.” In:\emph{Proceedings of the 35th International Conference on Machine Learning}. PMLR, 2018, pp. 2747–2755. The idea of using the Clebsch-Gordan transform is due to R. Kondor. The dissertation author was one of the primary contributors (jointly with R. Kondor and Z. Lin) and contributed ideas, the experimental evaluation and contributed to part of the write up. The text appearing in section \ref{sec-S2CNNexpts} is wholly excerpted from the following report: Risi Kondor, Zhen Lin, and Shubhendu Trivedi. “Clebsch-Gordan Nets: a Fully Fourier Space Spherical Convolutional Neural Network.” arXiv:1806.09231, Pre-print, 2018.
\end{enumerate}	
%	\item[2] Work presented in section 4.2 was joint with Behnam Neyshabur and Gregory Shakhnavorich. The dissertation author was the primary contributor, jointly with Behnam Neyshabur, and contributed to the main ideas, carried out experiments and did the write-up.
%	\item[3] Work presented in section 4.3 was joint with Gregory Shakhnarovich. The dissertation author was the primary contributor.
%	\item[4] Work presented in section Chapter 8 was joint work with Jialei Wang. The dissertation author was the primary contributor along with Jialei Wang, and contributed to the main proof ideas, carried out experiments did the writeup.
%\end{enumerate}
\endgroup

\vfill

\clearpage%*******************************************************
% Acknowledgments
%*******************************************************
\pdfbookmark[1]{Acknowledgments}{acknowledgments}

%\begin{flushright}{\slshape
%    We have seen that computer programming is an art, \\
%    because it applies accumulated knowledge to the world, \\
%    because it requires skill and ingenuity, and especially \\
%    because it produces objects of beauty.} \\ \medskip
%    --- \defcitealias{knuth:1974}{Donald E. Knuth}\citetalias{knuth:1974} \citep{knuth:1974}
%\end{flushright}

\bigskip
\begingroup
\let\clearpage\relax
\let\cleardoublepage\relax
\let\cleardoublepage\relax
\chapter*{Acknowledgments}
\small
It feels mildly disappointing to write this section \emph{ex post facto}; particularly in the anticlimactic aftertaste following the very brief but intense period of frenzy that went into putting this dissertation together. Nevertheless it is making me reflect on this journey and my time in Chicago. I came to Chicago and to TTI after a fulfilling and productive random walk, but soon enough, within a year, a combination of a lack of preparedness as well as a couple of extremely unusual personal events soon threatened to turn it into a nightmare. Wherefore, it gives me satisfaction that it turned to be a remarkable, intellectually stimulating and uplifting \emph{personal} experience. Surely, graduate school is not supposed to be easy for anyone, by definition and by design, and it might seem like an exercise in \emph{cheap} vanity to say that personal circumstance made it much harder than it ought to have been. What I intend to convey is that though I put a lot of sweat into this thesis, yet by itself, it does not mean \emph{anything} to me. Indeed, a few months here and there, and given the frenetic activity and pace, it just might have appeared completely different in character and in form, or even in its topic of focus. What is important to me is what the journey has taught me in its wake, and like most good journeys, the best parts of it:

\begin{flushright}
	\emph{Teach us to care and not to care \\ Teach us to sit still\footnote{\emph{Ash Wednesday, T. S. Eliot}}}
\end{flushright}
\noindent
Therefore, I will use this section to express my gratitude to everyone who has played a major part in it. I did wonder for a while if I were not being indulgent, immodest, or giving a supposedly common experience too much weight, thus flying in the face of my alleged avowal to stoicism. I apologize for breaking tradition and not keeping it the right measure of impersonal and stolid. I also apologize for its length, however, my closest friends, if they were to read it would understand why.

\vspace{1mm}
\noindent
I will begin with my advisor: Gregory Shakhnarovich. I think it would be preposterous to attempt to thank Greg for all that he has done for me and taught me, but I will try. I came to Chicago after an interview with Greg; impulsively changing my mind after having decided to enroll for graduate school in NYC. I was struck with his attention to detail: never allowing any minor detail to be swept under the proverbial rug, in fact often refusing to move forward till it was clarified, thus forcing me to think clearly as a result. Almost all my interactions with Greg seemed to have an inherent didactic value, perhaps by design, since it is something that also reflects in his excellent course. I learned a great deal from my early meetings with him, his wisdom, his flair for fairness, good humour, straight-shooting ways and aversion to bullshit. Other than my parents, Greg is the only person responsible for seeing me through graduate school. Often I meandered through various UChicago departments and thus technically he never had to care or bother, but I always knew that he had my back. I sometimes worry that I might have frequently disappointed Greg, other than testing his patience to the limit. Because of all that Greg has taught me and done for me, I hope I can make him proud someday. Greg was my primary advisor for Parts I \& II of this dissertation.

\vspace{1mm} 
\noindent
I am truly grateful for my interactions with Risi Kondor, who in many ways has been my second advisor. I was drawn to Risi because of his proclivity to gravitate towards deep problems, my own modest undergraduate training in signal processing, and his organizing a study group on the regularity lemma--which was a major component of my master's thesis, which I was curious about. After that I became a regular in all his classes and group meetings, and felt lucky to be associated with his group after summer 2013. Risi is a very deep thinker, with a wide range of knowledge, who has always tried to rub it on to his students, for which I am grateful. Risi was my primary advisor for part III of this dissertation.

\vspace{1mm}
\noindent
Next I would like to thank Samory Kpotufe and Brian D. Nord. Samory was my co-advisor for work presented in chapter \ref{ch:EGOP} of this dissertation. Through him I came to appreciate classical statistical theory and learning theory, as well as the art of thinking about machine learning problems theoretically. It is again difficult to express how grateful I feel for my interactions with Brian, which were both thoroughly enjoyable and uplifting. I thank him for his steady friendship and welcoming me to his astrophysics group at UChicago of which I have been a part of since the summer of 2017. Brian taught me a lot about problems in physics and the applicability of machine learning to them: through his weekly group meetings, through numerous one to one meetings as well our collaboration on a number of projects. In many ways Brian also was like my advisor to who I usually turned towards for counsel in case of professional issues during the last year of school as well the more prosaic side of being a grad student. I am also particularly grateful to Brian for serving on my committee, carefully reading through multiple iterations of this document and giving challenging comments on nearly every page.

\vspace{1mm}
\noindent
I am also thankful to Kevin Gimpel for agreeing to serve on my committee despite the expedited time-line, for his comments to improve the quality of his dissertation, as well as putting up with my ever shifting deadlines with patience. I would also like to thank Rohit Nagpal for being a great teacher, good friend and collaborator. In the stressful period of job applications when I found myself stuck, he was generous enough to take up my problem and not only helped me solve it, but also invested the time to teach me every week and shed my fear of representation theory with no expectation of return.

\vspace{1mm}
\noindent
I am grateful to my co-authors and collaborators with who I have worked on several interesting projects during my time in Chicago (listed in chronological order): Fei Song, Yutao Wang, G\'ab\"or N. S\'ark\"ozy, Neil T. Heffernan, Gregory Shakhnavorich, David McAllester, Samory Kpotufe, Jialei Wang, Behnam Neyshabur, Ryohei Fujimaki, Risi Kondor, Horace Pan, Truong Son Hy, Brandon M. Anderson, Kirk Swanson, Joshua Lequieu, Zhen Lin, Rohit Nagpal, Brian D. Nord, Camille Avestruz, Jo\~ao Caldeira, W. L. Kimmy Wu, Nick Huang and Kyle Story. 

\vspace{1mm}
\noindent
Amongst faculty members at TTI, I would particularly like to thank David McAllester, Karen Livescu, Madhur Tulsiani and Yury Makarychev. David and Karen were amongst my favorite people at the TTI. I have enjoyed almost all my interactions with David and learned a lot from them. In the ocean of the tough crowd that is TTI in scientific matters, I found David's encyclopedic knowledge and his easy warmth refreshing and inspiring. Despite his stature, he always listened to my very frequent and ill-posed ramblings, always patiently error-correcting and re-framing them. I also found his constant presence in the deep learning reading group, that I organized for 5 years, gratifying and learned a lot from his comments, as I frequently found myself to be the presenter. I regret not picking up speed faster and not collaborating with him more. I am grateful to Karen for help on many occasions (especially when I was required to take a course mid-quarter). I regret not writing up my speech course project for publication, despite her suggestion; for it would have been a fitting chapter in this dissertation. I am also thankful to Yury for going out of his way and spending a considerable amount of time, outside the purview of official coursework, to help improve my algorithmic thinking. I am thankful to Madhur for his help on various occasions as the director of graduate studies, as well as helping me with my random theory questions often.

\vspace{1mm}
\noindent
Other than David, Greg, Karen, Kevin, Madhur and Yury, I would also like to thank other permanent faculty members at the TTI: Sadaoki Furui, Avrim Blum, Julia Chuzhoy, Nathan Srebro, Matthew Walter and Jinbo Xu, for their efforts in making TTI a truly lively and vibrant unit, while maintaining the highest research standards. When I started, while it looked like a very interesting place from the outside, I have no hesitation in saying that it was tough for students. However, just in a few years I have seen the change as it matures, and now I see it as an ideal for how an academic unit ought to be organized.

\vspace{1mm}
\noindent
I have also learned a lot from my frequent interactions with the following research assistant professors at TTI: Mohit Bansal, Srinadh Bhojanapalli, Suriya Gunasekar,  Mehrdad Mahdavi, Michael Maire, Subhransu Maji, Mesrob Ohannessian, George Papandreou, Karl Stratos and Ryota Tomioka. In particular I would like to thank Michael and Mesrob for their help on many occasions and Mesrob for being generous with his time, and being game for working through problems and books together.

\vspace{1mm}
\noindent
While I have never been a "course person"; lacking the discipline to do well, I nevertheless made an attempt to sit through all sorts of courses that seemed interesting. In particular I learned a lot from the fantastic courses taught by L\'aszl\'o Babai, Alexander Razborov, Gregory Shaknavorich, David McAllester and Mary Silber. Laci is by far the best teacher I have encountered, and I made it a point to sit through whatever he taught. Greg's well prepared course (that I also had the privilege to TA) and his from-first-principles approach to teaching was the inspiration for my own graduate course at UChicago when I got the opportunity to teach. David's type theory course was more of a philosophy course, that I found both painful and thoroughly enjoyed. Although I was always behind by one week throughout, Mary's dynamical systems course remains the only course in my entire student life, where I have worked through the entire textbook.

\vspace{1mm}
\noindent
Amongst the students at TTI, I would begin with Haris Angelidakis (Geia sou Hari!
Xairomai poly pou eisai Chicago!). I am grateful to Haris for the inspiring company and for being one of my closest friends in Chicago. Haris was one of the few people who I could ask for help without my pride getting in the way, and who I could always ask to come for a cigar or a walk at even 4 in the morning. My equation with Haris was such that if we did not interact even for a day, it felt unusual and weird, and without him, my stay in Chicago would have been that much more drab and uninteresting than what it became. I also learned a lot from our study groups on proofs, measure theory and graph theory. I would also like to thank the early "deserters": Kaustav Kundu, Abhishek Sen and Vikas "Monty Parbat" Garg, who made the first couple of years pass in a jiffy. I would also like to thank Mrinalkanti Ghosh and Omar Montasser for their frequent help. I really respected Mrinal for his command over Kolmogorov complexity and ergodic theory and the frequent conversations about CS theory that I engaged with him. I thank him for putting up with my frequent mood swings, and the deluge of really bad jokes as one of my officemates. I would like to thank Behnam Neyshabur for our work together and Rachit Nimawat for his frequent help and sharing my appreciation of the lake. I would also like to thank Somaye Hashemifar, Avleen Bijral, Falcon Dai, Sudarshan Babu, Igor Vasiljevi\'c, Routian Luo, Kevin Stangl, Mohammadreza Mostajabi, Shane Settle, Lifu Tu, Andrea Daniele, Pedro Savarese, Payman Yadollahpour and Davis Yoshida for making day to day life during grad school fun and enjoyable. Out of the various interns that have passed through TTI, I would like to thank Akash Kumar, Abhishek Sharma and Dimitri Hanukaev. Dima has since become a good friend and it is rare even now for a fortnight to pass without a conversation on Israeli or Indian politics.

\vspace{1mm}
\noindent
I used to joke in my first three years, that if my graduate training were cast as a structured prediction problem, then TTI with its tough crowd, which I often found challenging, would constitute the loss augmented inference part; while UChicago, where I felt smart, would constitute the inference part. I have already referred to the role that Risi, Brian and Rohit have played in my graduate career. But amongst Stats/CS/Booth students and postdocs, I would like to thank Sabyasachi Chatterjee, who I interacted with almost daily since we worked at the same coffee shop; Naiqing Gu for his kindness, confidence in me and introducing me to many interesting problems in networks; Gustav Larsson for always being helpful and inspiring. I am also thankful to my frequent conversations about work, life and research with Goutham Rajendran, Liwen Zhang and Hin Yin Tsang. Out of the students in Risi's group, I am thankful for my interactions with Yi Ding, Brandon Anderson, Jonathan Eskreis-Winkler, Hanna Torrence, Horace Pan and in particular Pramod Kaushik Mudrakarta.

\vspace{1mm}
\noindent
In other reaches of UChicago and my meanderings through it, I would like to thank Ayelet Fine, Ana Ilievska, Katie Shapiro, Alexander Belikov, Pierre Gratia and Julia Thomas. I counted Ana as amongst my good friends in Chicago, and I always appreciated her veering every conversation about machine learning towards humanistic implications. I thank Pierre Gratia for sharing my obsessive love for books and Alexander Belikov for introducing me to many interesting problems in transport and graph curvature. I consider it an honor to count Julia Thomas as amongst my great friends. Despite being a professor (at Notre Dame) and a Japan expert, she always made me feel like the expert, and despite being twice my age she taught me a thing or two about youthfulness. I will miss Julia and our frequent walks circling the lake talking about literature and science. I would also like to thank all the friends I made due to my association with Doc Films, and Kagan Arik for being my Aikido sensei for many years, till I shattered my tarsals.

\vspace{1mm}
\noindent
It might seem out of place for a graduate student to say so, but I also had the privilege to interact with various students initially as a "teacher" (in capacities as TA, full instructor and my tendency to find people to teach privately). Finding it fulfilling, I put a lot of energy into teaching and eventually ended up learning a lot from the experience. In many cases some of the students ended up being great friends as time went by, or my teachers and even collaborators and co-authors. In particular I would like to thank Zhen Lin, who is perhaps one of the smartest and most hardworking people I have known; Kirk Swanson, for our collaboration and introducing me to interesting problems in glassy dynamics; Milica Popovi\'c, for her kindness and her penchant to surprise. I consider it a great honor to be able to call her one of my great friends; Xinguo Fan; Zimo "silent plum" Li; Nasr Maswood, for his friendship and our frequent conversations and meetings despite his moving out of UChicago and Philip Sparks, who I find inspiring and who makes me feel proud.

\vspace{1mm}
\noindent
I have also had three long research visits during my PhD. I would like to thank Ryohei Fujimaki and Yusuke Muruoka for their mentorship and Maxine Clochard and \'Akos Kovacs for the hospitality.

\vspace{1mm}
\noindent
Next I would like to thank all the past and present denizens of 22E; my various house-mates, who have had a major role in making the whole PhD life enjoyable. In particular, I would like to thank Ankan Saha, Pooya Hatami, Sarah Perou, Yuan Li, Emily Schofield, Adil Tobaa, Yael Levy, Thomas Gao and Shubham Toshniwal. I used to really appreciate my frequent conversations with Ankan and Pooya, usually on CS theory, mathematics and politics, extending late in the night till early morning. I am also grateful to Yuan for his generosity and willingness to help unpack my frequent theory questions. I am still psyched by the fact that Yael, who was a Buber scholar, didn't believe during the entirety of her stay that I did not study religion. I am also grateful to Thomas and (SLT specialist) Shubham for putting up with my extremely erratic schedules as might be expected in final year of graduate school. I also enjoyed my interactions with Shubham, who I saw little of at TTI before we became house-mates.

\vspace{1mm}
\noindent
I am very grateful to many of my friends who were around me most of the time I was in Chicago. In particular, I would like to thank Srikant Veeraraghavan for his constant and steady companionship, and our frequent and somehow unplanned adventures. I always looked forward to my weekly meetings with Vaibhav Pandit who somehow ended up in Chicago after a long random walk of his own, thus somehow recreating the time from our high school days. I was always thankful for my interactions with Predrag and Milica Popovi\'c, who often, unknowingly, provided a lot of emotional support. I would also like to thank Pramod Mudrakarta, Elyse, Anamika Acharya, Nora Pfeiffer, Adinath Narasgond, Aniket Joglekar, Gasthi, Brenda Oord and Gabriela J\"ager. 

\vspace{1mm}
\noindent
This acknowledgment section would be incomplete if it did not include a reference to the time I have spent in McGriffet House, where I spent as much time during graduate school as I spent at TTI, and did a sizable chunk of the work that is in this dissertation. Being a regular, I gradually came to know everyone who showed up there, and made many great friends. It helped that most people there, including some Math/Stat professors, assumed I was a professor at UChicago. But I would like to thank three in particular: Muriel Bernardi, Matt Jones and Ben Tianen. Muriel is one of the loveliest people I have met (not just in UChicago), and I am grateful to know her. I always looked forward to her inspiring company in midst of the madness and stress of my final year, and I think the graduate school experience would have been severely impoverished without knowing her. She was a breath of fresh air in the UChicago crowd and I really appreciated her kindness, absurdly funny sense of humor, and how she always inspired me to be a better person. Like in the case of Haris, my almost daily interactions, frequent walks and long conversations with Matt had a major role in keeping life in Chicago interesting.

\vspace{1mm}
\noindent
I am also grateful to my advisors from my earlier pit-stops: G\'abor S\'ark\"ozy, Neil Heffernan and Kalyani Joshi. Not a month went by when I did not hear from atleast one of them, checking on me, encouraging me and constantly making me feel supported.

\vspace{1mm}
\noindent
Outside of TTI-C and UChicago, I would also like to thank Taco S. Cohen, Caglar Gulcehre, Song Liu and Faruk Ahmed. Taco was kind enough to not only share pre-prints of two of his papers, relevant to some of my work, long before they appeared online, but also helped me with detailed instructions to place a chapter in this dissertation (which unfortunately had to be cropped out due to want of time towards the end). Caglar threw a few problems in dynamical systems at me, that were interesting enough for me to take relevant courses. Although I did not end up working on those specific problems, interactions with him have had a lasting impact on me and I foresee using the knowledge acquired in my research in the near feature. I am also thankful to Song for sharing several of his problems on the Stein estimator during his UChicago visit and discussing them, while we attempted a cross-Atlantic collaboration. I am thankful to Faruk for saving me when I somehow landed in Montr\'eal to present the work in chapter \ref{ch:mlng} at NIPS, without any money or a working phone.

\vspace{1mm}
\noindent
I am immensely grateful to all my friends from my time in Pune (a time, which, despite a complete lack of research resources and mentorship, I refer to as my \emph{halcyon seasons, solstice of my days}\footnote{Memoirs of Hadrian, Marguerite Yourcenar}); who I consider to be my best, closest and most dependable friends. Despite the thousands of miles in distance, our friendships only keep getting better with time. I am grateful for their constant support and love, and for being a steady source of strength. It would be impossible to name them without this section becoming as long as the dissertation itself, I could only name Pandit since he was in Chicago. I will however express my gratitude to Ritika, for the role in making me whoever I am today, as well as the early encouragement to pursue research, without which this dissertation would have never happened. 

\vspace{1mm}
\noindent
Finally, I think it would be presumptuous to even attempt to thank my parents and my siblings Divya and Dewanshu for everything that they have done for me; not the least for their sacrifices and encouragement--always supporting me no matter what I did and seeing me through the proverbial yellow brick road. I find it remarkable that despite my parents' backgrounds, the importance of all-round scholarly pursuits and the sacrifice it naturally entails is something they tried to instill in all their children from the very beginning. It is hard to find words to appreciate their dedication and disarming authenticity. I dedicate this thesis to them, as well as to my late grandfather: Prof. Ramesh Chandra Trivedi, whose immense serenity and wisdom I found awe-inspiring, and who might \emph{just} have been proud. 

\begin{flushleft}
Shubhendu Trivedi \\
Cambridge, MA \\
September 1, 2018.
\end{flushleft}

\endgroup

\clearpage%*******************************************************
% Table of Contents
%*******************************************************
\pagestyle{scrheadings}
%\phantomsection
\refstepcounter{dummy}
\pdfbookmark[1]{\contentsname}{tableofcontents}
\setcounter{tocdepth}{2} % <-- 2 includes up to subsections in the ToC
\setcounter{secnumdepth}{3} % <-- 3 numbers up to subsubsections
\manualmark
\markboth{\spacedlowsmallcaps{\contentsname}}{\spacedlowsmallcaps{\contentsname}}
\tableofcontents
\automark[section]{chapter}
\renewcommand{\chaptermark}[1]{\markboth{\spacedlowsmallcaps{#1}}{\spacedlowsmallcaps{#1}}}
\renewcommand{\sectionmark}[1]{\markright{\thesection\enspace\spacedlowsmallcaps{#1}}}
%*******************************************************
% List of Figures and of the Tables
%*******************************************************
\clearpage
\begingroup
    \let\clearpage\relax
    \let\cleardoublepage\relax
    %*******************************************************
    % List of Figures
    %*******************************************************
    %\phantomsection
    \refstepcounter{dummy}
    %\addcontentsline{toc}{chapter}{\listfigurename}
    \pdfbookmark[1]{\listfigurename}{lof}
    \listoffigures

    \vspace{8ex}

    %*******************************************************
    % List of Tables
    %*******************************************************
    %\phantomsection
    \refstepcounter{dummy}
    %\addcontentsline{toc}{chapter}{\listtablename}
    \pdfbookmark[1]{\listtablename}{lot}
    \listoftables

    \vspace{8ex}
    % \newpage

    %*******************************************************
    % List of Listings
    %*******************************************************
    %\phantomsection
    \refstepcounter{dummy}
    %\addcontentsline{toc}{chapter}{\lstlistlistingname}
    \pdfbookmark[1]{\lstlistlistingname}{lol}
    \lstlistoflistings

    \vspace{8ex}

    %*******************************************************
    % Acronyms
    %*******************************************************
    %\phantomsection
    \refstepcounter{dummy}
    \pdfbookmark[1]{Acronyms}{acronyms}
    \markboth{\spacedlowsmallcaps{Acronyms}}{\spacedlowsmallcaps{Acronyms}}
    \chapter*{Acronyms}
    \begin{acronym}[UMLX]
        \acro{DRY}{Don't Repeat Yourself}
        \acro{API}{Application Programming Interface}
        \acro{UML}{Unified Modeling Language}
    \end{acronym}

\endgroup

%********************************************************************
% Mainmatter
%*******************************************************
\clearpage
\pagestyle{scrheadings}
\pagenumbering{arabic}
%\setcounter{page}{90}
% use \cleardoublepage here to avoid problems with pdfbookmark
\cleardoublepage
%************************************************
\chapter{Introduction and Overview}\label{ch:thesisintro}
%************************************************
% Point 1: Learning to compare examples. Having the right notion of similarity/metric and the right representation and how they are linked to each ot
One of the most fundamental questions in machine learning is to compare examples: Given a pair of objects $(\xi_1, \xi_2)$, we want to automatically predict a value $\Psi(\xi_1, \xi_2) \in \mathbb{R}$, the magnitude of which indicates the degree of similarity or dissimilarity between $\xi_1$ and $\xi_2$. To underline the central nature of this problem, it is useful to consider the wide range of machine learning algorithms that explicitly or implicitly rely on a notion of pairwise similarity. Some such methods include: example based approaches such as $k$ nearest neighbors~\cite{NN1967}; clustering algorithms such as $k$-means~\cite{MacQueen}, mean-shift and centroid based methods, spectral clustering~\cite{Luxburg}; the various flavours of kernel methods such as support vector machines~\cite{boser1992}, kernel regression, Gaussian processes~\cite{GPBook} etc. 

\vspace{2mm}
\noindent
The similarity between a pair of objects is customarily obtained as a function of some pre-defined pairwise distance, which in turns depends on the nature of the objects $\xi_1$ and $\xi_2$. If the objects live in an explicit feature space, the Euclidean distance is a common choice; similarly, the $\chi^2$-squared distance is frequently used if the objects reside in a simplex; likewise, the Levenshtein distance may be employed if the objects are strings (see \cite{EncycDistances} for an exhaustive catalog of distance measures). As might be expected, such distances often fail to account for the quirks of a particular dataset and task at hand, and indeed, one might expect improved performance if the distance function is instead tailored to the task. Designing such distance functions automatically is the motivation behind the area of \emph{metric learning}~\cite{Xing02distancemetric}. 

\vspace{2mm}
\noindent
In its most general form, distance metric learning leverages examples provided for the task at hand, in order to wriggle out a better suited, task-specific distance function. For example: If the task is clustering, and we are provided with sets of items and complete clusterings over these sets, we would like to exploit this side information to estimate the distance function that can help cluster future sets better. Yet another example, which is by far most commonly addressed in the metric learning literature is when the task is classification or regression using a nearest neighbor method. The side information furnished to us comprises of labels of points, which are then used to learn a distance function that can improve $k$-NN performance. We explicate further on the latter example in what follows, in a relatively simple setting, to better motivate and build ground to summarize the main contributions of this thesis.

\vspace{2mm}
\noindent
Suppose we are working with a classification problem, which is specified by a suitable instance space $(\mathcal{X}, d)$, assumed to be a metric space, and a label space $\mathcal{Y}$. In particular, we assume that $\mathcal{X} \subset \mathbb{R}^d$, therefore $\xi_1, \xi_2 \in \mathbb{R}^d$. We also assign $\Psi(\xi_1, \xi_2) = +1$ if $\xi_1$ and $\xi_2$ are of the same class, and $\Psi(\xi_1, \xi_2) = -1$ otherwise. Furthermore, given a map $\xi \mapsto \Phi(\xi)$ parameterized by $\mathcal{W}$, let us suppose the distance between $\xi_1$ and $\xi_2$ is given as:

\begin{equation*}
\label{eq:D_W_general_intro}
D_\mathcal{W}(\xi_1, \xi_2) = \|\Phi(\xi_1;\mathcal{W}) - \Phi(\xi_2;\mathcal{W})\|_2^2
\end{equation*}

\begin{figure*}
	\centering
	\begin{center}
		\includepic{.25}{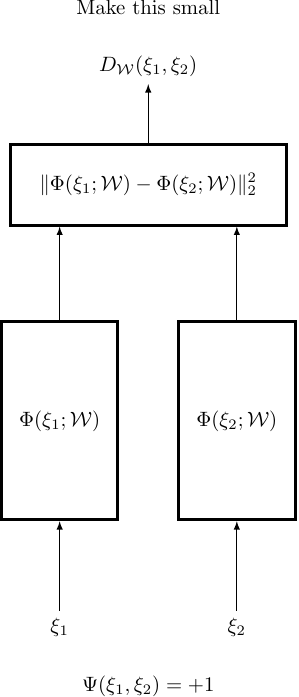}
		\hspace{3cm}
		\includepic{.25}{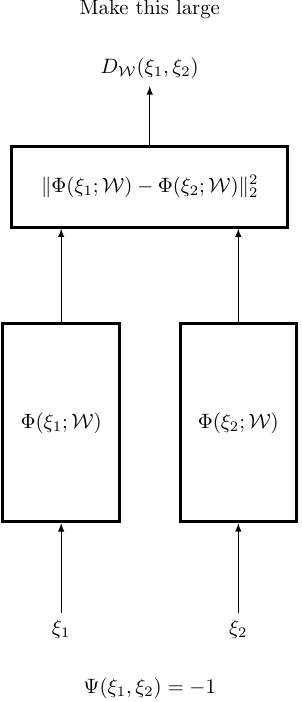}
	\end{center}
	\caption{A typical contrastive loss setup that learns mappings that are similarity sensitive}
	\label{fig:siamese} 
\end{figure*}

\vspace{2mm}
\noindent
We can set the optimization so as to update parameters $\mathcal{W}$ such that $D_\mathcal{W}(\xi_1, \xi_2)$ is increased if $\Psi(\xi_1, \xi_2) = -1$, and $D_\mathcal{W}(\xi_1, \xi_2)$ is decreased if $\Psi(\xi_1, \xi_2) = +1$. This is illustrated in figure \ref{fig:siamese}.

\section{The Interplay between Similarity and Representation Learning}
While the above example illustrates a simple method to learn a similarity-sensitive distance function, there are a few crucial issues that were swept under the proverbial rug, which we unpack below.

\vspace{2mm}
\noindent
First of all, notice that in the example we did not assume anything about the structure of $\xi_1$ and $\xi_2$, except that they were points in $\mathbb{R}^d$. In such a setting $\Phi: \mathbb{R}^d \mapsto \mathbb{R}^p$ (with $d=p$ not necessarily true) corresponds to a mapping, such that in the transformed space distances are more reflective of similarity. In short, we assume that we already have a good feature representation for our data, on top of which a distance function could be learned. Indeed, if the feature representation is poor i.e. has poor class discriminative ability, then learning similarity sensitive distances would be hard if not impossible. 

\vspace{2mm}
\noindent
However, the objects $\xi_1$ and $\xi_2$ might come endowed with richer structure, as is often the case in various applications of machine learning. For example, $(\xi_1, \xi_2)$ might be a pair of images, or a pair of sets, or a pair of spherical images, or a pair of point-clouds and so on. In such cases $\Phi: \xi \mapsto \mathbb{R}^d$ could instead be a module that learns a representation for the object that is inherently discriminative and models natural invariances in the data. 

\vspace{2mm}
\noindent
To drive home this point, consider the example illustrated in figure \ref{fig:siamese} again, but with the modification that the objects $\xi_1$ and $\xi_2$ are large $d \times d$ images, and $\mathcal{W}$ represents the parameters of a fully-connected feed-forward network. The system is then trained to be similarity sensitive as discussed. Such a system is likely to perform poorly, because the fully connected network is unlikely to generate good representations for the images. On the other hand, if $\mathcal{W}$ instead represents the parameters of a Convolutional Neural Network (CNN) \cite{LeCun1989}, the system is far more likely to succeed. That this should be the case is not hard to see, indeed, since CNNs are known to generate extremely good representations for images.

\vspace{2mm}
\noindent
As illustrated by the above example, in the context of similarity learning, there are two notions that are crucial to good performance:
\begin{enumerate}
	\item[1] Having a rich; discriminative representation for the type of input $\xi$, which models natural invariances and symmetries in the data. 
	\item[2] If the underlying task is nearest neighbor classification or regression, as is usually the case in similarity learning, we would want to devise a loss that is a more direct proxy to nearest neighbor performance.
\end{enumerate}

\noindent
Both these notions: having an appropriate representation for the data type at hand, as well as working with the right notion of loss for the learning of similarity reinforce each other, and can also be learned jointly end-to-end. Nevertheless, as already noted, getting both of these aspects in order is pivotal to good performance. In this dissertation, we make contributions towards both aspects, which we describe below, while also outlining the organization of this document. 
 
\section{Discriminative Metric Learning}

In \textbf{Part \ref{pt:discml}} of this dissertation, we wholly focus on the loss formulation for the discriminative learning of similarity and distance, while ignoring representational issues. That is, we assume that the inputs $\xi_i \in \mathbb{R}^d$, and that the representation is good enough for the task at hand. 

\vspace{2mm} 
\noindent
As already discussed, often, $k$-NN prediction performance is the real motivation for metric and similarity learning, on which there is a large literature. Typically such methods set the problem as an optimization problem, with the metric updated in such a way that good neighbors (say from the correct class for a query point) are pulled together, while bad neighbors are pushed away. We utilize \textbf{Chapter \ref{ch:discintro}} to review some popular methods for discriminative metric learning. 

\vspace{2mm} 
\noindent
In \textbf{Chapter \ref{ch:mlng}}, we propose a formulation for metric learning that makes a more direct attempt to optimize for the $k$-NN accuracy as compared to prior work. Our approach considers the choice of $k$ neighbors as a discrete valued latent variable, and casts the metric learning problem as a large margin structured prediction problem. This formulation allows us to use the arsenal of techniques for structural latent support vector machines for the problem of metric learning. We also devise procedures for exact inference and loss augmented inference in this model, and also report experimental results for our method, comparing to a suite of popular metric learning methods.

\vspace{2mm} 
\noindent
In \textbf{Chapter \ref{ch:extensions}}, we consider the direct loss minimization approach to metric learning from Chapter \ref{ch:mlng} and apply it in three different settings: Asymmetric similarity learning, discriminative learning of Hamming distance, and metric learning for improving $k$-NN regression.
 
\section{Metric estimation without learning}
In \textbf{Part \ref{pt:EGOP}} of the dissertation we consider a somewhat different tack: Suppose $\xi_i \in \mathbb{R}^d$ and that this is a good representation of the data. However, we now consider a situation where we are on a limited computational budget i.e. optimizing over a space of possible metrics would be infeasible. Nevertheless we still want access to a good metric that could improve $k$-NN classification and regression performance as compared to the plain Euclidean distance. 

\vspace{2mm}
\noindent
In \textbf{Chapter \ref{ch:EGOP}}, we consider the case of regression and binary classification i.e. when we have an unknown regression function $f: \mathbb{R}^d \to \mathbb{R}$, and consider the metric given by the Expected Gradient Outer Product (EGOP)
$$\expectation_{\mathbf{x}} G(\mathbf{x}) \triangleq \expectation_{\mathbf{x}}\paren{\nabla f(\mathbf{x}) \cdot \nabla f(\mathbf{x})^\top} .$$
We give a cheap estimator for the EGOP and prove that it remains statistically consistent under mild assumptions, while also showing empirically, that using the EGOP as a metric improves $k$-NN regression performance. 

\vspace{2mm}
\noindent
In \textbf{Chapter \ref{ch:EJOP}}, we consider the multi-class case i.e. when we have an unknown function $f: \mathbb{R}^d \to \mathbb{S}^c$ with $\mathbb{S}^c = \{ \mathbf{y} \in \mathbb{R}^c | \forall i \textbf{  } y_i \geq 0, \mathbf{y}^T \mathbf{1} = 1 \}$, and consider the metric given by the Expected Jacobian Outer Product (EJOP)
$$\mathbb{E}_X G(X) \triangleq \expectation_{\mathbf{x}}\paren{\mathbf{J}_{f}(\mathbf{x}) \mathbf{J}_{f}(\mathbf{x})^T}$$
where $\mathbf{J}_{f}$ is the Jacobian of $f$. Like in the case of EGOP, we give a rough estimator, that not only remains statistically consistent under reasonable assumptions, but also gives improvements in $k$-NN classification performance. 
 
\section{Group equivariant representation learning}
 
In \textbf{Part \ref{pt:GCNN}} of this thesis we address the representational issues discussed earlier in this chapter. Chapter \ref{ch:gcnnintro} introduces group equivariant neural networks and makes the case on how such neural networks exploit natural invariances in the data. We argue that group equivariance is an useful inductive bias in many domains. We start with the simple case of planar CNNs, and then review more recent efforts on generalizing classical CNNs in different settings. In chapter \ref{ch:S2CNN} we give an example of a group equivariant representation learning module: a SO(3) equivariant spherical convolution neural network that operates entirely in Fourier space. 

%In Chapter \ref{ch:steerability}, we prove a general theorem that shows that a neural network is equivariant to the action of a compact group on its inputs if and only 
%\subsection{Group equivariance as an inductive bias}
 
%\noindent
%\textbf{Point 5:} Focus on learning representations instead. Where $\xi_1, \xi_2 \in \mathcal{S}^2$ Suppose you only want to focus on the mapping between from input to $\mathbb{R}^d$. How do you do that? (like CNNs for images) Extend the group equivariant formalism to spheres. 
% 
% 
%\noindent
%\textbf{Point 6:} Prove a theorem about such representations in the more general setting i.e. as long as the input is a space that admits a transitive group action.

%*****************************************
%*****************************************
%*****************************************
%*****************************************
%*****************************************

\part{Discriminative Metric Learning}\label{pt:discml}
\cleardoublepage
%************************************************
\newcommand{\knn}{$k$NN\xspace}
\chapter{Introduction to Discriminative Metric Learning}\label{ch:discintro}
%************************************************

Amongst the oldest \cite{NN1967} and most widely used tools in machine learning are nearest neighbor methods (see \cite{GregBook} for a survey). Despite their simplicity, they are often successful and come with attractive properties. For instance, the $k$-NN classifier is universally consistent \cite{Stone77}, being the first learning rule for which this was demonstrated to be the case. Additionally, nearest neighbor methods use local information and are inherently non-linear; while also being relatively resilient to label noise, since prediction requires averaging across $k$ labels. Moreover, it is trivial to add new classes to the data without requiring any fresh model training.

\vspace{2mm}
\noindent
While nearest neighor rules can often be efficacious, their performance tends to be limited by two factors: the computational cost of searching for nearest neighbors and the choice of the metric (distance measure) defining ``nearest''.  The cost of searching for neighbors can be reduced with efficient indexing (see for example \cite{arya1998optimal,datar2004locality,beygelzimer2006cover}) or learning compact representations e.g.~\cite{kulis2009learning,wang2010sequential,norouzi2011minimal,gong2011iterative}. We will defer addressing this issue till Chapter \ref{ch:extensions}. In this part of the dissertation we instead focus on the choice of the metric. The metric is often taken to be Euclidean, Manhattan or $\chi^2$ distance.  However, it is well known that in many cases these choices are suboptimal in that they do not exploit statistical regularities that can be leveraged from labeled data. Here, we focus on supervised metric learning. In particular, we present a method of learning a metric so as to optimize the accuracy of the resulting nearest neighbor estimator.

\vspace{2mm}
\noindent
Existing works on metric learning (the overwhelming majority of which is for classification) formulate learning as an optimization task with various constraints driven by considerations of computational feasibility and reasonable, but often vaguely justified principles \cite{Xing02distancemetric,goldberger2004neighbourhood,globersoncollaps06,weinberger2009distance,weinberger2008fast,mcfee10_mlr,kedem2012non,dannytarlowkNCA}.  
A fundamental intuition is shared by most of the work in this area: an ideal distance for prediction is distance in the target space. Of course, that can not be measured, since prediction of a test example's target is what we want to use the similarities to begin with. Instead, one could learn a similarity measure with the goal for it to be a good proxy for the target similarity. Since the performance of $k$NN prediction often is the real motivation for similarity learning, the constraints typically involve ``pulling'' good neighbors (from the correct class for a given point in the case of classification) closer while ``pushing'' the bad
neighbors farther away. The exact formulation of ``good'' and ``bad'' varies but is defined as a combination of proximity and agreement between targets.  

\vspace{2mm}
\noindent
To give a flavor of some of these constraints and principles in order to improve nearest neighbor performance downstream, we review some well known metric learning algorithms in what follows. Yet another purpose for doing so will also be to set the ground for placing our approach in context. To begin to do so, we first suppose the classification problem is specified by a suitable instance space $(\mathcal{X}, d)$, which is assumed to be a metric space, and a label space $\mathcal{Y} \in \mathbb{Z}_+$. The distance between any two points $\mathbf{x}_i, \mathbf{x}_j \in \mathcal{X}$ is denoted as $D_{\mathcal{W}}(\mathbf{x}_i, \mathbf{x}_j )$, where $\mathcal{W}$ are the parameters that specify the distance measure. There might be considerable freedom in deciding what $\mathcal{W}$ should be, it could just represent the identity matrix, a low-rank projection matrix, or the parameters of a neural network. By far, the most popular family of metric learning algorithms involve learning a Mahalanobis distance. 

\begin{aside}{Mahalanobis Distances}
	Suppose $\mathbf{x}_i, \mathbf{x}_j \in \mathcal{X} \subset \mathbb{R}^d$, and $\mathbf{W} \in \mathbb{R}^{d \times d}$ as well as $ \mathbf{W} \succeq 0$. In the context of metric learning, the Mahalanobis distance has come to refer to all distances of the form $D_{\mathbf{W}}(\mathbf{x}_i, \mathbf{x}_j) = \sqrt{(\mathbf{x}_i - \mathbf{x}_j)^T \mathbf{W} (\mathbf{x}_i - \mathbf{x}_j)}$. However, its eponymous distance measure, proposed in 1936 in the context of anthropometry \cite{mahalanobis}, was defined in terms of a covariance matrix  $\Sigma$ as $D_{\Sigma}(\mathbf{x}_i, \mathbf{x}_j) = \sqrt{(\mathbf{x}_i, \mathbf{x}_j)^T \Sigma^{-1} (\mathbf{x}_i, \mathbf{x}_j)}$. It might also be worthwhile to note, that despite the prevalence of the term ``metric learning", it is somewhat of a misnomer. This is because $D_{\mathbf{W}}$ infact defines a pseudo-metric i.e. $\forall  \mathbf{x}_i, \mathbf{x}_j, \mathbf{x}_k \in \mathcal{X}$, it satisfies:
	\begin{enumerate}
		\item $D_{\mathbf{W}}(\mathbf{x}_i, \mathbf{x}_j) \geq 0 $
		\item $D_{\mathbf{W}}(\mathbf{x}_i, \mathbf{x}_i) = 0 $
		\item $D_{\mathbf{W}}(\mathbf{x}_i, \mathbf{x}_j) = D_{\mathbf{W}}(\mathbf{x}_j, \mathbf{x}_i)$
		\item $D_{\mathbf{W}}(\mathbf{x}_i, \mathbf{x}_k) \leq  D_{\mathbf{W}}(\mathbf{x}_i, \mathbf{x}_j) + D_{\mathbf{W}}(\mathbf{x}_j, \mathbf{x}_k)$
	\end{enumerate}
\end{aside}

\begin{figure*}
	\centering
	\begin{center}
		\includepic{.4}{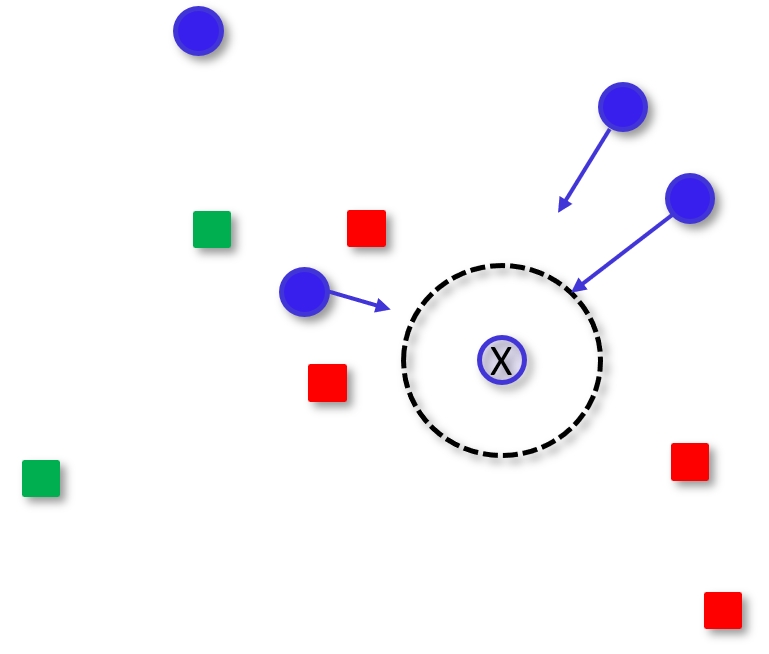}
	\end{center}
	\caption{An illustration of the approach to metric learning taken by ~\cite{Xing02distancemetric}. Look at the text for more details}
	\label{fig:Xing} 
\end{figure*}

\noindent
There is a large body of work on similarity learning done with the stated goal of improving $k$NN performance, which would be impossible to review justly. Therefore, we stick to reviewing some salient approaches that also help place our own approach in context. Some of the earliest work in what could be considered proto-metric learning goes back to Short and Fukunaga (1981)~\cite{Fukunaga1981}, with a string of follow up works in the 90s, for example see Hastie and Tibshirani \cite{HastieTibshirani1996}. However, in much of the recent work in the past decade and a half, the objective can be written as a combination of some sort of regularizer on the parameters of similarity, with loss reflecting the desired ``purity'' of the neighbors under learned similarity. Optimization then balances violation of these constraints with regularization. 

\vspace{2mm}
\noindent
In this sense the area of metric learning could be considered to have started with the influential work of Xing \emph{et al.}~\cite{Xing02distancemetric}. In this method, the ``good'' neighbors are defined as all similarly labeled points, while ``bad'' neighbors are defined as all points that have a different label.  During optimization, the metric is deformed such that each class is mapped into a ball of a fixed radius, but no separation is enforced between the classes (see figure \ref{fig:Xing}). Letting $\mathcal{S}$ and $\mathcal{D}$ denote the sets of pairs of similar and dissmilar points respectively, we may write this approach as the following optimization problem: 

\begin{align}
\label{eq:Xing1}
& \min_{\mathbf{W}} \sum_{(\mathbf{x}_i, \mathbf{x}_j) \in \mathcal{S}} \|\mathbf{x}_i - \mathbf{x}_j\|_{\mathbf{W}}^2 \\
\label{eq:Xing2}
& \textbf{s. t. } \min_{\mathbf{W}} \sum_{(\mathbf{x}_i, \mathbf{x}_j) \in \mathcal{D}} \|\mathbf{x}_i - \mathbf{x}_j\|_{\mathbf{W}}^2 \geq 1 \\
& \mathbf{W} \succeq 0
\end{align}

\noindent
Where, $\|\cdot\|_{\mathbf{W}}^2$ is the squared Mahalanobis distance parameterized by $\mathbf{W}$. Evidently, the immediate problem with this approach is that it has little relation to the actual $k$NN objective. Indeed, the $k$-NN objective does not require that similar points should be clustered together, and as a consequence methods of a similar flavour optimize an objective that is much harder than what is required for good $k$-NN performance.

\begin{figure*}
	\centering
	\begin{center}
		\includepic{.4}{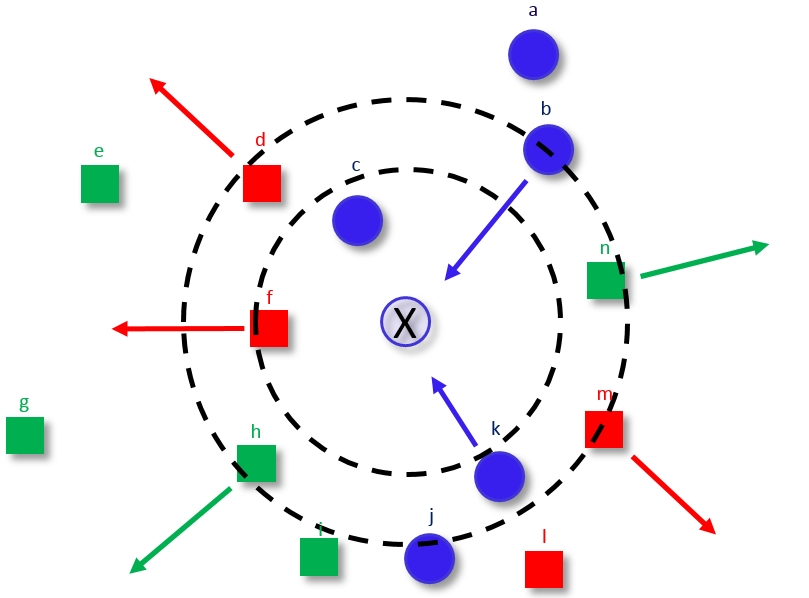}
	\end{center}
	\caption{An illustration of the approach to metric learning taken by ~\cite{weinberger2009distance}. We refer the reader to the text for more details}
	\label{fig:LMNN_new} 
\end{figure*}

\vspace{2mm}
\noindent
A popular family of approaches to metric learning that has a somewhat better motivated objective than the above, are based on the Large Margin Nearest Neighbor (LMNN) algorithm~\cite{weinberger2009distance}. In LMNN, the constraints for each training point involve a set of predefined ``target neighbors'' from the correct class, and ``impostors'' from other classes. The optimization is such that a margin is imposed between the ``target neighbors'' and the ``impostors''(see figure \ref{fig:LMNN_new}). Such an objective may be written as:

\begin{equation}
\min_{\mathbf{L}} \sum_{i,j: j \rightsquigarrow i} D_{\mathbf{L}} (\mathbf{x}_i, \mathbf{x}_j)^2 + \mu \sum_{k: y_i \neq y_k} [1 +  D_{\mathbf{L}} (\mathbf{x}_i, \mathbf{x}_j)^2 -  D_{\mathbf{L}} (\mathbf{x}_i, \mathbf{x}_k)^2]_{+}
\end{equation}

\noindent
where $\mathbf{W} = \mathbf{L}^T \mathbf{L}$; $\mu > 0$; $y_i$ denotes the label for $\mathbf{x}_i$; the notation $j \rightsquigarrow i$ indicates that $\mathbf{x}_j$ is a ``target neighbor''of $\mathbf{x}_i$ and $[z]_+ = \max(z,0)$ denotes the hinge loss.

\vspace{2mm}
\noindent
Despite being somewhat more suited to the underlying $k$-NN objective, the LMNN objective still has some issues that could affect its performance. To begin, the set of target neighbors are chosen at the onset based on the euclidean distance (in absence of a priori knowledge). Moreover as the metric is optimized, the set of 
``target neighbors'' is not dynamically updated. There is no reason to believe that the original choice of neighbors based on the euclidean distance is optimal while the metric is updated. Yet another issue is that in LMNN the target neighbors are forced to be of the same class. In doing so it does not fully leverage the power of the $k$NN objective, which only needs a majority of points to have the correct label. Extensions of LMNN~\cite{weinberger2008fast,kedem2012non} allow for non-linear metrics, but retain
the same general flavor of constraints.

\vspace{2mm}
\noindent
In Neighborhood Component Analysis (NCA)~\cite{goldberger2004neighbourhood} a different kind of proxy for
classification error is used: the piecewise-constant error of the
$k$NN rule is replaced by a soft version. This leads to a non-convex objective that is optimized via gradient descent. To write the objective, we denote the probability that a point $\mathbf{x}_i$ selects $\mathbf{x}_j$ as its neighbor by $p_{ij}$. In this set up, the point $\mathbf{x}_i$ will be assigned the class of point $\mathbf{x}_j$. Given $\mathbf{W} = \mathbf{L}^T\mathbf{L}$, we can define $p_{ij}$ as:

\begin{equation}
\displaystyle p_{ij} = \frac{\exp\Big(- \|\mathbf{L}\mathbf{x}_i - \mathbf{L}\mathbf{x}_j \|_2^2 \Big)}{\sum_{k \neq i}\exp\Big(- \|\mathbf{L}\mathbf{x}_i - \mathbf{L}\mathbf{x}_k \|_2^2 \Big)} \text{ and } p_{ii} = 0
\end{equation}

\noindent 
The probability that a point will be correctly classified is then given by: 

\begin{equation}
p_{i} = \sum_{\mathbf{x}_j \in C_i} p_{ij}
\end{equation}

\noindent
where $C_i$ denotes the set of all points that have the same class as $\mathbf{x}_i$. Finally, we can write the NCA objective (to be maximized) as follows: 

\begin{equation}
Obj(\mathbf{L}) = \sum_{i} \sum_{\mathbf{x}_j \in C_i} p_{ij}
\end{equation}

\noindent
One of the features of NCA is that it trades off convexity for attempting to directly optimize for the choice of nearest neighbor. This issue of non-convexity was partly remedied in~\cite{globersoncollaps06}, by optimization of a similar stochastic rule while attempting to collapse each class to one point. While this makes the optimization convex, collapsing classes to distinct points is unrealistic in practice. Another recent extension of NCA~\cite{dannytarlowkNCA} generalizes the stochastic classification idea to $k$NN classification with $k>1$. Out of all the methods reviewed so far, $k$-NCA is the only method that comes closest to optimize directly for the $k$-NN task loss. 

\vspace{2mm}

\noindent
There is also a wide plethora of metric learning methods that optimize for some kind of ranking loss. We discuss two examples here. In Metric Learning to Rank (MLR)\cite{mcfee10_mlr}, the constraints involve all the points: the goal is to push all the correct matches in front of all the incorrect ones. The idea essentially is: given a query point and a metric parameterized by $\mathbf{W}$ finding the distance with the database points should sort them in such a way that good neighbors end up in the front. While important for retrieval, this is again not the same as requiring correct classification. In addition to global optimization constraints on the rankings (such as mean average precision for target class), the authors allow localized evaluation criteria such as Precision at $k$, which can be
used as a surrogate for classification accuracy for binary classification, but is a poor surrogate for multi-way
classification. Direct use of $k$NN accuracy in optimization objective is briefly mentioned in~\cite{mcfee10_mlr}, but not pursued due to the difficulty in loss-augmented inference. This is because the interleaving technique of~\cite{Joachims05asupport} that is used to perform inference with other losses based inherently on contingency tables, fails for the multiclass case (since the number of data interleavings could possibly be exponential). A similar approach is taking in~\cite{norouzi2012hamming}, where the constraints are derived from triplets of points formed by a sample, correct and incorrect neighbors. Again, these are assumed to be set statically as an input  to the algorithm, and the optimization focuses on the distance ordering (ranking) rather than accuracy of classification.

\vspace{2mm}
\noindent
Before concluding, we must note that in this chapter we have not reviewed any of the deep metric learning techniques. This is because all such techniques that we are aware of are based on a flavour of loss as one of the above, with the only difference that the mapping of each point onto a metric space is done by a neural network. The focus and main novelty of the work presented in the next chapter lies in its loss as compared to the techniques discussed. Indeed, the function that is used to map the points to a suitable metric space is an orthogonal consideration. 

\vspace{2mm}
\noindent
Having considered a general background on the metric learning problem, along with various loss formulations that have been proposed to attack it, we now proceed to give a formulation that attempts to give a more direct proxy for $k$-NN classification.

\cleardoublepage
%************************************************
\chapter{Metric Learning by Neighborhood Gerrymandering}\label{ch:mlng}
%************************************************

\begin{aside}{Outline}
	In this chapter, we give a formulation for metric learning that facilitates a more direct attempt to optimize for the $k$NN accuracy as compared to previous work. We also show that our formulation makes it natural to apply standard learning methods for structural latent support vector machines (SVMs) to the problem of supervised metric learning. While we test this approach for the case of Mahalanobis metric learning, as emphasized in the previous chapter, the focus here is on obtaining a better proxy for the $k$-NN loss rather than the nature of mapping. 
	\end{aside}

To achieve our stated goal of formulating the metric learning problem such that it is more direct in optimizing for the underlying task: $k$-NN accuracy, we consider looking at the nearest neighbor problem a bit differently. 

\vspace{2mm}
\noindent
In the $k$NN prediction problem, given a query point and fixing the underlying metric, there is an implicit hidden variable: the choice of $k$ ``neighbors''. The inference of the predicted label from these $k$  examples is trivial: by simple majority vote among the associated labels for classification, and by taking a weighted average in the case of regression. In the case of classification, given a query point, there can possibly exist a very large number of choices of $k$ points that might correspond to zero loss: any set of $k$ points with the majority of correct class will do. Whereas, in the case of regression, there can exist a very large number of choices of $k$ points that might correspond to a loss less than a tolerance parameter (since zero loss would be impossible in most scenarios). We would like a metric to ``prefer'' one of these ``good'' example sets over any set of $k$ neighbors which would vote for a wrong class (or in the case of regression correspond to a high loss). Note that to win, it is not necessary for the right class to account for all the $k$ neighbors -- it just needs to get more votes than any other class. As the number of classes and the value of $k$ grow, so does the space of available good (and bad) example sets.

\vspace{2mm}
\noindent
These considerations motivate our approach to metric learning. It is akin to the common, albeit negatively viewed, practice of \emph{gerrymandering} in drawing up borders of election districts so as to provide advantages to desired political parties, e.g., by concentrating voters from that party or by spreading voters of opposing parties. In our case, the ``districts'' are the cells in the Voronoi diagram defined by the Mahalanobis metric, the ``parties'' are the class labels voted for by the neighbors falling in each cell, and the ``desired winner'' is the true label of the training points associated with the cell. This intuition is why we refer to our method as \emph{neighborhood gerrymandering} in the title.

\vspace{2mm}
\noindent
A bit more technically, we write $k$NN prediction as an inference problem with a structured latent variable being the choice of $k$ neighbors. Thus learning involves minimizing a sum of a structural latent hinge loss
and a regularizer \cite{boser1992}. Computing structural latent hinge loss involves loss-adjusted inference --- one must compute loss-adjusted values of both the output value (the label) and the latent items (the set of nearest neighbors). The loss augmented inference corresponds to a choice of worst $k$ neighbors in the sense that while having a high average similarity they also correspond to a high loss (``worst offending set of $k$ neighbors''). Given the inherent combinatorial considerations, the key to such a model is efficient inference and loss augmented inference. We give an efficient algorithm for {\em exact} inference. We also design an optimization algorithm based on stochastic gradient descent on the surrogate loss. Our approach achieves $k$NN accuracy higher than state of the art for most of the data sets we tested on, including some methods specialized for the relevant input domains.

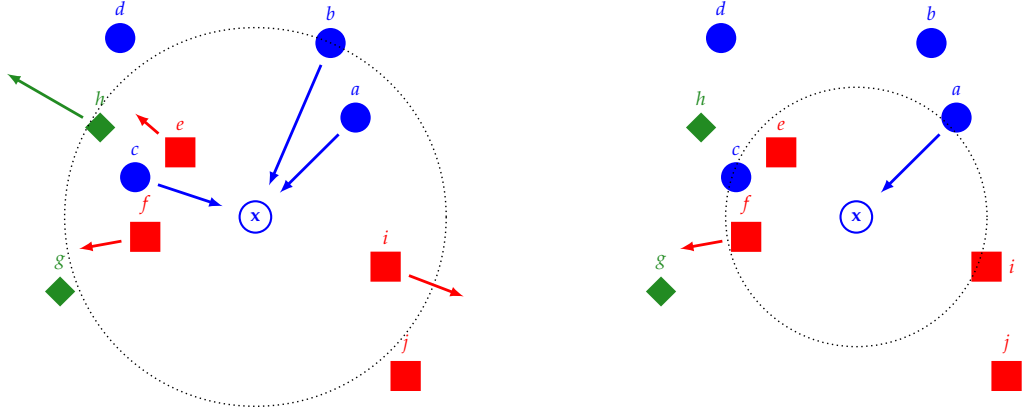
\begin{figure*}[!th]
	\centering
	\resizebox{.9\textwidth}{!}{%\documentclass{article}
%\usepackage[svgnames]{xcolor}
%\usepackage{pgf}
%\usepackage{tikz}
%\usepackage{amsmath,amssymb}
%\usepackage{colortbl}
%\usepackage{xspace}

%\begin{document}

%\usetikzlibrary{shapes,calc,positioning,patterns,decorations.pathmorphing}

\begin{tikzpicture}
\draw[blue,very thick] (0,0) node[circle,minimum size=16pt,draw] (x) {$\mathbf{x}$};
\draw[blue,very thick,fill=blue] (2,2) node[circle,minimum
size=16pt,draw,fill,label = $a$] (np1) {};
\draw[blue,very thick,fill=blue] (1.5,3.5) node[circle,minimum
size=16pt,draw,fill, label = $b$] (np2) {};
\draw[blue,very thick,fill=blue] (-2.4,.8) node[circle,minimum
size=16pt,draw,fill, label = $c$] (np3) {};
\draw[blue,very thick,fill=blue] (-2.7,3.6) node[circle,minimum
size=16pt,draw,fill, label = $d$] (np4) {};
\draw[red,very thick,fill=red] (-1.5,1.3) node[minimum
size=16pt,draw,fill, label = $e$] (nnr1) {};
\draw[red,very thick,fill=red] (-2.2,-.4) node[minimum
size=16pt,draw,fill, label = $f$] (nnr2) {};
%\draw[ForestGreen,very thick,fill=ForestGreen] (-4,-2.3) node[diamond,minimum
%size=16pt,draw,fill, label = $g$] (nng1) {};
\draw[ForestGreen,very thick,fill=ForestGreen] (-3.9,-1.5) node[diamond,minimum
size=16pt,draw,fill, label = $g$] (nng2) {};
\draw[ForestGreen,very thick,fill=ForestGreen] (-3.1,1.8) node[diamond,minimum
size=16pt,draw,fill, label = $h$] (nng3) {};
\draw[red,very thick,fill=red] (2.6,-1) node[minimum
size=16pt,draw,fill, label = $i$] (nnr3) {};
\draw[red,very thick,fill=red] (3.0,-3.2) node[minimum
size=16pt,draw,fill, label = $j$] (nnr4) {};
\draw[dotted,thick] (0,0) circle (3.81);
\foreach \n in {1,2,...,3} {
  \draw[blue,ultra thick,-latex,shorten >=10pt,shorten <=5pt] (np\n)
  -- (x);}
\foreach \n in {1,2,...,3} {
  \draw[red,ultra thick,-latex,shorten <=5pt] (nnr\n) -- ($
  (nnr\n)!.6!180:(x) $);}
\foreach \n in {3} {
  \draw[ForestGreen,ultra thick,-latex,shorten <=5pt] (nng\n) -- ($ (nng\n)!.6!180:(x) $);}
\draw[xshift=120mm,blue,very thick] (0,0) node[circle,minimum size=16pt,draw] (x) {$\mathbf{x}$};
\draw[xshift=120mm,blue,very thick,fill=blue] (2,2) node[circle,minimum
size=16pt,draw,fill, label = $a$] (np1) {};
\draw[xshift=120mm,blue,very thick,fill=blue] (1.5,3.5) node[circle,minimum
size=16pt,draw,fill, label = $b$] (np2) {};
\draw[xshift=120mm,blue,very thick,fill=blue] (-2.4,.8) node[circle,minimum
size=16pt,draw,fill, label = $c$] (np3) {};
\draw[xshift=120mm,blue,very thick,fill=blue] (-2.7,3.6) node[circle,minimum
size=16pt,draw,fill, label = $d$] (np4) {};
\draw[xshift=120mm,red,very thick,fill=red] (-1.5,1.3) node[minimum
size=16pt,draw,fill, label = $e$] (nn1) {};
\draw[xshift=120mm,red,very thick,fill=red] (-2.2,-.4) node[minimum
size=16pt,draw,fill, label = $f$] (nn2) {};
%\draw[ForestGreen,very thick,fill=ForestGreen] (-4,-2.3) node[diamond,minimum
%size=16pt,draw,fill, label = $g$] (nn3) {};
\draw[xshift=120mm,ForestGreen,very thick,fill=ForestGreen] (-3.9,-1.5) node[diamond,minimum
size=16pt,draw,fill, label = $g$] (nn4) {};
\draw[xshift=120mm,ForestGreen,very thick,fill=ForestGreen] (-3.1,1.8) node[diamond,minimum
size=16pt,draw,fill, label = $h$] (nn4) {};
\draw[xshift=120mm,red,very thick,fill=red] (2.6,-1) node[minimum
size=16pt,draw,fill, label=right:$i$] (nn5) {};
\draw[xshift=120mm,red,very thick,fill=red] (3.0,-3.2) node[minimum
size=16pt,draw,fill, label = $j$] (nn6) {};
\draw[xshift=120mm,dotted,thick] (0,0) circle (2.61);
\draw[xshift=120mm,blue,ultra thick,-latex,shorten >=10pt,shorten <=5pt] (np1) -- (x);
%\draw[blue,ultra thick,-latex,shorten >=10pt,shorten <=5pt] (np3) -- %(x);
%\draw[red,ultra thick,-latex,shorten <=5pt] (nn6) -- ($ %(nn6)!.6!180:(x) $);
\draw[xshift=120mm,red,ultra thick,-latex,shorten <=5pt] (nn2) -- ($ (nn2)!.6!180:(x) $);
\end{tikzpicture}}
	\caption{Illustration of objectives of LMNN (left) and our
		structured approach to ``neighborhood gerrymandering'' (right) for $k=3$. The point $\mathbf{x}$ of
		class blue is the query point. In LMNN, the target points are the
		nearest neighbors of the same class, which are points $a, b$ and
		$c$ (the circle centered at $\mathbf{x}$ has radius equal to th
		e farthest of the target points i.e. point b). The LMNN objective will push all the points of the wrong class that lie inside this circle out (points $e, f, h, i, \text{and} j$), while pushing in the target points to enforce the margin. On the other hand, for our structured approach (right), the circle around $\mathbf{x}$ has radius equal to the distance of the farthest of the three nearest neighbors irrespective of class. Our objective only needs to ensure zero loss. This would be achieved by pushing in point $a$ of the correct class (blue) while pushing out the point having the incorrect class (point $f$). Note that two points of the incorrect class lie inside the circle ($e, \text{and} f$), both being of class red. However $f$ is pushed out and not $e$ since it is farther from $\mathbf{x}$.}
	\label{fig:LMNNComparison}
\end{figure*}

\vspace{2mm}
\noindent
As stated toward the end of the previous chapter, although we initially restrict ourselves to learning a Mahalanobis distance in an explicit feature space, the formulation is easily extensible to nonlinear similarity measures such as those defined by nonlinear kernels, provided computing the gradients of similarities with respect to metric parameters is feasible. In such extensions, the inference and loss augmented inference steps remain unchanged. Our formulation can also naturally handle a user-defined loss matrix on labels rather than just a zero-one loss. We propose a series of extensions to the case of $k$NN regression, Asymmetric metric learning and the discriminative learning of Hamming distance in Chapter \ref{ch:extensions}. The extension to regression seems particularly foreboding given that in this case  the number of ``classes'' is uncountable. This is attacked by both modifying the objective suitably and presenting algorithms for inference and loss augmented inference to give a suitable approximation that is shown to perform well on standard benchmarks.

\section{Gerrymandering in Context}

In chapter \ref{ch:discintro}, we introduced the metric learning problem and discussed some canonical approaches to the problem. In this short section, we hark back to some of the approaches discussed there to put our framework in context. 

\begin{enumerate}
	\item[1] \textbf{Clustering type objectives: } Such methods for learning the metric are exemplified by the work of Xing \emph{et al.}~\cite{Xing02distancemetric}. In such approaches, ``good'' neighbors for a given query point are all points that have a similar label. The metric is learned so as to map all such ``good'' neighbors into a ball of fixed radius. However, the $k$-NN objective does not require such clustering behaviour for good performance. In that sense our approach is more direct in leveraging the $k$-NN objective. 
	\item[2] \textbf{LMNN type objectives: } As discussed earlier, in LMNN type algorithms~\cite{weinberger2009distance}, for a query point, the ``good'' neighbors are a set of ``target neighbors'' which are a) predefined and b) are all of the same class. In a way, the role of ``target neighbors'' in LMNN is not quite unlike the ``best correct set of $k$ neighbors'' ($h^\ast$	in Section~\ref{sec:learn}) in our method. Moreover, in LMNN type methods, the``target neighbors'' are predefined based on the Eucidean metric and then fixed throughout learning. In our method, the set of ``good'' neighbors $h^\ast$ are dynamically updated as the metric is learned. Yet another departure in our approach to that of LMNN is that for good $k$-NN performance, we don't need all the ``good'' neighbors to be of the same class. In our method we provide inference procedures that ensure leveraging the $k$-NN objective more directly by only focusing on having a majority of points to be of the correct class. This is also illustrated in an example in figure \ref{fig:LMNNComparison}. 
	\item[3] \textbf{NCA: } Our method is similar to NCA~\cite{goldberger2004neighbourhood} type methods in that it also trades off convexity (details in Section~\ref{sec:learn}) in order to directly optimize for the choice of nearest neighbor. However, traditional NCA type algorithms focus only on 1-NN. The work of ~\cite{dannytarlowkNCA} that generalizes ~\cite{goldberger2004neighbourhood} to instead focus on the right selection of $k$ nearest neighbors for classification is closest in spirit to our work, and as far as we are aware the only work attempts to optimize directly for the $k$-NN objective. Experimentally, we found our method gave superior performance.
	\item[4] \textbf{Ranking objectives: } The original inspiration for this work was \emph{metric learning to rank}\cite{mcfee10_mlr}, which optimizes for a ranking objective. As discussed this is not the same as requiring correct classification. Moreover, as discussed the approach of \cite{mcfee10_mlr} fails for the multiclass case given the inference used. We take a very different approach to loss augmented inference, using targeted inference
	and the classification loss matrix, and can easily extend it to	arbitrary number of classes
\end{enumerate}

\section{Discriminative loss minimization for classification}
\label{sec:setup}

In this section we formally set up the problem. Note that we first state the distance and similarity formulation in its full generality, to illustrate that it is more widely applicable, and not just to the case of learning global linear projections. We then modify it to work with the Mahalanobis distance that is eventually dealt with in the rest of this chapter, and for which detailed experimentation is carried out.  

\newcommand{\maha}[1]{D_{#1}\xspace}
\newcommand{\mahax}[3]{D_{#1}\left({#2},{#3}\right)\xspace}
\newcommand{\dscore}[1]{S_{#1}\xspace}
\newcommand{\shat}{\mathcal{D}\xspace}
\newcommand{\hstar}{\mathbf{h}^\ast}
\newcommand{\hhat}{\widehat{\mathbf{h}}}

\subsection{Problem setup}
We are given $N$ training examples
$X=\{\mathbf{x}_1,\ldots,\mathbf{x}_N\}$, represented by a ``native''
feature map, $\mathbf{x}_i\in\mathbb{R}^d$, and their 
 class labels $\mathbf{y}=[y_1,\ldots,y_N]^T$, with
$y_i\in[R]$, where $[R]$ stands for the set $\{1,\ldots,R\}$. We are
also given the loss matrix $\mathbf{\Lambda}$ with $\Lambda(r,r')$
being the loss incurred by predicting $r'$ when the correct class is $r$. We assume
$\Lambda(r,r)=0$, and $\forall (r,r')$, $\Lambda(r,r')\ge0$.
Most generally, we are interested in squared distances defined as:
\begin{equation}
  \label{eq:D_W_general}
  D_\mathcal{W}(\mathbf{x},\mathbf{x}_i) = \|\Phi(\mathbf{x};\mathcal{W}) - \Phi(\mathbf{x}_i;\mathcal{W})\|_2^2
\end{equation}
Where $\Phi(\mathbf{x};\mathcal{W})$ is a map (possibly non-linear), $\mathbf{x} \to \Phi(\mathbf{x})$, parameterized by $\mathcal{W}$.
Let $h\subset X$ be a set of examples in $X$. For a given $\mathcal{W}$ we define the distance
score of $h$ w.r.t. a point $\mathbf{x}$ as
\begin{equation}
  \label{eq:S_W_general}
  \dscore{\mathcal{W}}(\mathbf{x},h)\,=\, - \frac{\alpha}{K}\sum_{\mathbf{x}_j\in h} \|\Phi(\mathbf{x};\mathcal{W}) - \Phi(\mathbf{x}_j;\mathcal{W})\|_2^2 + \beta
\end{equation}
where $\alpha$, $\beta$ and $K$ are constants. This formulation of the distance score permits use of the dot product to measure similarity as well, as long as it is normalized to unit length.

\vspace{2mm}
\noindent
For the rest of this chapter, we are interested in the \emph{Mahalanobis metrics}
\begin{equation}
  \label{eq:D_W}
  \mahax{\mathbf{W}}{\mathbf{x}}{\mathbf{x}_i}\,=\,
  \left(\mathbf{x}-\mathbf{x}_i\right)^T\mathbf{W}\left(\mathbf{x}-\mathbf{x}_i\right),
\end{equation}
which are parameterized by positive semidefinite $d\times d$ matrices
$\mathbf{W}$, which can be seen as learning a linear map $\mathbf{x} \to \mathbf{L}\mathbf{x}$ where $\mathbf{W} = \mathbf{L}^T\mathbf{L}$, while satisfying fixed constraints (usually to optimize for $k$NN performance).
For a given $\mathbf{W}$ we define the distance
score of $h$ w.r.t. a point $\mathbf{x}$ as
\begin{equation}
  \label{eq:S_W}
  \dscore{\mathbf{W}}(\mathbf{x},h)\,=\,
-\sum_{\mathbf{x}_j\in h}
\mahax{\mathbf{W}}{\mathbf{x}}{\mathbf{x}_j}
\end{equation}
Hence, the set of $k$ nearest neighbors of $\mathbf{x}$ in $X$ is
\begin{equation}
  \label{eq:h_W}
  h_{\mathbf{W}}(\mathbf{x})\,=\,\argmax_{|h|=k}\dscore{\mathbf{W}}(\mathbf{x},h).
\end{equation}
For the remainder of this discussion, we will assume that $k$ is known and fixed. Note that, from any set $h$ of $k$ examples from $X$, we can predict the label of
$\mathbf{x}$ by 
(simple) majority vote:
\[\widehat{y}\left(h\right)\,=\,
\text{majority}\{y_j:\,\mathbf{x}_j\in h\},
\]
with ties resolved by a heuristic, e.g., according to 1NN vote.
In particular, the \knn classifier predicts
$\widehat{y}(h_{\mathbf{W}}(\mathbf{x}))$. 
Due to this deterministic dependence between $\widehat{y}$ and $h$, we
can define the classification loss incurred by a voting classifier when
using the set $h$ as
\begin{equation}
  \label{eq:lossh}
  \Delta(y,h)\,=\,\Lambda\left(y,\widehat{y}(h)\right).
\end{equation}

 \label{setup}

\section{Learning and inference}\label{sec:learn}

One might want to learn $\mathbf{W}$ to minimize training loss
$$\sum_i\Delta\left(y_i,h_{\mathbf{W}}(\mathbf{x}_i)\right)$$ 

\noindent 
However, this fails due to the intractable nature of classification loss $\Delta$. We will
follow the usual remedy: define a tractable surrogate loss. 

\vspace{2mm}
\noindent
While already discussed earlier, here we must note again that in our formulation, the output of the prediction is
a structured object $h_{\mathbf{W}}$, for which we eventually report the deterministically computed $\widehat{y}$. Structured prediction problems usually involve loss which is a generalization of the hinge loss; intuitively, it penalizes the gap between score of the correct structured output and the score of the ``worst offending'' incorrect
output (the one with the highest score \emph{and} highest $\Delta$).

\vspace{2mm}
\noindent
However, in our case, we have an additional complication in that there is no single correct output $h$, since in
general many choices of $h$ would lead to correct $\widehat{y}$ and zero classification loss: any $h$ in which the majority votes for the right class. Ideally, we want $\dscore{\mathbf{W}}$ to prefer \emph{at least
one} of these correct $h$s over all incorrect $h$s.

\vspace{2mm}
\noindent
This intuition leads to the following surrogate loss definition:
\begin{eqnarray}
  \label{eq:loss1}
  L(\mathbf{x},y,\mathbf{W})\,=\,
\max_{h}\left[
  \dscore{\mathbf{W}}(\mathbf{x},h)\,+\,\Delta(y,h)
\right]&\\
\label{eq:loss2}
-\,\max_{h:\Delta(y,h)=0}\dscore{\mathbf{W}}(\mathbf{x},h).&
\end{eqnarray}
This is quite different in spirit from the notion of margin sometimes
encountered in ranking problems where we want all the correct answers
to be placed ahead of all the wrong ones. Here, we only care to put
\emph{one} correct answer on top; it does not matter which one, hence
the $\max$ in~\eqref{eq:loss2}.

\section{Structured Formulation}
Our choice of the loss $L$ was motivated by intuitive arguments for what might correspond to a better proxy for the underlying task of $k$-NN prediction. However, it turns out that our problem is an instance of a familiar type of
problems: latent structured prediction~\cite{yu2009learning}, and thus
our choice of loss can be shown to form an upper bound on the
empirical task loss $\Delta$. 

\vspace{2mm}
\noindent
First, we note that the score $\dscore{\mathbf{W}}$ can be written as
\begin{equation}\label{eq:F}
\dscore{\mathbf{W}}(\mathbf{x},h)\,=\,
\left<\mathbf{W},-\sum_{\mathbf{x}_j\in h}(\mathbf{x}-\mathbf{x}_j)(\mathbf{x}-\mathbf{x}_j)^T\right>,
\end{equation}
where $\left<\cdot,\cdot\right>$ stands for the Frobenius inner product.
Defining 
the \emph{feature map}
\begin{equation}
  \label{eq:psi}
  \mathbf{\Psi}(\mathbf{x},h)\,\triangleq\,-\sum_{\mathbf{x}_j\in
  h}(\mathbf{x}-\mathbf{x}_j)(\mathbf{x}-\mathbf{x}_j)^T,
\end{equation}
we get a more compact expression
$\left<\mathbf{W},\mathbf{\Psi}(\mathbf{x},h)\right>$
for~\eqref{eq:F}.

\vspace{2mm}
\noindent 
Going a step further, we can
encode the deterministic dependence between $y$ and $h$ by a so-called
``compatibility''  function 

\[
A(y,h) = \Bigg\{\begin{array}{lr}
0 & \text{ if } y=\widehat{y}(h)\\
-\infty & \text{ otherwise }
\end{array}
\]

\vspace{2mm}
\noindent 
This notion of compatibility allows us to write the joint inference of $y$ and (hidden) $h$ 
performed by \knn classifier as
\begin{equation}
  \label{eq:hymax}
  \widehat{y}_\mathbf{W}(\mathbf{x}),\widehat{h}_\mathbf{W}(\mathbf{x})\,=\,
  \argmax_{h,y}\left[A(y,h)+\left<\mathbf{W},\mathbf{\Psi}(\mathbf{x},h)\right>\right].
\end{equation}
This is the familiar form of inference in a latent structured
model~\cite{yu2009learning,felzenszwalb2010object} with latent
variable $h$. So, notwithstanding the somewhat unusual property of our model where the latent $h$ completely
determines the inferred $y$, we can show the equivalence to ``normal'' latent structured prediction.

\subsection{Learning by gradient descent}
\label{sec:ssd}
We define the objective in learning $\mathbf{W}$ as
\begin{equation}
  \label{eq:obj}
  \min_{\mathbf{W}}\,\|\mathbf{W}\|_F^2\,+\,C\sum_i L\left(\mathbf{x}_i,y_i,\mathbf{W}\right),
\end{equation}
where $\|\cdot\|_F^2$ stands for Frobenius norm of a
matrix.\footnote{We discuss other choices of regularizer in
  Section~\ref{sec:discuss}.} The regularizer is convex, but as in other latent
structured models, the loss $L$ is non-convex due to the subtraction
of the $\max$ in~\eqref{eq:loss2}.  
To optimize~\eqref{eq:obj}, one can use the convex-concave procedure
(CCCP)~\cite{yuille2002concave} which has been proposed specifically
for latent SVM learning~\cite{yu2009learning}. However, CCCP tends to be slow on
large problems. Furthermore, its use is complicated here due to the
requirement that $\mathbf{W}$ be positive semidefinite (PSD). This means that
the inner loop of CCCP includes solving a semidefinite program, making
the algorithm slower still. Instead, we opt for a much faster, and
perhaps simpler, choice: stochastic gradient descent (SGD), described in Algorithm~\ref{alg:gd}.

\begin{algorithm}[!th]
  \caption{Stochastic gradient descent}\label{alg:gd}
  \SetKw{until}{until}
  \DontPrintSemicolon
  \KwIn{labeled data set $(X,Y)$, regularization parameter $C$,
    learning rate $\eta(\cdot)$}
  initialize $\mathbf{W}^{(0)}=\mathbf{0}$\;
  \For{$t=0,\ldots$, \text{while not converged}}{
    sample $i\,\sim\,[N]$\;
    $\widehat{h}_i\,=\,\argmax_h\left[\dscore{\mathbf{W}^{(t)}}(\mathbf{x}_i,h)+\Delta(y_i,h)\right]$\;
    $h^\ast_i\,=\,\argmax_{h:\Delta(y_i,h)=0}\dscore{\mathbf{W}^{(t)}}(\mathbf{x}_i,h)$\;
    $\displaystyle\delta\mathbf{W}\,=\,\left[
      \frac{\partial\dscore{\mathbf{W}}(\mathbf{x}_i,\widehat{h}_i)}
      {\partial\mathbf{W}}\,-\,
      \frac{\partial\dscore{\mathbf{W}}(\mathbf{x}_i,h^\ast_i)}
      {\partial\mathbf{W}}
    \right]\Bigg|_{\mathbf{W}^{(t)}}$\;
    $\mathbf{W}^{(t+1)}\,=\,(1-\eta(t))\mathbf{W}^{(t)}-C\delta\mathbf{W}$\;
    project $\mathbf{W}^{(t+1)}$ to PSD cone\;
  }
\end{algorithm}

\vspace{2mm}
\noindent 
The SGD algorithm requires solving two inference problems
($\widehat{h}$ and $h^\ast$), and computing the gradient of $\dscore{\mathbf{W}}$ which we
address below.\footnote{We note that both inference problems over $h$ are done
in leave one out settings, i.e., we impose an additional constraint
$i\notin h$ under the $\argmax$, not listed in the algorithm explicitly.}

\begin{algorithm}[!th]
\DontPrintSemicolon
  \KwIn{$\mathbf{x}$, $\mathbf{W}$, target class $y$,
    $\tau\triangleq\llbracket\text{ties forbidden}\rrbracket$}
  \KwOut{$\argmax_{h:\widehat{y}(h)=y}\dscore{\mathbf{W}}(\mathbf{x})$}
  \SetKw{Define}{define}
  \SetKwData{Score}{Score}
  Let
  $n^\ast=\lceil\frac{k+\tau(R-1)}{R}\rceil$\tcp*[r]{min. required number
    of neighbors from $y$}
  $h\,:=\,\varnothing$\;
  %\Define{$\#(r)\,\triangleq\,\sum_ih_i\llbracket
  %  y_i=r\rrbracket$}\tcp*[r]{count of selected neighbors from class $r$}
  \For{$j = 1, \ldots, n^\ast$}{
    $\displaystyle h:=h\,\cap\,\argmin_{\mathbf{x}_i:\,y_i=y,i\notin h} \mahax{\mathbf{W}}{\mathbf{x}}{\mathbf{x}_i}$\;
  }
  \For{$l = n^\ast + 1, \ldots, k$}{
    \Define{$\#(r)\,\triangleq\,|\{i:\,\mathbf{x}_i\in h,
      y_i=r\}|$}\tcp*[r]{\small count selected neighbors from class $r$}
      $\displaystyle h:=h\,\cap\,\argmin_{\mathbf{x}_i:\,y_i=y, \text{ or } \#(y_i)
        < \#(y)-\tau, \,i\notin h}\mahax{\mathbf{W}}{\mathbf{x}}{\mathbf{x}_i}$\;
  }
  \Return{$h$}
\caption{Targeted inference}\label{alg:hstar}
\end{algorithm}

\subsubsection{Targeted inference of $h^\ast_i$} 
Here we are concerned with finding the highest-scoring
$h$ constrained to be compatible with a given target class $y$. We give an
$O(N\log N)$ algorithm in Algorithm~\ref{alg:hstar}. Proof of its correctness and complexity analysis is in section~\ref{app:proof}.

\vspace{2mm}
\noindent
The intuition behind Algorithm~\ref{alg:hstar} is as follows. For a
given combination of $R$ (number of classes) and $k$ (number of
neighbors), the minimum number of neighbors from the target class $y$
required to allow (although not guarantee) zero loss,
is $n^\ast$ (see Proposition~\ref{prop:1} in section~\ref{app:proof}). The algorithm first includes $n^\ast$ highest scoring neighbors
from the target class. The remaining $k - n^\ast$ neighbors are picked
by a greedy procedure that selects the highest scoring neighbors (which might or might not be
from the target class) while making sure that no non-target class ends
up in a majority. 

\vspace{2mm}
\noindent
When using Alg.~\ref{alg:hstar} to find an element in
$H^\ast$, we forbid ties, i.e. set $\tau=1$.

\subsubsection{Loss augmented inference
  $\widehat{h}_i$} Calculating the $\max$ term in~\eqref{eq:loss1}
is known as loss augmented inference.
We note that
\begin{equation}
  \label{eq:hloss}
  \max_{h'}\left<\mathbf{W},\mathbf{\Psi(\mathbf{x},h')}\right>+\Delta(y,h')
\,=\,\qquad\max_{y'}\Bigl\{\underbracket{\max_{h'\in H^\ast(y')}
\left<\mathbf{W},\mathbf{\Psi(\mathbf{x},h')}\right>}_{=\,\left<\mathbf{W},\mathbf{\Psi\left(\mathbf{x},h^\ast(\mathbf{x},y')\right)}\right>}
\,+\,
\Lambda(y,y')
\Bigr\}
\end{equation}
which immediately leads to Algorithm~\ref{alg:haugmented}, relying on
Algorithm~\ref{alg:hstar}. The intuition: perform targeted inference
for each class (as if that were the target class), and the choose the
set of neighbors for the class for which the loss-augmented score is
the highest. In this case, in each call to
Alg.~\ref{alg:hstar} we set $\tau=0$, i.e., we allow ties, to make sure
the $\argmax$ is over all possible $h$'s.

\begin{algorithm}[!th]
  \DontPrintSemicolon
  \KwIn{$\mathbf{x}$, $\mathbf{W}$,target class $y$}
  \SetKwData{Value}{Value}  
  \KwOut{$\argmax_h\left[\dscore{\mathbf{W}}(\mathbf{x},h)+\Delta(y,h)\right]$}
  \For{$r\in \{1,\ldots,R\}$}{
    $h^{(r)}\,:=\,h^\ast(\mathbf{x},\mathbf{W},r,1)$
    \tcp*{using Alg.~\ref{alg:hstar}}
    Let \Value$(r)\,:=\,\dscore{\mathbf{W}}(\mathbf{x},h^{(r)}),+\,\Lambda(y,r)$\;
  }
  Let $r^\ast=\argmax_r$\Value$(r)$\;
  \Return{$h^{(r^\ast)}$}
  \caption{Loss augmented inference}\label{alg:haugmented}
\end{algorithm}

\subsubsection{Some more intuition behind inference procedures}\label{sec:intuition}
\begin{figure*}
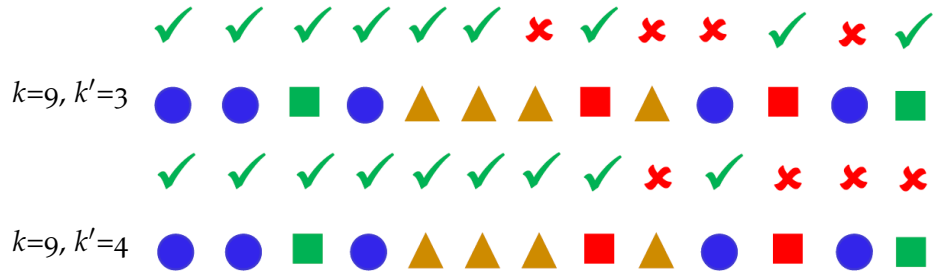

	\centering
	\begin{center}
		\raisebox{1em}{$k$=9, $k'$=3}~~\includepic{.7}{./Chapters/Chapter03/inf_k9-3}
		~~~~
		\raisebox{1em}{$k$=9, $k'$=4}~~\includepic{.7}{./Chapters/Chapter03/inf_k9-4}
	\end{center}
	\caption{Inference as packing k neighbors with smallest distance while ensuring correct vote. For more details see sections \ref{sec:ssd} and \ref{sec:intuition}}
	\label{fig:inferenceintuition} 
\end{figure*}

While in the previous section we have provided inference procedures and section \ref{app:proof} contains the analysis and proof of correctness, in this section we briefly provide more intuition. We can view of inference in this model as wanting to pack $k$ neighbors with the smallest distance, while ensuring the correct vote. For instance, suppose we want $k \leq k'$ neighbors from a target class. In that case, we start with $k'$ nearest neighbors from the target class, and then proceed in the order of decreasing distance. While doing so, we also pick neighbors from the wrong class, but such that we also ensure that do not let any class have $\geq k'$ points. This is then done for all feasible values of $k'$, and from this we select the best set. This is illustrated further in figure \ref{fig:inferenceintuition} 

\subsubsection{Analysis and Proof of correctness of Algorithm~\ref{alg:hstar}}\label{app:proof}

First of all it is easy to see that Algorithm \ref{alg:hstar}
terminates. There are $k - n^\ast$ iterations after initialization (of the first $n^\ast$
points) and this amounts to
at most a linear scan of $X$. We need $O(N\log N)$ time to
sort the data and then finding $\mathbf{h}^\ast$ involves $O(N)$,
thus the algorithm runs in time $O(N \log N)$.

\vspace{2mm}
\noindent
We need to prove that the algorithm returns $h^\ast$ as defined
earlier. First, we establish the correctness of setting $n^\ast$:

\begin{proposition} \label{prop:1}
	Let $R$ be the number of classes, and let $\#(h,y)$ be
	the count of neighbors from target
	class $y$ included in the assignment $h$.
	Then, $\Delta(y^\ast,h)=0$ only if
	$\#(h,y^\ast)\geq n^\ast$, where
	\[n^\ast\,=\,\begin{cases}
	\left\lceil\frac{k+R-1}{R}\right\rceil & \text{if ties not allowed},\\
	\left\lceil\frac{k}{R}\right\rceil & \text{if ties allowed}.\\
	\end{cases}
	\]
	We prove it below for the case with no ties; the proof when ties are
	allowed is very similar.
	\begin{proof}
		Suppose by contradiction that
		$\Delta(y^\ast,h)=0$ and
		$\#(h,y^\ast)\le\lceil\frac{k+R-1}{R}\rceil-1$. Then, since
		no ties are allowed, for all
		$y\ne y^\ast$, we have
		$\#(h,y^\ast)\le\lceil\frac{k+R-1}{R}\rceil-2$, and
		\begin{align}
		\sum_{y}\#(h,y)\,&\le\,(R-1)\left(\left\lceil\frac{k+R-1}{R}\right\rceil-2\right)\\
		&\qquad\qquad+\,
		\left\lceil\frac{k+R-1}{R}\right\rceil-1\\
		&<\,k,
		\end{align}
		a contradiction to $|h|=k$. 
	\end{proof}
\end{proposition}

\vspace{2mm}
\noindent
Next, we prove that the algorithm terminates and produces a correct
result. For the purposes of complexity analysis,
we consider $R$ (but not $k$) to be
constant, and number of examples from each class to be $O(N)$.

\begin{claim}
	Algorithm~\ref{alg:hstar} terminates after at most $O((N+k)\log N)$
	operations and produces an $h$ such that $|h|=k$.
	\begin{proof}
		The elements of $X$ can be held in $R$ priority queues, keyed by
		$\maha{\mathbf{W}}$ values, one queue per class. Construction of this data
		structure is an $O(N\log N)$ operation, carried out before the
		algorithm starts. To initialize $h$ with $n^\ast$ values, the 
		algorithm retrieves $n^\ast$ top elements from the priority 
		queue for class $y^\ast$. An $O(n^\ast\log N)$ operation. Then, for 
		each of the iterations over $l$, the algorithm needs to examine
		at most one top element from $R$ queues, which costs $O(\log N)$; 
		each such iteration increases $|h|$ by one. Thus after $k-n^\ast$
		iterations $|h| = k$; the total cost is thus  $O(k\log N)$. Combined with
		the complexity of data structure construction mentioned above, this concludes the proof.
		
	\end{proof}
\end{claim}

\vspace{2mm}
\noindent
Note that for typical scenarios in which $N\gg k$, the cost will be
dominated by the $N\log N$ data structure setup.

\begin{claim}\label{claim:1}
	Let $h^\ast$ be returned by Algorithm~\ref{alg:hstar}. Then,
	\begin{equation}\label{eq:hclaim}
	h^\ast\,=\,\argmax_{h:|h|=k,\,\Delta(y^\ast,h)=0)}\dscore{\mathbf{W}}(\mathbf{x},h),
	\end{equation}
	i.e., the algorithm finds the highest scoring $h$ with total
	of $k$ neighbors among those $h$ that attain zero loss.
	
	\begin{proof}
		From Proposition~\ref{prop:1} we know that if $\#(h,y)<n^\ast$, 
		then $h$ does not satisfy the $\Delta(\mathbf{x},h)=0$ condition.
		$|h|\ge n^\ast$ to~\eqref{eq:hclaim} without altering the definition.
		
		We will call $h$ ``optimal for $l$'' if
		\[h\,=\,\argmax_{h:|h|=n^\ast+l,\,\#(h,y) \ge n^\ast,\,\Delta(y^\ast,h))=0}\dscore{\mathbf{W}}(\mathbf{x},h).
		\]
		We now prove
		by induction over $l$ that this property is maintained through the
		loop over $l$ in the algorithm.
		
		\vspace{2mm}
		\noindent
		Let $h^{(j)}$ denote choice of 
		$h$ after $j$ iterations of the loop, i.e., $|h|=n^\ast
		+ j$. Suppose that $h^{(l-1)}$ is optimal for $l-1$.
		Now the algorithm selects $\mathbf{x}_a \in X$, such that
		
		\begin{equation}\label{eq:amax}
		\displaystyle \mathbf{x}_a=\argmin_{\mathbf{x}_i:\,y_i=y, \text{ or } \#(y_i) < \#(y)-\tau,
			\,\mathbf{x}_i\notin h}\mahax{\mathbf{W}}{\mathbf{x}}{\mathbf{x}_i}.\end{equation}
		Suppose that $h^{(l)}$
		is not optimal for $l$. Then there exists an $\mathbf{x}_b \in X$ for which
		$\mahax{\mathbf{W}}{\mathbf{x}}{\mathbf{x}_b} < \mahax{\mathbf{W}}{\mathbf{x}}{\mathbf{x}_a}$ such that
		picking $\mathbf{x}_b$ instead of $\mathbf{x}_a$ would produce
		$h$ optimal for $l$. 
		But $\mathbf{x}_b$ is not picked by the algorithm; this can only 
		happen if conditions on the $\argmin$ in~\eqref{eq:amax} are
		violated, namely, if $\#(y_b) = \#(y)-\tau$; therefore picking $\mathbf{x}_b$ 
		would violate conditions of optimality of $h^{(l)}$, and we
		get a contradiction.
		
		\vspace{2mm}
		\noindent
		It is also clear that after initialization with $k$ highest scoring
		neighbors in $y^\ast$, $h$ is optimal for $l=0$, which forms
		the base of induction. We conclude that $h^{(k-n^\ast)}$,
		i.e. the result of the algorithm, is optimal for $k-n^\ast$, which is
		equivalent to definition in~\eqref{eq:hclaim}.
		
	\end{proof}
\end{claim}

\subsubsection{Gradient update}
Finally, we need to compute the gradient of the distance score. Since
it is linear in $\mathbf{W}$ as shown in~\eqref{eq:F}, we have
\begin{equation}
  \label{eq:grad}
  \frac{\partial\dscore{\mathbf{W}}(\mathbf{x},h)}{\partial\mathbf{W}}\,=\,
  \mathbf{\Psi}(\mathbf{x},h)\,=\,
-\sum_{\mathbf{x}_j\in
  h}(\mathbf{x}-\mathbf{x}_j)(\mathbf{x}-\mathbf{x}_j)^T.
\end{equation}

\begin{figure}
  \centering
 \includegraphics[width=0.35\linewidth]{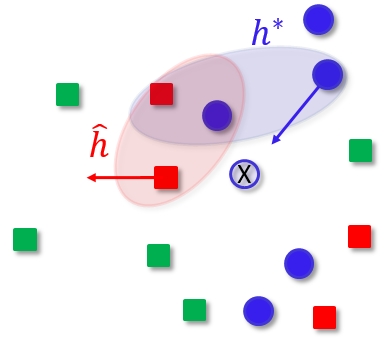}  
 \caption{Illustration of interpretation of the gradient update, more details in the text}\label{fig:gradientupdate}
\end{figure}

\vspace{2mm}
\noindent
Thus, the update in Alg~\ref{alg:gd} has a simple interpretation,
illustrated in Fig~\ref{fig:LMNNComparison}. For every
$\mathbf{x}_i\in h^\ast\setminus \widehat{h}$, it ``pulls''
$\mathbf{x}_i$ closer to $\mathbf{x}$. For every $\mathbf{x}_i\in
\widehat{h}\setminus h^\ast$, it ``pushes'' it farther from
$\mathbf{x}$; these push and pull refer to increase/decrease of
Mahalanobis distance under the updated $\mathbf{W}$. Any other
$\mathbf{x}_i$, including any $\mathbf{x}_i\in
h^\ast \cap \widehat{h}$, has no influence on the update. This is a
difference of our approach from LMNN, MLR etc. This is illustrated in 
Figure \ref{fig:LMNNComparison}. In particular $h^\ast$ corresponds to points 
$a,\,c\,\text{and}\,e$, whereas $\widehat{h}$ corresponds to points $c,\,e\,\text{and}\,f$. Thus point $a$
is pulled while point $f$ is pushed.

\vspace{2mm}
\noindent
Since the update does not necessarily preserve $\mathbf{W}$ as a PSD matrix, we enforce it by projecting
$\mathbf{W}$ onto the PSD cone, by zeroing negative eigenvalues. Note
that since
we update (or ``downdate'') $\mathbf{W}$ each time by  matrix of rank
at most $2k$, the eigendecomposition can be accomplished more
efficiently than the na\"ive $O(d^3)$ approach, e.g., as in~\cite{stange2008efficient}.

\vspace{2mm}
\noindent
Using first order methods, and in particular gradient  methods for
optimization of non-convex functions, has been common across machine
learning, for instance in training deep neural networks. Despite lack
(to our knowledge) of satisfactory guarantees of convergence, these
methods are often successful in practice; we will show in the next
section that this is true here as well. 
However, care should be taken to ensure validity of the method, and we
discuss this briefly before reporting on experiments.

\vspace{2mm}
\noindent
A given $\mathbf{x}$ imposes a Voronoi-type partition of the 
space of $\mathbf{W}$ into a finite number of cells; each cell is associated with a
particular combination of $\widehat{h}(\mathbf{x})$ and
$h^\ast(\mathbf{x})$ under the values of $\mathbf{W}$ in that cell.
The score $\dscore{\mathbf{W}}$ is differentiable (actually
linear) on the interior of the cell, but may be non-differentiable
(though continuous) on the boundaries. Since the boundaries between a
finite number of cells form a set of measure zero, we see that the
score is differentiable almost everywhere.

 \label{learning}

\section{Experiments}\label{sec:exp}
\vspace{-.7em}
We compare the error of $k$NN classifiers using metrics learned with our
approach to that with other learned metrics. For this evaluation we replicate
the protocol in~\cite{kedem2012non}, using the
seven data sets in Table~\ref{tab:errors_zscore}. For all data sets, we
report error of $k$NN classifier for a range of values of $k$; for each
$k$, we test the metric learned for that $k$. Competition to our
method includes Euclidean Distance, LMNN~\cite{weinberger2009distance}, NCA, ~\cite{goldberger2004neighbourhood}, ITML ~\cite{itml2007davis}, MLR~\cite{mcfee10_mlr} and
GB-LMNN~\cite{kedem2012non}. The latter learns non-linear
metrics rather than Mahalanobis. 

\vspace{2mm}
\noindent
For each of the competing methods, we used the code
provided by the  authors. In each case we
tuned the parameters of each method, including ours, in the same cross-validation
protocol. We omit a few other methods that were consistently shown in literature to be dominated by the ones we compare to, such as $\chi^2$ distance, MLCC, M-LMNN. We also could not include $\chi^2$-LMNN since code for it is not available; however published results for $k=3$~\cite{kedem2012non} indicate that our method would win against $\chi^2$-LMNN as well. % Citation for Jun Wang is JunWangMultiMetric

\vspace{2mm}
\noindent
Isolet and USPS have a standard training/test partition, for the other five data sets, we report the mean and standard errors of 5-fold cross validation (results for all methods are on the same folds). We experimented with different methods for initializing our method (given the non-convex objective), including the euclidean distance, all zeros etc. and found the euclidean initialization to be always worse. We initialize each fold with either the diagonal matrix learned by ReliefF~\cite{relieff1992} (which gives a scaled euclidean distance) or all zeros depending on whether the scaled euclidean distance obtained using ReliefF was better than unscaled euclidean distance. In each experiment, $\mathbf{x}$ are scaled by mean and standard deviation of the training portion.\footnote{For Isolet we also reduce dimensionality to 172 by PCA computed on the training portion.}
The value of $C$ is tuned on on a 75\%/25\% split of the training portion. Results using different scaling methods are also reported. 

\newcommand{\smpm}[1]{\hspace{.3em}\scriptsize{$\pm${#1}}}
\renewcommand{\arraystretch}{1.1} 

\begin{table}[!th]
	\small
	\centering
	\begin{tabu}{l}
		k = 3
	\end{tabu}
	\begin{tabu}{|l|l|r|r|r|r|r|r|r|}
		\hline
		Dataset & Isolet & USPS & letters & DSLR & Amazon & Webcam &
		Caltech\\
		\hline 
		\hline
		\rowfont{\small}
		$d$ & 170 & 256 & 16 & 800 & 800 & 800 & 800\\
		\rowfont{\small}
		$N$ & 7797 & 9298 & 20000 & 157 & 958 & 295 & 1123\\
		\rowfont{\small}
		$C$ & 26 & 10 & 26 & 10 & 10 & 10 & 10\\
		\hline
		\small{Euclidean} & 8.66 & 6.18  &  4.79 \smpm 0.2 & 75.20 \smpm 3.0  & 60.13\smpm 1.9 & 56.27 \smpm 2.5 & 80.5 \smpm 4.6 \\
		\hline
		\small{LMNN} & 4.43 & 5.48  & 3.26 \smpm 0.1  & 24.17 \smpm 4.5 & 26.72 \smpm 2.1 & 15.59 \smpm 2.2 & 46.93 \smpm 3.9 \\
		\hline
		\small{GB-LMNN} & \bf{4.13} & 5.48 & 2.92 \smpm 0.1 & 21.65 \smpm 4.8 & 26.72 \smpm 2.1 & 13.56 \smpm 1.9 & 46.11 \smpm 3.9 \\
		\hline
		\small{MLR} & 6.61 & 8.27 & 14.25 \smpm 5.8 & 36.93 \smpm 2.6 & 24.01 \smpm 1.8 & 23.05 \smpm 2.8 &  46.76 \smpm 3.4 \\
		\hline 
		\small{ITML} & 7.89  & 5.78  & 4.97 \smpm 0.2  & 19.07 \smpm 4.9  & 33.83 \smpm 3.3  & 13.22 \smpm 4.6 & 48.78 \smpm 4.5 \\
		\hline
		\small{1-NCA} &  6.16 & 5.23 & 4.71 \smpm 2.2 & 31.90 \smpm 4.9 & 30.27 \smpm 1.3 & 16.27 \smpm 1.5  & 46.66 \smpm 1.8\\    
		\hline
		\small{k-NCA} &  4.45 & \bf{5.18} & 3.13 \smpm 0.4  & 21.13 \smpm 4.3 & 24.31 \smpm 2.3  &  13.19 \smpm 1.3  & 44.56 \smpm 1.7 \\    
		\hline
		\small{ours} & 4.87 & \bf{5.18} & \bf{2.32 \smpm 0.1} & \bf{17.18\smpm4.7} & \bf{21.34\smpm 2.5} & \bf{10.85\smpm 3.1} & \bf{43.37\smpm 2.4} \\ 
		\hline
	\end{tabu}
	\begin{tabu}{l}
		k = 7 
	\end{tabu}
	\begin{tabu}{|l|l|r|r|r|r|r|r|r|}
		\hline
		Dataset & Isolet & USPS & letters & DSLR & Amazon & Webcam &
		Caltech\\
		\hline 
		\hline
		\small{Euclidean} & 7.44 & 6.08 &  5.40 \smpm 0.3 & 76.45 \smpm 6.2 &  62.21 \smpm 2.2 & 57.29 \smpm 6.3 & 80.76 \smpm 3.7 \\
		\hline
		\small{LMNN} &  3.78 & \bf{4.9}  & 3.58 \smpm 0.2  & 25.44 \smpm 4.3 & 29.23 \smpm 2.0 & 14.58 \smpm 2.2 & 46.75 \smpm 2.9 \\
		\hline
		\small{GB-LMNN} & \bf{3.54} & \bf{4.9} & 2.66 \smpm 0.1 & 25.44 \smpm 4.3 & 29.12 \smpm 2.1 & 12.45 \smpm 4.6 & 46.17 \smpm 2.8 \\
		\hline
		\small{MLR} & 5.64 & 8.27 & 19.92 \smpm 6.4 & 33.73 \smpm 5.5 & 23.17 \smpm 2.1 & 18.98 \smpm 2.9 &  46.85 \smpm 4.1 \\
		\hline 
		\small{ITML} & 7.57 & 5.68 & 5.37 \smpm 0.5 &  22.32 \smpm 2.5 & 31.42 \smpm 1.9 & \bf{10.85 \smpm 3.1} & 51.74 \smpm 2.8\\
		\hline
		\small{1-NCA} & 6.09 & 5.83 & 5.28 \smpm 2.5 & 36.94 \smpm 2.6 & 29.22 \smpm 2.7 & 22.03 \smpm 6.5 & 45.50 \smpm 3.0 \\    
		\hline
		\small{k-NCA} &  4.13 & 5.1 & 3.15 \smpm 0.2 & 22.78 \smpm 3.1 & 23.11 \smpm 1.9 & 13.04 \smpm 2.7  & 43.92 \smpm 3.1 \\    
		\hline
		\small{ours} & 4.61 & \bf{4.9} & \bf{2.54\smpm 0.1} & \bf{21.61 \smpm 5.9} & \bf{22.44 \smpm 1.3} & 11.19\smpm 3.3 & \bf{41.61 \smpm 2.6} \\ 
		\hline
	\end{tabu}
	
	\begin{tabu}{l}
		k = 11
	\end{tabu}
	\begin{tabu}{|l|l|r|r|r|r|r|r|r|}
		\hline
		Dataset & Isolet & USPS & letters & DSLR & Amazon & Webcam &
		Caltech\\
		\hline 
		\hline
		\small{Euclidean} & 8.02 & 6.88 & 5.89 \smpm 0.4 & 73.87 \smpm 2.8  & 64.61 \smpm 4.2 & 59.66 \smpm 5.5 & 81.39 \smpm 4.2\\
		\hline
		\small{LMNN} & \bf{3.72} & \bf{4.78} & 4.09 \smpm 0.1 & 23.64 \smpm 3.4 & 30.12 \smpm 2.9 & 13.90 \smpm 2.2 & 49.06 \smpm 2.3 \\
		\hline
		\small{GB-LMNN} & 3.98 & \bf{4.78} & \bf{2.86} \smpm 0.2 & 23.64 \smpm 3.4 & 30.07 \smpm 3.0 & 13.90 \smpm 1.0 & 49.15 \smpm 2.8 \\
		\hline
		\small{MLR} & 5.71 & 11.11 & 15.54 \smpm 6.8 & 36.25 \smpm 13.1 & 24.32 \smpm 3.8 & 17.97 \smpm 4.1 &  44.97 \smpm 2.6 \\
		\hline 
		\small{ITML} & 7.77  & 6.63 & 6.52 \smpm 0.8  & \bf{22.28 \smpm 3.1} & 30.48 \smpm 1.4 & 11.86 \smpm 5.6 & 50.76 \smpm 1.9 \\
		\hline
		\small{1-NCA} &  5.90 & 5.73 & 6.04 \smpm 2.8 & 40.06 \smpm 6.0 & 30.69 \smpm 2.9 & 26.44 \smpm 6.3 & 46.48 \smpm 4.0 \\    
		\hline
		\small{k-NCA} &  4.17 & 4.81 & 3.87 \smpm 0.6 & 23.65 \smpm 4.1 & 25.67 \smpm 2.1 & 11.42 \smpm 4.0  & 43.8 \smpm 3.1 \\    
		\hline
		\small{ours} & 4.11 & 4.98 & 3.05\smpm 0.1 & \bf{22.28 \smpm 4.9} & \bf{24.11\smpm 3.2} & \bf{11.19 \smpm 4.4} & \bf{40.76 \smpm 1.8}  \\ 
		\hline
	\end{tabu}   
	\caption{$k$NN errors for  $k$=3, 7 and 11. Features were scaled by z-scoring.}
	\label{tab:errors_zscore}
\end{table}

\vspace{2mm}
\noindent
Our SGD algorithm stops when the running average of the surrogate loss
over most recent epoch no longer decreases substantially, or after
max. number of iterations. We use learning rate $\eta(t)=1/t$.

\begin{figure}
	\centering
	\includegraphics[width=.325\linewidth]{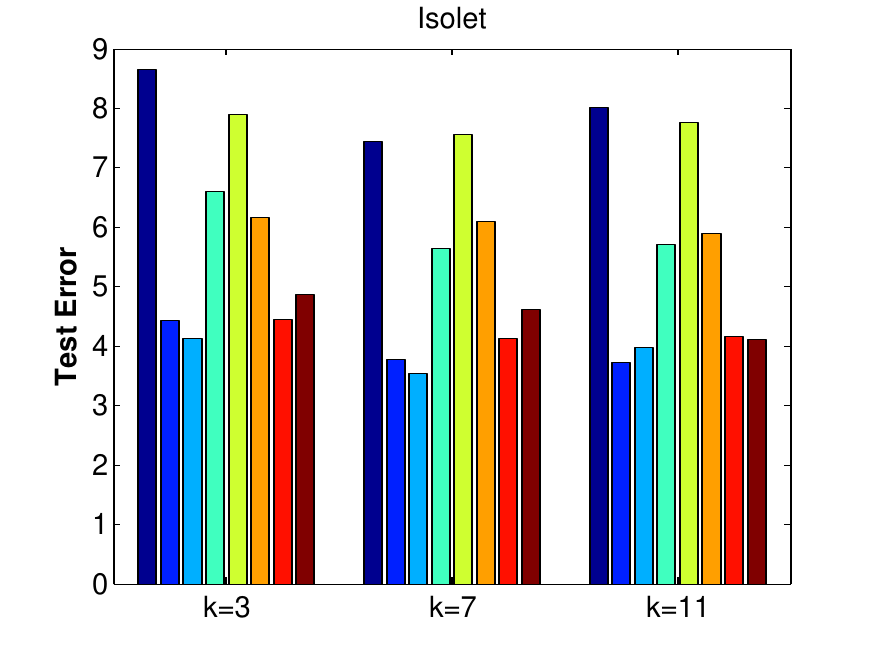} \includegraphics[width=.325\linewidth]{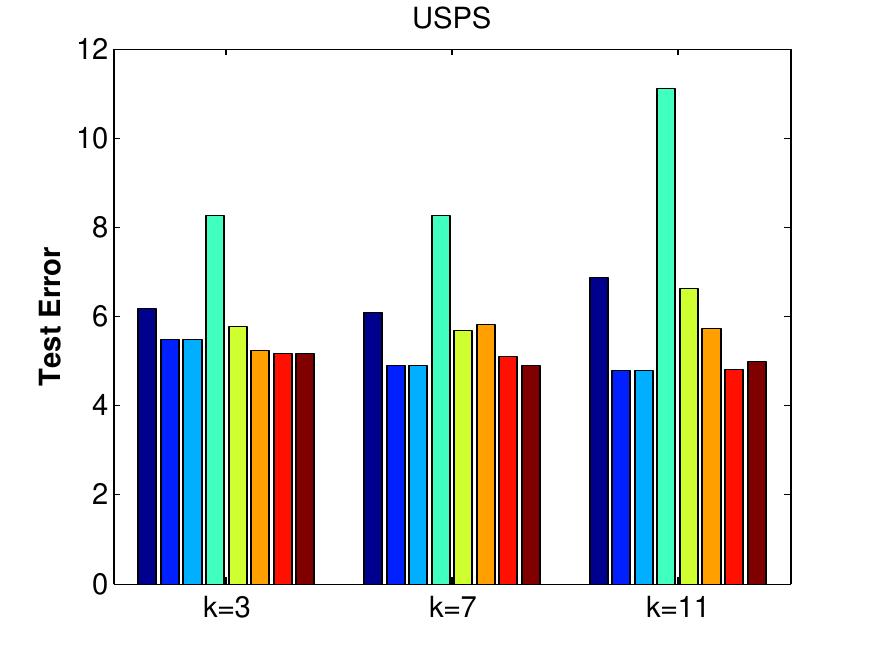} \includegraphics[width=.325\linewidth]{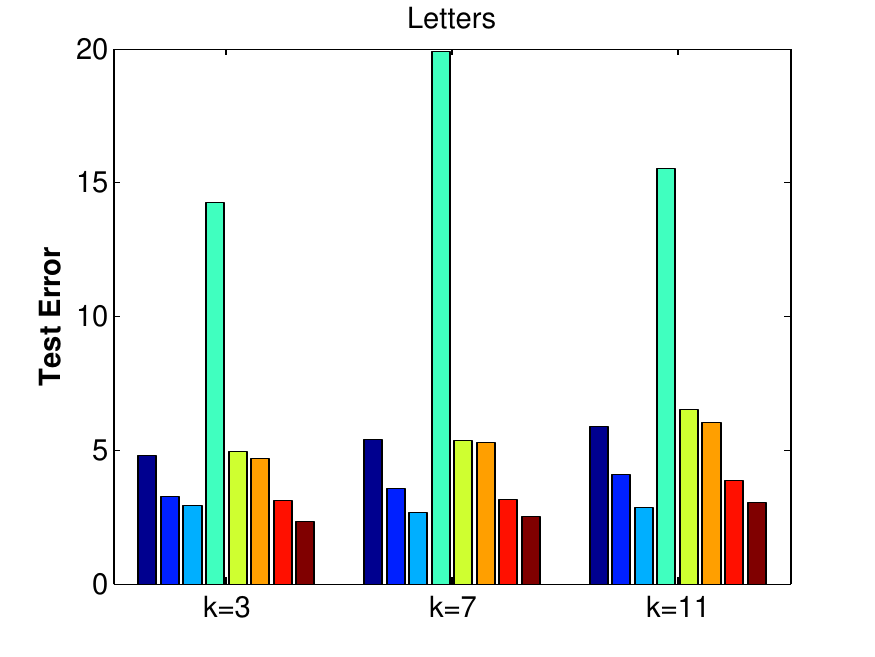}  %
	
	\includegraphics[width=.325\linewidth]{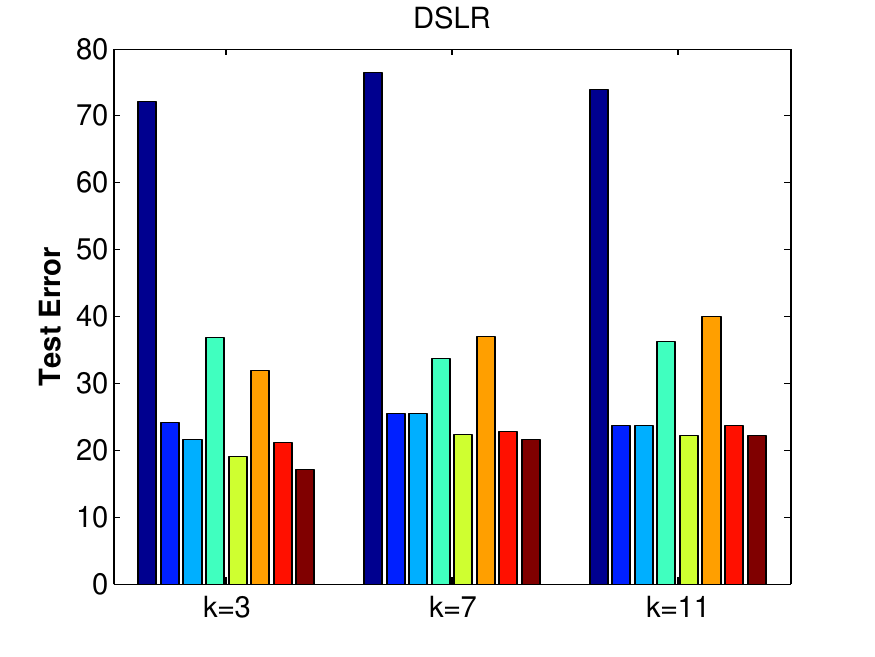} \includegraphics[width=.325\linewidth]{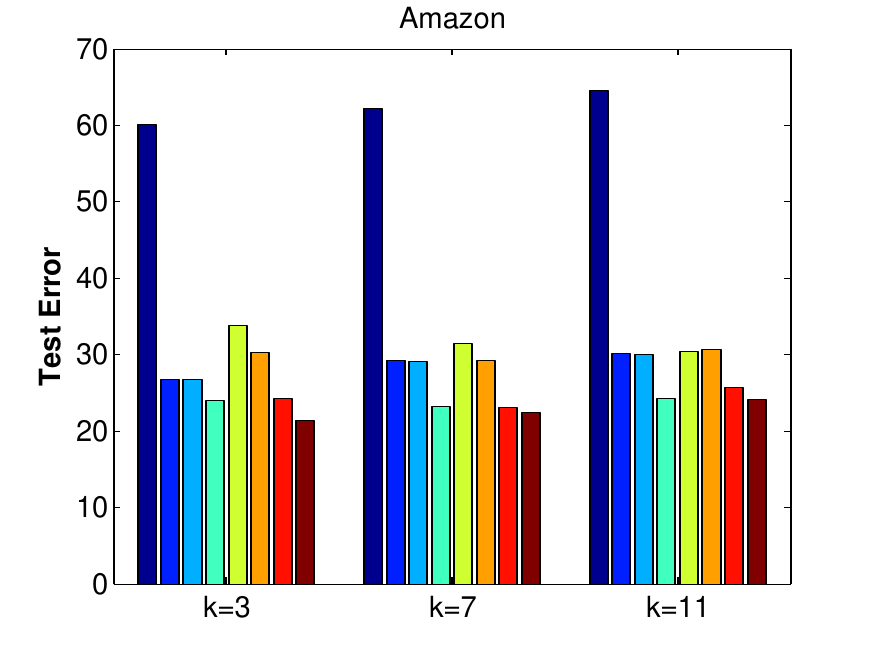} \includegraphics[width=.325\linewidth]{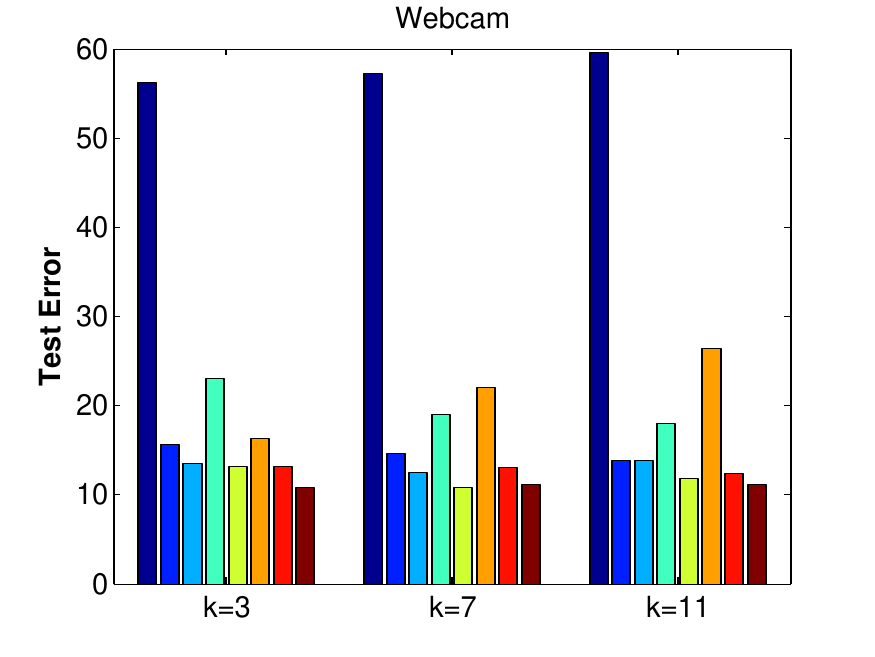}  
	
	\includegraphics[width=.325\linewidth]{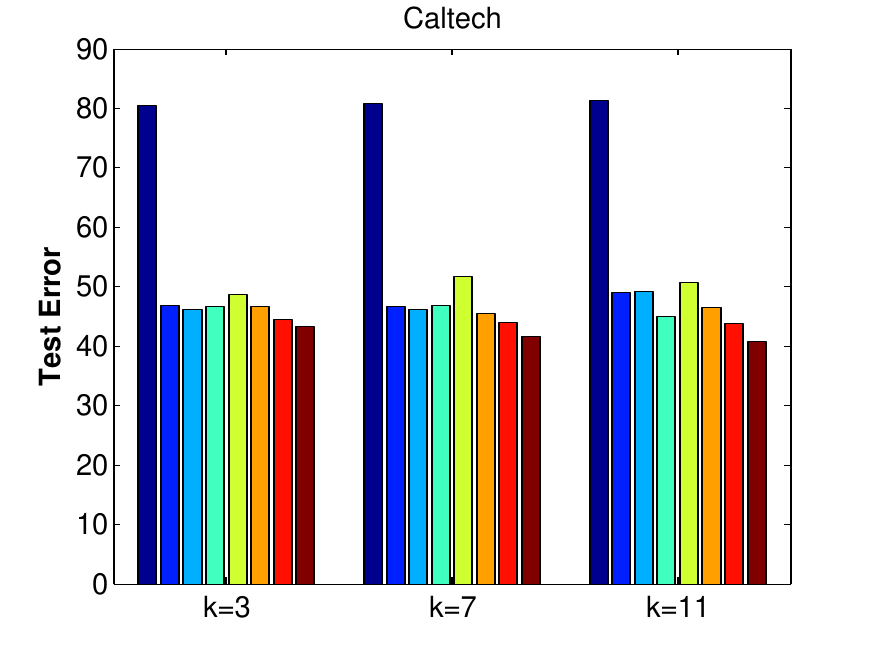} \hspace{2.5cm}\includegraphics[width=.2\linewidth]{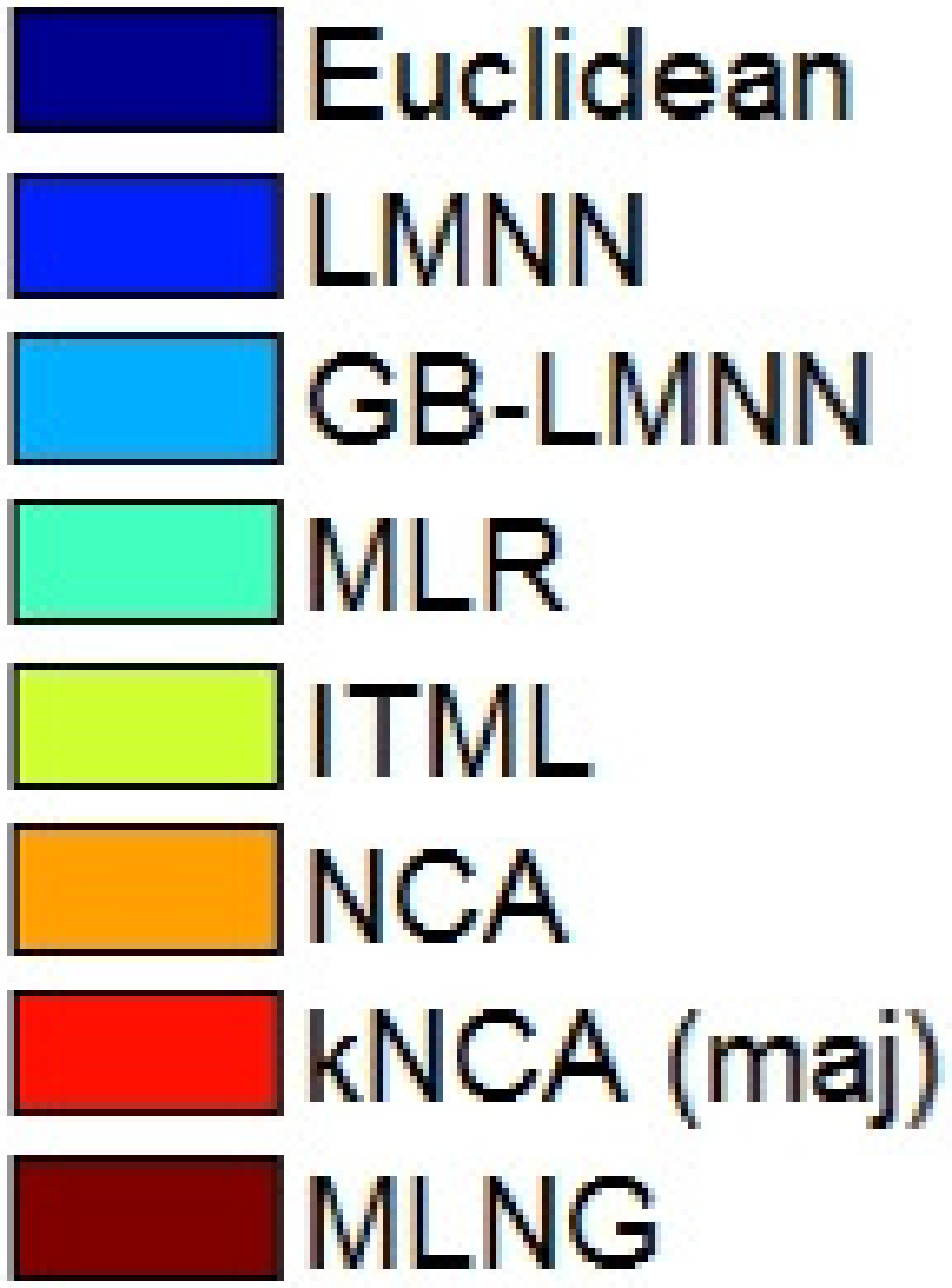}		
	%\includegraphics[width=.33\linewidth]{../Figures/WineepsKNN} 
	%  \caption{Error profile for regression as a function of sample size}\label{viina8}
		\caption{$k$NN errors for  $k$=3, 7 and 11 on various datasets when the features are scaled by z-scoring}
	\label{fig:zscoreerrors}
\end{figure}

\vspace{2mm}
\noindent
The results show that our
method dominates other competitors, including non-linear metric learning
methods, and in some cases achieves results
significantly better than those of the competition. The results for the initialization and set-up mention above are illustrated in table \ref{tab:errors_zscore}, as well as figure \ref{fig:zscoreerrors}

% Manual newpage inserted to improve layout of sample file - not
% needed in general before appendices/bibliography.

\subsection{Runtimes using different methods}

Here we include the training times in seconds for one fold of each dataset. These timings are for a single partition, for optimal parameters for $k=7$. These experiments were run on a 12-core Intel Xeon E5-2630 v2 @ 2.60GHz and are reported in \ref{tb:runtimes}. We notice that the approach, while competitive is quite slow due to exact inference and loss augmented inference, each of which are expensive steps. The run time increases linearly both in the number of classes as well as $k$.

\begin{table}
\begin{tabular}{l*{6}{c}r}
	Dataset             & DSLR & Caltech & Amazon & Webcam & Letters  & USPS & Isolet \\
	\hline
	LMNN & 358.11 & 1812.1 & 1545.1 & 518.7 & 179.77 & 782.66 & 1762.1  \\
	GB-LMNN & 410.13 & 1976.4 & 1680.9 & 591.29 &  272.87 & 3672.9 &  2882.6  \\
	MLR & 4.93 & 124.42 & 88.96 & 85.02 &  838.13 & 1281 &  33.20  \\
	MLNG & 413.36 & 1027.6 & 2157.2 & 578.74 &  6657.3 & 3891.7 &  3668.9  \\
\end{tabular}
\caption{Comparison of runtimes}
\label{tb:runtimes}
\end{table}

\subsection{Experimental results using different feature normalizations}
For the sake of completeness, we also experimented with different feature normalization other than z-scoring to see if it impacted the results significantly. Somewhat surprisingly, we did notice a deterioration in performance when no feature normalization was done. We report the results for the case of no feature normalization, for the same competition in table \ref{tab:errors_native}. We also ran the same set of experiments for the case when we did histogram normalization. In this case obviously, we could only run experiments in the case of 4 datasets: DSLR, Amazon, Webcam and Caltech, which allowed for such normalization. Results for experiments with such a feature normalization are reported in table \ref{tab:errors_histogram}

\begin{table}[!th]
	\small
	\centering
	\begin{tabu}{l}
		k = 3
	\end{tabu}
	\begin{tabu}{|l|l|r|r|r|r|r|r|r|}
		\hline
		Dataset & Isolet & USPS & letters & DSLR & Amazon & Webcam &
		Caltech\\
		\hline 
		\hline
		\rowfont{\small}
		$d$ & 170 & 256 & 16 & 800 & 800 & 800 & 800\\
		\rowfont{\small}
		$N$ & 7797 & 9298 & 20000 & 157 & 958 & 295 & 1123\\
		\rowfont{\small}
		$C$ & 26 & 10 & 26 & 10 & 10 & 10 & 10\\
		\hline
		Euclidean & - &  - & - &  26.71 \smpm 11 & 37.26 \smpm 2.3 & 23.39 \smpm 5.3 & 58.42 \smpm 3.7\\
		\hline
		LMNN & - & - & - & 23.53 \smpm 7.6 & \bf{26.30 \smpm 1.6} & \bf{11.53 \smpm 6.7} & 43.72 \smpm 3.5 \\
		\hline
		GB-LMNN & - & - & - & 23.53 \smpm 7.6 & \bf{26.30 \smpm 1.6} & \bf{11.53 \smpm 6.7} & \bf{43.54 \smpm 3.5} \\
		\hline
		MLR & - & - & - & 24.78 \smpm 14.2 & 32.35 \smpm 4.5 & 14.58 \smpm 3.5  &  52.18 \smpm 2.0\\
		\hline 
		ITML & - & - & - & 22.22 \smpm 9.9 & 32.67 \smpm 3.2 & 12.88 \smpm 6.1 & 51.74 \smpm 4.2 \\
		\hline
		NCA & - & - & - & 29.84 \smpm 8.1 & 33.72 \smpm 2.1 & 21.36 \smpm 4.9 &  54.50 \smpm 2.0\\    
		\hline
		ours & - & - & - & \bf{21.63 \smpm 6.1} & 28.08 \smpm 2.4 & 14.58 \smpm 5.4 & 45.33 \smpm 2.8\\ 
		\hline
	\end{tabu}
	\begin{tabu}{l}
		k = 7 
	\end{tabu}
	\begin{tabu}{|l|l|r|r|r|r|r|r|r|}
		\hline
		Dataset & Isolet & USPS & letters & DSLR & Amazon & Webcam &
		Caltech\\
		\hline 
		\hline
		Euclidean & - & - & - & 32.46 \smpm 8.3 &  38.2 \smpm 1.6 & 27.46 \smpm 5.9 & 56.9 \smpm 2.9\\
		\hline
		LMNN & - & - & - & 26.11 \smpm 8.6 & 25.47 \smpm 1.6 & \bf{10.51 \smpm 4.9} & 41.77 \smpm 4.0 \\
		\hline
		GB-LMNN & - & - & - & \bf{25.48 \smpm 10.9} & \bf{25.36 \smpm 1.7} & \bf{10.51 \smpm 4.9} & \bf{41.59 \smpm 3.6} \\
		\hline
		MLR & - & - & - & 27.94 \smpm 9.0 & 30.16 \smpm 3.0 & 16.95 \smpm 3.4 &  49.51 \smpm 3.6 \\
		\hline 
		ITML & - & - & - & 22.28 \smpm 8.8 & 32.88 \smpm 3.3 & 13.90 \smpm 6.3 & 50.59 \smpm 4.7\\
		\hline
		NCA & - & - & - & 37.48 \smpm 8.2 & 33.09 \smpm 1.9 & 23.39 \smpm 5.3 & 51.74 \smpm 2.6 \\    
		\hline
		ours & - & - & - & \bf{25.65 \smpm 7.1} & 27.24 \smpm 2.7 & 17.29 \smpm 5.0 & 44.62\smpm 2.6 \\ 
		\hline
	\end{tabu}
	
	\begin{tabu}{l}
		k = 11
	\end{tabu}
	\begin{tabu}{|l|l|r|r|r|r|r|r|r|}
		\hline
		Dataset & Isolet & USPS & letters & DSLR & Amazon & Webcam &
		Caltech\\
		\hline 
		\hline
		Euclidean & - & - & - & 35.02 \smpm 8.9 & 37.57 \smpm 2.3 & 30.51 \smpm 4.8 & 56.55\smpm 2.4\\
		\hline
		LMNN & - & - & - & 49.64 \smpm 5.7 & \bf{24.84 \smpm 2.1} & \bf{10.17 \smpm 3.8} & 43.19 \smpm 2.7\\
		\hline
		GB-LMNN & - & - & - & 43.89 \smpm 5.6 & 25.16 \smpm 2.0 & \bf{10.17 \smpm 3.8} & \bf{43.10\smpm 3.1}  \\
		\hline
		MLR & - & - & - & 28.63 \smpm 7.7 & 30.48 \smpm 2.4 & 17.63 \smpm 5.3 & 48.18 \smpm 3.8 \\
		\hline 
		ITML & - & - & - &  \bf{24.82 \smpm 5.1} & 31.10 \smpm 2.6 & 15.25 \smpm 6.3 & 50.32 \smpm 3.9 \\
		\hline
		NCA & - & - & - & 41.37 \smpm 4.7 & 32.88 \smpm 1.5 & 24.07 \smpm 8.4 & 51.20 \smpm 3.9 \\    
		\hline
		ours & - & - & - & 31.79 \smpm 7.2 & 28.49 \smpm 2.8 & 17.65 \smpm 3.5 & 45.95 \smpm 4.8 \\ 
		\hline
	\end{tabu}   
	\caption{$k$NN error,for  $k$=3, 7 and 11. Mean and standard deviation are shown
		for data sets on which 5-fold partition was used. These experiments were done after histogram normalization. Best performing methods are shown in bold. Note that the only non-linear metric learning method in the above is GB-LMNN}
	\label{tab:errors_histogram}
\end{table}

\begin{table}[!th]
	\small
	\centering
	\begin{tabu}{l}
		k = 3
	\end{tabu}
	\begin{tabu}{|l|l|r|r|r|r|r|r|r|}
		\hline
		Dataset & Isolet & USPS & letters & DSLR & Amazon & Webcam &
		Caltech\\
		\hline 
		\hline
		\rowfont{\small}
		$d$ & 170 & 256 & 16 & 800 & 800 & 800 & 800\\
		\rowfont{\small}
		$N$ & 7797 & 9298 & 20000 & 157 & 958 & 295 & 1123\\
		\rowfont{\small}
		$C$ & 26 & 10 & 26 & 10 & 10 & 10 & 10\\
		\hline
		Euclidean & 8.98 & 5.03 & 4.31 \smpm 0.2 & 58.01 \smpm 5.0 & 56.89 \smpm 2.4 & 40.34 \smpm 4.2 & 74.89 \smpm 3.2 \\
		\hline
		LMNN & 4.17 & 5.38 & 3.26 \smpm 0.1 & 23.53 \smpm 5.6 & 28.08 \smpm 2.2 & \bf{11.19 \smpm 5.6} & 44.97 \smpm 2.6\\
		\hline
		GB-LMNN & \bf{3.72} & 5.03 & 2.50 \smpm 0.2 & 23.53 \smpm 5.6 & 28.08 \smpm 2.2 & 11.53 \smpm 5.5 & 44.70 \smpm 2.4 \\
		\hline
		MLR & 17.32 & 8.42 & 45.70 \smpm 18.7 & 35.69 \smpm 7.6 & \bf{23.40 \smpm 1.7} & 20 \smpm 4.6 & 47.11 \smpm 1.7 \\
		\hline 
		ITML &  6.86 & \bf{4.78} & 4.35 \smpm 0.2 & 24.82 \smpm 10.9 & 34.77 \smpm 4.7 & 12.20 \smpm 4.1 & 53.97 \smpm 3.2 \\
		\hline
		NCA &  5.07 & 5.18 & 4.39 \smpm 1.1  & 24.19 \smpm 5.8 & 29.54 \smpm 1.4 & 12.88 \smpm 4.9 & 46.84 \smpm 2.0\\    
		\hline
		ours & 4.11 & 5.13 & \bf{2.24 \smpm 0.1} & \bf{21.01 \smpm 4.1} & 26.20 \smpm 2.6 & 13.56 \smpm 4.6 & \bf{44.54 \smpm 2.9} \\ 
		\hline
	\end{tabu}
	\begin{tabu}{l}
		
		k = 7 
	\end{tabu}
	\begin{tabu}{|l|l|r|r|r|r|r|r|r|}
		\hline
		Dataset & Isolet & USPS & letters & DSLR & Amazon & Webcam &
		Caltech\\
		\hline 
		\hline
		Euclidean & 6.93 & 5.08 & 4.69 \smpm 0.2 & 60.46 \smpm 5.2 & 59.07 \smpm 4.5 &  43.05 \smpm 3.7 & 72.3 \smpm  3.3 \\
		\hline
		LMNN & 4.04 & 5.28 & 3.53 \smpm 0.2 & 24.15 \smpm 9.0 & 28.19 \smpm 2.8 & 13.56 \smpm 4.5 & 43.90 \smpm 2.4\\
		\hline
		GB-LMNN & \bf{3.72} & 5.03 & \bf{2.32 \smpm 0.2} & 24.80 \smpm 8.1 & 28.29 \smpm 3.1 & 13.14 \smpm 5.8 & 43.54 \smpm 2.2  \\
		\hline
		MLR & 23.28 & 8.12 & 33.61 \smpm 16.8 & 38.17 \smpm 10.9 & \bf{23.79 \smpm 3.9} & 20.34 \smpm 2.9 & 45.60 \smpm 4.8 \\
		\hline 
		ITML & 5.90 & 5.23 & 4.93 \smpm 0.5 & \bf{23.57 \smpm 9.6} & 32.46 \smpm 3.2 & \bf{11.19 \smpm 5.7} & 52.63 \smpm 3.3  \\
		\hline
		NCA &  5.52 & 4.98 & 5.06 \smpm 1.1  & 37.58 \smpm 5.7 & 31.01 \smpm 2.0 & 16.81 \smpm 5.9 & 43.90 \smpm 2.4\\    
		\hline
		ours & 4.07 & \bf{4.93} & \bf{2.49 \smpm 0.1} & 29.94 \smpm 7.6 & 26.10 \smpm 2.1 & 13.24 \smpm 3.1 & \bf{42.83 \smpm 3.1}\\ 
		\hline
	\end{tabu}
	
	\begin{tabu}{l}
		k = 11
	\end{tabu}
	\begin{tabu}{|l|l|r|r|r|r|r|r|r|}
		\hline
		Dataset & Isolet & USPS & letters & DSLR & Amazon & Webcam &
		Caltech\\
		\hline 
		\hline
		Euclidean & 7.95 & 5.68 & 5.26 \smpm 0.2 & 61.71 \smpm 6.4 & 61.48 \smpm 3.7 & 49.15 \smpm 3.9 & 73.1 \smpm 3.6\\
		\hline
		LMNN & \bf{3.85} & 5.73 & 4.09 \smpm 0.2 & 49.6 \smpm 5.5 & 27.04 \smpm 1.8 & 14.58 \smpm 4.6 & 44.61 \smpm 1.3 \\
		\hline
		GB-LMNN & 3.98 & 6.33 & 2.96 \smpm 0.1 & 45.18 \smpm 10.5 & 27.25 \smpm 2.2 & 14.58 \smpm 4.6 & 45.55 \smpm 6.9 \\
		\hline
		MLR & 33.61 & 10.26 & 35.50 \smpm 16.5 &34.40 \smpm 8.2 & \bf{24.21 \smpm 3.4} & 18.31 \smpm 5.3 & 46.04 \smpm 1.9 \\
		\hline 
		ITML & 7.18 & 5.88 & 5.35 \smpm 0.3 & \bf{28.04 \smpm 7.7} & 33.09 \smpm 2.1 & \bf{12.54  \smpm 5.4} & 51.91 \smpm 3.3 \\
		\hline
		NCA & 5.52 & 5.03 & 5.8 \smpm 1.3 & 45.18 \smpm 6.5 & 32.47 \smpm 1.7 & 19.32 \smpm 7.5 & \bf{44.17 \smpm 2.6}\\    
		\hline
		ours & 3.87 & \bf{4.98} & \bf{2.8 \smpm 0.2} & 33.00 \smpm 5.7 & 26.10 \smpm 2.7 & 14.24 \smpm 6.5 & 45.76 \smpm 2.9\\ 
		\hline
	\end{tabu}   
	\caption{$k$NN error,for  $k$=3, 7 and 11. No feature scaling was applied in these experiments. Mean and standard deviation are shown
		for data sets on which 5-fold partition was used. Best performing methods are shown in bold. Note that the only non-linear metric learning method in the above is GB-LMNN.}
	\label{tab:errors_native}
\end{table}

\section{Conclusion and summary of work}\label{sec:discuss}
In this part of the dissertation we proposed a formulation of the metric learning for $k$NN classifier
as a structured prediction problem, with discrete latent variables representing the selection of $k$ neighbors. We also provided efficient algorithms for exact inference in this model, including for loss augmented inference. While proposed in the context of metric learning, these procedures might be of wider interest. We also devised a stochastic gradient descent based procedure for learning in this model. The proposed approach allows us to learn a Mahalanobis metric with an objective which is a more direct proxy for the stated goal (improvement of classification
by $k$NN rule) than previously proposed similarity learning methods. Our learning algorithm is simple yet efficient, converging on all the data sets we have experimented with in run-times significantly lesser or comparable than other methods, such as LMNN and MLR.

\vspace{2mm}
\noindent
We used the Frobenius norm as the choice of the regularizer in our experiments. This was motivated by the intuition of wanting to do capacity control but without biasing our model towards any particular form. In our experiments, we have also experiments with other schemes for regularization such as using the trace norm of $\mathbf{W}$ and the shrinkage towards Euclidean distance, $\|\mathbf{W}-\mathbf{I}\|^2_F$, but found both to be inferior to $\|\mathbf{W}\|_F^2$. Our suspicion for such behavior is that often the optimal matrix that parameterizes the distance function i.e. $\mathbf{W}$ corresponds to a highly anisotropic scaling of data dimensions, and thus an initial bias towards $\mathbf{I}$ may be unhealthy.

\vspace{2mm}
\noindent
The results in this section of the dissertation are restricted to learning the Mahalanobis metric, which is an appealing choice for a number of reasons. In particular, learning such metrics is equivalent to learning linear
embedding of the data, allowing very efficient methods for metric search. As mentioned in section \ref{sec:setup}, we can consider more general notions of distance:
\begin{equation*}
D_\mathcal{W}(\mathbf{x},\mathbf{x}_i) = \|\Phi(\mathbf{x};\mathcal{W}) - \Phi(\mathbf{x}_i;\mathcal{W})\|_2^2
\end{equation*}
Where $\Phi(\mathbf{x};\mathcal{W})$ is a map (possibly non-linear), $\mathbf{x} \to \Phi(\mathbf{x})$, parameterized by $\mathcal{W}$.

\vspace{2mm}
\noindent
In such cases, learning $S$ when the map is non-linear can be seen as optimizing for a kernel with discriminative objective of improving $k$NN performance. Such a model would be more expressive, and using methods for automatic differentiation should be straightforward to optimize.

\vspace{2mm}
\noindent
In the next chapter of this disseration, we explore some extensions of this approach. First, we consider the case when the query and database point are mapped to different subspaces. This was inspired by the work of Neyshabur \emph{et al.} on asymmetric hashing \cite{neyshabur15} \cite{neyshabur13}. Next, we leverage the objective for the discriminative learning of Hamming distance, which gives us compact binary representations for fast retrieval. Lastly, we modify the inference procedures to make the approach amenable to learn suitable metrics for the (harder) case of $k$-NN regression.

%*****************************************
%*****************************************
%*****************************************
%*****************************************
%*****************************************

\cleardoublepage
%*****************************************
\chapter{Extensions}\label{ch:extensions}
%*****************************************

\begin{aside}{Outline}
	The main contribution of the \emph{Neighborhood Gerrymandering} method was primarily in its \emph{loss} and inference procedures; its use for Mahalanobis metric learning being just one use case. In this chapter this wider applicability is demonstrated in three different settings: Asymmetric similarity learning, metric learning when the labels are continuous and finally learning compact binary codes that are similarity sensitive in Hamming space. 
\end{aside}

In the previous chapter, the problem of metric learning for $k$-NN classification was formulated as a large margin structured prediction problem, with the choice of neighbors represented by discrete latent variables. Efficient algorithms for exact inference and loss-augmented inference in this model (dubbed as \emph{neighborhood gerrymandering}) were provided, and it was argued; with supporting experiments, that this formulation gave a more direct proxy for nearest neighbor classification as compared to prior art in metric learning. It was also noted that while the method was only tested in the case of learning Mahalanobis distances for points living in an explicit feature space, the methodology was more generally applicable.

\vspace{2mm}
\noindent
To impress upon this point, we again consider the distance computation that might be used:

\begin{equation}\label{eq:distanceform}
D_\mathcal{W}(\mathbf{x},\mathbf{x}_i) = \|\Phi(\mathbf{x};\mathcal{W}) - \Phi(\mathbf{x}_i;\mathcal{W})\|_2^2
\end{equation}
Note that there is considerable freedom in choosing the map $\Phi(\mathbf{x};\mathcal{W})$, and while the experiments reported in Chapter \ref{ch:mlng} were specifically for the Mahalanobis distance i.e.  $\mathbf{x} \to \mathbf{L}\mathbf{x}$ such that $\mathbf{L}^T\mathbf{L} = \mathbf{W} \succeq 0$, the map $\mathbf{x} \to \Phi(\mathbf{x};\mathcal{W})$ could be non-linear, with $\mathcal{W}$ representing the parameters of a deep neural network. With the same structured formulation, the procedures for inference and loss-augmented inference would remain the same, and the optimization would involve training a Siamese-like network while optimizing the \emph{gerrymandering} objective. This direction was explored by dissertation author, however, in this chapter we present work on three somewhat different directions. 

\begin{enumerate}
	\item[1] \textbf{Asymmetric Similarity Computation: } Note that in equation \ref{eq:distanceform}, the \emph{query} point $\mathbf{x}$ and the \emph{database} point $\mathbf{x}_i$ involve the same transformation parameterized by $\mathcal{W}$. But it is clear that this need not be the case, indeed, we might modify \ref{eq:distanceform} to the following form:
	\begin{equation}\label{eq:distanceformasym}
	D_\mathcal{W}(\mathbf{x},\mathbf{x}_i) = \|\Phi(\mathbf{x};\mathcal{W}) - \Phi'(\mathbf{x}_i;\mathcal{W'})\|_2^2
	\end{equation}
	while ensuring that $\Phi$ and $\Phi'$ maps to the same metric space i.e. $\Phi, \Phi': \mathbb{R}^d \to \mathbb{R}^p$. In particular, we work with linear transformations: we transform the query point as $\mathbf{x} \to \mathbf{U}\mathbf{x}$, and the database point as $\mathbf{x} \to \mathbf{V}\mathbf{x}$, with $\mathbf{U}, \mathbf{V} \in \mathbb{R}^{d \times d}$. This approach, its motivation and empirical validation is discussed in section \ref{sec:asym}.
	\item[2] \textbf{Similarity Computation in Hamming Space: } As discussed in chapter \ref{ch:discintro}, nearest neighbor methods are limited by two factors: first, the choice of the metric defining ``nearest'', and second, efficient indexing facilitating fast retrieval from large datasets. In chapter \ref{ch:mlng}, as well as in sections \ref{sec:asym} and \ref{sec:mlknnregression}, we only focus on the former. The latter, however, is important for the success of nearest neighbor methods as well. This is often done by generating compact binary codes for the data in either a supervised or unsupervised fashion. In section \ref{sec:hamming} we discuss an approach that generates binary codes while optimizing for the nearest neighbor performance in Hamming space using the gerrymandering objective. In particular, $\Phi: \mathbb{R}^d \to \mathcal{H}$, where $\mathcal{H}$ is the Hamming space, containing $2^c$ binary codes of length $c$. The Hamming space is endowed with a metric, the Hamming distance, which implies that with a good choice of $\Phi$, we can use the same inference procedures as in chapter \ref{ch:mlng} to optimize for binary codes such that the nearest neighbor classification performance in Hamming space is improved. 
	\item[3] \textbf{Metric Learning with Continuous Labels: } In preceding discussions we have worked with an instance space $(\mathcal{X},d)$, which is a metric space, and a discrete label space $\mathcal{Y} \in \mathbb{Z}_+$. We assume an unknown, smooth function $f: \mathcal{X} \to \mathcal{Y}$, and try to optimize for the metric such that accuracy of the $k$-nearest neighbor classifier is improved. We now change tack and work with the case when the labels are not discrete i.e. $\mathcal{Y} \in \mathbb{R}$. Therefore, the problem becomes that of optimizing for a metric such that $k$-nearest neighbor regression performance improves. This problem is also somewhat different than the previous two in that the inference and loss-augmented inference procedures, which worked for discrete labels, are no longer directly applicable. Thus we need to suitably modify them in order to make inference tractable. This approach is discussed further in section \ref{sec:mlknnregression}.
\end{enumerate}

\vspace{2mm}
\noindent
In the following three sections, we develop the ideas outlined above in detail. 

\section{Asymmetric Metric Learning}\label{sec:asym}

Recall that for some $\mathbf{W} \succeq 0 $, we could factorize it as $\mathbf{W} = \mathbf{L}^T\mathbf{L}$. Thus, the squared distance may be written as:

\begin{equation}\label{symmetric-distance}
D_\W(\x, \x_i) = (\x - \x_i)^T \mathbf{W} (\x - \x_i) = \|\mathbf{L}\x -\mathbf{L}\x_i\|_2^2 = (\mathbf{L}\x -\mathbf{L}\x_i)^T(\mathbf{L}\x -\mathbf{L}x_i)
\end{equation}

\noindent
We can thus think of the metric learning problem as learning the same projection matrix $\mathbf{L}$ for both the query (denoted $\x$) and the database points (denoted $\x_i$). In this section, we instead consider the following alternative formulation instead:
\begin{equation}\label{asymmetric-distance}
D_\W(\x, \x_i) = \|\mathbf{U}\x -\mathbf{V}\x_i\|_2^2 = (\mathbf{U}\x -\mathbf{V}\x_i)^T(\mathbf{U}\x -\mathbf{V}x_i)
\end{equation}
\noindent
In this formulation, all the points are not subject to the same global transformation: the query and the database points are (linearly) projected separately, thus making the distance computation asymmetric.

\vspace{2mm}
\noindent
The above (Eq. \ref{asymmetric-distance}) may be rewritten as follows:

\begin{equation}\label{asymmetric-distance2}
D_\mathbf{W}(\mathbf{x}, \mathbf{x}_i) = \begin{bmatrix} \mathbf{x} & \mathbf{x}_i\end{bmatrix} \begin{bmatrix} \mathbf{U}^T\mathbf{U} & -\mathbf{U}^T\mathbf{V} \\ -\mathbf{V}^T\mathbf{U} & \mathbf{V}^T\mathbf{V} \end{bmatrix} \begin{bmatrix} \mathbf{x} \\ \mathbf{x}_i\end{bmatrix}
\end{equation}

\noindent
Therefore, the problem of learning $\mathbf{U}$ and $\mathbf{V}$ as specified in the similarity computation of \ref{asymmetric-distance} is equivalent to learning a matrix $\mathbf{W} \in \mathbb{R}^{2d \times 2d}$ such that $\mathbf{W} \succeq 0 $, with

\begin{equation*}
\mathbf{W} = \begin{bmatrix} \mathbf{U}^T\mathbf{U} & -\mathbf{U}^T\mathbf{V} \\ -\mathbf{V}^T\mathbf{U} & \mathbf{V}^T\mathbf{V} \end{bmatrix}
\end{equation*}

\noindent
We will return to the formulation of the metric learning problem in this setting in the next section. But before doing so, it perhaps might be pertinent to point out the motivation for this approach. 

\vspace{2mm}
\noindent
The idea of subjecting the query and database points to different projections was inspired by work on hashing \cite{neyshabur15} \cite{neyshabur13}, and was explored by the dissertation author \cite{trivediAsym15} with the first author of \cite{neyshabur15}\cite{neyshabur13}, immediately after the publication of \cite{trivediNIPS14}. The main (and somewhat counterintuitive) message of \cite{neyshabur13} was the following: Usually, the similarity $S(\x, \x_i)$ for query point $\x$ and database point $\x_i$, is approximated by the Hamming distance between the outputs of the same hash function $f(\x)$ and $f(\x_i)$, for some $f \in \{\pm 1\}^k$. Now, instead of using the same hash function for both the query and database points, suppose two distinct functions $f(\x)$ and $g(\x_i)$ were used instead, then, even in cases where the target similarity happens to be symmetric, this asymmetry in the similarity computation affords representational advantages and reduces code length. Thus a natural question to consider was to see if asymmetry offered any advantages in the case of discriminative metric learning as well. With this brief background on the motivation, we now proceed to formulate the problem using the \emph{gerrymandering} formalism described in the previous chapter.

\subsection{Formulation}

\noindent
Coming back to the distance computation in equations \ref{asymmetric-distance} and \ref{asymmetric-distance2}: given $\mathbf{U}, \V \in \mathbb{R}^{d \times d}$, for any $h \subset \X$ with $|h| = k$, we can define the similarity between $\x$ and $h$ in direct analogy with that in the previous chapter:

\begin{equation}\label{eq:asysimilarity}
S_{\mathbf{U}, \mathbf{V} }(\mathbf{x}, h) = - \sum_{\mathbf{x}_i \in h} (\mathbf{U} \mathbf{x} - \mathbf{V} \mathbf{x}_i)^T (\mathbf{U} \mathbf{x} - \mathbf{V}\mathbf{x}_i)
\end{equation}

\noindent
Likewise, we can use the above measure of similarity to define the following surrogate loss for $k$-NN classification:

\begin{equation}
L(\mathbf{x},y,\{\mathbf{U},\mathbf{V}\}) = \max_{h} \left[S_{\mathbf{U},\mathbf{V}}(\mathbf{x},h) + \Delta(y,h)\right] - \max_{h:\Delta(y,h)=0} S_{\mathbf{U},\mathbf{V}}(\mathbf{x},h)
\end{equation}

\noindent
Given the loss formulation, we are now left with an appropriate penalty for capacity control. While there are many options to consider, a natural choice is to penalize for the Frobenius  norm of the full matrix $\W$. The objective then becomes:

\begin{equation}
\min_{\mathbf{U}, \mathbf{V}} \|\W\|_{F} + C \sum_i L(\mathbf{x}_i, y_i, \{\mathbf{U}, \mathbf{V}\})
\end{equation}

\noindent
The derivatives of the loss and the regularizer with respect to $\mathbf{U}$ (query gradient) and $\mathbf{V}$ (database gradient) are worked out to be:

\begin{equation}
\frac{\partial S_{\mathbf{U}, \mathbf{V}}(\mathbf{x},h) }{\partial \mathbf{U}} = -2 \left( \sum_{\mathbf{x}_i \in h} \mathbf{U}(\mathbf{x} \mathbf{x}^T) - (\mathbf{V} \mathbf{x}_i)\mathbf{x}^T \right)
\end{equation}

\begin{equation}
\frac{\partial S_{\mathbf{U}, \mathbf{V}}(\mathbf{x},h) }{\partial \mathbf{V}} = -2 \left( \sum_{\mathbf{x}_i \in h} \mathbf{V}(\mathbf{x}_i \mathbf{x}_i^T) - (\mathbf{U} \mathbf{x})\mathbf{x}_i^T \right)
\end{equation}

\begin{equation}
\frac{\partial \|W\|_{F}}{\partial \mathbf{U}} = 2 (\mathbf{U} \mathbf{U}^T)\mathbf{U} + 4(\mathbf{V} \mathbf{V}^T)\mathbf{U}
\end{equation}

\begin{equation}
\frac{\partial \|W\|_{F}}{\partial \mathbf{V}} = 2 (\mathbf{V} \mathbf{V}^T)\V + 4(\mathbf{U}\mathbf{U}^T)\mathbf{V}
\end{equation}

\noindent
Other possibly well motivated regularizers (that were also tried during experimentation) are $\|\mathbf{U}^T\mathbf{U}\|_{F}$ or $\|\mathbf{U}\|_{F}$ in the update equation for $\mathbf{U}$ and $\|\mathbf{V}^T\mathbf{V}\|_{F}$ or $\|\mathbf{V}\|_{F}$ in the update equation for $\mathbf{V}$.

\subsection{Optimization}

\noindent
With all the machinery stated and out of the way, we are finally in a position to write how an iteration of the learning algorithm proceeds. Each iteration $t$ of the algorithm consists of three steps:
\begin{enumerate}
	\item Targeted inference of $h_i^*$ for each sample $\x_i$:
	\begin{equation}
	h_i^* = \argmax_{h:\Delta(y_i,h)=0} S_{\mathbf{U}^{(t)},\mathbf{V}^{(t)}} (\x_i,h)
	\end{equation}
	This can be done by algorithm \ref{alg:hstar} in time $O(N \log N)$, but with the slight modification of using equation \ref{eq:asysimilarity} for the similarity computation instead.
	\item Loss augmented inference of $\hat{h}_i$ for each sample $\x_i$:
	\begin{equation}
	\hat{h}_i = \argmax{h} \left[S_{\mathbf{U}^{(t)},\mathbf{V}^{(t)}} (\x_i,h) + \Delta(y_i,h)\right]
	\end{equation}
	This can be done by algorithm \ref{alg:haugmented}, again by using \ref{eq:asysimilarity} for the similarity instead.
	\item Gradient updates for $\mathbf{U}$ and $\mathbf{V}$.
	\begin{align*}
	\mathbf{U}^{(t+1)} &= \mathbf{U}^{(t)} - \eta^{(t)} \left[\frac{\partial S_{\mathbf{U}, \mathbf{V}}(\x,\hat{h}_i) }{\partial \mathbf{U}} - \frac{\partial S_{\mathbf{U}, \mathbf{V}}(\x,h^*_i) }{\partial \mathbf{U}}  + 2 (\mathbf{U} \mathbf{U}^T)\mathbf{U}+ 4(\mathbf{V} \mathbf{V}^T)\mathbf{U} \right]\\
	\mathbf{V}^{(t+1)} &= \mathbf{V}^{(t)} - \eta^{(t)} \left[\frac{\partial S_{\mathbf{U}, \mathbf{V}}(\x,\hat{h}_i) }{\partial \mathbf{V}} - \frac{\partial S_{\mathbf{U}, \mathbf{V}}(\x,h^*_i) }{\partial \mathbf{V}}  + 2 (\mathbf{V} \mathbf{V}^T)\mathbf{V} + 4(\mathbf{U} \mathbf{U}^T)\mathbf{V} \right]
	\end{align*}
	
\end{enumerate}

\subsection{Experiments and Conclusion}

The experimental setup is the same as in \ref{sec:exp}: We consider the same datasets, replicate the protocol in~\cite{kedem2012non}, consider the same test train splits, cross validation procedure and report $k$-NN errors for $k= 3, 7 ,11$ for different methods on exactly the same folds. However, we do not repeat the experiments for different feature normalizations other than z-scoring, as they were consistently found to not help. This was also reflected in the results reported in \ref{sec:exp}. Therefore the numbers reported in \ref{tab:asym_errors_zscore} are identical to that in \ref{tab:errors_zscore}, except for the last column. 

\vspace{2mm}
\noindent 
The only difference as compared to \ref{sec:exp} is that the initialization using ReliefF~\cite{relieff1992} was done such that $\mathbf{U}$ and $\mathbf{V}$ were initialized to the same diagonal matrix; with the diagonal elements being the square-root of the weights obtained by using ReliefF. The experiments show no clear trend, although one thing is clear: there is marginal improvement over \cite{trivediNIPS14}, and at least the results obtained by using the asymmetric distance as consistently better than the other methods in the competition. These results demonstrate that using this asymmetric similarity metric in conjunction with the \emph{gerrymandering} formulation did slightly better than the case where the distance computation was symmetric. 

\begin{table}
	\small
	\centering
	\begin{tabu}{l}
		k = 3
	\end{tabu}
	\begin{tabu}{|l|l|r|r|r|r|r|r|r|}
		\hline
		Dataset & Isolet & USPS & letters & DSLR & Amazon & Webcam &
		Caltech\\
		\hline 
		\hline
		\rowfont{\small}
		$d$ & 170 & 256 & 16 & 800 & 800 & 800 & 800\\
		\rowfont{\small}
		$N$ & 7797 & 9298 & 20000 & 157 & 958 & 295 & 1123\\
		\rowfont{\small}
		$C$ & 26 & 10 & 26 & 10 & 10 & 10 & 10\\
		\hline
		\small{Euclidean} & 8.66 & 6.18  &  4.79 \smpm 0.2 & 75.20 \smpm 3.0  & 60.13\smpm 1.9 & 56.27 \smpm 2.5 & 80.5 \smpm 4.6 \\
		\hline
		\small{LMNN~\cite{weinberger2009distance}} & 4.43 & 5.48  & 3.26 \smpm 0.1  & 24.17 \smpm 4.5 & 26.72 \smpm 2.1 & 15.59 \smpm 2.2 & 46.93 \smpm 3.9 \\
		\hline
		\small{GB-LMNN~\cite{kedem2012non}} & \bf{4.13} & 5.48 & 2.92 \smpm 0.1 & 21.65 \smpm 4.8 & 26.72 \smpm 2.1 & 13.56 \smpm 1.9 & 46.11 \smpm 3.9 \\
		\hline
		\small{MLR~\cite{mcfee10_mlr}} & 6.61 & 8.27 & 14.25 \smpm 5.8 & 36.93 \smpm 2.6 & 24.01 \smpm 1.8 & 23.05 \smpm 2.8 &  46.76 \smpm 3.4 \\
		\hline 
		\small{ITML~\cite{itml2007davis}} & 7.89  & 5.78  & 4.97 \smpm 0.2  & 19.07 \smpm 4.9  & 33.83 \smpm 3.3  & 13.22 \smpm 4.6 & 48.78 \smpm 4.5 \\
		\hline
		\small{1-NCA~\cite{goldberger2004neighbourhood}} &  6.16 & 5.23 & 4.71 \smpm 2.2 & 31.90 \smpm 4.9 & 30.27 \smpm 1.3 & 16.27 \smpm 1.5  & 46.66 \smpm 1.8\\    
		\hline
		\small{k-NCA} &  4.45 & \bf{5.18} & 3.13 \smpm 0.4  & 21.13 \smpm 4.3 & 24.31 \smpm 2.3  &  13.19 \smpm 1.3  & 44.56 \smpm 1.7 \\    
		\hline
		\small{MLNG~\cite{trivediNIPS14}} & 4.87 & \bf{5.18} & \bf{2.32 \smpm 0.1} & \bf{17.18\smpm4.7} & \bf{21.34\smpm 2.5} & \bf{10.85\smpm 3.1} & \bf{43.37\smpm 2.4} \\ 
		\hline
		\small{Asym-MLNG} & 4.65 & \bf{5.17} & \bf{2.39 \smpm 0.1} & 19.01\smpm3.6 & 23.45\smpm 1.9 & \bf{10.53\smpm 4.7} & \bf{43.6\smpm 2.1} \\ 
		\hline		
	\end{tabu}
	\begin{tabu}{l}
		k = 7 
	\end{tabu}
	\begin{tabu}{|l|l|r|r|r|r|r|r|r|}
		\hline
		Dataset & Isolet & USPS & letters & DSLR & Amazon & Webcam &
		Caltech\\
		\hline 
		\hline
		\small{Euclidean} & 7.44 & 6.08 &  5.40 \smpm 0.3 & 76.45 \smpm 6.2 &  62.21 \smpm 2.2 & 57.29 \smpm 6.3 & 80.76 \smpm 3.7 \\
		\hline
		\small{LMNN~\cite{weinberger2009distance}} &  3.78 & \bf{4.9}  & 3.58 \smpm 0.2  & 25.44 \smpm 4.3 & 29.23 \smpm 2.0 & 14.58 \smpm 2.2 & 46.75 \smpm 2.9 \\
		\hline
		\small{GB-LMNN~\cite{kedem2012non}} & \bf{3.54} & \bf{4.9} & 2.66 \smpm 0.1 & 25.44 \smpm 4.3 & 29.12 \smpm 2.1 & 12.45 \smpm 4.6 & 46.17 \smpm 2.8 \\
		\hline
		\small{MLR~\cite{mcfee10_mlr}} & 5.64 & 8.27 & 19.92 \smpm 6.4 & 33.73 \smpm 5.5 & 23.17 \smpm 2.1 & 18.98 \smpm 2.9 &  46.85 \smpm 4.1 \\
		\hline 
		\small{ITML~\cite{itml2007davis}} & 7.57 & 5.68 & 5.37 \smpm 0.5 &  22.32 \smpm 2.5 & 31.42 \smpm 1.9 & \bf{10.85 \smpm 3.1} & 51.74 \smpm 2.8\\
		\hline
		\small{1-NCA~\cite{goldberger2004neighbourhood}} & 6.09 & 5.83 & 5.28 \smpm 2.5 & 36.94 \smpm 2.6 & 29.22 \smpm 2.7 & 22.03 \smpm 6.5 & 45.50 \smpm 3.0 \\    
		\hline
		\small{k-NCA} &  4.13 & 5.1 & 3.15 \smpm 0.2 & 22.78 \smpm 3.1 & 23.11 \smpm 1.9 & 13.04 \smpm 2.7  & 43.92 \smpm 3.1 \\    
		\hline
		\small{MLNG~\cite{trivediNIPS14}} & 4.61 & \bf{4.9} & 2.54\smpm 0.1 & \bf{21.61 \smpm 5.9} & \bf{22.44 \smpm 1.3} & 11.19\smpm 3.3 & \bf{41.61 \smpm 2.6} \\ 
		\hline
		\small{Asym-MLNG} & 4.63 & \bf{4.9} & \bf{2.34 \smpm 0.1} & 23.65\smpm3.9 & 23.84\smpm 2.8 & 11.4\smpm 2.3 & 41.35\smpm 2.2 \\ 
		\hline		
	\end{tabu}
	
	\begin{tabu}{l}
		k = 11
	\end{tabu}
	\begin{tabu}{|l|l|r|r|r|r|r|r|r|}
		\hline
		Dataset & Isolet & USPS & letters & DSLR & Amazon & Webcam &
		Caltech\\
		\hline 
		\hline
		\small{Euclidean} & 8.02 & 6.88 & 5.89 \smpm 0.4 & 73.87 \smpm 2.8  & 64.61 \smpm 4.2 & 59.66 \smpm 5.5 & 81.39 \smpm 4.2\\
		\hline
		\small{LMNN~\cite{weinberger2009distance}} & \bf{3.72} & \bf{4.78} & 4.09 \smpm 0.1 & 23.64 \smpm 3.4 & 30.12 \smpm 2.9 & 13.90 \smpm 2.2 & 49.06 \smpm 2.3 \\
		\hline
		\small{GB-LMNN~\cite{kedem2012non}} & 3.98 & \bf{4.78} & 2.86 \smpm 0.2 & 23.64 \smpm 3.4 & 30.07 \smpm 3.0 & 13.90 \smpm 1.0 & 49.15 \smpm 2.8 \\
		\hline
		\small{MLR~\cite{mcfee10_mlr}} & 5.71 & 11.11 & 15.54 \smpm 6.8 & 36.25 \smpm 13.1 & 24.32 \smpm 3.8 & 17.97 \smpm 4.1 &  44.97 \smpm 2.6 \\
		\hline 
		\small{ITML~\cite{itml2007davis}} & 7.77  & 6.63 & 6.52 \smpm 0.8  & \bf{22.28 \smpm 3.1} & 30.48 \smpm 1.4 & 11.86 \smpm 5.6 & 50.76 \smpm 1.9 \\
		\hline
		\small{1-NCA~\cite{goldberger2004neighbourhood}} &  5.90 & 5.73 & 6.04 \smpm 2.8 & 40.06 \smpm 6.0 & 30.69 \smpm 2.9 & 26.44 \smpm 6.3 & 46.48 \smpm 4.0 \\    
		\hline
		\small{k-NCA} &  4.17 & 4.81 & 3.87 \smpm 0.6 & 23.65 \smpm 4.1 & 25.67 \smpm 2.1 & 11.42 \smpm 4.0  & 43.8 \smpm 3.1 \\    
		\hline
		\small{MLNG~\cite{trivediNIPS14}} & 4.11 & 4.98 & 3.05\smpm 0.1 & \bf{22.28 \smpm 4.9} & \bf{24.11\smpm 3.2} & \bf{11.19 \smpm 4.4} & \bf{40.76 \smpm 1.8}  \\ 
		\hline
		\small{Asym-MLNG} & 4.0 & \bf{4.81} & \bf{2.4 \smpm 0.1} & 23.78\smpm4.3 & \bf{24.11\smpm 3.9} & \bf{11.1\smpm 3.7} & 43.7\smpm 2.7 \\ 
		\hline		
	\end{tabu}   
	\caption{$k$NN errors for  $k$=3, 7 and 11 (asymmetric metric learning versus other methods). Features were scaled by z-scoring.}
	\label{tab:asym_errors_zscore}
\end{table}

%***************************************** end of section on Asymmetric Metric Learning. 

\section{Hamming Distance Metric Learning}\label{sec:hamming}

As discussed earlier, performance of nearest neighbor classification methods are often limited by two factors: 
\begin{enumerate}
	\item[1] Computational cost of searching for nearest neighbors in a large database.
	\item[2] The choice of the underlying metric that defines ``nearest''.
\end{enumerate}

\noindent
In preceding discussions in this dissertation, we have exclusively focused on addressing (2): the choice of metric. In this chapter we turn our attention to (1), while still maintaining the flavor of solutions that were used to address (2).

\vspace{2mm}
\noindent
The cost searching for nearest neighbors is usually addressed by efficient indexing and searching for approximate nearest neighbors instead (see \cite{arya1998optimal,datar2004locality,beygelzimer2006cover,IndykMotwani, Rabani} and references therein). The motivation for some such methods is simple: Usually for some task, approximate nearest neighbors rather than exact nearest neighbors should be good enough, assuming the data is well behaved. Relaxing the requirement for exact nearest neighbors can allow for sub-linear time search, which can be substantial for extremely large dataset sizes.

\vspace{2mm}
\noindent
Yet another (but related to the above) approach is instead searching in Hamming space. That is, the data is projected onto a (potentially) lower dimensional Hamming space, where fast exact or approximate nearest neighbor search might be carried out. Needless to say, the embeddings generated must reflect some properties of the data. To this end, the original work on Locality Sensitive Hashing \cite{IndykMotwani, datar2004locality}, the codes were found using random projections, such that two points in the Hamming space were likely to be close if they were close in the original feature space. It has been observed that while random projections generate codes that are faithful to the pairwise distance, the code lengths can become prohibitively large. 

\vspace{2mm}
\noindent
Moreover, such approaches are completely label oblivious: the codes generated do not reflect any semantic structure, and often the purpose of nearest neighbor search is some downstream task-specific application (like classification). The approach to instead machine learn the binary codes such that they are \emph{similarity sensitive} to the underlying semantic structure was taken in the pioneering works of Shakhnarovich \cite{GregThesis}, the \emph{Semantic Hashing} or \emph{supermarket search} of Salakhutdinov and Hinton \cite{SemanticHashing}, and Weiss \emph{et al.} \cite{weiss08}. To motivate machine learning of binary codes, consider we have to search an image from a large database which is similar to a given query image. We can then ask a number of binary questions to make search easier: Is the query a color image or a grayscale image? Is there is a dog in the query or not? Does the query represent an indoor scene or an outdoor scene and so on. We can think of the codes as encoding a set of binary choices that are generated with the explicit goal to reflect label structure. If the underlying task is $k$-NN classification, we would want the code to be able to reflect the class of the image.  

\vspace{2mm}
\noindent
Following \cite{GregThesis, SemanticHashing, weiss08}, in the last decade there has been an explosion in research in this area of learning compact, similarity-sensitive binary codes, which is nearly impossible to review. In any case, for the purpose of this chapter, it is not necessary either. For some prominent works we direct the reader to ~\cite{kulis2009learning,wang2010sequential,norouzi2011minimal,gong2011iterative} and the references therein, or more recent works that cite these.

\vspace{2mm}
\noindent
To motivate our approach to the problem of learning binary codes that are similarity sensitive, we first hark back to work discussed in this dissertation thus far: We considered methods for learning the underlying metric or notion of similarity, this was achieved by projecting the data into a real space such that the underlying nearest neighbor classification accuracy in this space was improved. Indeed, the loss function employed was such that it was a more direct proxy to the $k$-NN classification error. 

\vspace{2mm}
\noindent
In this section, we suitably modify the framework proposed earlier for learning binary codes such that nearness in the Hamming space is a better proxy for classification accuracy. In other words, we learn a mapping from points in $\mathbb{R}^d$ to $\mathcal{H}$ in such a way that the accuracy of the $k$-NN classifier in the Hamming space is improved, by using the gerrymandering loss. Our main competitor in this regard is the method of Nourouzi \emph{et al.} \cite{norouzi2012hamming}, also inspired by Latent Structural SVMs, where a triplet loss was used to learn a Hamming distance between two points that is reflective of similarity. As already noted, the triplets are assumed to be set statically as an input to the algorithm, and the optimization focuses on the distance ordering rather than accuracy of classification. We use a loss function akin to that proposed in chapter \ref{ch:mlng}, but now adapted for the case of optimizing for the similarity directly in Hamming space. The gerrymandering loss being a more direct proxy for $k$-NN classification performance, can perhaps aid in the learning of compact binary codes that directly reflect nearest neighbor accuracy, and thus is perhaps better motivated than the approach of \cite{norouzi2012hamming}. In the next section, we formulate the problem and our approach to attack it.

\subsection{Formulation}
\noindent
We are interested in the problem of discriminative learning of Hamming distance between points. To this end, we first consider the Hamming distance between asymmetric linear binary hashes with code length $c$: Once we have used this to introduce the framework, we will consider other variants, such as using non-linear binary hashes as well as symmetric hashes. 

\begin{equation}
D_{\mathbf{U},\mathbf{V}}(\mathbf{x},\mathbf{x}_i) = \sum_{j=1}^c {\mathbf{1}}_{\sign(\mathbf{U}_j\mathbf{x}) \neq \sign(\mathbf{V}_j\mathbf{x}_i)}
\end{equation}
where $\mathbf{U},\mathbf{V} \in \mathbb{R}^{c\ast d}$ are the parameters of the model. For any $h\in \mathbf{X}$, we define the distance score of $h$ w.r.t. a point $\mathbf{x}$ as:
\begin{equation}
S_{\mathbf{U},\mathbf{V}}(\mathbf{x},h) = hc - \sum_{\mathbf{x}_i\in h} D_{\mathbf{U},\mathbf{V}}(\mathbf{x},\mathbf{x}_i) = \inner{\sign(\mathbf{U}\mathbf{x})}{\sum_{\mathbf{x}_i\in h} \sign(\mathbf{V}\mathbf{x}_i)}
\end{equation}
Therefore, the set of $k$ nearest neighbors of $\mathbf{x}$ in $\mathbf{X}$ is:
\begin{equation}
h_{\mathbf{U},\mathbf{V}}(\mathbf{x}) = \argmax_{\abs{h}=k} S_{\mathbf{U},\mathbf{V}}(\mathbf{x},h)
\end{equation}
We will assume that $k$ is known. Given any set $h$, we can predict the labels of $\mathbf{x}$ by the majority vote among members of $h$. We use a surrogate loss similar to
\begin{equation}
L(\mathbf{x},y,\{\mathbf{U},\mathbf{V}\}) = \max_{h} \left[S_{\mathbf{U},\mathbf{V}}(x,h) + \Delta(y,h)\right] - \max_{h:\Delta(y,h)=0} S_{\mathbf{U},\mathbf{V}}(\mathbf{x},h)
\end{equation}
\subsection{Optimization}
Each iteration $t$ of the algorithm consists of four steps:
\begin{enumerate}
	\item Targeted inference of $h_i^\ast$ for each sample $\mathbf{x}_i$:
	\begin{equation}
	h_i^\ast = \argmax_{h:\Delta(y_i,h)=0} S_{\mathbf{U}^{(t)},\mathbf{V}^{(t)}} (\mathbf{x}_i,h)
	\end{equation}
	\item Loss augmented inference of $\hat{h}_i$ for each sample $\mathbf{x}_i$:
	\begin{equation}
	\hat{h}_i = \argmax_{h} \left[S_{\mathbf{U}^{(t)},\mathbf{V}^{(t)}} (\mathbf{x}_i,h) + \Delta(y_i,h)\right]
	\end{equation}
	\item Gradient updates for $\mathbf{U}$ and $\mathbf{V}$. Let $b$ be the mini batch at iteration $t$. Since the sign function is not differentiable, we can approximate it by another function $f$. In this case, the updates will be
	\begin{align*}
	\mathbf{U}^{(t+1)} &= \mathbf{U}^{(t)} - \eta \sum_{\mathbf{x}_i\in b}\left[ f'(\mathbf{U}\mathbf{x}_i) \circ \left( \sum_{\mathbf{x}_j\in \hat{h}_i} \sign(\mathbf{V}\mathbf{x}_j) -\sum_{\mathbf{x}_j\in h^*_i} \sign(\mathbf{V}\mathbf{x}_j)\right)\right] \mathbf{x}_i^\top\\
	\mathbf{V}^{(t+1)} &= \mathbf{V}^{(t)} - \eta \sum_{\mathbf{x}_i\in b}\left[ \sum_{\mathbf{x}_j\in \hat{h}_i}\bigg(\sign(\mathbf{U}\mathbf{x}_i)\circ f'(\mathbf{V}\mathbf{x}_j)\bigg)\mathbf{x}_j^\top -\sum_{\mathbf{x}_j\in h^*_i}\bigg(\sign(\mathbf{U}\mathbf{x}_i)\circ f'(\mathbf{V}\mathbf{x}_j)\bigg)\mathbf{x}_j^\top\right]\
	\end{align*}
	where possible options for the function $f$ are:
	\begin{enumerate}
		\item $f(x)=x$ and therefore $f'(x)=1$.
		\item $f(x) = \tanh(x)$ and therefore $f'(x) = 1 - \tanh^2(x)$
	\end{enumerate}
	\item Normalization:
	\begin{align}
	\mathbf{U}^{(t+1)} &= \frac{ \mathbf{U}^{(t+1)} }{  \norm{\mathbf{U}^{(t+1)} }_F}\\
	\mathbf{V}^{(t+1)} &= \frac{\mathbf{V}^{(t+1)} }{ \norm{\mathbf{V}^{(t+1)} }_F}
	\end{align}
\end{enumerate}

\subsection{Symmetric Variant}

We have (with $\W \in \R^{c*d}$): 
\begin{equation}
S_{\W}(\x,h) = hc - \sum_{\x_i\in h} D_{\W}(\x,\x_i) = \inner{\sign(\W\x)}{\sum_{\x_i\in h} \sign(\W\x_i)}
\end{equation}

Gradient:

\begin{equation}
\frac{\partial S_{\W}(\x,h) }{\partial \W} = \left[ \left( \sum_{\x_i\in h} \sign(\W\x_i)\right) \circ f'(\W\x)\right] \x^\top + \left[ \sum_{\x_i\in h} \left( \sign(\W\x) \circ f'(\W\x_i) \right) \x_i^\top  \right]
\end{equation}

We also add the following penalty to the objective function as suggested in \cite{weiss08}. This encourages each bit, averaged over the training data, to be zero mean before quantization

\begin{equation}\label{eq:WeissZeroMean}
\frac{1}{2} \| mean_{\x} \sign(\W\x)\|_2^2
\end{equation}

Which adds the following term to the update of $\W$

\begin{equation}
mean_{\x} \left( \left[\sign(\W\x) \circ f'(\W\x)\right]\x^\top\right)
\end{equation}

\vspace{2mm}
\noindent
Note that we also add a term akin to \ref{eq:WeissZeroMean} to the asymmetric variant of our algorithm in the experiments with two additional terms suitably added in the gradient updates of $\mathbf{U}$ and $\mathbf{V}$. 

\vspace{2mm}
\noindent
Before we begin to describe our experiments, we first mention how we carried out nearest neighbor search in Hamming space. 

\subsection{Distance Computation in Hamming Space}\label{sec:asymHD}
When we consider a query point $\mathbf{x} \in \mathcal{H}$ and look for nearest neighbors in the database, there is a high probability of a tie, especially when code lengths are shorter. We experimented with several options for tie breaking, but eventually settled on using what is referred to in the literature as \emph{asymmetric hamming distance}. Note that this nomenclature is rather unfortunate given our formulation of Hamming metric learning that uses different hash functions for the query and the database points. This notion of the asymmetric hamming distance used in the literature (see \cite{norouzi2012hamming, charikarasym, gordoasym}) for retrieving points in Hamming space is quite different. We use the approach used by \cite{norouzi2012hamming} as we found it to consistently give superior results. For a given query $\mathbf{x} \in \mathbb{R}^d$, we do not binarize it while searching for neighbors in the database. The database points however live in $\mathcal{H}$. The distance between the query and the database point is given by:

$$AsymH(\mathbf{x}, \mathbf{x}_i; \mathbf{s}) = \frac{1}{4} \|\mathbf{x}_i - \tanh(Diag(\mathbf{s})\mathbf{x})\|_2^2$$

\vspace{2mm}
\noindent
As also observed by \cite{norouzi2012hamming}, the distance computation is relatively insensitive to the choice of scaling parameters $\mathbf{s} \in \mathbb{R}^d$. However, after some experimentation, we set the scale parameters to be such that the real valued projection of the query vector has an average absolute value of $0.4$.

\subsection{Experiments and Conclusion}

To test the efficacy of our method, we compare it directly with the experiments of \cite{norouzi2012hamming} on MNIST. For training, we initialize $\mathbf{U}$ and $\mathbf{V}$ (for the asymmetric case) and $\mathbf{W}$ (for the symmetric case) as random gaussian matrices with mean zero and standard deviation of 1. We stop training when the running average of the surrogate loss over the last epoch does not decrease substantially, or when the maximum number of epochs is reached. We use a decaying learning rate schedule as $\eta(t) = \frac{1}{t}$, and set the momentum parameter to 0.9. For training, to compare with the approach \cite{norouzi2012hamming}, we fix $k$ to be $3$ and $30$ while training our system. Finally, we set aside 10,000 points from the MNIST training set for the purpose of tuning the regularization constant. 

\vspace{2mm}
\noindent
We compare results that report the $k$NN error, and compare directly with the results of \cite{norouzi2012hamming}, as well as results reported using baseline methods that do not use binary codes to find nearest neighbors. 

\begin{table*}[htpb]
	\centering % used for centering table
	\begin{tabular}{l l l l l} % centered columns (4 columns)
		\hline %inserts double horizontal lines
		
		Hash Function/Loss & $k$ & 32 bits & 64 bits & 128 bits \\
		\hline \hline
		Linear/pairwise hinge & 3 & 4.3 & 2.78 & 2.46  \\
		Linear/triplet & 3 & 3.88 & 2.90 & 2.51  \\
		Two-Layer ($\tanh$)/pairwise  & 30 & 1.50 & 1.36 & 1.35 \\
		Two-Layer ($\tanh$)/triplet & 30 & 1.45 & 1.29 & 1.20  \\
		\hline
		\hline
		Linear/MLNG (SH) & 3 & 5.7 & 3.6 & 2.7  \\
		$\tanh$/MLNG (SH) & 30 & 2.3 & 2.17 & 1.91  \\
		Two-Layer ($\tanh$)/MLNG (SH)  & 30 & 1.39 & 1.31 & 1.28 \\
		\hline
		\hline
		Linear/MLNG (ASH) & 3 & 4.9 & 3.2 & 2.49  \\
		$\tanh$/MLNG (ASH) & 30 & 2.15 & 2.11 & 1.76  \\
		Two-Layer ($\tanh$)/MLNG (ASH)  & 30 & 1.2 & 1.18 & 1.25 \\
		\hline
		\hline
	\end{tabular}
	\footnotesize
	\caption{$k$ NN classification errors on MNIST using Hamming distance metric learning by gerrmandering, compared to the approach of \cite{norouzi2012hamming}. All results reported use the distance computation explicated in Section \ref{sec:asymHD}. MLNG refers to the Gerrymandering loss, SH refers to the use of symmetric hashes, while ASH to asymmetric hashes.} % title of Table
	\label{table:HDMLMnist} % is used to refer this table in the text
\end{table*}

\vspace{2mm}
\noindent
We observe the following clear trends: The linear hash function used in the gerrymandering framework, both in the symmetric and asymmetric case performs worse than the linear setting in  \cite{norouzi2012hamming}. We also observe that the performance improves in our case when the code length increases, almost equaling the performance of the linear setting in \cite{norouzi2012hamming} at higher code lengths. It is also noticeable that the asymmetric variant consistently gives better performance than the symmetric variant, suggesting that the asymmetry indeed helps in learning better codes. Next, in the two-layer setting: The asymmetric variant of the Gerrymandering loss consistently performs better than all the settings explored in \cite{norouzi2012hamming}, while the symmetric variant is comparable to \cite{norouzi2012hamming}, but usually slightly worse. Another interesting observation is that the gerrymandering loss in the two layer setting gives performance roughly similar as the code length increases, as contrasted to \cite{norouzi2012hamming} where it steadily improves. This shows that the loss is able to learn more compact and discriminative binary codes.

\vspace{2mm}
\noindent
In conclusion, in this section, we have presented a novel method for the discriminative learning of Hamming distance, and on experiments demonstrated that it works better than the competition in this space. One drawback of our method as compared to \cite{norouzi2012hamming} is that the inference procedures are quite expensive to solve exactly, making it quite slow. We leave the development of approximate inference procedures for future work, which will enable to scale our method to much larger dataset sizes.

%****************** End of the discrminiative part

\section{Metric Learning for k-NN Regression}\label{sec:mlknnregression}

To motivate this part of this chapter, we first review the basic setting in kernel regression, and use it to motivate the work of Weinberger and Tesauro \cite{MLKR}. We then use this to place our contribution in contrast.

\vspace{2mm}
\noindent
Recall the standard regression setting: We have an unknown, smooth $f: \mathbb{R}^d \to \mathbb{R}$, that we have to estimate based on labeled data $\{ (\mathbf{x}_1, y_1), (\mathbf{x}_2, y_2), \dots, (\mathbf{x}_N, y_N)\}$, with $\mathbf{x}_i \in \mathbb{R}^d$ and $y_i \in \mathbb{R}$. The labels are a noisy version of the actual function outputs i.e. $y_i = f(\mathbf{x}_i) + \epsilon$, where $\epsilon$ is unknown. The task then is to use this sample to get an estimate $\hat{f}$ of $f$ that minimizes some loss function. In the non-parametric setting, a standard regression technique is kernel regression that predicts an output $\hat{y}_i$ for an input $\mathbf{x}_i$ based on a weighted average of some of the inputs that are selected as its neighbors, the weighing is done as a function of the distance. This is given as below:

\begin{equation}\label{eq:nonparamreg}
\hat{y}_i = \sum_{j\neq i}^{N} w(\mathbf{x}_i,\mathbf{x}_j) y_j
\end{equation}

\noindent
To define the choice of weights $w(\mathbf{x}_i,\mathbf{x}_j)$, we first define a kernel function $K: \mathbb{R} \to \mathbb{R}$ which satisfies the following conditions:

$$\int K(x) dx = 1 \quad \text{ and }\quad K(x) = K(-x)$$

\noindent
While there is a vast literature on the choice of kernels, we content ourselves by defining only the Gaussian kernel, which suffices for the purpose of our discussion:

$$K(x) = \exp\Big( \frac{-x^2}{2\sigma^2}\Big)$$

\noindent
Then, for any pair of points $\mathbf{x}_i$ and $\mathbf{x}_j$ $\in \mathbb{R}^d$, we can define the kernel as:

$$K(\mathbf{x}_i, \mathbf{x}_j) = \exp\Big( \frac{-\|\mathbf{x}_i - \mathbf{x}_j\|_2^2}{2\sigma^2}\Big)$$

\noindent
We can then define the weights $w(\mathbf{x}_i,\mathbf{x}_j)$ in equation \ref{eq:nonparamreg} simply as:

$$w(\mathbf{x}_i,\mathbf{x}_j) = \frac{K(\mathbf{x}_i,\mathbf{x}_j)}{\sum_{j\neq i}^{N} K(\mathbf{x}_i,\mathbf{x}_j)}$$

\noindent
More generally, we can write the kernel for some distance function $d(\mathbf{x}_i,\mathbf{x}_j)$
$$K(\mathbf{x}_i, \mathbf{x}_j) = \exp\Big( \frac{-d(\mathbf{x}_i - \mathbf{x}_j)^2}{2\sigma^2}\Big)$$

\noindent
For our familiar parameterization of $d$ as the Mahalanobis distance i.e. $d(\mathbf{x}_i,\mathbf{x}_j) = \sqrt{(\mathbf{x}_i - \mathbf{x}_j) \mathbf{W} (\mathbf{x}_i - \mathbf{x}_j)}$ with $\mathbf{W} \succeq 0$ and $\mathbf{W} = \mathbf{L}^T\mathbf{L}$, the kernel might be written as:
\begin{equation}\label{eq:MLKRkernel}
K(\mathbf{x}_i, \mathbf{x}_j) = \exp\Big( \frac{-\|\mathbf{L}\mathbf{x}_i - \mathbf{L}\mathbf{x}_j\|}{2\sigma^2}\Big)
\end{equation}

\noindent
Working with \ref{eq:MLKRkernel} and \ref{eq:nonparamreg}, optimizing for $\mathbf{L}$ to minimize the loss $L = \sum_{i=1}^{N} (y_i - \hat{y}_i)^2$ is the main contribution of \emph{metric learning for kernel regression} \cite{MLKR}. This work remains to be the only major work in the literature that optimizes for the metric under a regression objective. The authors in \cite{MLKR} consider a gaussian kernel, which facilitates computing gradients with ease, and thus making optimization straightforward. In many scenarios, we are interested in using $k$ nearest neighbors, however, the problem of metric learning in such a setting is not straightforward considering the combinatorial nature of the problem, as well as taking into account the fact that defining sets of pairs of similar and dissimilar points is no longer straightforward as was in the classification case (where we could deem a pair of points to be similar if they belonged to the same class). The only work that we are aware of that optimizes for a metric under a nearest neighbor based regression objective is that of \cite{Mannor}. However, it works with a NCA \cite{goldberger2004neighbourhood} type objective, minimizing the expected squared loss, and thus only working with a $1$-NN regressor. In the next section, we outline our method that attempts to bridge this gap and learns a metric when the downstream task is $k$-NN regression.

\vspace{2mm}
\subsection{Problem formulation and Optimization}
\noindent
In this section, we are interested in the problem of learning a Mahalanobis metric for the case of $k$-NN regression rather than kernel regression. Note that in $k$-NN regression \ref{eq:nonparamreg} only has a particular choice of weights, we consider the following choice: if $\mathbf{x}_j$ is a neighbor of $\mathbf{x}_i$, then $w(\mathbf{x}_i,\mathbf{x}_j) = \frac{1}{k}$ and $0$ otherwise. Here, we show that with a suitable modification of the approach proposed in \ref{ch:mlng}, we can obtain an efficient metric learning algorithm for the case of $k$-NN regression as well. 

\vspace{2mm}
\noindent
In order to describe the approach, we briefly state the setting again (which is similar to \ref{ch:mlng}). That is, suppose we are given $N$ training examples $\X=\{\x_1,\dots,\x_N\}$ and their outputs $\y=[y_1,\dots,y_n]^T$, where $\x_i\in \mathbb{R}^d$ and $y_i\in \mathbb{R}$. For a subset $h \subset X$ with $|h| = k$, we can define its measure of similarity with a query point $\mathbf{x}$ as:

\begin{equation}
\label{eq:S_W)reg}
\dscore{\mathbf{W}}(\mathbf{x},h)\,=\,
-\sum_{\mathbf{x}_j\in h}
\mahax{\mathbf{W}}{\mathbf{x}}{\mathbf{x}_j}
\end{equation}

where 

\begin{equation}
\label{eq:D_W_reg}
\mahax{\mathbf{W}}{\mathbf{x}}{\mathbf{x}_i}\,=\,
\left(\mathbf{x}-\mathbf{x}_i\right)^T\mathbf{W}\left(\mathbf{x}-\mathbf{x}_i\right),
\end{equation}

\noindent 
Thus, we work with the same measure of similarity as in \ref{ch:mlng}, but a departure from the formulation occurs here as we move forward to state a reasonable objective. This is because, in the case of regression, the outputs $y_i$ are no longer discrete, but real valued. Therefore, unlike in \ref{ch:mlng}, the loss $\Delta(y,h)$ may not necessarily be zero for any set $h \subset X$. Note that, given $h$, the squared loss for a query point $(\mathbf{x},y)$ is:

\begin{equation}\label{eq:reghardloss}
\Delta(y,h) = \left(y - \frac{1}{k} \sum_{\x_i \in h } y_i\right)^2
\end{equation}
Armed with the notion of similarity as well as the loss, we can make a first attempt to define a loss for the task of metric learning for $k$-NN regression as follows. For some $\gamma > 0$:
\begin{equation}
L(\x,y,\W) = \max_{h} \left[S_{\W}(\mathbf{x},h) + \gamma \Delta(y,h)\right] - \max_{h} \left[S_{\W}(\x,h) - \gamma \Delta(y,h)\right]
\end{equation}

\noindent
Adding the regularizer, this supplies us with the following objective

\begin{equation}
\min_{\mathbf{W}} \|\W\|_{F} + C \sum_i L(\mathbf{x}_i, y_i, \mathbf{W})
\end{equation}

\noindent
At first blush, the optimization for this objective is similar to the variants of the \emph{gerrymandering} approach discussed earlier. That is, 

Each iteration $t$ of the algorithm consists of the following steps:
\begin{enumerate}
	\item Targeted inference of $h_i^*$ for each sample $\x_i$:
	\begin{equation}\label{eq:targetedregression}
	h_i^* = \argmax_{h} \left[S_{\W} (\x_i,h) - \gamma\Delta(y_i,h)\right]
	\end{equation}
	\item Loss augmented inference of $\hat{h}_i$ for each sample $\x_i$:
	\begin{equation}\label{eq:lossaugregression}
	\hat{h}_i = \argmax_{h} \left[S_{\W} (\x_i,h) + \gamma\Delta(y_i,h)\right]
	\end{equation}
	\item Gradient update for $\mathbf{W}$.
\end{enumerate}

\noindent
However, for the loss defined in \ref{eq:reghardloss}, this optimization is hard to solve exactly. Indeed, we can make the following claim:

\begin{claim}
	Given data $\X$ and corresponding labels $\y$, matrix $\W$ and query point $\x_i$, $h \subset X$ with $|h| > 1$ and $\gamma > 0$  the problem of finding $h_i^* $ and $\hat{h}_i$ is NP Hard.
\end{claim}

\noindent 
While the proof of this claim is omitted, this can be shown by an appropriate reduction to a modified version of the subset sum problem. We could resort to relaxations of the above problem, and thus work with suitable approximation algorithms to optimize for \ref{eq:targetedregression} and \ref{eq:lossaugregression}. Instead, we work with the following simple modification to our loss:

\begin{equation}\label{eq:regeasyloss}
\hat{\Delta}(y,h) = \frac{1}{k} \sum_{\x_i \in h}\left(y - y_i\right)^2
\end{equation}

\noindent
Then, we have the following modified inference problems:

\begin{equation}\label{eq:modifiedhstar}
h_i^* = \argmax_{h} \left[S_{\W} (\x_i,h) - \gamma\hat{\Delta}(y_i,h)\right]
\end{equation}

\begin{equation}\label{eq:modifiedhhat}
\hat{h}_i = \argmax_{h} \left[S_{\W} (\x_i,h) + \gamma\hat{\Delta}(y_i,h)\right]
\end{equation}

\noindent
Note that both of these problems are easy to solve: Fixing $\mathbf{W}$ and $\gamma > 0$, for each query point $(\mathbf{x},y)$, we simply need to sort the data in ascending order by the sum of their loss and similarity to the query and then pick the top and bottom $k$ points to be $h_i^*$ and $\hat{h}_i$ respectively.

\vspace{2mm}
\noindent
Before proceeding to report experiments on this approach and compare it with \cite{MLKR}, it is instructive to see what the relation between the losses \ref{eq:reghardloss} and \ref{eq:regeasyloss} is. This is made explicit in the following claim.

\begin{claim}
	$\hat{\Delta}(y,h)$ upper bounds the loss $\Delta(y,h)$
\end{claim}

\begin{proof}
	
We have: $\displaystyle \hat{\Delta}(y,h) = \frac{1}{k} \sum_{i \in h}\left(y - y_i\right)^2$ \\

\noindent Expanding: $\displaystyle \hat{\Delta}(y,h) = \frac{1}{k} \sum_{i \in h} \left(y^2 + y_i^2 - 2yy_i\right)  = \left( y^2 + \frac{1}{k^2} \sum_{i \in h} y_i^2 \sum_{i \in h} 1^2 - \frac{2y}{k} \sum_{i \in h} y_i\right)$

\noindent Now, using Cauchy Schwarz on the second term and rearranging, we have: \\
 $\displaystyle \hat{\Delta}(y,h) \ge \left( y^2 + \left( \frac{1}{k} \sum_{i \in h} y_i \cdot 1 \right)^2 - \frac{2y}{k} \sum_{i \in h} y_i\right) = \left(y - \frac{1}{k} \sum_{i \in h} y_i\right)^2$

\noindent or, $\displaystyle \hat{\Delta}(y,h) \ge \Delta(y,h)$

\end{proof}

\subsection{Alternate notions of $h^*$}

The set $h^*$ may also be defined in other ways. We consider two such definitions here. First uses the notion of an $\epsilon$-insensitive loss (similar to the notion used in Support Vector Regression \cite{SVR}):
We can define a set $\mathcal{H}$ where for a query $(\x_i,y_i)$ $$\mathcal{H}_i = \{h| \Delta(y_i,h) \leq \epsilon \}$$ Then we may define $h^*$ as:

\begin{equation}
h^*_{\epsilon i} = h^*_i = \arg\max_{h\in \mathcal{H}} \left[ S_{\W}(\x_i,h)\right]
\end{equation}

\noindent
$\epsilon$ then becomes a hyper-parameter to be found by cross-validation. 

\vspace{2mm}
\noindent
Yet another notion of $h^*$ may be defined as follows:

\begin{equation}
h^*_{Mi}  = h^*_i = \arg\min_{h} \Delta(y_i, h)
\end{equation}
Note that using this definition makes the overall objective convex.
\noindent
The inference and loss augmented procedures are straightforward for all these definitions. 

\subsection{Experiments}
For the purpose of experimental evaluation of our approach, we make a direct comparison to the experiments reported in \cite{MLKR} by working with the DELVE (Data for Evaluating Learning in Valid Experiments) datasets \footnote{Available online at \href{https://www.cs.toronto.edu/~delve/data/datasets.html}{https://www.cs.toronto.edu/~delve/data/datasets.html}}. In particular we work with 16 datasets, 8 from the Kin family and 8 from Pumadyn family of datasets. The Kin datasets were obtained from realistic simulations of a 8-link all robot arm, and consist of four datasets having 8 dimensions and four 32 dimensional datasets. Likewise, the Pumadyn datasets were obtained from the simulations of dynamics of a Puma 560 robot arm, and also consist of four 8 dimensional and four 32 dimensional datasets. The various features represent the torque, angular momentum, angular positions of the robotic arm, and the output is a real number. 

\vspace{2mm}
\noindent
Each of these datasets have 8092 points. Following the protocol in \cite{MLKR}, each dataset is split into four disjoint training sets, each of size 1024 and one unique test set, also of size 1024. For each training set we do a 5 fold cross validation for parameter tuning. In the experiments reported here we use the following naming protocol: <NAME>$-$axy. Where <NAME> is the name of the dataset. 'a' denotes an integer that signifies the dimensionality of the dataset. 'x' takes on the character values of either'f' or 'n', where 'f' denotes that the dataset is largely linear, while 'n' denotes non-linearity. 'y' takes on character values of 'm' or 'h', where 'm' denotes medium noise, while 'h' denotes high noise. Thus, for an example, the nomenclature Puma-32nh would imply we are referring to a dataset from the Pumadyn family with 32 features, in which the regression target is a non-linear function of the input and that the dataset has high noise. 

\vspace{2mm}
\noindent
For testing, we consider the following baselines: linear regression, a nearest neighbor model in which the $k$ is chosen over a range of values by cross-validation, a nearest neighbor model in which the coordinates are weighted by ReliefF~\cite{relieff1992} and the appropriate value of $k$ is chosen by cross validation, Gaussian process regression \cite{GPR}, the MLKR approach of \cite{MLKR}. All the numbers reported are relative to a baseline method that uses the mean of the training regression targets in case the squared error is minimized (for more details on the protocol and the DELVE suite, we point the reader to the manual \cite{delve}. For the approach in \cite{MLKR}, we use code provided by the authors). In all cases we minimize for the squared error. 

\vspace{2mm}
\noindent
For our approach, we work with two models: One in which the metric is learned by optimizing for $\mathbf{L}$ (such that $\mathbf{L}^T\mathbf{L} = \mathbf{W}$) using the formulation implicit through equations \ref{eq:modifiedhstar} and \ref{eq:modifiedhhat}, while initializing $\mathbf{L} = \mathbf{I}$. The second model is similar except that we use the formulation described in \ref{sec:asym}, and initialize the matrices $\mathbf{U}$ and $\mathbf{V}$ as diagonal matrices, where the diagonal carries the square root of the ReliefF coefficients after they are normalized to be between $0$ and $1$. For both these models we cross validate for the batch size (from values 1, 3, 5, 11), the parameter $\gamma$ (for values increasing logarithmically from 0.00001 to 100), and the number of neighbors $k$ (for values from 1, 3, 5, 7, 11, 21 and 31). 

\begin{figure*}
	\centering
	\begin{center}
		\includepic{.45}{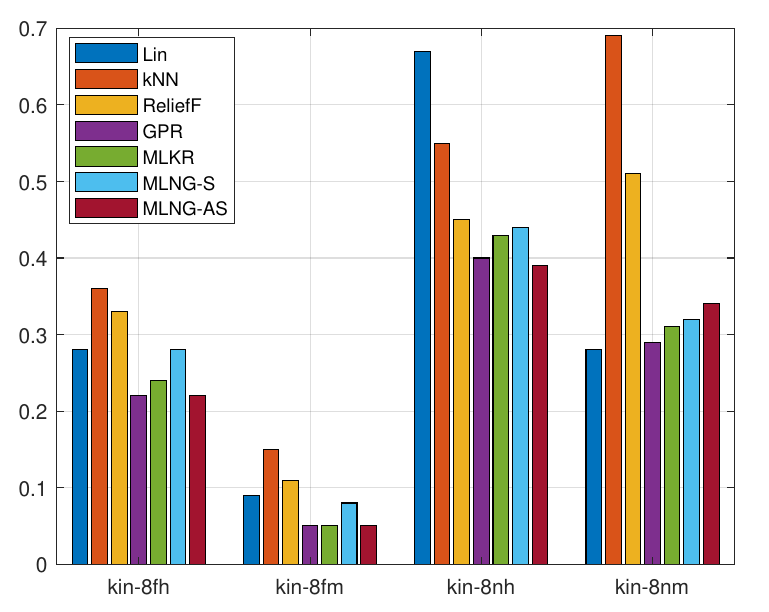}
		\includepic{.45}{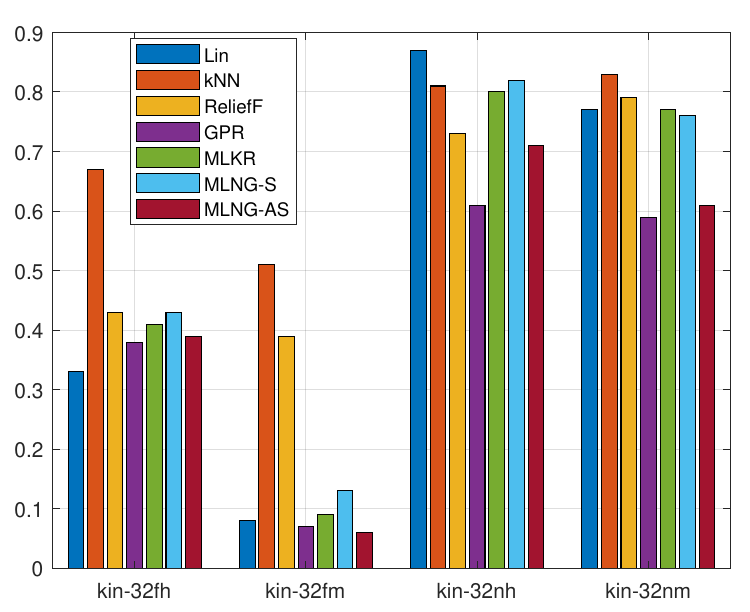}
	\end{center}
	\caption{Regression errors on the Kin datasets. (Here Lin stands for linear regression, ReliefF for $k$NN regression after feature weighing by ReliefF, GPR for Gaussian Process Regression, MLKR for Metric Learning for Kernel Regression, MLNG-S for metric learning by neighborhood gerrymandering for regression using symmetric similarity computation, while MLNG-AS refers to the same method but with an asymmetric notion of similarity)}
	\label{fig:kin} 
\end{figure*}

\begin{figure*}
	\centering
	\begin{center}
		\includepic{.45}{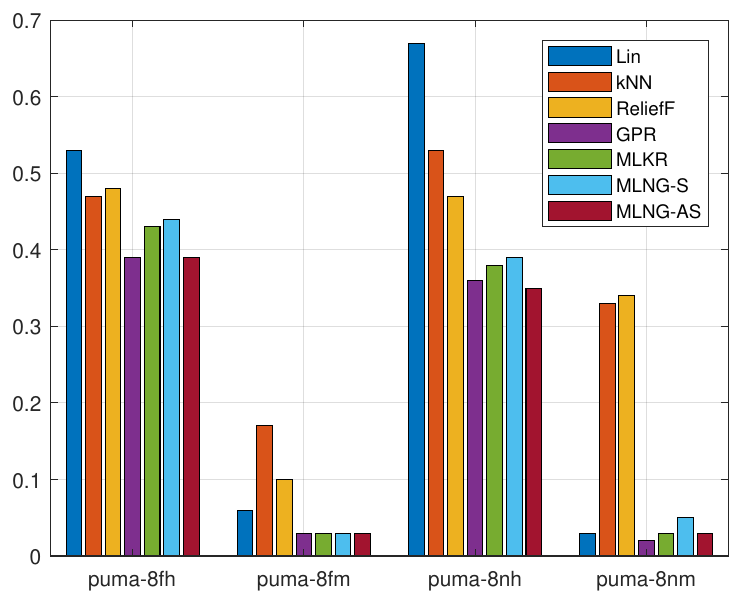}
		\includepic{.45}{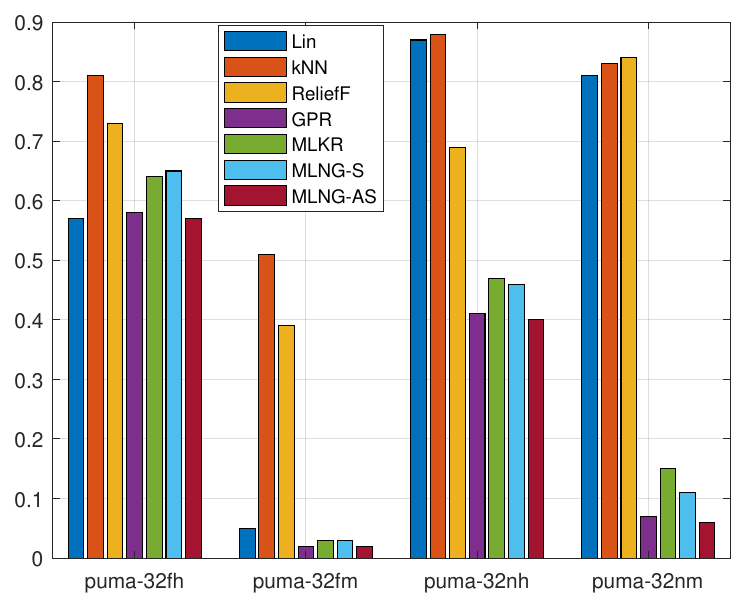}
	\end{center}
	\caption{Regression errors on the Puma datasets (Here Lin stands for linear regression, ReliefF for $k$NN regression after feature weighing by ReliefF, GPR for Gaussian Process Regression, MLKR for Metric Learning for Kernel Regression, MLNG-S for metric learning by neighborhood gerrymandering for regression using symmetric similarity computation, while MLNG-AS refers to the same method but with an asymmetric notion of similarity)}
	\label{fig:puma} 
\end{figure*}

\noindent
Figures \ref{fig:kin} and \ref{fig:puma} illustrate the results. On the 'linear' datasets (marked with 'f'), a recurring trend is that plain linear regression does quite well as compared to nearest neighbors. Thus it is interesting to see that all three metric learning methods evaluated here perform better than it. In the non-linear datasets on the other hand, all other methods other than the metric learning methods and Gaussian process regression perform poorly. The main trend that we see in this set of experiments is that Metric Learning for Kernel Regression \cite{MLKR} usually performs better than the symmetric variant of our algorithm, while being consistently worse than the asymmetric variant with ReliefF initialization. In summary, Gaussian Process Regression and the asymmetric variant of our algorithm consistently perform better, with our algorithm often performing slightly better.  

\subsection{Conclusion of the Regression Experiments}

\noindent
In summary, in this section we proposed an approach for metric learning for $k$-NN Regression. As noted earlier, literature on the topic is quite sparse, despite the fact that in many settings having a good metric while the downstream task is $k$NN regression might be desirable. In our experiments, we observed that the symmetric approach performs better than most baselines while underperforming the approach of \cite{MLKR} and Gaussian process regression, while the asymmetric approach performs at par with Gaussian Process Regression, often out-performing it, while consistently outperforming the approach of \cite{MLKR}. Our algorithm has very simple and efficient inference procedures; with the optimization for each dataset completing in a matter of a few minutes. Therefore it is perhaps worthwhile to consider the problem of metric learning in the continuous label setting in more detail, while also considering more applications where it could be useful.

%***************************************** end of section on regression ************* 

\section{Conclusion of Part I and Summary}

The conclusion of the regression experiments also brings us to the conclusion of this part of the dissertation. Below we summarize the main contributions made:

\begin{aside}{Summary of Part I}
	\begin{enumerate}
		\item[1] In Chapter \ref{ch:mlng}, we presented an approach to metric learning that is more direct in trying to optimize for the $k$-NN accuracy than methods previously proposed. The approach formulates the problem of metric learning for $k$-NN classification as a large margin structured prediction problem, with the choice of neighbors represented by discrete latent variables, making it natural to use machinery from latent structural support vector machines to the task of metric learning. We also provided exact procedures for inference and loss-augmented inference in this model, and validated the approach by comparing to a range of Mahalanobis distance metric learning methods. 
		\item[2] In this chapter we explored the formalism explicated on in \ref{ch:mlng} in different settings. In section \ref{sec:asym}, we explored an approach to similarity learning in which the similarity computation is asymmetric: the query and database points are subjected to different transformations. In section \ref{sec:asymHD} we combined the approach of Chapter \ref{ch:mlng} and Section \ref{sec:asym} to learn the Hamming distance such that it is a better proxy for $k$-NN classification performance. Lastly, in section \ref{sec:mlknnregression}, we presented an approach to metric learning for $k$-NN regression that is consistently shown to perform better than its main competitors.
	\end{enumerate}
\end{aside}

\noindent
Often, while doing $k$-NN classification and regression, we might be on a limited computational budget-- we might not be able to optimize for a metric over a space of possible metrics. However, we would still like to have access to a metric that is not completely label agnostic, but can be estimated cheaply, as performance using simply the Euclidean metric might be quite poor. Estimation of such a metric for improving $k$-NN classification and regression performance is the focus of the next part of this dissertation. 

%*****************************************
%*****************************************
%*****************************************
%*****************************************

\clearpage
%\ctparttext{Write something}
\part{Single Pass Metric Estimation}\label{pt:EGOP}
%\addtocontents{toc}{\protect\clearpage} % <--- just debug stuff, ignore
\cleardoublepage
%************************************************
\chapter{Metric Estimation via Gradients}\label{ch:egopintro}
%************************************************

In the preceding part of this dissertation we proposed and worked with a more direct approach to metric learning; direct in that it tries to optimize for the metric using a differentiable loss that aims to be a more reasonable proxy for the
underlying task: k-NN classification or regression. In this part, we take another, if somewhat indirect approach to the problem of identifying a good metric based only on gradient estimates of the unknown classification or regression function $f$. While somewhat roundabout from the perspective of the underlying task, this approach has the advantage that the metric can be estimated by a single pass over the dataset, while affording significant improvements in non-parametric regression and classification tasks. The rest of this section is used to motivate the problem, as well as to stage ground for the following two chapters which propose two variants of a gradient based metric estimator, which despite their simplicity remain statistically consistent under fairly mild assumptions.

\vspace{2mm}
\noindent
To begin, recall that in high dimensional classification and regression problems, the task is to infer the unknown function $f$. We are given a set of observations $(\mathbf{x}, \mathbf{y})_i, i = 1, 2, \dots n.$, with $\mathbf{x}_i \in \mathcal{X} \subset \mathbb{R}^d$ and the labels $\mathbf{y}_i$ are noisy versions of the function values $f(\mathbf{x}_i)$. We are interested in distance based (non-parametric) regression, which provides our function estimate: 
$$f_n (\mathbf{x}) = \sum_{i=1}^{n} w(\mathbf{x}, \mathbf{x}_i) \mathbf{y}_i$$

\noindent
$w(\mathbf{x}, \mathbf{x}_i)$ depends on the distance $\rho$. Where $\displaystyle \rho(\mathbf{x}, \mathbf{x}') = \sqrt{(\mathbf{x} - \mathbf{x}') \mathbf{W} (\mathbf{x} - \mathbf{x}') } $ and $\mathbf{W} \succeq 0$. Note that, when $\mathbf{W} = \mathbf{I}$, $\rho$ is the Euclidean distance.

\vspace{2mm}
\noindent
The problem of estimating unknown $f$ becomes significantly harder as $d$ increases due to the \emph{curse of dimensionality}. To remedy this situation, several pre-processing techniques are used, each of which rely on a suitable assumption about the data and/or about $f$. For instance, a conceptually simple, yet often reasonable assumption that can be made is that $f$ might not vary equally along all coordinates of $\mathbf{x}$. Letting $f'_i = \nabla f^T e_i $ denote the derivative along coordinate $i$, and $\|f'_i\|_{1,\mu} \equiv \mathbb{E}_{\mathbf{x} \sim \mu} f'_i(\mathbf{x})$, we can use the above distance based estimator by setting $\rho$ such that
\[   
\mathbf{W}_{i,j} = 
\begin{cases}
\|f'_i\|_{1,\mu} &\quad\text{ if } i = j\\
0 &\quad\text{ if } i \neq j\\
\end{cases}
\]

\begin{figure*}[!th]
	\centering
	\resizebox{.55\textwidth}{!}{\begin{tikzpicture}
\filldraw[color=black!60, fill=black!5, very thick](-3,0) circle (2.125);
\draw[thick, ->] (-3,0) -- (-3,2.01)node[midway, left] {$e_1$};
\draw[thick, ->] (-3,0) -- (-0.99,0)node[midway, below] {$e_2$};

\filldraw[color=black!60, fill=black!5, very thick](3.5,0) ellipse (3 and 1.5);
\draw[thick, ->] (3.5,0) -- (3.5,1.4)node[midway, left] {$e_1$};
\draw[thick, ->] (3.5,0) -- (6.45,0)node[midway, below] {$e_2$};

\end{tikzpicture}}
	\caption{Left: Euclidean ball $\mathcal{B}$ which assigns equal importance to both directions $e_1$ and $e_2$. Right: ball $\mathcal{B}_{\rho}$ such that $\|f'_1\|_{1,\mu} \gg \|f'_2\|_{1,\mu}$, giving its ellipsoidal shape. Relative to the $\mathcal{B}$; $\mathcal{B}_{\rho}$ will have more mass in direction $e_2$}
	\label{fig:GWrescale}
\end{figure*}
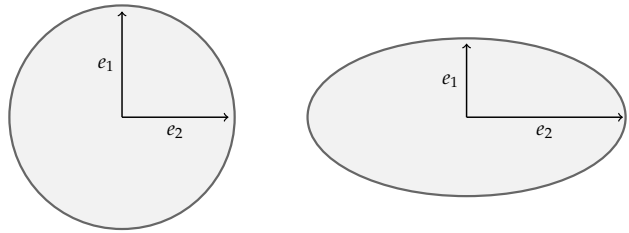

\noindent
This \emph{gradient weighting} rescales the space such the ball $\mathcal{B}_{\rho}$ contains more points relative to the Euclidean ball $\mathcal{B}$ (figure \ref{fig:GWrescale} for an example in $\mathbb{R}^2$). This is the intuition pursued in works such as \cite{DBLP:conf/nips/KpotufeB12}, \cite{GWJMLR} (which are also the inspiration for what follows in this part of the dissertations), with an emphasis on deriving an efficient, yet consistent estimator for the gradient. Using gradient weights for coordinate scaling in this manner has strong theoretical grounding as is shown in these works, in that it has the effect of reducing the regression variance, while keeping the bias in control. 

\begin{figure*}[!th]
	\centering
	\resizebox{.8\textwidth}{!}{\begin{tikzpicture}
%\filldraw[color=black!60, fill=black!5, very thick](-3,0) circle (2.125);
%\draw[thick, ->] (-3,0) -- (-3,2.01)node[midway, left] {$e_1$};
%\draw[thick, ->] (-3,0) -- (-0.99,0)node[midway, below] {$e_2$};

\hspace{15mm}

\filldraw[color=black!60, fill=black!5, very thick](3.5,0) ellipse (3 and 1.5);
\draw[thick, ->] (3.5,0) -- (3.5,1.4)node[midway, left] {$e_1$};
\draw[thick, ->] (3.5,0) -- (6.45,0)node[midway, below] {$e_2$};

\hspace{-25mm}

\filldraw[rotate=45,color=black!60, fill=black!5, very thick](10,-10) ellipse (3 and 1.5);
\draw[rotate=45,thick, ->] (10,-10) -- (10,-8.6)node[midway, left] {$e_1$};
\draw[rotate=45,thick, ->] (10,-10) -- (12.95,-10)node[midway, below] {$e_2$};

\end{tikzpicture}}
	\caption{Left: Rescaled ball $\mathcal{B}_{\rho}$ which is still axis aligned but rescales the coordinates. Right: Ball that not only rescales the data but also rotates it}
	\label{fig:GWrescale2}
\end{figure*}
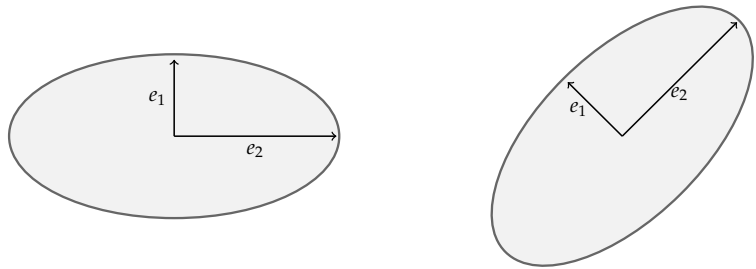

\vspace{2mm}
\noindent 
While appealing in its simplicity, gradient weighing has an obvious drawback: the metric only involves diagonal $\mathbf{W}$. In general $\mathbf{W} \succeq 0 $, need not be diagonal and may be decomposed as $\mathbf{W} = \mathbf{V} \Sigma \mathbf{U}^T$ where $\mathbf{U}, \mathbf{V}$ are orthogonal matrices. In such a case the data would not only be rescaled but also rotated. This is illustrated in figure \ref{fig:GWrescale2}.

\vspace{2mm}
\noindent
Using the above motivation to construct a covariance type matrix but only using gradients via an iterative algorithm (which involved taking the outer product of the gradients $\nabla f(X) \cdot \nabla f(X)^\top$ and summing over all the points) was used by us to derive an operator, with the following property: If the function $f$ does not vary along some direction $\mathbf{v} \in \mathbb{R}^d$, then $\mathbf{v}$, must lie in the nullspace of the operator. But we later discovered that this operator was already known in the literature in a different context. To define things clearly, put it in proper context and also outline our contributions, we first take a step back and consider the motivation that we mentioned earlier. The unknown classification or regression function $f$ might not vary equally in all coordinates. We used this fact to review the approach of \cite{DBLP:conf/nips/KpotufeB12}, \cite{GWJMLR} above. 

\vspace{2mm}
\noindent
This simple observation is also motivation for a plethora of variable selection methods. In variable selection, the assumption that is used is that for $f$, we have $f(\x) = g(P\x)$,  where $P\in \braces{0, 1}^{k \times d}$ projects $X$ down to $k<d$ coordinates that are most relevant to predicting the output $y$. This assumption is generalized further in \emph{multi-index regression} e.g.\cite{li1991sliced,powell1989semiparametric, hardle1993optimal, xia2002adaptive}). This is done by letting $P\in \real^{k \times d}$ project $\x$ down to a $k$-dimensional subspace of $\real^d$. Put differently, this is a generalization because here it is assumed that while $f$ might vary along all coordinates, it actually only depends on an unknown $k$-dimensional subspace. Such a subspace is called a \emph{relevant} subspace. The task then becomes finding the said relevant subspace rather than chopping coordinates since they all might be relevant in predicting the output $y$.

\vspace{2mm}
\noindent
Work to recover this relevant subspace (which is sometimes also referred to in the literature as \emph{effective dimension reduction} \cite{li1991sliced}) gives rise to the expected gradient outerproduct (EGOP):
$$\expectation_{\mathbf{x}} G(\mathbf{x}) \triangleq \expectation_\mathbf{x}\paren{\nabla f(\mathbf{x}) \cdot \nabla f(\mathbf{x})^\top} .$$

\vspace{2mm}
\noindent
This operator (which superficially seems similar to the Fisher information matrix) is useful beyond the multi-index motivation mentioned above. That is, even when there is no clearly relevant dimension-reduction $P$, as is usually likely in practice, one might still expect that $f$ does not vary equally in all directions. Therefore, beyond the use of EGOP for dimension-reduction, we might use it instead to weight any direction $v\in \real^d$ according to its relevance as captured by the average variation of $f$ along $v$ (encoded in the EGOP). The weighting approach will be the main use of EGOP considered in this work. That is, we use the EGOP in the following way: let $VDV^\top$ be a spectral decomposition of the estimated EGOP, we use it to transform the input $\mathbf{x}$ as $D^{1/2}V^\top \mathbf{x}$. Also, for constructing the EGOP we need to compute gradient estimates. Just as in the case of gradient estimation, optimal estimators of the EGOP can be expensive in practice. In this part of the thesis we also show that a simple, efficient difference based estimator suffices in that it remains statistically consistent under mild assumptions. 

\vspace{2mm}
\noindent
It is important to note that estimating the EGOP and using it to transform the inputs as $\mathbf{x} \mapsto D^{1/2}V^\top \mathbf{x}$ and then using it for $k$-classification and regression does not involve any \emph{learning}. Thus this approach is related to but distinct from metric learning in that a metric is not optimized for over a space of possible metrics parametrized by positive semi-definite matrices. This approach is also online and cheap: we only require $2d$ estimates of the function $f$ at $\mathbf{x}$, and can also be used for preprocessing for standard metric learning methods. Work on the EGOP is explicated upon in Chapter \ref{ch:EGOP}. 

\vspace{2mm}
\noindent
As will be described later, the EGOP can be used for metric weighing in the setting where $f: \mathbb{R}^d \to \mathbb{R}$ and thus only in the case of regression and binary classification. For the multi-class case, we could treat it as a multinomial regression problem, where the unknown function $f: \mathbb{R}^d \to \mathbb{S}^c$ where $\mathbb{S}^c = \{ \mathbf{y} \in \mathbb{R}^c | \mathbf{y} \geq 0, \mathbf{y}^T \mathbf{1} = 1 \}$. This leads to a similar operator based on computing the Jacobian of this vector valued function, which we call the Expected Jacobian Outer Product (EJOP). 
$$\mathbb{E}_{\mathbf{x}} G(\mathbf{x}) \triangleq  \expectation_\mathbf{x}\paren{\mathbf{J}_{f}(\mathbf{x}) \mathbf{J}_{f}(\mathbf{x})^T}$$
\noindent
We describe metric weighing experiments for non-parametric classification and also show that a simple estimator for the EJOP remains statistically consistent under mild assumptions in Chapter \ref{ch:EJOP}. 

\begin{aside}{}
	To summarize, in this part of the thesis, we make the following contributions:
	\begin{enumerate}
		\item[1] We describe a simple estimator for the Expected Gradient Outerproduct (EGOP) 
		$$\expectation_{\mathbf{x}} G(\mathbf{x}) \triangleq \expectation_\mathbf{x}\paren{\nabla f(\mathbf{x}) \cdot \nabla f(\mathbf{x})^\top} .$$  and show that it remains statistically consistent under mild assumptions.
		\item[2] We use the EGOP (with a spectral decomposition $VDV^\top$) in non-parameteric regression by using it to transform the inputs as $\mathbf{x}$ as $D^{1/2}V^\top \mathbf{x}$ and show it improves performance in several real world datasets.
		\item[3] We extend the EGOP to the multiclass case, proposing a variant called the Expected Jacobian Outer Product (EJOP)
		$$\mathbb{E}_{\mathbf{x}} G(\mathbf{x}) \triangleq  \expectation_\mathbf{x}\paren{\mathbf{J}_{f}(\mathbf{x}) \mathbf{J}_{f}(\mathbf{x})^T}$$
		\item[4] For the EJOP, we propose a simple estimator and also prove that it remains statistically consistent under reasonable assumptions.
		\item[5] Similarly to the case of the EGOP, we use the EJOP for transforming the input space and also demonstrate that it improves performance in various non-parametric classification tasks. 
	\end{enumerate}
	
\end{aside}

%*****************************************
%*****************************************
%*****************************************
%*****************************************
%*****************************************

\cleardoublepage
%************************************************
\chapter{The Expected Gradient Outer Product}\label{ch:EGOP} % $\mathbb{ZNR}$
%************************************************
In high dimensional classification and regression problems, the task is to infer the unknown, smooth function $f:\mathcal{X} \to \mathcal{Y}$. To this end, we are provided $n$ of functional estimates that comprises our data. In other words, we have $\{ (\mathbf{x}, \mathbf{y})_1, (\mathbf{x}, \mathbf{y})_2, \dots, (\mathbf{x}, \mathbf{y})_n  \}$, with $\mathbf{x}_i \in \mathcal{X} \subset \mathbb{R}^d$ and the labels $\mathbf{y}_i \approx f(\mathbf{x}_i)$ i.e. they are noisy versions of the function values. We are interested in distance based regression, which provides our function estimate: 
$$f_n (\mathbf{x}) = \sum_{i=1}^{n} w(\mathbf{x}, \mathbf{x}_i) \mathbf{y}_i$$

\vspace{2mm}
\noindent
The weights $w(\mathbf{x}, \mathbf{x}_i)$, depend on the underlying metric. In Chapter \ref{ch:egopintro} we made the case for metric estimation from gradients in case of a restricted computational budget, where optimizing over a space of metrics might not be feasible. In particular, in the case of regression and binary classification, we consider the metric given by the Expected Gradient Outerproduct, which is written as. 
$$\expectation_{\mathbf{x}} G(\mathbf{x}) \triangleq \expectation_{\mathbf{x}}\paren{\nabla f(\mathbf{x}) \cdot \nabla f(\mathbf{x})^\top} .$$

\vspace{2mm}
\noindent
Originally proposed in the context of multi-index regression, the EGOP recovers the average variation of $f$ in all directions. To see this: For some $\mathbf{v}\in \real^d$, the directional derivative at $\mathbf{x}$ along $\mathbf{v}$
is given by $f_{\mathbf{v}}'(\mathbf{x}) = \nabla f(\mathbf{x})^\top \mathbf{v}$, in other words
$$ \expec_X \abs{f_{\mathbf{v}}'(\mathbf{x})}^2 = \expec_{\mathbf{x}} \paren{\mathbf{v}^\top G(\mathbf{x}) \mathbf{v}}=\mathbf{v}^\top \paren{\expec_{\mathbf{x}} G(\mathbf{x})} \mathbf{v}$$

\noindent
From the above, it follows that, if $f$ does not vary along $\mathbf{v}$, $\mathbf{v}$ must be in the null-space of the EGOP matrix $\expec_X G(X)$, since $\expec_X \abs{f_v'(X)}^2 =0$. \cite{Mukherjee_wu2010learning} infact show that considering $f$ is continuously differentiable on a compact space $\Xx$, the column space of $\expec_X G(X)$ is exactly the \emph{relevant} subspace defined by $P$ (recall that $P$ is the relevant subspace defined in Chapter \ref{ch:egopintro} i.e. $f(\x) = g(P\x)$ with $P\in \real^{k \times d}$, with $P$ being the subspace most relevant to predicting the output $y$)

\vspace{2mm}
\noindent
As already discussed in Chapter \ref{ch:egopintro}, the EGOP is useful beyond the multi-index setting, where its utility is to recover the relevant subspace $P$. That is, we might expect that in most practical, real world settings, a clear relevant subspace might not exist. Nevertheless, we can still expect that $f$ does not vary uniformly in all directions. The usefulness of the EGOP in such settings can be to weight any direction $\mathbf{v}\in \real^d$ according to its relevance as captured by the average variation of $f$ along $\mathbf{v}$
(encoded in the EGOP). It is this weighting use of the EGOP that we will consider in this chapter. 

\vspace{2mm}
\noindent
The estimation of the EGOP can be done in various sophisticated ways, which can however be prohibitively expensive. 
For instance an optimal way of estimating $\nabla f(x)$, and hence the EGOP, is to estimate the slope of a linear approximation to $f$ locally at each $x=X_i$ in an $n$-sample $\braces{(X_i, Y_i)}_{i=1}^n$. Local linear fits can however be prohibitively expensive since it involves multiplying and inverting large-dimensional matrices at all $X_i$.  This can render the approach impractical although it is otherwise well motivated.

\vspace{2mm}
\noindent
One of the main messages of the work discussed in this chapter is that the EGOP does need to be estimated optimally for the utility of it that we discussed above. That is, it just needs to be estimated well enough to use towards improving classification and regression. To this end, we consider the following cheap, albeit very rough estimator. Let $f_n$ denote an initial estimate of $f$ (we use a kernel estimate); for the $i$-th coordinate of
$\nabla f(x)$, we use the rough estimate
$$\Delta_{t,i} f_n(x) = \frac{(f_n(x+t e_i) - f_n(x-t e_i))}{2t},\, t>0.$$
Now, let $G_n(x)$ be the outer-product of the resulting gradient estimate $\hat{\nabla}f_n(x)$, the EGOP is estimated as $\expec_n G_n(X)$, the empirical average of $G_n$. The exact procedure is given in Section \ref{sec:estimate}.

\begin{figure*}[!th]
	\centering
	\includegraphics[width=.5\linewidth]{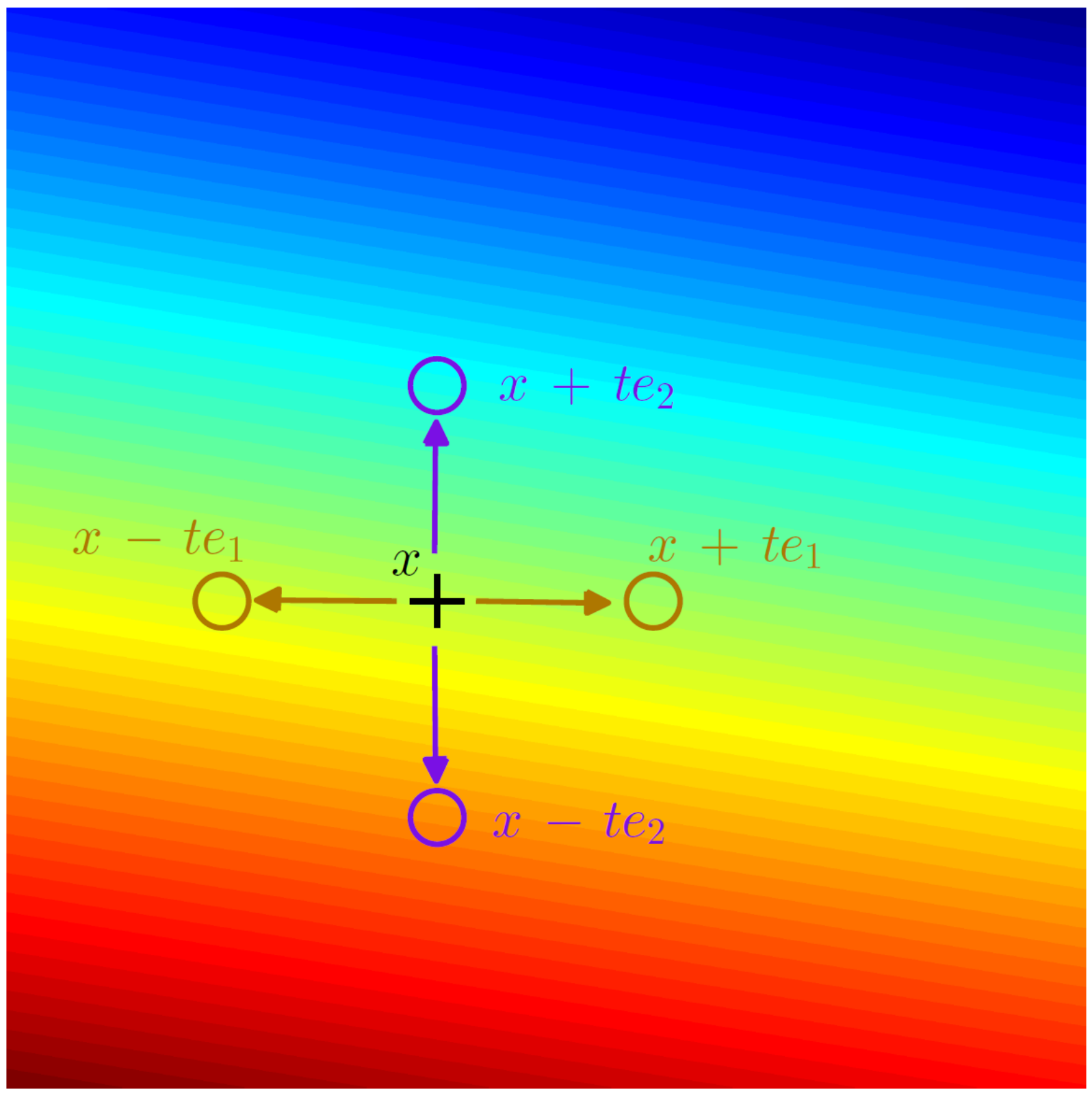}
	\caption{A simple illustration of the difference based gradient estimator when $\mathbf{x} \in \mathbb{R}^2$. We perturb the input along each coordinate, record the value of $f_{n,h} (\mathbf{x})$, and get a finite difference estimate. The background color represents the functional values}
	\label{fig:gradient_estimator}
\end{figure*}

\noindent
We must first demonstrate that this rough estimator is, in fact, a sound estimator. To this end, we show it remains a statistically consistent estimate of the EGOP under very general distributional conditions. These assumptions being milder than the usual conditions on proper gradient estimation (for detailed assumptions see Section \ref{sec:Assumptions}). The main consistency result and key difficulties (having to do with interdependencies in the estimate) are discussed in Section \ref{sec:consistency}.

\vspace{2mm}
\noindent
We also show, through extensive experiments that preprocessing the data with this cheap EGOP estimate can still significantly improve the performance of non-parametric classification and regression procedures in many real-world datasets. The experimentation is described in Section \ref{sec:experiments}. In the next Section \ref{sec:relevant}, we first give a quick overview of some of the relevant work and place ours in context. 

\section{Related Work}
\label{sec:relevant}

The work of Kpotufe \emph{et al.} \cite{DBLP:conf/nips/KpotufeB12}, \cite{GWJMLR} already briefly discussed in Chapter \ref{ch:egopintro} was the direct inspiration for the work described in this chapter. Kpotufe \emph{et al.} consider estimating the coordinates $f_i'$ of $\nabla f$ in a similar fashion as described here. However, there is a notable difference, in that \cite{DBLP:conf/nips/KpotufeB12}, \cite{GWJMLR} are only concerned with a variable selection setting. That is, each coordinate $i$ of $X$ is to be weighted by an estimate of $\expec_X \abs{f_i' (X)}$,
which is their quantity of interest. In this chapter we consider the more general approach of estimating the Expected Gradient Outerproduct. We also study its consistency and applicability in the context of non-parametric classification and regression.

\vspace{2mm}
\noindent
Another body of literature, which is quite closely related to the work described in this chapter, is in the context of methods for multi-index regression. Many methods developed for doing multi-index regression use the so-called \emph{inverse regression} approach (e.g. \cite{li1991sliced}), and many of them operate by incorporating estimates of derivate functionals of the unknown $f$. These approaches can be found in works as early as \cite{powell1989semiparametric}, and typically estimate $\nabla f$ as the slope of local linear approximations of $f$.

\vspace{2mm}
\noindent
Comparatively recent works of \cite{Mukherjee_wu2010learning, mukherjee2010learning} build a much clearer bridge between various approaches to multi-index regression. In particular, they also related the EGOP to the \emph{covariance}-type matrices typically estimated in inverse regression. Besides, \cite{Mukherjee_wu2010learning, mukherjee2010learning} also propose an alternative estimator for the gradient, rather than using local linear slopes. Their approach estimates $\nabla f$ via a regularized least-squares objective over an Reproduced Kernel Hibert Space. This approach is still expensive, since the least-square solution involves inverting an $n\times n$ feature matrix. In contrast our less sophisticated approach will take time in the order of $n$ times the time to estimate $f_n$ (in practice, we could employ fast range search methods when $f_n$ is a fast kernel regressor).

\vspace{2mm}
\noindent
As already described, the primary utility of the EGOP in multi-index regression is to recover the \emph{relevant} subspace given by $P$ in the model $f(\mathbf{x})=g(P \mathbf{x})$. The data can first be projected to this subspace before doing predicting on the projected data. 

\vspace{2mm}
\noindent
In this chapter, we do not make a case for any particular methodology that leverages the EGOP for preprocessing the data. Instead, our experiments focus on the use of EGOP as a metric for distance based non-parametric regression and classification. That is, suppose $VDV^\top$ is the spectral decomposition of the estimated EGOP, we then use this to transform the input as follows $\mathbf{x} \mapsto D^{1/2}V^\top \mathbf{x}$. Our use of the EGOP does not rely on the multi-index model holding, but rather on a more general model where $P$ might be a full-dimensional rotation (i.e. all directions are relevant), but $g$ varies more in some coordinate than in others. The diagonal element $D_{i,i}$ recovers $\expec_{\mathbf{x}} (g_i'(\mathbf{x}))^2$ where $g_i'$ denotes coordinate $i$ of $\nabla g$,  while
$V^\top$ recovers $P$.

\section{Setup and Definitions}
\label{sec:egopsetup}
We consider a regression or classification setting where the input $X$ belongs to a space $\Xx\subset \real^d$,
of bounded diameter $1$. The output $Y$ is real. We are interested in the unknown \emph{regression function}
$f(x)\triangleq \expectation [Y| X= x]$ (in the case of classification with $Y\in \braces{0, 1}$, this is just the
probability of $1$ given $x$).

\vspace{2mm}
\noindent
For a vector $x\in \real^d$, let $\norm{x}$ denote the Euclidean norm,
while for a matrix $A$, let $\norm{A}_2$ denote the spectral norm, i.e. the largest singular value $\sigma_{\max}(A)$.

\vspace{2mm}
\noindent
We use $\text{im}(A)$ to denote the column space of matrix
$A \in R^{n \times m}$: $\text{im}(A) = \{\y \in \R^n :\y = A\x \text{ for some } \x \in \R^m\}$, and the $\text{ker}(A)$
to denote the null space of matrix $A \in R^{n \times m}$: $\text{ker}(A) = \{\x \in \R^m : A\x = 0 \}$.
We use $A \circ B$ to denote the entry-wise product of matrices $A$ and $B$.

\vspace{2mm}
\noindent
As a little aside, we use both $\mathbf{x}$ and $X$ to refer to $d$ dimensional vectors ($\mathbf{X}$ to a matrix) in this chapter henceforth as well as the next chapter. The purpose of latching onto this notational freedom will be clear from the proofs, where working with $\mathbf{x}$ can cause confusion. 

\subsection{Estimation procedure for the Expected Gradient Outerproduct}
\label{sec:estimate}
We let $\mu$ denote the marginal of $P_{X, Y}$ on $\Xx$ and we let $\mu_n$
denote its empirical counterpart on a random sample $\Xspl = \braces{X_i}_{i=1}^n$.
Given a labeled sample $(\Xspl, \Yspl)= \braces{(X_i, Y_i)}_{1}^n$
from $P_{X, Y}^n$, we estimate the EGOP as follows.

\vspace{2mm}
\noindent
We consider a simple kernel estimator defined below, using a Kernel $K$ satisfying the following admissibility conditions:
\begin{definition}[Admissible Kernel]
	$K:\real_+\mapsto\real_+$ is nonincreasing, $K>0$ on $[0, 1)$, and $K(1)=0$.
\end{definition}

\noindent
Using such an admissible kernel $K$, and a bandwidth $h>0$,
we consider the regression estimate $f_{n,h}(\x) = \sum_{i} \omega_i(\x) Y_i$ where
\begin{align*}
\omega_i(x) &= \frac{K(\norm{x-X_i}/h)}{\sum_j K(\norm{x-X_j}/h)}\text{ if } B(x, h)\cap \Xspl \neq \emptyset,\\
\omega_i(x) &= \frac{1}{n} \text{ otherwise} .
\end{align*}

%$\tilde{f}_{n,h}(\x)$ as $\tilde{f}_{n,h}(\x) = \sum_{i} \omega_i(\x) f(\x_i)$;
\noindent
For any dimension $i\in[d]$, and $t>0$,  we first define
$$\Delta_{t,i}f_{n,h}(\x) \triangleq \frac{f_{n,h}(\x + t e_i) - f_{n,h}(\x - t e_i)}{2t}.$$
This is a rough estimate of the line-derivative along coordinate $i$. However, for a robust estimate we also need to ensure that enough sample
points contribute to the estimate. To this end, given a confidence parameter $0<\delta<1$
(this definiton for $\delta$ is assumed in the rest of this work), define $A_{n,i}(X)$ as the event that
$$\min_{s\in \{t,-t\}} \mu_n(B(X+s e_i, h/2)) \geq \frac{2d \ln 2n + \ln(4/\delta)}{n}.$$

\noindent
The gradient estimate is then given by the vector
$$\hat{\nabla} f_{n,h}(\x) = \paren{\Delta_{t,i}f_{n,h}(\x) \cdot \ind{A_{n,i} (\x)}}_{i\in [d]}.$$
Note that, in practice we can just replace $A_{n, i}(X)$ with the event that the balls
$B(X+se_i, h), s\in \braces{-t, t}$, contain samples.

\noindent
Finally, define $G_n(x)$ as the outer-product of $\hat{\nabla} f_{n,h}(\x)$,
we estimate $\expec_X G(X)$ as
$$\expec_n G_n(X) \triangleq \frac{1}{n}\sum_{i=1}^n  \hat{\nabla} f_{n,h}(X_i) \cdot \hat{\nabla} f_{n,h}(X_i)^\top.$$

\subsection{Distributional Quantities and Assumptions}
\label{sec:Assumptions}
For the analysis, our assumptions are quite general.
In fact we could simply assume, as is common, that $\mu$ has lower-bounded density on
a compact support $\Xx$, and that $f$ is continuously differentiable; all the assumptions below will then hold.
We list these more general detailed assumptions to better understand the minimal distributional requirements for consistency
of our EGOP estimator.

\begin{assumption}[Noise]
	Let $\eta(X) \triangleq Y-f(X)$.
	We assume the following general noise model:
	$\forall \delta>0 \text{ there exists } c>0 \text{ such that } \sup_{x\in \X} \prf{Y | X= x}{\abs{\eta(x)}>c}\leq \delta.$
	We denote by $C_Y(\delta)$ the infimum over all such $c$.  For instance, suppose $\eta(X)$ has exponentially decreasing tail, then
	$\forall \delta>0$, $C_Y(\delta)\leq O(\ln 1/\delta)$.
	
	Last the variance of $(Y|X=x)$ is upper-bounded by a constant $\sigma^2_Y$ uniformly over $x\in \X$.
\end{assumption}
The next assumption is standard for nonparametric regression/classification.

\noindent
\begin{assumption}[Bounded Gradient]
	\label{ass:boundongrad}
	Define the $\tau$-\emph{envelope} of $\Xx$ as $\Xx+B(0, \tau) \triangleq \braces{z\in B(x, \tau), x\in \Xx}$. We assume there exists
	$\tau$ such that $f$ is continuously differentiable on the $\tau$-envelope $\Xx+B(0, \tau)$. Furthermore, for all
	$x\in \Xx+B(0, \tau)$, we have $\norm{\nabla f(x)}\leq R$ for some $R>0$, and $\nabla f$ is uniformly continuous on $\Xx+B(0, \tau)$ (this is automatically the case if the support $\Xx$ is compact).
\end{assumption}

\vspace{2mm}
\noindent
The next assumption generalizes common smoothness assumptions: it is typically required for gradient estimation that the
gradient itself be H\"{o}lder continuous (or that $f$ be second-order smooth). These usual assumptions imply
the more general assumptions below.

\begin{assumption}[Modulus of continuity of $\nabla f$]
	Let $\epsilon_{t,i} = \sup_{\x \in \Xx, s \in [-t,t]} |f'_i(\x) - f'_i(\x + s e_i)|$. We assume $\epsilon_{t,i} \xrightarrow{t \rightarrow 0} 0$ which is for instance the case when $\nabla f$ is uniformly continuous on an envelope $\Xx + B(0, \tau)$.
\end{assumption}

\noindent
The next two assumptions capture some needed regularity conditions on the marginal $\mu$.
To enable local approximations of $\nabla f(x)$ over $\Xx$, the marginal $\mu$ should not concentrate on
the boundary of $\Xx$. This is captured in the following assumption.

\begin{assumption}[Boundary of $\Xx$]
	\label{ass:boundary}
	Define the {\bf $(t,i)$-boundary} of $\mathcal{X}$ as $\partial_{t,i}(\Xx) = \{\x: \{\x+ t e_i, x- te_i\} \not\subseteq \Xx\}$.
	Define the vector $\mu_{\partial_t} = \paren{\mu(\delta_{t, i}(\Xx))}_{i\in [d]}$. We assume that
	$\mu_{\partial_t}\xrightarrow{t\to 0}{\mathbf 0}$. This is for instance the case if $\mu$ has a continuous density on $\Xx$.
\end{assumption}

\noindent
Finally we assume that $\mu$ has mass everywhere, so that for samples $X$ in dense regions,  $X\pm te_i$ is also likely to be in a dense region.

\begin{assumption}[Full-dimensionality of $\mu$]
	For all $x\in \Xx$ and $ h>0$,  we have $\mu(B(x, h))\geq C_\mu h^{d}$. This is for instance the case if $\mu$ has a lower-bounded density on $\Xx$.
\end{assumption}

\section{Consistency of the Estimator \texorpdfstring{$\e_n G_n(X)$}{ } %
	 of \texorpdfstring{$\e_X G(X)$} %
	 { }}
\label{sec:consistency}
We establish consistency by bounding $\| \e_n G_n(X) - \e_X G(X) \|_2$ for finite sample size $n$.
The main technical difficulties in establishing the main result below have to do with the fact that each
gradient approximation $\Delta_{t, h} f_{n, h}(X)$ at a sample point $X$ depends on all other samples in
$\Xspl$. These inter-dependencies are circumvented by proceeding in steps
which consider related quantities that are less sample-dependent.

\begin{aside}{}
\begin{theorem}[Main]
	%Recall the envelope parameter $\tau$ of assumption A\ref{ass:boundongrad}, and the gradient bound $R$.
	Assume A1, A2 and A5.
	Let $t< \tau$ and suppose  $h \geq (\log^2(n/\delta)/n)^{1/d}$. There exist $C = C(\mu,K(\cdot))$ and $N = N(\mu)$ such that the following holds with probability at least $1 - 2\delta$. Define $A(n) = \sqrt{Cd \cdot \log(n/\delta)} \cdot C_Y^2(\delta/2n) \cdot \sigma_Y^2/\log^2(n/\delta)$.
	Suppose $n \geq N$, we have:
	\begin{align*}
	&\| \e_n G_n(X)] - \e_X G(X) \|_2 \leq \frac{6 R^2}{\sqrt{n}}\paren{\sqrt{\ln d} + \sqrt{\ln \frac{1}{\delta}}} + \\
	&\, \left(3R + \norm{\epsilon_t}   + \sqrt{d}\paren{\frac{hR + C_Y(\delta/n)}{t}} \right) \cdot
	\left[ \norm{\epsilon_t} + \vphantom{\sqrt{\frac{A(n)}{nh^d} + 2 h^2 R^2}} \right.\\
	& \left. \frac{\sqrt{d}}{t} \sqrt{\frac{A(n)}{nh^d} + 2 h^2 R^2} + R \left( \sqrt{\frac{d \ln \frac{d}{\delta}}{2n}} + \norm{\mu_{\partial_t}} \right) \right]
	\end{align*}
	\label{theorem:main}
\end{theorem}
\end{aside}

\begin{proof}
	Start with the decomposition
	\begin{align}
	\| \e_n G_n(X) - \e_X G(X) \|_2 \leq& \| \e_n G(X) - \e_X G(X) \|_2 \nonumber\\
	+& \| \e_n G_n(X) - \e_n G(X) \|_2. \label{eq:decomp}
	\end{align}
	The two terms of the r.h.s. are bounded separately in Lemma \ref{lemma:EGOPrm} and \ref{lem:bound_decomp0}.
\end{proof}

\noindent
{\it Remark.}
Under the additional assumptions A3 and A4, the theorem implies consistency for $t \xrightarrow{n \rightarrow \infty} 0$, $h \xrightarrow{n \rightarrow \infty} 0$, $h/t^2 \xrightarrow{n \rightarrow \infty} 0$, and $(n / \log n)h^d t^4 \xrightarrow{n \rightarrow \infty} \infty$, this is satisfied for many settings, for example $t \propto h^{1/4}$, $h \propto (1/n)^{1/(2(d+1))}$.

\vspace{2mm}
\noindent
The bound on the first term of (\ref{eq:decomp}) is a direct result of the below concentration bound for random matrices:
\begin{lemma}
	\cite{randommatrix,kakadenotes}.
	Consider a random matrix $A \in \R^{d \times d}$ with bounded spectral norm $\norm{A}_2 \leq M$.
	Let $A_1,A_2,...,A_n$ be i.i.d. copies of $A$.  With probability at least $1 - \delta$, we have
	\begin{align*}
	\norm{\frac{1}{n} \sum_{i=1}^n A_i - \e A }_2 \leq \frac{6 M}{\sqrt{n}}\paren{\sqrt{\ln d} + \sqrt{\ln \frac{1}{\delta}}}.
	\end{align*}
\end{lemma}

\noindent
We apply the above concentration to the i.i.d. matrices $G(X), X\in \Xspl$,  using the fact that
$\| G(X) \|_2 = \| \nabla f(X) \|^2 \leq R^2$.
\begin{lemma}
	Assume A2. With probability at least $1 - \delta$ over the i.i.d sample $\Xspl\triangleq \braces{X_i}_{i=1}^n$, we have
	\begin{align*}
	\| \e_n G(X) - \e_X G(X) \|_2 \leq \frac{6 R^2}{\sqrt{n}}\paren{\sqrt{\ln d} + \sqrt{\ln \frac{1}{\delta}}}.
	\end{align*}
	\label{lemma:EGOPrm}
\end{lemma}

\noindent
The next Lemma provides an initial bound on the second term of (\ref{eq:decomp}).
\begin{lemma}
	Fix the sample $(\Xspl, \Yspl)$. We have:
	\begin{align}
	\| \e_n G_n(X)  - \e_n G(X) \|_2 \leq&
	\e_n {\|\nabla f(X) - \hat{\nabla} f_{n,h}(X)\|} \nonumber \\
	& \cdot  \max_{x\in \Xspl}\|\nabla f(x) + \hat{\nabla} f_{n,h}(x)\| .
	\label{eq:redux}
	%\left(3R + \norm{\epsilon_t}_2   + \sqrt{d}(\frac{hR + C_Y(\delta)}{t})\right) \e_n \|\nabla f(X) - \hat{\nabla} f_{n,h}(X)\|_2
	\end{align}
	\label{lemma:EGOPmatrix}
\end{lemma}
\begin{proof}
	We have by a triangle inequality $\| \e_n G_n(X)  - \e_n G(X) \|_2$ is bounded by:
	\begin{align*}
	\e_n\norm{\paren{\hat{\nabla} f_{n,h}(X) \cdot   \hat{\nabla} f_{n,h}(X)^\top  - \nabla f(X) \cdot \nabla f(X)^\top}}_2.
	\end{align*}
	
	\noindent
	To bound the r.h.s above, we use the fact that, for vectors $a, b$, we have
	$$aa^\top - bb^\top = \frac{1}{2}(a-b)(b+a)^\top + \frac{1}{2}(b+a)(a-b)^\top,$$
	implying that
	\begin{align*}
	\norm{aa^\top -bb^\top}_2 \leq& \frac{1}{2}\norm{(a-b)(b+a)^\top}_2 \\
	&+  \frac{1}{2}\norm{(b+a)(a-b)^\top}_2 \\
	=& \norm{(b+a)(a-b)^\top}_2
	\end{align*}
	since the spectral norm is invariant under matrix transposition.
	
	\noindent
	We therefore have that $\| \e_n G_n(X)  - \e_n G(X) \|_2$ is at most
	\begin{align*}
	&\e_n \| (\nabla f(X) - \hat{\nabla} f_{n,h}(X)) \cdot (\nabla f(X) + \hat{\nabla} f_{n,h}(X))^\top  \|_2 \\
	&= \e_n \|\nabla f(X) - \hat{\nabla} f_{n,h}(X)\| \cdot \|\nabla f(X) + \hat{\nabla} f_{n,h}(X)\|  \\
	&\leq \e_n {\|\nabla f(X) - \hat{\nabla} f_{n,h}(X)\|} \cdot \max_{x\in \Xspl}\|\nabla f(x) + \hat{\nabla} f_{n,h}(x)\|.
	% &\leq \left(3R + \norm{\epsilon_t}_2   + \sqrt{d}(\frac{hR + C_Y(\delta)}{t})\right) \e_n \|\nabla f(X) - \hat{\nabla} f_{n,h}(X)\|_2
	\end{align*}
\end{proof}
\noindent
Thus the matrix estimation problem is reduced to that of an average gradient estimation. The two terms of (\ref{eq:redux})
are bounded in the following two subsections. These sections thus contain the bulk of the analysis. All
omitted proofs are found in the supplementary.

\subsection{Bound on \texorpdfstring{$\e_n \|\nabla f(X) - \hat{\nabla} f_{n,h}(X)\|$}{} %
}
The analysis of this section relies on a series of approximations.
In particular we relate the vector $\hat{\nabla} f_{n,h}(\x)$ to
the vector $$\hat{\nabla} f(\x) \triangleq \paren{\Delta_{t,i}f(\x) \cdot \ind{A_{n,i} (\x)}}_{i\in [d]}.$$

\noindent
In other words we start with the decomposition:
\begin{align}
\e_n \|\nabla f(X) - \hat{\nabla} f_{n,h}(X)\| \leq & \e_n \|\nabla f(X) - \hat{\nabla} f(X)\| \nonumber \\
+& \e_n \|\hat{\nabla} f(X) - \hat{\nabla} f_{n,h}(X)\|. \label{eq:decomp1}
\end{align}
We bound each term separately in the following subsections.

\subsubsection{Bounding \texorpdfstring{$\e_n \|\nabla f(X) - \hat{\nabla} f(X)\|$}{} %
}
We need to introduce vectors $\mathbf{I}_n(x)\triangleq\paren{\mathbf{1}_{A_{n, i} (\x)}}_{i\in [d]}$, and
$\overline{\mathbf{I}_n(x)}\triangleq \paren{\mathbf{1}_{\bar{A}_{n,d} (\x)}}_{i \in [d]}$. We then have:
\begin{align}
\e_n \|\nabla f(X) - \hat{\nabla} f(X)\| \leq  \e_n \|\nabla f(X) \circ \overline{\mathbf{I}_n(X)}\| \nonumber \\
+ \e_n \|\nabla f(X) \circ \mathbf{I}_n(X) - \hat{\nabla} f(X)\| . \label{eq:decomp2}
\end{align}

\noindent
The following lemma bounds the first term of (\ref{eq:decomp2}).
\begin{lemma}
	Assume A2 and A5. Suppose $h \geq (\log^2(n/\delta)/n)^{1/d}$.
	With probability at least $1 - \delta$ over the sample of $\Xspl$:
	\begin{align*}
	\e_n \norm{\nabla f(X) \circ \overline{\mathbf{I}_n(X)}} \leq R\cdot \left( \sqrt{\frac{d \ln \frac{d}{\delta}}{2n}} + \norm{\mu_{\partial_t}} \right).
	\end{align*}
	\label{lemma1}
\end{lemma}

\begin{proof}
	By assumption, $\|\nabla f(\x)\| \leq R$, so we have
	\begin{align}
	\e_n \norm{\nabla f(X) \circ \overline{\mathbf{I}_n(X)}} \leq R \cdot \e_n \norm{\overline{\mathbf{I}_n(X)}}. \label{eq:1}
	\end{align}
	We bound $\norm{\overline{\mathbf{I}_n(X)}}$ as follows. For any $i\in [d]$, define the events
	$A_{i}(X) \equiv \min_{\{t,-t\}} \mu(B(X+s e_i, h/2)) \geq 3 \cdot \frac{2d \ln 2n + \ln(4/\delta)}{n},$
	and define the vector $\overline{\mathbf{I}(X)} \triangleq \paren{\ind{\bar{A_i}(X)}}_{i\in [d]}$.
	
	\noindent
	By relative VC bounds \cite{doi:10.1137/1116025}, let $\alpha_n = \frac{2d \ln 2n + \ln(4/\delta)}{n}$, then with probability at least $1 - \delta$ over the choice of $\Xspl$, for all balls $B \in R^d$ we have $\mu(B) \leq \mu_n(B) + \sqrt{\mu_n(B) \alpha_n} + \alpha_n$.
	Therefore, with probability at least $1 - \delta$, $\forall i \in [d]$ and $x$ in the sample $\Xspl$, $\bar{A}_{n,i}(x) \Rightarrow \bar{A}_{i}(x)$.
	
	\noindent
	Moreover, since $\|\overline{\mathbf{I}(X)}\| \leq \sqrt{d}$, by Hoeffding's inequality,
	\begin{align*}
	\mathbb{P}(\e_n \|\overline{\mathbf{I}(X)}\| - \e_X \|\overline{\mathbf{I}(X)}\| \geq \epsilon) \leq e^{-\frac{2n \epsilon^2}{d}}.
	\end{align*}
	It follows that, with probability at least $1-\delta$,
	\begin{align}
	\e_n \|\overline{\mathbf{I}_n(X)}\| \leq& \e_n \|\overline{\mathbf{I}(X)}\| \nonumber \\
	\leq& \e_X \|\overline{\mathbf{I}(X)}\| + \sqrt{\frac{d \ln \frac{1}{\delta}}{2n}} \nonumber \\
	\leq& \sqrt{\e_X \|\overline{\mathbf{I}(X)}\|^2} + \sqrt{\frac{d \ln \frac{1}{\delta}}{2n}}, \label{eq:2}
	\end{align}
	by Jensen's inequality. We bound each of the $d$ terms of
	$\e_X \|\overline{\mathbf{I}(X)}\|^2 = \sum_{i\in [d]} \e_X \mathbf{1}_{\bar{A}_i(X)}$ as follows.
	
	%Fix any $i\in [d]$ and let
	%$\bar\Xx_{i}\triangleq \braces{x\in \Xx: \mathbf{1}_{\bar{A}_i(X)} = 1}$. We want to bound
	%$\e_X \mathbf{1}_{\bar{A}_i(X)} = \mu(\bar\Xx_i)$. Consider
	
	\noindent
	Fix any $i\in [d]$. We have $\e_X \mathbf{1}_{\bar{A}_i(X)} \leq \e_X[\mathbf{1}_{\bar{A}_i(X)} | X \in \mathcal{X} \backslash \partial_{t,i}(\mathcal{X})] + \mu(\partial_{t,i}(\mathcal{X}))$. Notice that $\e_X[\mathbf{1}_{\bar{A}_i(X)} | X \in \mathcal{X} \backslash \partial_{t,i}(\mathcal{X})] = 0$ since, by assumption, $\mu(B(x+se_i,h/2)) \geq C_{\mu} (h/2)^d \geq 3 \alpha$ whenever $h \geq (\log^2(n/\delta)/n)^{1/d}$.
	Hence, we have
	\begin{align*}
	\sqrt{\e_X \|\overline{\mathbf{I}(X)}\|^2} \leq \sqrt{\sum_{i \in [d]} \mu^2(\partial_{t,i}(\mathcal{X}))}.
	\end{align*}
	Combine this last inequality with (\ref{eq:1}) and (\ref{eq:2}) and conclude.
	%Thus we have with probability at least $1 - \delta$,
	%\begin{eqnarray*}
	%\e_n \|\nabla f(X) \circ \overline{\mathbf{I}_n(X)}\|_2 \leq R \left( \sqrt{\frac{d \ln \frac{d}{\delta}}{2n}} + \norm{\mu_{\partial_t}}_2 \right)
	%\end{eqnarray*}
\end{proof}

\noindent
The second term of (\ref{eq:decomp2}) is bounded in the next lemma.

\begin{lemma}
	\label{lem:somename}
	Fix the sample $\Xspl$. We have $\max_{X\in \Xspl} \|\nabla f(X) \circ \mathbf{I}_n(X) - \hat{\nabla} f(X)\| \leq \norm{\epsilon_t}$.
	\label{lemma2}
\end{lemma}

\begin{proof}
	For a given coordinate $i\in [d]$, let $f_i'$ denote the directional derivative $e_i^\top \nabla f$ along $i$.
	Pick any $x\in \Xx$.
	Since $f(\x + t e_i) - f(\x - t e_i) = \int_{-t}^t  f'_i(\x + s e_i) ds$, we have
	\begin{align*}
	2t (f'_i(\x) - \epsilon_{t,i}) \leq& f(\x + t e_i) - f(\x - t e_i) \\
	\leq& 2t (f'_i(\x) + \epsilon_{t,i})
	\end{align*}
	Thus $|\frac{1}{2t} (f(\x + t e_i) - f(\x - t e_i)) - f'_i(\x)| \leq \epsilon_{t,i}$.
	We therefore have that $\|\nabla f(\x) \circ \mathbf{I}_n(x) - \hat{\nabla} f(\x)\|$ equals
	\begin{align*}
	&\sqrt{\sum_{i=1}^d \paren{f'_i(\x) \cdot  \mathbf{1}_{A_{n,i}(\x)} - \Delta_{t,i} f(\x) \cdot \mathbf{1}_{A_{n,i}(\x)}}^2} \\
	&= \sqrt{\sum_{i=1}^d \paren{\frac{1}{2t} (f(\x + t e_i) - f(\x - t e_i)) - f'_i(\x)}^2} \\
	&\leq \norm{\epsilon_t}  .
	\end{align*}
	%Taking empirical expectation on both sides concludes the proof.
\end{proof}

\noindent
The last two lemmas can then be combined using equation (\ref{eq:decomp2}) into the final bound of this subsection.

\begin{lemma}
	\label{lem:bound_decomp2}
	Assume A2 and A5. Suppose  $h \geq (\log^2(n/\delta)/n)^{1/d}$. With probability at least $1-\delta$ over the sample $\Xspl$:
	\begin{align*}
	\e_n \|\nabla f(X) - \hat{\nabla} f(X)\| \leq&  R\cdot \left( \sqrt{\frac{d \ln \frac{d}{\delta}}{2n}} + \norm{\mu_{\partial_t}} \right) \\
	&+ \norm{\epsilon_t}.
	\end{align*}
\end{lemma}

\subsubsection{Bounding \texorpdfstring{$\e_n \|\hat{\nabla} f(X) - \hat{\nabla} f_{n,h}(X)\|$}{} %
}
We need to consider bias and variance functionals of estimates $f_{n, h}(x)$. To this end we introduce the expected estimate
$\tilde{f}_{n, h}(x) = \expec_{\Yspl | \Xspl} f_{n, h}(x) = \sum_{i=1}^n w_i(x) f(X_i).$
\noindent
The following lemma bounds the bias of estimates $f_{n, h}$. The proof relies on standard ideas.
\begin{lemma}[Bias of $f_{n, h}$]
	\label{lem:bias}
	Assume A2. Let $t<\tau$.
	We have for all $X\in \Xspl$, all $i \in [d]$, and $s \in \{-t,t\}$:
	\begin{align*}
	|\tilde{f}_{n,h}(X + s e_i) - f(X + s e_i)| \cdot  \mathbf{1}_{A_{n,i}(\x)} \leq hR.
	\end{align*}
\end{lemma}

\begin{proof}
	Let $x = X + se_i$. Using a Taylor approximation on $f$ to bound $\abs{f(X_i) - f(x)}$, we have
	\begin{align*}
	\hspace{-0.05 in} |\tilde{f}_{n,h}(\x ) - f(\x)| &\leq \sum_{i \in [d]} w_i(x) |f(X_i) - f(x)| \\
	&\leq \sum_{i \in [d]} w_i(x) \|X_i - x\| \cdot \sup_{\Xx + B(0, \tau)}\norm{\nabla f} \\
	& \leq hR.
	\end{align*}
\end{proof}

\noindent
The following lemma bounds the variance of estimates $f_{n, h}$ averaged over the sample $\Xspl$.
To obtain a high probability bound, we relie on results of Lemma 7 in \cite{DBLP:conf/nips/KpotufeB12}.
However in \cite{DBLP:conf/nips/KpotufeB12}, the variance of the estimator if evaluated at a point, therefore requiring local density assumptions.
The present lemma has no such local density requirements given that we are interested in an average quantity over a collection of points.

\begin{lemma}[Average Variance]
	\label{lem:var}
	Assume A1. There exist $C = C(\mu,K(\cdot))$, such that the following holds with probability at least $1 - 2 \delta$ over the choice of the sample
	$(\Xspl, \Yspl)$. Define $A(n) = \sqrt{C d \cdot \ln(n/\delta)} \cdot C_Y^2(\delta/2n) \cdot \sigma_Y^2 $, for all $i \in [d]$, and all $s \in \{-t,t\}$:
	\begin{align*}
	\e_n |\tilde{f}_{n,h}(X + s e_i) - f_{n,h}(X + s e_i)|^2 \cdot  \mathbf{1}_{A_{n,i}(X)} \leq \frac{A(n)}{nh^d}
	\end{align*}
\end{lemma}

\begin{proof}[Proof of Lemma \ref{lem:var}]
	Fix the sample $\Xspl$ and consider only the randomness in $\Yspl$.
	The following result is implicit to the proof of Lemma 7 of \cite{DBLP:conf/nips/KpotufeB12}:
	with probability at least $1-2\delta$, for all $X\in \Xspl$, $i\in [d]$, and $s\in \braces{-t, t}$, we have
	(where, for simplicity, we write $x = X+ se_i$) $|\tilde{f}_{n,h}(x) - f_{n,h}(x)|^2 \cdot  \mathbf{1}_{A_{n,i}(X)}$ is at most
	\begin{align*}
	\frac{Cd\cdot\log(n/\delta)C_Y^2(\delta/2n)\cdot \sigma^2_Y}{n\mu_n((B(x, h/2))}.
	\end{align*}
	Fix $i\in [d]$ and $s\in \braces{-t , t}$.
	Taking empirical expectation, we get $\e_n |\tilde{f}_{n,h}(x) - f_{n,h}(x)|^2$ is at most
	\begin{align*}
	\frac{\sqrt{C d \cdot \ln(n/\delta)} \cdot C_Y^2(\delta/2n) \cdot \sigma_Y^2}{n}
	\sum_{j \in [n]} \frac{1}{n(x_j,h/2)}
	\end{align*}
	where $x_j= X_j + se_i$, and
	$n(x_i,h/2) = n\mu_n(B(x_i,h/2))$ is the number of samples in $B(x_i,h/2)$. Let $\mathcal{Z}\subset \real^d$ denote a
	minimal $h/4$-cover of
	$\braces{X_1,...,X_n}$. Since $\Xx$ has bounded diameter, such a cover has size at most
	$C_\Xx(h/4)^d$ for some $C_\Xx$ depending on the support $\Xx$ of $\mu$.
	
	\noindent
	Assume every $x_j$ is assigned to the closest $z \in \mathcal{Z}$, where ties can be broken any way,
	and write $x_j \rightarrow z$ to denote such an assignment.
	By definition of $Z$, $x_j$ is contained in the ball $B(z,h/4)$, and we therefore have $B(z,h/4) \subset B(x_j,h/2)$.
	
	Thus
	\begin{align*}
	\sum_{j \in [n]} \frac{1}{n(x_j,h/2)} &= \sum_{z \in \mathcal{Z}} \sum_{x_j \rightarrow z} \frac{1}{n(x_j,h/2)} \\
	&\leq \sum_{z \in \mathcal{Z}} \sum_{x_j \rightarrow z} \frac{1}{n(z,h/4)} \\
	&\leq \sum_{z \in \mathcal{Z}} \frac{n(z,h/4)}{n(z,h/4)}
	= |\mathcal{Z}| \leq C_\Xx (h/4)^{-d}.
	\end{align*}
	Combining with the above analysis finishes the proof.
\end{proof}

\noindent
The main bound of this subsection is given in the next lemma which combines the above bias and variance results.
\begin{lemma}
	Assume A1 and A2. There exist $C = C(\mu,K(\cdot))$, such that the following holds with probability at least $1 - 2 \delta$ over the choice of $(\Xspl, \Yspl)$. Define $A(n) = \sqrt{C d \cdot \ln(n/\delta)} \cdot C_Y^2(\delta/2n) \cdot \sigma_Y^2 $:
	\begin{align*}
	\e_n \|\hat{\nabla} f(X) - \hat{\nabla} f_{n,h}(X)\| \leq \frac{\sqrt{d}}{t} \sqrt{\frac{A(n)}{nh^d} + 2 R^2 h^2}.
	\end{align*}
	\label{lemma7}
\end{lemma}
\begin{proof}
	In what follows, we first apply Jensen's inequality, and the fact that $(a+b)^2 \leq 2a^2 + 2b^2$. We have:
	\begin{align}
	&\e_n \|\hat{\nabla} f(X) - \hat{\nabla} f_{n,h}(X)\| \nonumber \\
	&= \e_n \paren{\sum_{i \in [d]} |\Delta_{t,i}f_{n,h}(X) - \Delta_{t,i} f(X)|^2 \cdot  \mathbf{1}_{A_{n,i}(X)}}^{1/2} \nonumber\\
	&\leq \paren{\sum_{i \in [d]} \e_n |\Delta_{t,i}f_{n,h}(X) - \Delta_{t,i} f(X)|^2 \cdot  \mathbf{1}_{A_{n,i}(X)}}^{1/2} \nonumber\\
	&\leq \frac{\sqrt{d}}{2t}\paren{\max_{i\in [d], s \in \{-t,t\}} 4\e_n |f_{n,h}(\tilde{X}) - f(\tilde{X})|^2 \cdot  \mathbf{1}_{A_{n,i}(X)}}^{1/2}
	\label{eq:biasvar}
	\end{align}
	where $\tilde{X} = X + s e_i$.
	Next, use the fact that for any $s\in \braces{-t, t}$, we have the following decomposition into variance and bias terms
	\begin{align*}
	&|f_{n,h}(X + s e_i) - f(X + s e_i)|^2 \\
	&\leq 2|f_{n,h}(X + s e_i) - \tilde{f}_{n, h}(X + s e_i)|^2 \\
	& + 2|\tilde{f}_{n,h}(X + s e_i) - f(X + s e_i)|^2.
	\end{align*}
	Combine this into (\ref{eq:biasvar}) to get a bound in terms of the average bias and variance of estimates $f_{n, h}(X+ se_i)$.
	Apply Lemma \ref{lem:bias} and \ref{lem:var} and conclude.
\end{proof}

\subsubsection{Main Result of this Section}
The following theorem provides the final bound of this section on $\e_n \|\nabla f(X) - \hat{\nabla} f_{n,h}(X)\|$.
It follows directly from the decomposition of equation \ref{eq:decomp1}
and Lemmas \ref{lem:bound_decomp2} and \ref{lemma7}.
\begin{lemma}
	Assume A1, A2 and A5. Let $t< \tau$ and suppose  $h \geq (\log^2(n/\delta)/n)^{1/d}$. With probability at least $1 - 2 \delta$ over the choice of the sample $(\Xspl, \Yspl)$, we have
	\begin{align*}
	\e_n \|\nabla f(X) - \hat{\nabla} f_{n,h}(X)\| \leq \frac{\sqrt{d}}{t} \sqrt{\frac{A(n)}{nh^d} +2R^2 h^2 } \\
	+ R \left( \sqrt{\frac{d \ln \frac{d}{\delta}}{2n}} + \norm{\mu_{\partial_t}} \right) + \norm{\epsilon_t}.
	\end{align*}
	\label{theorem:f}
\end{lemma}

\subsection{Bounding \texorpdfstring{$\max_{X\in \Xspl}\| \nabla f(X) + \hat{\nabla} f_{n,h}(X) \|$}{} %
}
\begin{lemma}
	Assume A1 and A2. With probability at least $1 - \delta$, we have
	\begin{align*}
	\| \nabla f(X) + \hat{\nabla} f_{n,h}(X) \| \leq& 3R + \norm{\epsilon_t}   \\
	&+ \sqrt{d}\paren{\frac{hR + C_Y(\delta/n)}{t}}.
	\end{align*}
	\label{lem:onemore}
\end{lemma}
\begin{proof}
	Fix $X\in \Xspl$.We have
	\begin{align}
	\| \nabla f(X) + \hat{\nabla} f_{n,h}(X) \| \leq& 2 \|\nabla f(X) \| \nonumber \\
	&+ \|\nabla f(\x) - \hat{\nabla} f_{n,h}(X)\| \nonumber\\
	\leq & 2R + \|\nabla f(X) - \hat{\nabla} f(\x) \| \nonumber \\
	&+ \|\hat{\nabla} f(X) - \hat{\nabla} f_{n,h}(X)\|. \label{eq:something}
	\end{align}
	We can bound the second term of (\ref{eq:something}) above as follows.
	\begin{align*}
	\|\nabla f(X) - \hat{\nabla} f(X) \| \leq& \|\nabla f(X) \circ \mathbf{I}_n(X) - \hat{\nabla} f(X)\| \nonumber\\
	&+ \|\nabla f(X) \circ \overline{\mathbf{I}_n(X)}\| \\
	\leq&  \norm{\epsilon_t} + R,
	\end{align*}
	where we just applied Lemma \ref{lem:somename}.
	
	\noindent
	For the third term of (\ref{eq:something}), $\|\hat{\nabla} f(\x) - \hat{\nabla} f_{n,h}(\x)\|$ equals
	\begin{align*}
	\sqrt{\sum_{i \in [d]} (|\Delta_{t,i}f_{n,h}(\x) - \Delta_{t,i} f(\x)| \cdot  \mathbf{1}_{A_{n,i}(\x)})^2}.
	\end{align*}
	As in the proof of Lemma \ref{lemma7}, we decompose the above summand into bias and variance terms, that is:
	\begin{align*}
	&|\Delta_{t,i}f_{n,h}(\x) - \Delta_{t,i} f(\x)| \\
	&\leq \frac{1}{t} \max_{s \in \{-t,t\}} |\tilde{f}_{n,h}(\x + s e_i) - f(\x + s e_i)| \\
	&+ \frac{1}{t} \max_{s \in \{-t,t\}} |\tilde{f}_{n,h}(\x + s e_i) - f_{n,h}(\x + s e_i)|.
	\end{align*}
	By Lemma \ref{lem:bias}, $|\tilde{f}_{n,h}(\x + s e_i) - f(\x + s e_i)|\leq Rh$ for any $s\in \braces{-t, t}$.
	
	\noindent
	Next, by definition of $C_Y(\delta/n)$, with probaility at least $1-\delta$, for each $j\in [n]$, $Y_j$ has value within
	$C_Y(\delta)$ of $f(X_j)$.
	It follows that  $|\tilde{f}_{n,h}(X + s e_i) - f_{n,h}(X + s e_i)| \leq C_Y(\delta/n)$ for $s\in \braces{-t, t}$.
	
	\noindent
	Thus, with probability at least $1 - \delta$, we have
	\begin{eqnarray*}
		\|\hat{\nabla} f(X) - \hat{\nabla} f_{n,h}(X)\| \leq \sqrt{d}\paren{\frac{hR + C_Y(\delta/n)}{t}}.
	\end{eqnarray*}
	
	\noindent
	Combine these bounds in (\ref{eq:something}) and conclude.
\end{proof}

\subsection{Final Bound \texorpdfstring{$\| \e_n G_n(X) - \e_n G(X) \|_2$}{} %
}
We can now combine the results of the last two subsections, namely Lemma \ref{theorem:f} and \ref{lem:onemore}, into the next lemma, using the bound of Lemma \ref{lemma:EGOPmatrix}.
\begin{lemma}
	\label{lem:bound_decomp0}
	Assume A1, A2 and A5. Let $t< \tau$ and suppose  $h \geq (\log^2(n/\delta)/n)^{1/d}$. With probability at least $1 - 2 \delta$ over the choice of the sample $(\Xspl, \Yspl)$, we have that $\| \e_n G_n(X) - \e_n G(X) \|_2$ is at most
	\begin{small}
		\begin{align*}
		&\left(3R + \norm{\epsilon_t}   + \sqrt{d}\left(\frac{hR + C_Y(\delta/n)}{t}\right) \right) \cdot \\
		&\left[\frac{\sqrt{d}}{t} \sqrt{\frac{A(n)}{nh^d} + 2 h^2 R^2} + R \left( \sqrt{\frac{d \ln \frac{d}{\delta}}{2n}} + \norm{\mu_{\partial_t}} \right) + \norm{\epsilon_t}   \right].
		\end{align*}
	\end{small}
\end{lemma}

\section{Experiments}
\label{sec:experiments}

In this section we describe experiments aimed at evaluating the
utility of EGOP as a metric estimation technique for regression or
classification. We consider a family of non-parametric methods that
rely on the notion of distance under a given Mahalanobis metric
$\mathbf{M}$, computed as $(\mathbf{x}-\mathbf{x}')^T\mathbf{M}(\mathbf{x}-\mathbf{x}')$.

\vspace{2mm}
\noindent
In this setup, we consider three choices of $\mathbf{M}$: (i)
identity, i.e., Euclidean distance in the original space; (ii) the
estimated gradient weights (GW) matrix as in~\cite{DBLP:conf/nips/KpotufeB12}, i.e., Euclidean distance weighted by the
estimated $\Delta_{t, i}f_n$, and (iii) the estimated EGOP matrix $\expectation_n G_n(X)$. The
latter corresponds to Euclidean distance in the original space under
linear transform given by $\left[\expectation_n G_n(X)\right]^{1/2}$.
Note that a major distinction between the metrics based on GW and EGOP
is that the former only scales the Euclidean distance, whereas the
latter introduces a rotation.

\vspace{2mm}
\noindent
Each choice of $\mathbf{M}$ can define the set of neighbors of an
input point $x$ in two ways: (a) $k$ nearest neighbors ($k$NN) of
$x$ for a fixed $k$, or (b) neighbors with distance
$\le h$ for a fixed $h$; we will refer to this as $h$NN. When the task is regression, the output values of the
neighbors are simply averaged; for classification, the class
label for $x$ is decided by majority vote among neighbors. Note that $h$NN corresponds to
kernel regression with the boxcar kernel.

\vspace{2mm}
\noindent
Thus, we will consider six methods, based on combinations of the
choice of metric $M$ and the definition of neighbhors: $k$NN,
$k$NN-GW, $k$NN-EGOP, $h$NN,
$h$NN-GW, and $h$NN-EGOP.

\subsection{Synthetic Data}\label{synth}

In order to understand the effect of varying the dependence of $f$ on the input coordinates, on the quality of the metric estimated by the EGOP as well as other approaches, we first consider experiments on synthetic data. 
For the purpose of these experiments, the output is generated as follows: We set $y = \sum_i sin(c_ix_i)$, with the sum over all the dimensions of $\mathbf{x} \mathbb{R}^d$. The profile of the $\mathbf{c}$ vector is responsible for the degree upto which the value of $\mathbf{x}_i$ affects the output $y$.

\begin{figure*}
	\begin{center}
		\includegraphics[width=.48\linewidth]{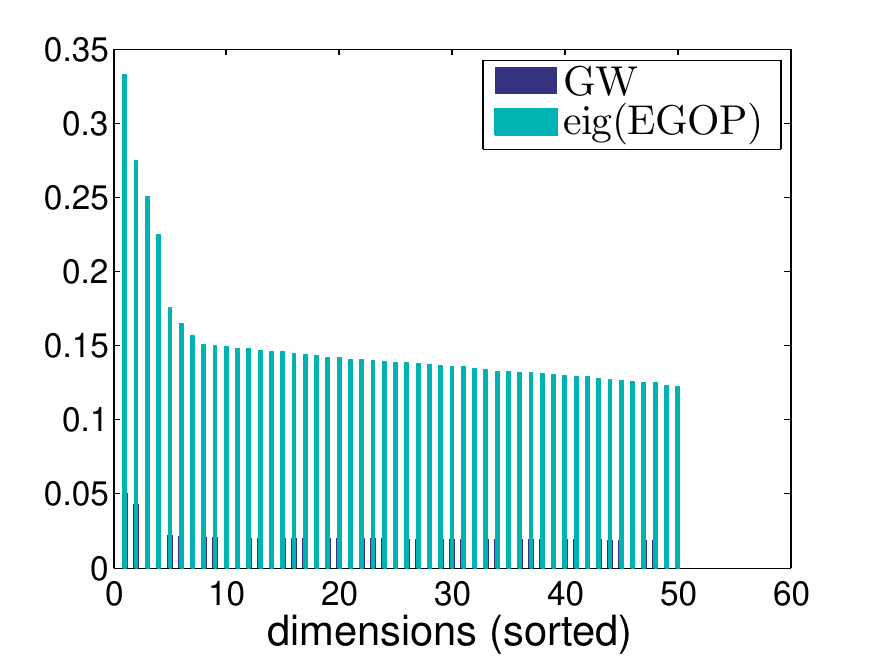}
		\includegraphics[width=.48\linewidth]{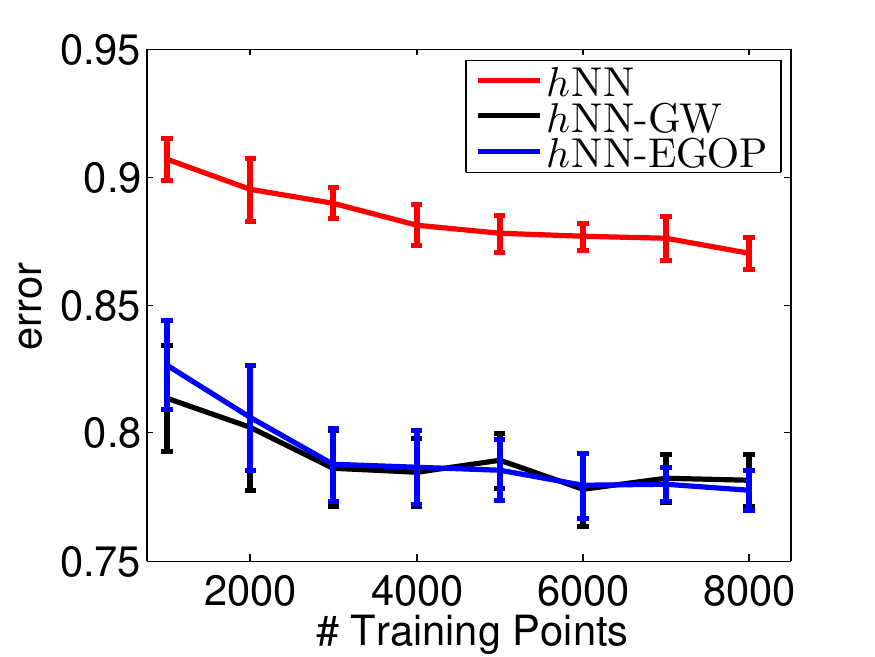}
		\caption*{No rotation}
	\end{center}
	\caption{Synthetic data, $d$=50, without rotation applied after generating $y$ from $\mathbf{x}$. The figure shows error of $h$NN with different metrics and the profile of derivatives recovered by GW and EGOP. In the case when there is no rotation, the performance of GW is similar to that returned by the EGOP}
	\label{fig:synthNoRot}
\end{figure*}

\vspace{2mm}
\noindent
We set $c[1]=50$ and then $c[i] = 0.6*c[i-1]$ for $i=2:50$, and sampled $d=50$-dimensional input over a bounded domain. In this data, we consider two cases: The first denoted (R), in which the input features are transformed by a random rotation in $\mathbb{R}^d$, \emph{after} $y$ has been generated; and the second, denoted (I) in which the input features are preserved. Under these conditions we evaluate the out of sample regression accuracy with
original metric, GW and EGOP-based metrics, for different value of $n$; in each experiment, the values
of $h$ and $t$ are tuned by cross-validation on the training set.

%\lipsum

\begin{figure*}
	\begin{center}
		\includegraphics[width=.49\linewidth]{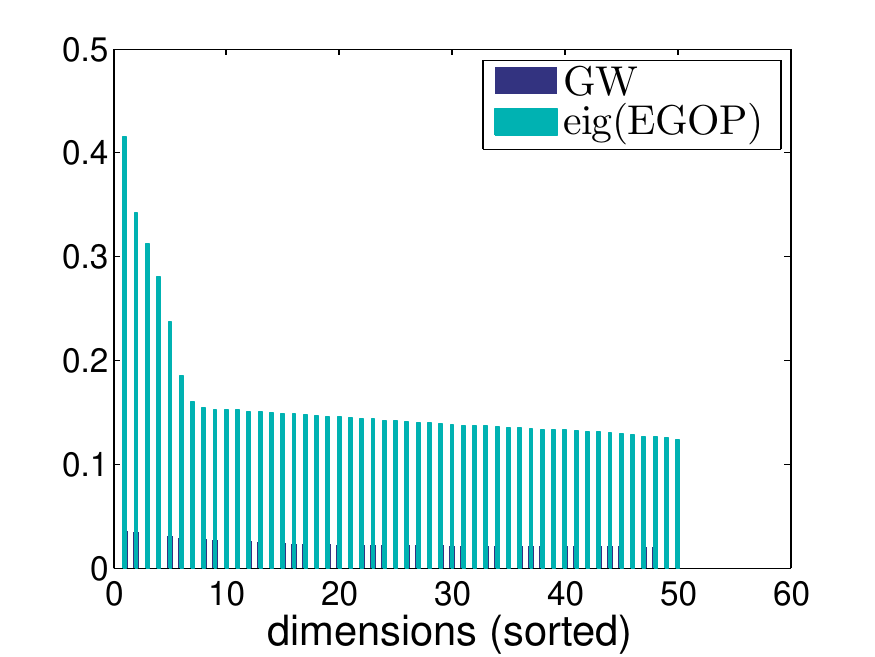}
		\includegraphics[width=.49\linewidth]{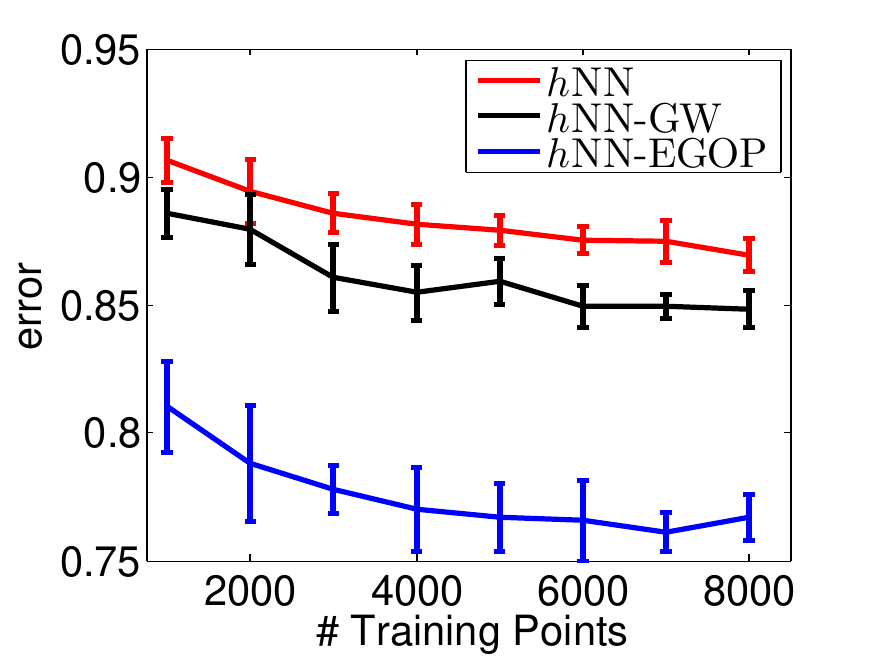}
		\caption*{Rotation}
		\caption{Synthetic data, $d$=50, with rotation applied after generating $y$ from $\mathbf{x}$. As in the companion figure \ref{fig:synthNoRot} we show error of $h$NN with different metrics, along with the profile of derivatives recovered by both GW and EGOP. The deterioration of the error performance of the Gradient Weights approach after the feature space is subject to a random rotation is noteworthy.}
		\end{center}
	\label{fig:synthRot}
\end{figure*}
%\lipsum

\vspace{2mm}
\noindent
From the results that we can see in figures \ref{fig:synthNoRot} and \ref{fig:synthRot} is that reweighting examples by either gradient weights or by using the expected gradient outerproduct helps in performance in all cases. However, in the case when the synthetic data is rotated, as might be expected, the performance of the case when the EGOP is used for the reweighing, is not significantly affected as compared to the no rotation case. This is in sharp contrast to the case of gradient weights: which is able to recover a good metric (as can be seen by the accuracy) in the no-rotation case, however, its performance falls steeply when the data is rotated. 

\vspace{2mm}
\noindent
In order to get some insight into the nature of the metrics that were estimated from this synthetic data, we also plot profiles of the estimated feature relevance. For the gradients weights approach, these are just the weights obtained. For the EGOP, we use the eigenvalues of the matrix as a measure of feature importance. In other words, for gradient weights this corresponds to values on the diagonal of $\mathbf{M}$, and for EGOP of the (square roots) of the eigenvalues of $\mathbf{M}$. Plots in figures \ref{fig:synthNoRot} and \ref{fig:synthRot} also show these profiles (sorted in descending order). By inspecting at these profiles, it is clear that the EGOP is largely invariant to rotation of the feature space, and is much better at recovering the relevance of the features according to what was prescribed by the $\mathbf{c}$ vector described above.

\subsection{Regression Experiments}
After the experiments on synthetic data, we now present some results on real world datasets. The name of the datasets, along with information such as their dimensionality, number of training and test points etc., is mentioned in Table~\ref{table:KernelRegression}. For each data set, we report the results averaged over ten random training/test splits.

\noindent
As a measure of performance we compute for each experiment the \emph{normalized mean squared error} (nMSE): mean squared error over test set, divided by target variance over that set. This can be interpreted as fraction of variance in the target unexplained by the regressor.

\noindent
In each experiment the input was normalized by the mean and standard deviation of the training set. For each method, the values of $h$ or $k$ as wel as $t$ (the bandwidth used to estimate finite differences for GW and EGOP) were set by two fold cross-validation on the training set.

\subsection{Classification Experiments}

The setup for classification data sets is very similar for regression, except that the task is binary classification, and the labels of the neighbors selected by each prediction method are aggregated by simple
majority vote, rather than averaging as in regression. The performance measure of interest here is classification
error. As in regression experiments, we normalized the data, tuned all relevant parameters by cross validation on training data, and repeated the entire experimental procedure ten times with random training/test splits.

In addition to the baselines listed above, in classification experiments we considered another competitor: the popular feature relevance determination method called ReliefF~\cite{conf/aaai/KiraR92,Kononenko97overcomingthe}. A highly engineered method that includes heuristics honed over considerable time by practitioners, it has the same general form of assigning weights to features as do GW and EGOP.

\begin{figure*}[!th]
\begin{center}
\includegraphics[width=.5\linewidth]{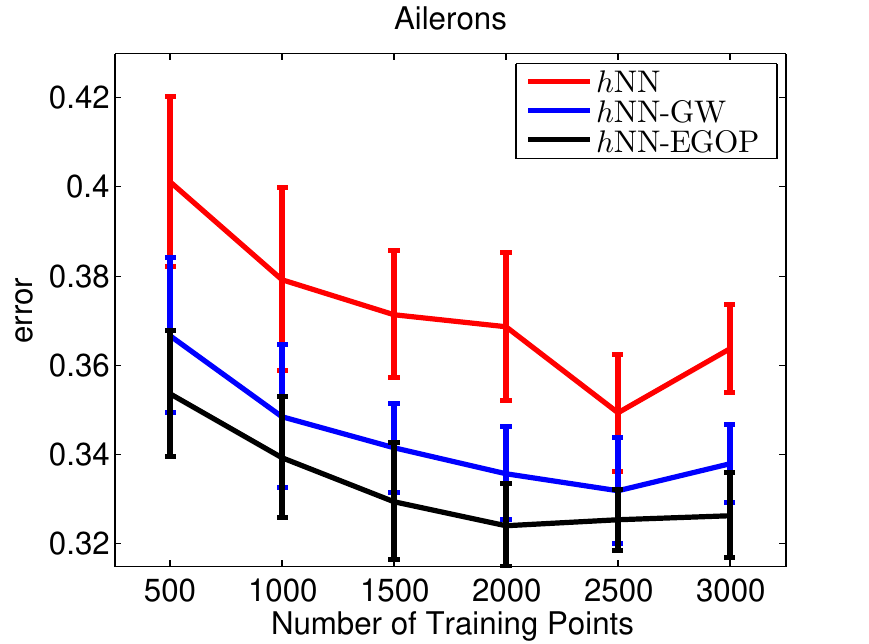} % first figure itself
\includegraphics[width=.5\linewidth]{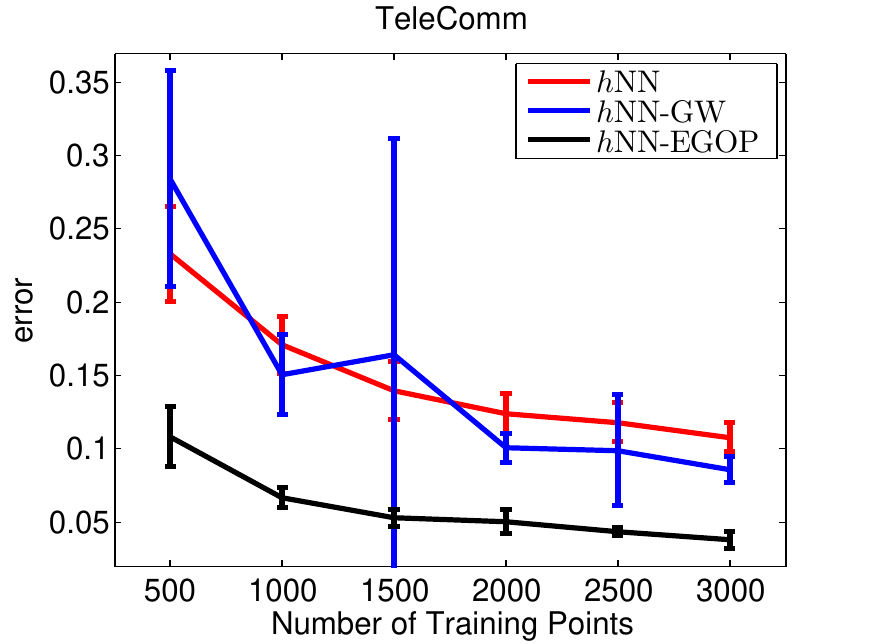}% second figure itself
\includegraphics[width=.5\linewidth]{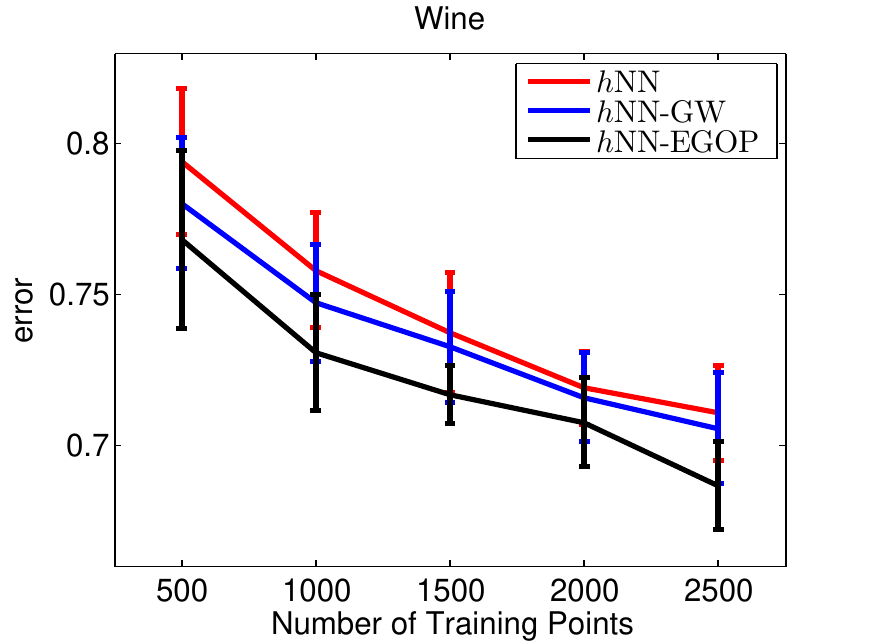} % first figure itself
\end{center}
\caption{Regression error (nMSE) as a function of training set size
  for Ailerons, TeleComm, Wine data sets.}
\label{fig:RegVariation}
\end{figure*}

\begin{figure*}[!th]
\begin{center}
\includegraphics[width=.5\linewidth]{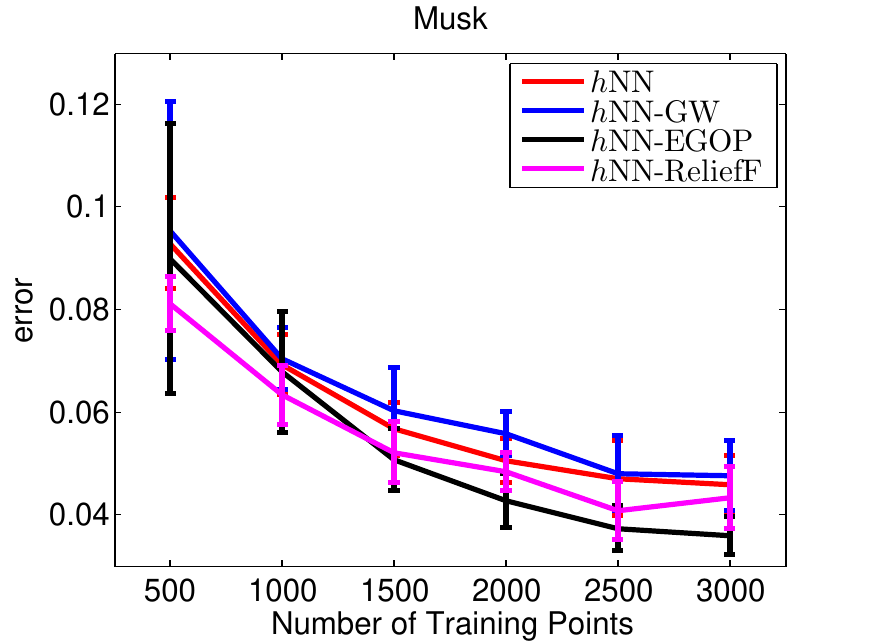} % first figure itself
\includegraphics[width=.5\linewidth]{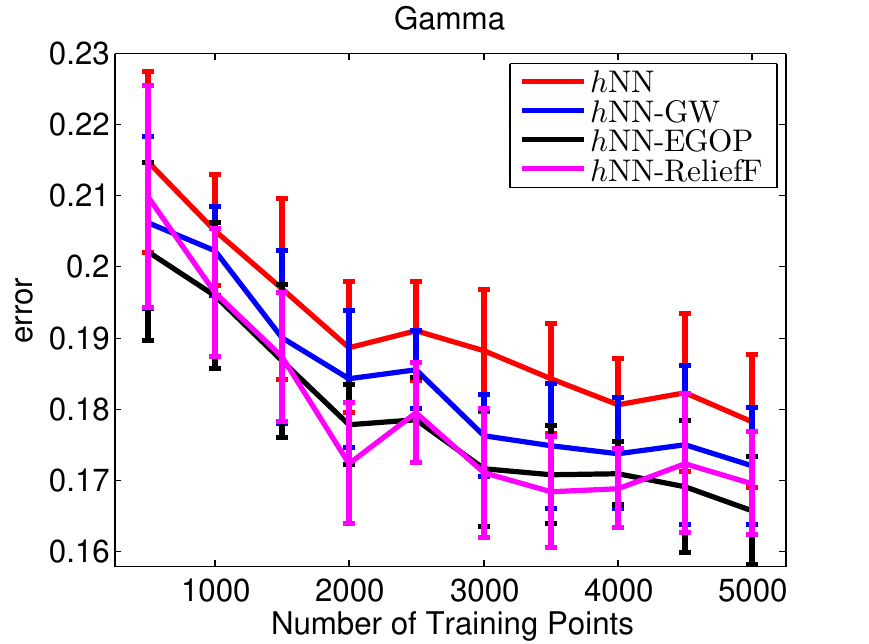}% second figure itself
\includegraphics[width=.5\linewidth]{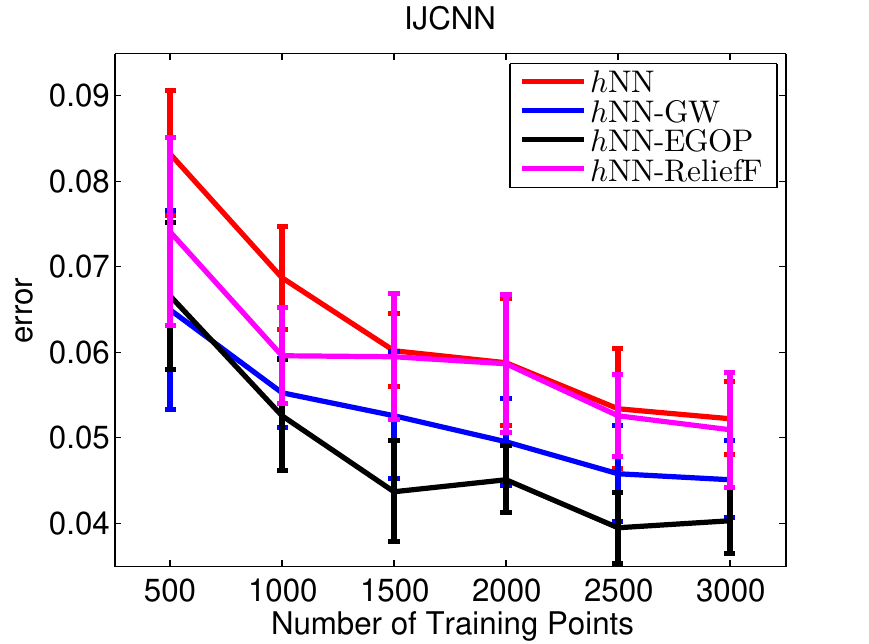} % first figure itself
\end{center}
\caption{Classification error as a function of training set size for
  Musk, Gamma, IJCNN data sets.}
\label{figure:classificationvar}
\end{figure*}

\begin{figure*}[!th]
\begin{center}
\includegraphics[width=.5\linewidth]{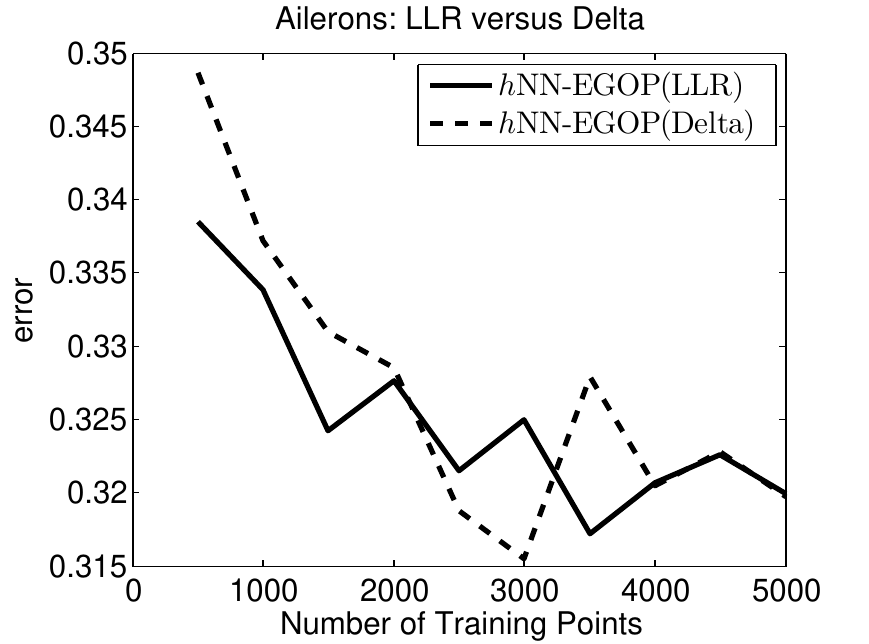}% first figure itself
\includegraphics[width=.5\linewidth]{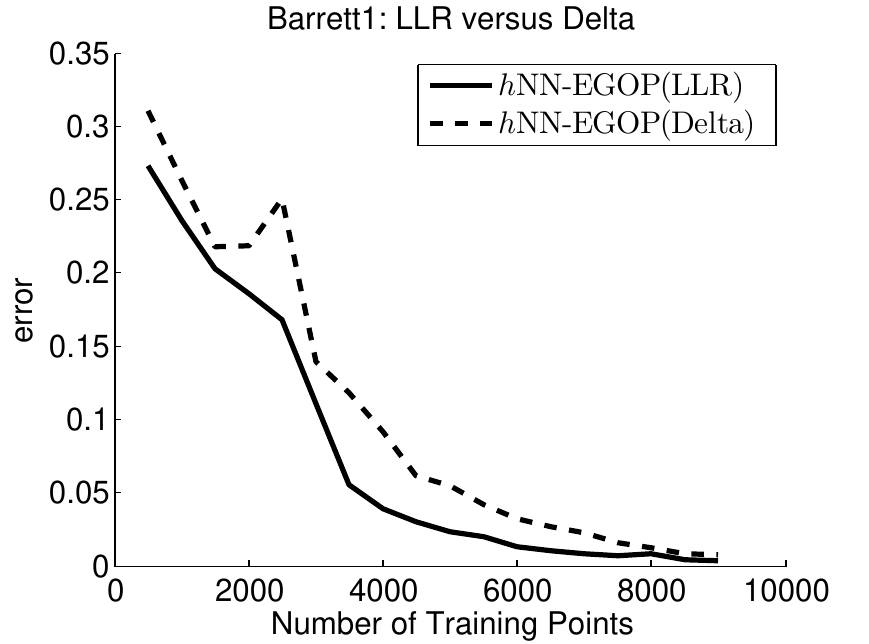}% second figure itself
\end{center}
\caption{Comparison of EGOP estimated by our proposed method
  vs. locally linear regression, for Ailerons and Barrett1 datasets. See the text for more details including runtime}
\label{figure:llr}
\end{figure*}

\begin{figure*}[!th]
	\begin{center}
		\includegraphics[width=.5\linewidth]{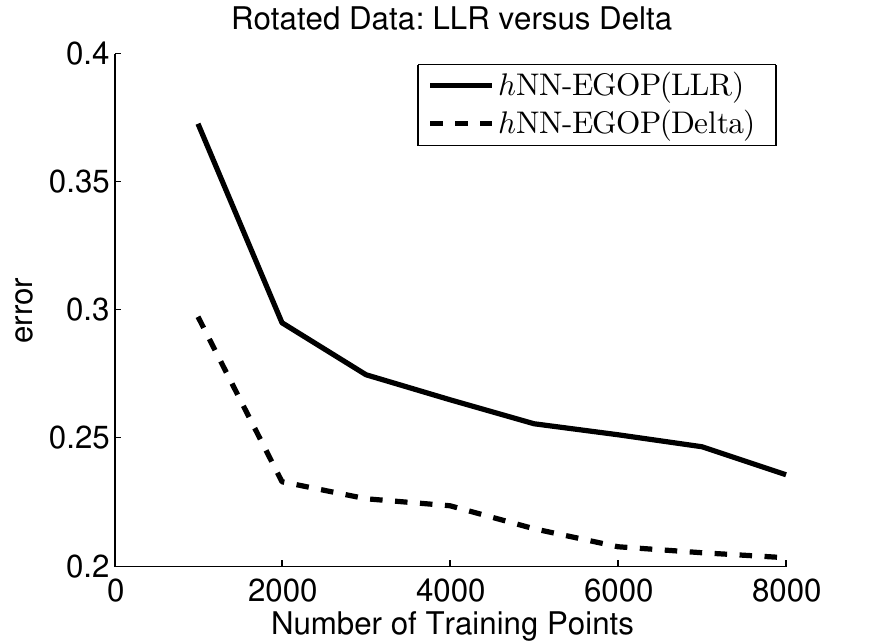}% first
	\end{center}
	\caption{Comparison of EGOP estimated by our proposed method
		vs. locally linear regression for a synthetic dataset (with rotation). This synthetic data is similar to the one used in section \ref{synth} but with $d=12$ and c = [5, 3, 1, .5, .2, .1, .08, .06, .05, .04, .03, .02].}
	\label{figure:llr_rotation}
\end{figure*}

\vspace{-0.05 in}
\subsection{Results}

The detailed results are reported in Tables \ref{table:KernelRegression} and~\ref{table:Classification}. These correspond to a single value of training set size. Plots in Figures~\ref{fig:RegVariation} and~\ref{figure:classificationvar} show a few representative cases for regression and classification, respectively, of performance of different methods as a function of training set size; it is evident from these that while the
performance of all methods tends to improve if additional training data are available, the gaps methods persist across the range of training set sizes.

\vspace{2mm}
\noindent
From the results in Tables \ref{table:KernelRegression} and~\ref{table:Classification}, we can see that the -EGOP
variants dominate the -GW ones, and that both produce gains relative to using the original metric. This is true both for $k$NN and for kernel regression ($h$NN) methods, suggesting general utility of EGOP-based metric, not tied to a particular non-parametric mechanism. We also see that the metrics based on estimated EGOP are competitive
with ReliefF.

\subsection{Experiments with Local Linear Regression}
As mentioned earlier in the paper, our estimator for EGOP is an
alternative to an estimator based on computing the slope of locally
linear regression (LLR)~\cite{cleveland1988locally} over the training data. We have compared these
two estimation methods on a number of data sets, and the results are
plotted in Figure~\ref{figure:llr}.
In these experiments, the bandwidth of LLR was tuned by a 2-fold
cross-validation on the training data.

\vspace{2mm}
\noindent
We observe that despite its simplicity, the accuracy of predictors
using EGOP-based metric
estimated by our approach is competitive with or even better than the
accuracy with EGOP estimated using LLR. As the sample size increases,
accuracy of LLR improves. However, the computational expense of
LLR-based estimator also grows with the size of data, and in our experiments it became
dramatically slower than our estimator of EGOP for the larger data
sizes. This confirms the intuition that our estimator is an appealing
alternative to LLR-based estimator, offering a good tradeoff of speed
and accuracy. 

\vspace{2mm}
\noindent
To impress upon the reader the computational advantage of our simple estimator over LLR, we also report the following running times (averaged over the ten random runs) for the same using our method and LLR respectively for the highest sample size used in the above real world datasets: Ailerons (128.13s for delta and 347.48s for LLR), Barrett (377.03s for delta and 1650.55s for LLR). Showing that our rough estimator is significantly faster than Local Linear Regression while giving competitive performance. These timings were recorded on an Intel i7 processor with CPU @ 2.40 GHz and 12 GB of RAM.

\begin{table*}[htpb]\vspace{-1.5em}
\footnotesize
\caption{Regression results, with ten random runs per data set.} % title of Table
\centering % used for centering table
\begin{tabular}{l l l l l l} % centered columns (4 columns)
\hline %inserts double horizontal lines

Dataset & d & train/test & $h$NN & $h$NN-GW & $h$NN-EGOP \\
\hline \hline
Ailerons & 5 & 3000/2000 & 0.3637 $\pm$ 0.0099 & 0.3381 $\pm$ 0.0087 & {\bf 0.3264} $\pm$ 0.0095 \\
Concrete & 8 & 730/300 & 0.3625 $\pm$ 0.0564 & 0.2525 $\pm$ 0.0417 & {\bf 0.2518} $\pm$ 0.0418 \\
Housing & 13 & 306/200 & 0.3033 $\pm$ 0.0681 & {\bf 0.2628} $\pm$ 0.0652 & 0.2776 $\pm$ 0.0550 \\
Wine & 11 & 2500/2000 & 0.7107 $\pm$ 0.0157 & 0.7056 $\pm$ 0.0184  & {\bf 0.6867} $\pm$ 0.0145 \\
Barrett1 & 21 & 3000/2000 & 0.0914 $\pm$ 0.0106 & {\bf 0.0740} $\pm$ 0.0209 & 0.0927 $\pm$ 0.0322  \\
Barrett5 & 21 & 3000/2000 & 0.0906 $\pm$ 0.0044 & {\bf 0.0823} $\pm$ 0.0171 & 0.0996 $\pm$ 0.0403 \\
Sarcos1 & 21 & 3000/2000 & 0.1433 $\pm$ 0.0087 & {\bf 0.0913} $\pm$ 0.0054 & 0.1064 $\pm$ 0.0101 \\
Sarcos5 & 21 & 3000/2000 & 0.1101 $\pm$ 0.0033 & 0.0972 $\pm$ 0.0044 & {\bf 0.0970} $\pm$ 0.0064 \\
ParkinsonM & 19 & 3000/2000 & 0.4234 $\pm$ 0.0386  & 0.3606 $\pm$ 0.0524 & {\bf 0.3546} $\pm$ 0.0406\\
ParkinsonT & 19 & 3000/2000 & 0.4965 $\pm$ 0.0606 & {\bf 0.3980 $\pm$ 0.0738} & 0.4168 $\pm$ 0.0941 \\
TeleComm & 48 & 3000/2000 & 0.1079 $\pm$ 0.0099 & 0.0858 $\pm$ 0.0089
& {\bf 0.0380} $\pm$ 0.0059 \\
\vspace{-.3em}&\\
\hline
Dataset &  &  & $k$NN  & $k$NN-GW & $k$NN-EGOP \\
\hline \hline
Ailerons &  &  & 0.3364 $\pm$ 0.0087 & 0.3161 $\pm$ 0.0058 & {\bf 0.3154} $\pm$ 0.0100  \\
Concrete &  &  & 0.2884 $\pm$ 0.0311 &  {\bf 0.2040} $\pm$ 0.0234& 0.2204 $\pm$ 0.0292 \\
Housing &  & & 0.2897 $\pm$ 0.0632 & {\bf 0.2389} $\pm$ 0.0604 & 0.2546 $\pm$ 0.0550 \\
Wine &  & & 0.6633 $\pm$ 0.0119 &  0.6615 $\pm$ 0.0134 & {\bf 0.6574} $\pm$ 0.0171 \\
Barrett1 &  &  & 0.1051 $\pm$ 0.0150 & {\bf 0.0843} $\pm$ 0.0229 & 0.1136 $\pm$ 0.0510  \\
Barrett5 &  &  & 0.1095 $\pm$ 0.0096 & {\bf 0.0984} $\pm$ 0.0244 &  0.1120 $\pm$ 0.0315 \\
Sarcos1 &  & &  0.1222 $\pm$ 0.0074 & {\bf 0.0769} $\pm$ 0.0037 &  0.0890 $\pm$ 0.0072 \\
Sarcos5 &  &  &  0.0870 $\pm$ 0.0051 & 0.0779 $\pm$ 0.0026&  {\bf 0.0752} $\pm$ 0.0051 \\

ParkinsonM &  &  & 0.3638 $\pm$ 0.0443 & {\bf 0.3181} $\pm$ 0.0477 & 0.3211 $\pm$ 0.0479 \\
ParkinsonT &  &  & 0.4055 $\pm$ 0.0413 & 0.3587 $\pm$ 0.0657 &  {\bf 0.3528} $\pm$ 0.0742 \\
TeleComm &  &  &  0.0864 $\pm$ 0.0094 & 0.0688 $\pm$ 0.0074 & {\bf 0.0289} $\pm$ 0.0031 \\
\hline \hline

\end{tabular}
\label{table:KernelRegression} % is used to refer this table in the text
\end{table*}

\begin{table*}[htpb]
\vspace{-1.5em}
\footnotesize
\caption{Classification results with 3000 training/2000 testing.} % title of Table
\label{table:Classification}
\centering % used for centering table
\begin{tabular}{l l l l l l} % centered columns (4 columns)
\hline %inserts double horizontal lines

Dataset & d &  $h$NN  & $h$NN-GW & $h$NN-EGOP &  $h$NN-ReliefF \\
\hline \hline
Cover Type & 10 & 0.2301 $\pm$ 0.0104   & 0.2176 $\pm$ 0.0105 & 0.2197 $\pm$ 0.0077 & {\bf 0.1806} $\pm$ 0.0165 \\
Gamma & 10 &  0.1784 $\pm$ 0.0093 & 0.1721 $\pm$ 0.0082 &  {\bf 0.1658} $\pm$ 0.0076 & 0.1696 $\pm$ 0.0072 \\
Page Blocks & 10 & 0.0410 $\pm$ 0.0042   & 0.0387 $\pm$ 0.0085 &  {\bf 0.0383} $\pm$ 0.0047 & 0.0395 $\pm$ 0.0053 \\
Shuttle & 9  & 0.0821 $\pm$ 0.0095 &   0.0297 $\pm$ 0.0327 & {\bf 0.0123} $\pm$ 0.0041 & 0.1435 $\pm$ 0.0458 \\
Musk & 166 & 0.0458 $\pm$ 0.0057   & 0.0477 $\pm$ 0.0069  & {\bf 0.0360} $\pm$ 0.0037   & 0.0434 $\pm$ 0.0061 \\
IJCNN & 22 &  0.0523 $\pm$ 0.0043  & 0.0452 $\pm$ 0.0045  & {\bf 0.0401} $\pm$ 0.0039   & 0.0510 $\pm$ 0.0067\\
RNA & 8 &   0.1128 $\pm$ 0.0038 &  0.0710 $\pm$ 0.0048 &  {\bf 0.0664} $\pm$ 0.0064 &  0.1343 $\pm$ 0.0406\\
\vspace{-.3em}&\\
\hline
Dataset &  &  $k$NN  & $k$NN-GW & $k$NN-EGOP &  $k$NN-ReliefF \\
\hline \hline
Cover Type &  & 0.2279 $\pm$ 0.0091   & 0.2135 $\pm$ 0.0064 & 0.2161 $\pm$ 0.0061 & {\bf 0.1839} $\pm$ 0.0087 \\
Gamma &  &  0.1775 $\pm$ 0.0070 & 0.1680 $\pm$ 0.0075 &  0.1644 $\pm$ 0.0099 & {\bf 0.1623} $\pm$ 0.0063 \\
Page Blocks &  & 0.0349 $\pm$ 0.0042   & 0.0361 $\pm$ 0.0048 &  {\bf 0.0329} $\pm$ 0.0033 & 0.0347 $\pm$ 0.0038 \\
Shuttle &  & 0.0037 $\pm$ 0.0025 &   0.0024 $\pm$ 0.0016 & {\bf 0.0021} $\pm$ 0.0011 & 0.0028 $\pm$ 0.0021 \\
Musk &  & 0.2279 $\pm$ 0.0091   & 0.2135 $\pm$ 0.0064  & 0.2161 $\pm$ 0.0061   & {\bf 0.1839} $\pm$ 0.0087 \\
IJCNN &  &  0.0540 $\pm$ 0.0061  & 0.0459 $\pm$ 0.0058 & {\bf 0.0413} $\pm$ 0.0051   &  0.0535 $\pm$ 0.0080\\
RNA &  &   0.1042 $\pm$ 0.0063  &  0.0673 $\pm$ 0.0062&  {\bf 0.0627} $\pm$ 0.0057 &  0.0828 $\pm$ 0.0056\\
\hline \hline

\end{tabular}
%\label{table:epsClassification} % is used to refer this table in the text
\end{table*}

%*****************************************
%*****************************************
%*****************************************
%*****************************************
%*****************************************

\cleardoublepage
%\include{multiToC} % <--- just debug stuff, ignore for your documents
%************************************************
\chapter{The Expected Jacobian Outer Product}\label{ch:EJOP} % $\mathbb{ZNR}$
%************************************************

\begin{aside}{Outline}
	The Expected Gradient Outer Product, which was the focal point of the previous chapter, is an interesting operator that emerges naturally from the theory of multi-index regression and \emph{effective dimension reduction}. While it is straightforward to estimate while working with an unknown regression function $f: \mathbb{R}^d \to \mathbb{R}$, it is unclear how such an operator could be estimated when the unknown regression function is vector valued $f: \mathbb{R}^d \to \mathbb{R}^c$, while retaining the original multi-index motivation. In this chapter we give a generalization of the traditional EGOP for this case. We also show that a rough estimator for it remains statistically consistent under natural assumptions, while also providing gains in real world non-parametric classification tasks when used as a distance metric.
\end{aside}

In the previous chapter, we worked with the following object, namely, the Expected Gradient Outer Product (EGOP):
$$\expectation_{\mathbf{x}} G(\mathbf{x}) \triangleq \expectation_\mathbf{x}\paren{\nabla f(\mathbf{x}) \cdot \nabla f(\mathbf{x})^\top}$$
Where $f$ was an unknown regression function $f: \mathbb{R}^d \to \mathbb{R}$. The EGOP has the attractive property that it captures the average variation of $f$ in all directions. As has been discussed earlier, in practice the function $f$ might not vary equally along all coordinates: some features might be more important than the others, this being the motivation for variable selection methods as well as feature weighing methods such as \cite{DBLP:conf/nips/KpotufeB12}, \cite{GWJMLR}. More generally, even if all the features have a bearing toward predicting the output $y \in \mathbb{R}$, there might exist an unknown $k$ dimensional subspace on which $y$ \emph{effectively} depends upon. Such a \emph{relevant} subspace can be recovered by doing a singular value decomposition of the EGOP. Even more generally, as might be the case frequently in practice, even a \emph{relevant} subspace $P$ might not exist. However, the EGOP is still useful as $f$ is unlikely to vary equally in all directions: it can be employed to weight different directions according to their relevance. This was the motivation for using the EGOP as a metric in the previous chapter for non-parametric regression. Indeed, given the spectral decomposition $\mathbf{V}D\mathbf{V}^\top$ of the EGOP, we can transform the input $\mathbf{x}$ as $\mathbf{x} \mapsto \mathbf{D}^{1/2}\mathbf{V}^\top \mathbf{x}$. $\mathbf{V}$ rotates the data, while $\mathbf{D}^{1/2}$ weighs the coordinates. Using this transformation of the input was shown to improve regression performance on almost all datasets. 

\vspace{2mm}
\noindent
However, as was apparent, all the experimental results reported in Chapter \ref{ch:EGOP} were for regression and binary classification. Around the time of the publication of \cite{trivedi2014UAI}, it was unclear if a similar metric could be estimated for the multiclass case. We noticed this to uniformly be the case in the use of the EGOP throughout the multi-index regression literature (for instance see the the experiments reported by \cite{Mukherjee_wu2010learning}, which also involve regression and binary classification only).

\vspace{2mm}
\noindent
In this part of the dissertation, we generalize the EGOP such that it can also be estimated efficiently in the multi-class setting, and similarly be used to reweigh features in non-parametric multi-class classification tasks. Like in the case of the EGOP, we propose a rough estimator, which is cheap to estimate. We also prove that under similarly mild assumptions as for the EGOP, that it remains statistically consistent. We also provide experimental evidence that this generalization, which we call the expected Jacobian outer product (EJOP), can give significant improvements on classification error in real-world datasets, when used as the underlying metric in nonparametric classifiers. 

\vspace{2mm}
\noindent
Before we develop further on the EJOP, it might be instructive to first consider the EGOP for the case of binary classification. It might not be immediately obvious to the reader that the EGOP, which is well grounded for the case of nonparametric regression, carries through seamlessly for binary classification. We provide reasoning below that shows why this is the case, which also serves to motivate our approach to proposing an estimator for the EJOP.

\section{EGOP and Binary Classification}
To demonstrate that arguments used to motivate the EGOP in the case of nonparametric regression also work for binary classification, we first show that $k$-NN and $\epsilon$-NN are plug-in classifiers. The same reasoning also works for other nonparametric regression methods, but we keep ourselves to nearest neighbors. For the sake of completeness, we begin by a standard definition.

\begin{definition}[Bayes Classifier]
	Suppose $\eta(x) = \mathbb{P}(Y = 1 | X = \mathbf{x})$ and 
	\[   
	f(x) = 
	\begin{cases}
	1 &\quad\text{ if } \eta(x) > \frac{1}{2}\\
	0 &\quad\text{ if } \eta(x) \leq \frac{1}{2}\\
	\end{cases}
	\]
	Then $f(x)$ is called the Bayes classifier.
\end{definition}

\begin{definition}[Plug-in Classifier]
	Suppose $\hat{\eta}(x)$ is an estimate of $\eta(x)$ obtained from $\{(x_i, y_i)\}_{i=1}^{N}$, and
	\[   
	\hat{f}(x) = 
	\begin{cases}
	1 &\quad\text{ if } \hat{\eta}(x) > \frac{1}{2}\\
	0 &\quad\text{ if } \hat{\eta}(x) \leq \frac{1}{2}\\
	\end{cases}
	\]
	Then $\hat{f}(x)$ is called a plug-in classifier.
\end{definition}

\noindent
Consider $\hat{\eta}(x) = \sum_{i=1}^{N} w_i \mathbb{1}[y_i = 1]$ with $\sum_{i=1}^{N} w_i = 1$. If $C$ is the set of selected neighbors. Then, in the case of $k$-NN:

\[   
w_i(x) = 
\begin{cases}
\frac{1}{k} &\quad\text{ if } i \in C\\
0 &\quad\text{ otherwise } \\
\end{cases}
\]

\noindent
Likewise, in the case of $\epsilon$-NN
\[   
w_i(x) = 
\begin{cases}
\frac{1}{|\mathcal{B}(x, \epsilon)|} &\quad\text{ if } i \in \mathcal{B}(x, \epsilon)\\
0 &\quad\text{ otherwise } \\
\end{cases}
\]

\begin{proposition}
	$k$-NN and $\epsilon$-NN are plug-in classifiers
\end{proposition}

\begin{proof}
	The plug-in classifier might be rewritten as: 
	\begin{align*}
	\hat{f}(x)  
	&= \mathbb{1} \Bigg(\hat{\eta}(x) > \frac{1}{2} \Bigg)u 
	\\ & = \mathbb{1} \Bigg(\sum_{i=1}^{N} w_i \mathbb{1}[y_1 = 1] > \frac{1}{2} \Bigg)
	\\ & = \mathbb{1} \Bigg(\sum_{i=1}^{N} w_i (2\mathbb{1}[y_1 = 1] - 1) > 0 \Bigg)
	\\ & = \mathbb{1} \Bigg(\sum_{i=1}^{N} w_i (\mathbb{1}[y_1 = 1] - \mathbb{1}[y_1 = 0]) > 0 \Bigg)
	\\ & = \mathbb{1} \Bigg(\sum_{i=1}^{N} w_i \mathbb{1}[y_1 = 1] > \sum_{i=1}^{N} w_i \mathbb{1}[y_1 = 0] \Bigg)		
	\end{align*}
\end{proof}

\noindent
While laying out this trivial reasoning might seem unnecessarily excessive, the main message that we want to impress upon the reader is that $k$-NN and $\epsilon$-NN are plug-in classifiers. Note that, $\hat{\eta}(x)$, which is the plug-in classifier is a regression estimate	of the Bayes classifier $\eta(x)$, and it is well known that the 0-1 classification error of the plug-in methods is related to the regression estimate (see Devroye \emph{et al.} \cite{Lugosi}). Thus the reasoning used for non-parametric regression in the case of gradient weights\cite{DBLP:conf/nips/KpotufeB12}, \cite{GWJMLR}, EGOP \cite{trivedi2014UAI}, \cite{Mukherjee_wu2010learning} etc., also carries to the case of binary classification; one can just use $\hat{\eta}$ to find the derivatives.

\section{The Multiclass Case}
For the multi-class case, we can consider $$\hat{\eta}_k (x) = \sum_{i=1}^{N} w_i \mathbb{1}[y_i = k]; \text{   where   } \sum_{i=1}^{N} w_i = 1 \text{  and  } k \in \{1, \dots, r\}$$ and define $\hat{f}(x) = \max_k \hat{\eta}_k(x)$. 

\vspace{2mm}
\noindent
We can consider using $\hat{f}(x)$ for finding derivatives, but that is prevented by the appearance of the max. Alternatively, we could consider a vector valued function ($c$ being the number of classes) $$\tilde{f}(x) = [\hat{\eta}_1, \dots, \hat{\eta}_c]$$ and then use the differences to find the derivatives. The latter approach seems in direct analogy to the case of binary classification, thus we use it to define the Jacobian Outer Product. Note that the properties of this plug-in and whether it is similar to the standard plug-in defined above is beyond the scope of this chapter (see Devroye \emph{et al.} \cite{Lugosi}). We simply content ourselves with using it to define the EJOP, and this intuition is borne out by being able to use it to prove a consistency result akin to the EGOP. We now conclude these meanderings to better motivate the EJOP, and proceed to define it more formally in the following section.

\section{The Expected Jacobian Outerproduct}
Recall that in high dimensional classification problems over $\mathbb{R}^d$, the unknown (multinomial regression) function $f$ could be considered to be a vector-valued function mapping to a probability simplex $\mathbb{S}^c = \{ \mathbf{y} \in \mathbb{R}^c | \forall i \textbf{  } y_i \geq 0, \mathbf{y}^T \mathbf{1} = 1 \}$, where $c$ is the number of classes.  Or, more concisely, $f: \mathbb{R}^d \to \mathbb{S}^c$. The prediction for some point $\mathbf{x}$ is then given by: $y = \argmax_{i = 1, \dots, c} f_i(\mathbf{x})$. 

\vspace{2mm}
\noindent
For $f$, at point $\mathbf{x}$, we can define the Jacobian as:

$$ \displaystyle \mathbf{J}_{f}(\mathbf{x}) = \begin{bmatrix} \frac{\partial f_1(\mathbf{x})}{\partial x_1} & \frac{\partial f_2(\mathbf{x})}{\partial x_1} & \dots & \frac{\partial f_c(\mathbf{x})}{\partial x_1} \\ \vdots & \vdots & \ddots & \vdots \\ \frac{\partial f_1(\mathbf{x})}{\partial x_d} & \frac{\partial f_2 (\mathbf{x})}{\partial x_d} & \dots & \frac{\partial f_c (\mathbf{x})}{\partial x_d}\end{bmatrix}$$ 

\noindent
We are interested in the quantity $$\mathbb{E}_X G(X) = \mathbf{J}_{f}(\mathbf{x}) \mathbf{J}_{f}(\mathbf{x})^T$$ Let $f_n$ be an initial estimate of $f$, for which we use a kernel estimate, then for the $(i,j)^{th}$ element of $\mathbf{J}_f(\mathbf{x})$, we can use the following rough estimate:
$$\displaystyle \Delta_{t,i,j} f_n(\mathbf{x}) = \frac{f_{n,i} (\mathbf{x} + t \mathbf{e}_j) - f_{n,i} (\mathbf{x} - t \mathbf{e}_j) }{2t}, t > 0$$

\noindent
Let $\mathbf{J}_n (\mathbf{x})$ be the Jacobian estimate at $\mathbf{x}$. The Jacobian outer product is then estimated as $\mathbb{E}_n \mathbf{J}_n (\mathbf{x}) \mathbf{J}_n (\mathbf{x})^T$, which is the empirical average of $\mathbf{J}_n (\mathbf{x})\mathbf{J}_n (\mathbf{x})^T$. 

\vspace{2mm}
\noindent
\textbf{Note:} Many of the assumptions and notation used overlap with that employed in chapter \ref{ch:EGOP}. We introduce new notation as needed, and if occasionally dictated for ease of exposition, redefine some term already defined in chapter \ref{ch:EGOP}. 

\subsection{Function Estimate}

First, we need to specify the function estimate, that is used both for the theoretical analysis, and the experiments reported. 

\vspace{2mm}
\noindent
Again, considering $c$ to be the number of classes, let the $c$ dimensional vector valued function estimate be denoted by $\bar{f}_{n,h} (\mathbf{x})$, such that $\bar{f}_{n,h} (\mathbf{x}) \in \mathbb{S}^c$
$$ \bar{f}_{n,h} (\mathbf{x}) = [ \bar{f}_{n,h,1} (\mathbf{x}) ,\dots, \bar{f}_{n,h,c} (\mathbf{x})] \text{ s.t. } \bar{f}_{n,h,i} (\mathbf{x}) > 0 \text { and } \sum_i \bar{f}_{n,h,i} (\mathbf{x}) = 1$$ 

\noindent
The prediction in that case is given by:

 $$ \hat{y} = \arg \max_i \bar{f}_{n,h,i} (\mathbf{x})$$. 

\noindent
We use the following kernel estimate: $\bar{f}_{n,h,c} (\mathbf{x}) = \sum_i w_i \mathbbm{1} \{ Y_i = c\}$, where:

\begin{align*}
w_i(\mathbf{x}) &= \frac{K(\|\mathbf{x} - \mathbf{x}_i\| / h)}{\sum_j K(\|\mathbf{x} - \mathbf{x}_j\| / h)} \text{ if } B(\mathbf{x},h) \cap \mathbf{x} \neq \phi, \\
w_i(\mathbf{x}) &= \frac{1}{n} \text{ otherwise }
\end{align*}

\noindent
Note that for $k$-NN, $w_i(\mathbf{x}) = \frac{1}{k}$ and for $\epsilon$-NN, $w_i(\mathbf{x}) = \frac{1}{|B(\mathbf{x},h)|}$

\vspace{2mm}
\noindent
While estimating gradients, we actually work with the softmaxed output 
$$ \bar{f}_{n,h,i} (\mathbf{x}) = \frac{\exp (\bar{f}_{n,h,i} (\mathbf{x})) }{\sum_j \exp (\bar{f}_{n,h,j} (\mathbf{x}))}   $$

\vspace{2mm}
\noindent
Additionally, in the experiments we use a temperature term in the softmax for affording ease in gradient computation. But we omit this aspect from the discussion to keep the theoretical analysis simple, in any case, appearance of the temperature term does not affect any of the discussion to follow. 

%\vspace{2mm}
%\noindent
%
%Applications: Use of $\mathbf{J}_n (\mathbf{x}) \mathbf{J}_n (\mathbf{x})^T$ for metric transformation i.e. since $\mathbf{J}_n (\mathbf{x})\mathbf{J}_n (\mathbf{x})^T = \mathbf{V} \mathbf{D}\mathbf{V}^T$, we can transform $\mathbf{x} \to \mathbf{D}^{\frac{1}{2}}\mathbf{V}^T \mathbf{x}$ to show improvements on $k$-NN classification over the usual euclidean distance (see experiments). We can also use the matrix $\mathbf{V}$ for class-aware dimensionality reduction (see experiments). 

\subsection{Note on the nomenclature}
A natural question to ask is regarding the use of the name: Expected Jacobian Outer Product (EJOP), as compared to simply the Expected Gradient Outer Product (EGOP), after all we still find gradients, even in the multiclass case. We simply use different terminology to distinguish the two cases, especially given the lack of work on \emph{effective dimension reduction} and multi-index regression for multinomial regression.

\section{Notation and Setup}
For a vector $x\in \mathbb{R}^d$, we denote the euclidean norm as $\|x\|$. For a matrix, we denote the spectral norm, which is the largest singular value of the matrix $\sigma_{\max}(A)$ as $\|A\|_2$. The column space of a matrix $A \in R^{n \times m}$ is denoted as $\text{im}(A)$ where $\text{im}(A) = \{\y \in \R^n | \y = A\x \text{ for some } \x \in \R^m\}$, and $\text{ker}(A)$ is used to denote the null space of matrix $A \in R^{n \times m}$: $\text{ker}(A) = \{\x \in \R^m | A\x = 0 \}$. We use $A \circ B$ to denote the Hadamard product of matrices $A$ and $B$.

\vspace{2mm}
\noindent
Let the estimated nonparametric function be $f_{n,h,c}(\x) = \sum_{i} \omega_i(\x) \mathbb{1} \{y_i = c\}$, and $\tilde{f}_{n,h,c}(\x) = \sum_{i} \omega_i(\x) \P(y_i = c|x_i)$. Our estimated gradient at dimension $i$ is given as $$\displaystyle \Delta_{t,i}f_{n,h,c}(\x) = \frac{f_{n,h,c}(\x + t e_i) - f_{n,h,c}(\x - t e_i)}{2t},$$ and the estimated and true gradients for class $c$ are given as: 
$$\hat{\nabla} f_{n,h,c}(\x) = \begin{bmatrix}
\Delta_{t,1}f_{n,h,c}(\x) \cdot \mathbb{1}_{A_{n,1} (\x)} \\
\Delta_{t,2}f_{n,h,c}(\x) \cdot \mathbb{1}_{A_{n,2} (\x)} \\
...\\
\Delta_{t,d}f_{n,h,c}(\x) \cdot \mathbb{1}_{A_{n,d} (\x)}
\end{bmatrix}, \hat{\nabla} f_c(\x) = \begin{bmatrix}
\Delta_{t,1} f_c(\x) \cdot \mathbb{1}_{A_{n,1} (\x)} \\
\Delta_{t,2} f_c(\x) \cdot \mathbb{1}_{A_{n,2} (\x)} \\
...\\
\Delta_{t,d} f_c(\x) \cdot \mathbb{1}_{A_{n,d} (\x)}
\end{bmatrix}$$

\vspace{2mm}
\noindent
Where $A_{n,i}(X)$ is the event that enough samples contribute to the estimate $\Delta_{t,i} f_{n,h}(X)$: $$A_{n,i}(X) \equiv \min_{\{t,-t\}} \mu_n(B(X+s e_i, h/2)) \geq \frac{2d \ln 2n + ln(4/\delta)}{n}$$ and likewise $$A_{i}(X) \equiv \min_{\{t,-t\}} \mu(B(X+s e_i, h/2)) \geq 3 \cdot \frac{2d \ln 2n + ln(4/\delta)}{n}$$ note that $\mu_n$, $\mu$ are empirical mass and mass of a ball, respectively. 

\vspace{2mm}
\noindent
We denote indicators of the events $A_{n,i}(X)$ and $A_{i}(X)$  in the following form

$\mathbb{I}_n(x) = \begin{bmatrix}
\mathbb{1}_{A_{n,1} (\x)} \\
\mathbb{1}_{A_{n,2} (\x)} \\
...\\
\mathbb{1}_{A_{n,d} (\x)}
\end{bmatrix}$, $\overline{\mathbb{I}_n(x)} = \begin{bmatrix}
\mathbb{1}_{\bar{A}_{n,1} (\x)} \\
\mathbb{1}_{\bar{A}_{n,2} (\x)} \\
...\\
\mathbb{1}_{\bar{A}_{n,d} (\x)}
\end{bmatrix}$, $\mathbb{I}(x) = \begin{bmatrix}
\mathbb{1}_{A_{1} (\x)} \\
\mathbb{1}_{A_{2} (\x)} \\
...\\
\mathbb{1}_{A_{d} (\x)}
\end{bmatrix}$, $\overline{\mathbb{I}(x)} = \begin{bmatrix}
\mathbb{1}_{\bar{A}_{1} (\x)} \\
\mathbb{1}_{\bar{A}_{2} (\x)} \\
...\\
\mathbb{1}_{\bar{A}_{d} (\x)}
\end{bmatrix}$.

\vspace{2mm}
\noindent
Let the Jacobian matrix $\mathbf{J}_f(\x) \in \R^{d \times c}$ to be
$$\mathbf{J}_{f}(\x) = 
\begin{bmatrix} 
\frac{\partial f_1(\x)}{\partial x_1} & \frac{\partial f_2(\x)}{\partial x_1} & \dots & \frac{\partial f_c(\x)}{\partial x_1} \\ 
\vdots & \vdots & \ddots & \vdots \\ 
\frac{\partial f_1(\x)}{\partial x_d} & \frac{\partial f_2 (\x)}{\partial x_d} & \dots & \frac{\partial f_c (\x)}{\partial x_d}
\end{bmatrix}$$
where $c$ is the number of classes. And $f_k(x) = \P(y=k|x), \forall k \in [c]$ represents the conditional distribution of the class labels.

\vspace{2mm}
\noindent
Then the Jacobian outer product matrix $G(\x)$ is $G(\x) = \mathbf{J}_f(\x) \mathbf{J}_f(\x)^T$.The estimated Jacobian matrix is
$$\hat{\mathbf{J}}_{f}(\x) = 
\begin{bmatrix} 
\hat{\nabla} f_{n,h,1}(\x) & \hat{\nabla} f_{n,h,2}(\x) & \dots & \hat{\nabla} f_{n,h,k}(\x)
\end{bmatrix}$$ the estimated Jacobian product matrix is denoted $\hat{G}(\x) = \hat{\mathbf{J}}_f(\x) \hat{\mathbf{J}}_f(\x)^T$.

\subsection{Assumptions}
The assumptions are the same as in \ref{sec:Assumptions}, with the following modifications: first to the bounded gradient assumption such that it extends to each class. 
\begin{itemize}
	\item Bounded Gradient: $\|\nabla f_k(\x)\|_2 \leq R, \forall \x \in \Xx, k \in [c]$.
\end{itemize}
\vspace{2mm}
\noindent
Secondly, we modify the assumption on the modulus of continuity of $\nabla f_k$ similarly

\noindent
Let $\epsilon_{t,k,i} = \sup_{\x \in \Xx, s \in [-t,t]} \left| \frac{\partial f_k(\x)}{\partial x_i} - \frac{\partial f_k(\x + s e_i)}{\partial x_i} \right|$ and $\epsilon_{t,i} = \max_{k} \epsilon_{t,c,i}$, define the $(t,i)$-boundary of $\mathcal{X}$ as $\partial_{t,i}(\Xx) = \{\x: \{\x+ t e_i, x- te_i\} \not\subseteq \Xx\}$. When $\mu$ has continues density on $\mathcal{X}$ and $\nabla f_k$ is uniformly continuous on $\Xx + B(0,\tau)$, we have $\mu(\partial_{t,i}(\Xx)) \xrightarrow{t \rightarrow 0} 0$ and $\epsilon_{t,k,i} \xrightarrow{t \rightarrow 0} 0$.

\section{Consistency of Estimator \texorpdfstring{$ \e_n \hat{G}(X)$} {} %
	of the Jacobian Outerproduct \texorpdfstring{$\e_X G(X)$}{} %
}

To show that the estimator $ \e_n \hat{G}(X)$ is consistent, we proceed to bound $ \| \e_n \hat{G}(X) - \e_X G(X) \|$ for finite $n$, which is encapsulated in the theorem that follows. There are two main difficulties in the proof, which are addressed by a sequence of lemmas. One has to do with the fact that the gradient estimate at any point depends on all other points, and second, having gradient estimates for $c$ classes.

\begin{aside}{Main Result}
\begin{theorem}
	Let $t + h \leq \tau$, and let $0 \leq \delta \leq 1$. There exist $C = C(\mu,K(\cdot))$ and $N = N(\mu)$ such that the following holds with probability at least $1 - 2\delta$. Define $A(n) = \sqrt{Cd \cdot \log(kn/\delta)} \cdot 0.25/\log^2(n/\delta)$. Let $n \geq N$, we have:
	\begin{small}
		\begin{eqnarray*}
			&&\| \e_n \hat{G}(X)] - \e_X G(X) \|_2 \leq \frac{6 R^2}{\sqrt{n}}\left(\sqrt{\ln d} + \sqrt{\ln \frac{1}{\delta}} \right) + k \left(3R + \sqrt{\sum_{i \in [d]} \epsilon^2_{t,i}} + \sqrt{d}\left(\frac{hR + 1}{t} \right) \right)\\
			&&\left[\frac{\sqrt{d}}{t} \sqrt{\frac{A(n)}{nh^d} + h^2 R^2} + R \left( \sqrt{\frac{d \ln \frac{d}{\delta}}{2n}} + \sqrt{\sum_{i \in [d]} \mu^2(\partial_{t,i}(\mathcal{X}))} \right) + \sqrt{\sum_{i \in [d]} \epsilon^2_{t,i}} \right]
		\end{eqnarray*}
	\end{small}
	\label{thm:main}
\end{theorem}
\end{aside}
\begin{proof}
	We begin with the following decomposition:
	\begin{eqnarray*}
		\| \e_n \hat{G}(X) - \e_X G(X) \|_2 \leq \| \e_n G(X) - \e_X G(X) \|_2 + \| \e_n \hat{G}(X) - \e_n G(X) \|_2
	\end{eqnarray*}
	The first term on the right hand side i.e. $\| \e_n G(X) - \e_X G(X) \|_2$ is bounded using Lemma \ref{lemma:rm}; by using Lemma \ref{lemma:matrix} we bound the second term $\| \e_n \hat{G}(X)  - \e_n G(X) \|_2$, this is done with respect to $\sum_{k \in [c]} \e_n \|\nabla f_k(X) - \hat{\nabla} f_{n,h,k}(X)\|_2$; therefore we need to bound $\sum_{k \in [c]} \e_n \|\nabla f_k(X) - \hat{\nabla} f_{n,h,k}(X)\|_2$, which is done by employing Theorem \ref{thm:f} which concludes the proof.
\end{proof}

\noindent
{\it Remark.}
The theorem implies consistency for $t \xrightarrow{n \rightarrow \infty} 0$, $h \xrightarrow{n \rightarrow \infty} 0$, $h/t^2 \xrightarrow{n \rightarrow \infty} 0$, and $(n / \log n)h^d t^4 \xrightarrow{n \rightarrow \infty} \infty$, this is satisfied for many settings, for example $t \propto h^{1/4}$, $h \propto \frac{1}{\ln n}$.

\subsection{Bounding \texorpdfstring{$\| \e_n G(X) - \e_X G(X) \|_2$}{} %
}
To bound this term, like in the case of the EGOP, we use the following random matrix concentration result.
\begin{lemma}
	\cite{randommatrix,kakadenotes}. For the random matrix $\X \in \R^{d_1 \times d_2}$ with bounded spectral norm $\|\X\|_2 \leq M$, let $d = \min\{d_1,d_2\}$, and $\X_1,\X_2,...,\X_n$ are i.i.d. samples, with probability at least $1 - \delta$, we have
	\begin{eqnarray*}
		\left\| \frac{1}{n} \sum_{i=1}^n \X_i - \e \X \right\|_2 \leq \frac{6 M}{\sqrt{n}} \left(\sqrt{\ln d} + \sqrt{\ln \frac{1}{\delta}} \right)
	\end{eqnarray*}
\end{lemma}
\vspace{2mm}
\noindent
Recall the bounded gradient assumption $\| G(X) \|_2 = \| \nabla f(X) \|_2^2 \leq R^2$. Using this assumption we can apply the above lemma to i.i.d matrices $G(X), X \in \mathbf{X}$, yielding the following lemma. 
\begin{lemma}
	With probability at least $1 - \delta$
	\begin{eqnarray*}
		\| \e_n G(X) - \e_X G(X) \|_2 \leq \frac{6 R^2}{\sqrt{n}} \left(\sqrt{\ln d} + \sqrt{\ln \frac{1}{\delta}} \right)
	\end{eqnarray*}
	\label{lemma:rm}
\end{lemma} 

\vspace{2mm}
\noindent
Next we proceed to bound the second term in the decomposition mentioned in the proof of theorem \ref{thm:main}.

\subsection{Bounding \texorpdfstring{$\| \e_n \hat{G}(X) - \e_n G(X) \|_2$}{} %
}
A first bound is provided by the following lemma:
\begin{lemma}
	Exist constant $c$, with probability at least $1 - \delta$:
	\begin{align*}
	\| \e_n \hat{G}(X)  - \e_n G(X) \|_2 \leq \sum_{k \in [c]} \e_n \|\nabla f_k(X) - \hat{\nabla} f_{n,h,k}(X)\|_2 \cdot \max_{x \in \mathbf{X}} \|\nabla f_k(X) + \hat{\nabla} f_{n,h,k}(X)\|_2
	\end{align*}
	\label{lemma:matrix}
\end{lemma}
\begin{proof}
	First we can write the term on the l.h.s in terms of the gradients for each class:
	\begin{align*}
	\| \e_n \hat{G}(X)  - \e_n G(X) \|_2 =& \| \e_n [\hat{G}(X) - G(X)] \|_2 \\
	=& \left\| \sum_{k \in [c]} \e_n [  \nabla f_k(X) \cdot \nabla f_k(X)^T - \hat{\nabla} f_{n,h,k}(X) \cdot   \hat{\nabla} f_{n,h,k}(X)^T ] \right\|_2 \\
	\leq&
	\sum_{k \in [c]}  \left\| \e_n [  \nabla f_k(X) \cdot \nabla f_k(X)^T - \hat{\nabla} f_{n,h,k}(X) \cdot   \hat{\nabla} f_{n,h,k}(X)^T ] \right\|_2 \\
	\end{align*}
	\noindent
	next, we notice that $\nabla f_k(\x) \cdot \nabla f_k(\x)^T - \hat{\nabla} f_{n,h,k}(\x) \cdot   \hat{\nabla} f_{n,h,k}(\x)^T$ can be rewritten as:
	
	\begin{align*}
	\nabla f_k(\x) \cdot \nabla f_k(\x)^T - \hat{\nabla} f_{n,h,k}(\x) \cdot   \hat{\nabla} f_{n,h,k}(\x)^T =& \frac{1}{2} \cdot (\nabla f_k(\x) + \hat{\nabla} f_{n,h,k}(\x)) \cdot (\nabla f_k(\x) - \hat{\nabla} f_{n,h,k}(\x))^T \\
	+ &\frac{1}{2} \cdot (\nabla f_k(\x) - \hat{\nabla} f_{n,h,k}(\x)) \cdot (\nabla f_k(\x) + \hat{\nabla} f_{n,h,k}(\x))^T 
	\end{align*}	
	
	\vspace{2mm}
	\noindent
	Using this yields:
	\begin{align*}
	\| \e_n \hat{G}(X)  - \e_n G(X) \|_2 \leq& \frac{1}{2} \sum_{k \in [c]} \|\e_n [  (\nabla f_k(X) + \hat{\nabla} f_{n,h,k}(X)) \cdot (\nabla f_k(X) - \hat{\nabla} f_{n,h,k}(X))^T ] \|_2 \\
	+& \frac{1}{2} \sum_{k \in [c]} \|\e_n [ (\nabla f_k(X) - \hat{\nabla} f_{n,h,k}(X)) \cdot (\nabla f_k(X) + \hat{\nabla} f_{n,h,k}(X))^T ]\|_2 \\
	=& \sum_{k \in [c]} \| \e_n [(\nabla f_k(X) - \hat{\nabla} f_{n,h,k}(X)) \cdot (\nabla f_k(X) + \hat{\nabla} f_{n,h,k}(X))^T]\|_2
	\end{align*}
	By using Jensen's inequality, we have:
	\begin{multline*}
	\noindent
	\e_n [(\nabla f_k(X) - \hat{\nabla} f_{n,h,k}(X)) \cdot (\nabla f_ck(X) + \hat{\nabla} f_{n,h,k}(X))^T]\|_2  \leq \\
	\e_n \| (\nabla f_k(X) - \hat{\nabla} f_{n,h,k}(X)) \cdot (\nabla f_k(X) + \hat{\nabla} f_{n,h,k}(X))^T \|_2
	\end{multline*}
	\vspace{2mm}
	\noindent
	combining the above, gives us the following bound on $\| \e_n \hat{G}(X)  - \e_n G(X) \|_2$
	\begin{align*}
	\| \e_n \hat{G}(X)  - \e_n G(X) \|_2 \leq& \sum_{k \in [c]} \e_n \|  (\nabla f_k(X) - \hat{\nabla} f_{n,h,k}(X)) \cdot (\nabla f_k(X) + \hat{\nabla} f_{n,h,k}(X))^T\|_2 \\
	=& \sum_{k \in [c]} \e_n \|\nabla f_k(X) - \hat{\nabla} f_{n,h,k}(X)\|_2 \cdot \|\nabla f_k(X) + \hat{\nabla} f_{n,h,k}(X)\|_2. \\
	\leq& \sum_{k \in [c]} \e_n \|\nabla f_k(X) - \hat{\nabla} f_{n,h,k}(X)\|_2 \cdot \max_{X \in \mathbf{X}} \|\nabla f_k(X) + \hat{\nabla} f_{n,h,k}(X)\|_2.
	\end{align*}
\end{proof}

\vspace{2mm}
\noindent
The above bound has a dependence on $\| \nabla f_k(X) + \hat{\nabla} f_{n,h,k}(X) \|_2$, which we now proceed to bound below:

\subsection{Bounding \texorpdfstring{$\| \nabla f_k(X) + \hat{\nabla} f_{n,h,k}(X) \|_2$}{} %
}
We first bound the max term, by the following lemma:

\begin{lemma}
	$\forall c \in [k]$, we have
	\begin{eqnarray*}
		\max_{X \in \mathbf{X}} \| \nabla f_k(X) + \hat{\nabla} f_{n,h,k}(X) \|_2 \leq 3R + \sqrt{\sum_{i \in [d]} \epsilon^2_{t,i}} + \sqrt{d}(\frac{hR + 1}{t})
	\end{eqnarray*}
\end{lemma}
\begin{proof}
	$\forall x \in \mathbf{X}$, we have
	\begin{eqnarray*}
		\| \nabla f_k(\x) + \hat{\nabla} f_{n,h,k}(\x) \|_2 &\leq& \| \nabla f_k(\x) \|_2 + \| \hat{\nabla} f_{n,h,k}(\x) \|_2 \\
		&\leq& 2 \|\nabla f_k(\x) \|_2 + \|\nabla f_k(\x) - \hat{\nabla} f_{n,h,k}(\x)\|_2 \\
		&\leq& 2R + \|\nabla f_k(\x) - \hat{\nabla} f_k(\x) \|_2 + \|\hat{\nabla} f_k(\x) - \hat{\nabla} f_{n,h,k}(\x)\|_2 \\
	\end{eqnarray*}
	Next, we adopt the steps as in the proof for Lemma \ref{lemma7}, and get the following bound: 
	\[
	\|\hat{\nabla} f_k(\x) - \hat{\nabla} f_{n,h,k}(\x)\|_2 \leq \sqrt{\sum_{i \in [d]} (|\Delta_{t,i}f_{n,h,k}(\x) - \Delta_{t,i} f_k(\x)| \cdot  \mathbb{1}_{A_{n,i}(\x)})^2},
	\]
	this is because
	\begin{align*}
	|\Delta_{t,i}f_{n,h,k}(\x) - \Delta_{t,i} f_k(\x)| \cdot  \mathbb{1}_{A_{n,i}(\x)} \leq& \frac{1}{t} 
	\max_{s \in \{-t,t\}} |\tilde{f}_{n,h,k}(\x + s e_i) - f_k(\x + s e_i)| \cdot  \mathbb{1}_{A_{n,i}(\x)} \\
	&+ \frac{1}{t} \max_{s \in \{-t,t\}} |\tilde{f}_{n,h,k}(\x + s e_i) - f_{n,h,k}(\x + s e_i)| \cdot  \mathbb{1}_{A_{n,i}(\x)},
	\end{align*}
	we also know that
	\[
	\max_{s \in \{-t,t\}} |\tilde{f}_{n,h,k}(X + s e_i) - f_{n,h,k}(X + s e_i)|
	\leq 1.
	\] 
	Thus we obtain the following bound:
	\begin{align*}
	\|\hat{\nabla} f_k(X) - \hat{\nabla} f_{n,h,k}(X)\|_2 \leq \sqrt{d}(\frac{hR + 1}{t})
	\end{align*}
	While, we also have that
	\begin{align*}
	\|\nabla f_k(X) - \hat{\nabla} f_k(X) \|_2 \leq& \|\nabla f_k(X) \circ \mathbf{I}_n(X) - \hat{\nabla} f_k(X)\|_2 + \|\nabla f_k(X) \circ \overline{\mathbf{I}_n(X)}\|_2 \\
	\leq& R + \sqrt{\sum_{i \in [d]} \epsilon^2_{t,i}}
	\end{align*}
	Combining the above completes the proof
\end{proof} 

\vspace{2mm}
\noindent
Next we need to bound $\e_n \|\nabla f_k(X) - \hat{\nabla} f_{n,h,k}(X)\|_2$, which we do so in the next subsection:

\subsection{Bound on \texorpdfstring{$\e_n \|\nabla f_k(X) - \hat{\nabla} f_{n,h,k}(X)\|_2$}{} %
}

We first decompose $\e_n \|\nabla f_c(X) - \hat{\nabla} f_{n,h,k}(X)\|_2$ as:
\begin{multline*}
	\e_n \|\nabla f_k(X) - \hat{\nabla} f_{n,h,k}(X)\|_2 \leq \\
	\e_n \|\nabla f_k(X) - \hat{\nabla} f_k(X)\|_2 + \e_n \|\hat{\nabla} f_k(X) - \hat{\nabla} f_{n,h,k}(X)\|_2
\end{multline*}
the first term in the r.h.s of the above i.e. $\e_n \|\nabla f_k(X) - \hat{\nabla} f_k(X)\|_2$ can in turn be decomposed as:
\begin{multline*}
	\e_n \|\nabla f_k(X) - \hat{\nabla} f_k(X)\|_2 \leq  \\ \e_n \|\nabla f_k(X) \circ \mathbb{I}_n(X) - \hat{\nabla} f_k(X)\|_2 + \e_n \|\nabla f_k(X) \circ \overline{\mathbb{I}_n(X)}\|_2
\end{multline*}

\vspace{2mm}
\noindent
We need to bound both terms that appear on the r.h.s of the above, which we do so in the next two subsections, starting with the second term. 

\subsubsection{Bounding $\e_n \|\nabla f_k(X) \circ \overline{\mathbb{I}_n(X)}\|_2$}

The bound is encapsulated in the following lemma:
\begin{lemma}
	With probability at least $1 - \delta$ over the choice of $X$:
	\begin{eqnarray*}
		\e_n \|\nabla f_k(X) \circ \overline{\mathbb{I}_n(X)}\|_2 \leq R \left( \sqrt{\frac{d \ln \frac{d}{\delta}}{2n}} + \sqrt{\sum_{i \in [d]} \mu^2(\partial_{t,i}(\mathcal{X}))} \right)
	\end{eqnarray*}
	\label{lemma1}
\end{lemma}
\begin{proof}
	We begin by recalling the bounded gradient assumption: $\|\nabla f(\x)\|_2 \leq R$, using which we get
	\begin{eqnarray*}
		\e_n \|\nabla f(X) \circ \overline{\mathbb{I}_n(X)}\|_2 \leq R \e_n \|\overline{\mathbb{I}_n(X)}\|_2
	\end{eqnarray*}
	By relative VC bounds \cite{doi:10.1137/1116025}, if we set $\alpha_n = \frac{2d \ln 2n + \ln(4/\delta)}{n}$, then with probability at least $1 - \delta$ over the choice of $X$, for all balls $B \in R^d$ we have $\mu(B) \leq \mu_n(B) + \sqrt{\mu_n(B) \alpha_n} + \alpha_n$. Thus, with probability at least $1 - \delta$, $\forall i \in [d]$, $\bar{A}_{n,i}(X) \Rightarrow \bar{A}_{i}(X)$. Moreover, since $\|\overline{\mathbb{I}(X)}\|_2 \leq \sqrt{d}$, then by Hoeffding's inequality,
	\begin{eqnarray*}
		\mathbb{P}(\e_n \|\overline{\mathbb{I}(X)}\|_2 - \e_X \|\overline{\mathbb{I}(X)}\|_2 \geq \epsilon) \leq e^{-\frac{2n \epsilon^2}{d}}
	\end{eqnarray*}
	applying the union bound, we have the following with probability at least $1 - \delta$
	\begin{eqnarray*}
		\e_n \|\overline{\mathbb{I}_n(X)}\|_2 \leq \e_n \|\overline{\mathbb{I}(X)}\|_2
		\leq \e_X \|\overline{\mathbf{I}_n(X)}\|_2 + \sqrt{\frac{d \ln \frac{d}{\delta}}{2n}}
	\end{eqnarray*}
	But note that we have: $$\e_X \mathbb{1}_{\bar{A}_i(X)} \leq \e_X[\mathbb{1}_{\bar{A}_i(X)} | X \in \mathcal{X} \backslash \partial_{t,i}(\mathcal{X})] + \mu(\partial_{t,i}(\mathcal{X}))$$
	\noindent
	to see why this is true observe that $\e_X[\mathbf{1}_{\bar{A}_i(X)} | X \in \mathcal{X} \backslash \partial_{t,i}(\mathcal{X})] = 0$ because $\mu(B(x+se_i,h/2)) \geq C_{\mu} (h/2)^d \geq 3 \alpha$ when we set $h \geq (\log^2(n/\delta)/n)^{1/d}$. 
	
	\vspace{2mm}
	\noindent
	So, we have:
	\begin{eqnarray*}
		\e_X \|\overline{\mathbb{I}_n(X)}\|_2 \leq \sqrt{\sum_{i \in [d]} \mu^2(\partial_{t,i}(\mathcal{X}))}
	\end{eqnarray*}
	Thus with probability at least $1 - \delta$, we obtain the following:
	\begin{eqnarray*}
		\e_n \|\nabla f_k(X) \circ \overline{\mathbf{I}_n(X)}\|_2 \leq R \left( \sqrt{\frac{d \ln \frac{d}{\delta}}{2n}} + \sqrt{\sum_{i \in [d]} \mu^2(\partial_{t,i}(\mathcal{X}))} \right)
	\end{eqnarray*}
\end{proof}

\vspace{2mm}
\noindent
Next we need to bound the first term that appeared on the r.h.s. of the decomposition of $\e_n \|\nabla f_k(X) - \hat{\nabla} f_k(X)\|_2$, reproduced below for ease of exposition:

\begin{multline*}
\e_n \|\nabla f_k(X) - \hat{\nabla} f_k(X)\|_2 \leq  \\ \e_n \|\nabla f_k(X) \circ \mathbb{I}_n(X) - \hat{\nabla} f_k(X)\|_2 + \e_n \|\nabla f_k(X) \circ \overline{\mathbb{I}_n(X)}\|_2
\end{multline*}

\subsubsection{Bounding  \texorpdfstring{$\e_n \|\nabla f_k(X) \circ \mathbb{I}_n(X) - \hat{\nabla} f_k(X)\|_2$}{} %
} 
This bound is encapsulated in the following lemma
\begin{lemma}
	We have
	\begin{eqnarray*}
		\e_n \|\nabla f_k(X) \circ \mathbb{I}_n(X) - \hat{\nabla} f_k(X)\|_2 \leq \sqrt{\sum_{i \in [d]} \epsilon^2_{t,c,i}}
	\end{eqnarray*}
	\label{lemma2}
\end{lemma}
\begin{proof}
	We start with the simple observation regarding the envelope: $$f_k(\x + t e_i) - f_k(\x - t e_i) = \int_{-t}^t \frac{\partial f_k(\x + s e_i)}{\partial x_i} ds$$ using this we have
	\begin{eqnarray*}
		2t \left( \frac{\partial f'_k(\x)}{\partial x_i} - \epsilon_{t,k,i} \right) \leq f_k(\x + t e_i) - f_k(\x - t e_i) \leq 2t \left(\frac{\partial f'_k(\x)}{\partial x_i} + \epsilon_{t,k,i} \right)
	\end{eqnarray*}
	Thus we have $$\left|\frac{1}{2t} (f_c(\x + t e_i) - f_c(\x - t e_i)) - \frac{\partial f'_c(\x)}{\partial x_i} \right| \leq \epsilon_{t,c,i}$$
	using which we have the following
	\begin{eqnarray*}
		\|\nabla f_k(\x) \circ \mathbb{I}_n(x) - \hat{\nabla} f_k(\x)\|_2 &=& \sqrt{\sum_{i=1}^d \left| \frac{\partial f'_k(\x)}{\partial x_i} \cdot  \mathbb{1}_{A_{n,i}(\x)} - \Delta_{t,i} f_k(\x) \cdot \mathbb{1}_{A_{n,i}(\x)} \right|^2} \\
		&\leq& \sqrt{\sum_{i=1}^d \left| \frac{1}{2t} (f_k(\x + t e_i) - f_k(\x - t e_i)) - \frac{\partial f'_k(\x)}{\partial x_i} \right|^2} \\
		&\leq& \sqrt{\sum_{i \in [d]} \epsilon^2_{t,k,i}}
	\end{eqnarray*}
	Taking empirical expectation on both sides finishes the proof.
\end{proof}

\noindent
Taking a step back, recall again the decomposition of $\e_n \|\nabla f_c(X) - \hat{\nabla} f_{n,h,k}(X)\|_2$:
\begin{multline*}
\e_n \|\nabla f_k(X) - \hat{\nabla} f_{n,h,k}(X)\|_2 \leq \\
\e_n \|\nabla f_k(X) - \hat{\nabla} f_k(X)\|_2 + \e_n \|\hat{\nabla} f_k(X) - \hat{\nabla} f_{n,h,k}(X)\|_2
\end{multline*}
the first term in the r.h.s of the above i.e. $\e_n \|\nabla f_k(X) - \hat{\nabla} f_k(X)\|_2$ was in turn decomposed as:
\begin{multline*}
\e_n \|\nabla f_k(X) - \hat{\nabla} f_k(X)\|_2 \leq  \\ \e_n \|\nabla f_k(X) \circ \mathbb{I}_n(X) - \hat{\nabla} f_k(X)\|_2 + \e_n \|\nabla f_k(X) \circ \overline{\mathbb{I}_n(X)}\|_2
\end{multline*}

\vspace{2mm}
\noindent
The analysis in the previous subsection was bounding these two terms individually. Now we turn our attention towards bounding $\e_n \|\hat{\nabla} f_k(X) - \hat{\nabla} f_{n,h,k}(X)\|_2$

\subsubsection{Bounding \texorpdfstring{$\e_n \|\hat{\nabla} f(X) - \hat{\nabla} f_{n,h}(X)\|_2$}{} %
}
First we introduce a lemma which is a modification of Lemma 6 appearing in \cite{DBLP:conf/nips/KpotufeB12}
\begin{lemma}
	Let $t + h \leq \tau$. We have for all $i \in [d]$, and all $s \in \{-t,t\}$:
	\begin{eqnarray*}
		|\tilde{f}_{n,h,c}(\x + s e_i) - f_c(\x + s e_i)| \cdot  \mathbb{1}_{A_{n,i}(\x)} \leq hR
	\end{eqnarray*}
\end{lemma}
\begin{proof}
	The proof follows the same logic as in \cite{DBLP:conf/nips/KpotufeB12}, with the last step modified appropriately. To be more specific, let $x = X + se_i$, let $v_i = \frac{X_i - x}{\|X_i - x\|_2}$, then we have
	\begin{eqnarray*}
		|\tilde{f}_{n,h,c}(\x + s e_i) - f_c(\x + s e_i)| &\leq& \sum_{i \in [d]} w_i(x) |f(X_i) - f(x)| \\ 
		&=& \sum_{i \in [d]} w_i(x) |\int_{0}^{\|X_i - x\|_2} v_i^T \nabla f(x + t v_i) dt | \\
		&\leq& \sum_{i \in [d]} w_i(x) \|X_i - x\|_2 \cdot \max_{x' \in \Xx + B(0,\tau)} \|v_i^T \nabla f(x)\|_2 \\
		&\leq& \sum_{i \in [d]} w_i(x) \|X_i - x\|_2 R \\
		&\leq& hR
	\end{eqnarray*}
\end{proof}

\begin{lemma}
	There exist a constant $C = C(\mu,K(\cdot))$, such that the following holds with probability at least $1 - 2 \delta$ over the choice of $X$. Define $A(n) = 0.25 \cdot \sqrt{C d \cdot \ln(kn/\delta)}$, for all $i \in [d], k \in [c]$, and all $s \in \{-t,t\}$:
	\begin{eqnarray*}
		\e_n |\tilde{f}_{n,h,k}(X + s e_i) - f_{n,h,k}(X + s e_i)|^2 \cdot  \mathbb{1}_{A_{n,i}(X)} \leq \frac{A(n)}{nh^d}
	\end{eqnarray*}
\end{lemma}
\begin{proof}
	The proof follows a similar line of argument as made for the proof of Lemma 7 in \cite{DBLP:conf/nips/KpotufeB12}. First fix any $k \in [c]$,
	Assume $A_{n,i}(X)$ is true, and fix $x = X + s e_i$. Taking conditional expectation on $\mathbf{Y}^n = Y_1,...,Y_n$ given $\mathbf{X}^n = X_1,...,X_n$, we have
	\begin{eqnarray*}
		\e_{\mathbf{Y}^n | \mathbf{X}^n} |f_{n,h,k}(x) - \tilde{f}_{n,h,k}(x)|^2 \leq 0.25 \cdot \sum_{i \in [n]} (w_i(x))^2 \leq 0.25 \cdot \max_{i \in [n]} w_i(x)
	\end{eqnarray*}
	Use $\mathbf{Y}^n_x$ to denote corresponding $Y_i$ of samples $X_i \in B(x,h)$. 
	
	\vspace{2mm}
	\noindent
	Next, we consider the random variable $$\psi(\mathbf{Y}^n_x) = |f_{n,h,k}(x) - \tilde{f}_{n,h,k}(x)|^2$$ Let $\mathcal{Y}_\delta$ denote the event that for all $Y_i \in \mathbf{Y}^n$, $|Y_i - f(X_i)|^2 \leq 0.25$. We know $\mathcal{Y}_\delta$ happens with probability at least $1/2$. Thus
	\begin{eqnarray*}
		\P_{\mathbf{Y}^n | \mathbf{X}^n} (\psi(\mathbf{Y^n}_x) > 2 \e_{\mathbf{Y}^n | \mathbf{X}^n} \psi(\mathbf{Y^n}_x) + \epsilon ) &\leq&  \P_{\mathbf{Y}^n | \mathbf{X}^n} (\psi(\mathbf{Y^n}_x) >  \e_{\mathbf{Y}^n | \mathbf{X}^n, \mathcal{Y}_\delta} \psi(\mathbf{Y^n}_x) + \epsilon ) \\
		&\leq& \P_{\mathbf{Y}^n | \mathbf{X}^n, \mathcal{Y}_\delta} (\psi(\mathbf{Y^n}_x) >  \e_{\mathbf{Y}^n | \mathbf{X}^n, \mathcal{Y}_\delta} \psi(\mathbf{Y^n}_x) + \epsilon ) + \delta /2
	\end{eqnarray*}
	By McDiarmid's inequality, we have
	\begin{eqnarray*}
		\P_{\mathbf{Y}^n | \mathbf{X}^n, \mathcal{Y}_\delta} (\psi(\mathbf{Y^n}_x) >  \e_{\mathbf{Y}^n | \mathbf{X}^n, \mathcal{Y}_\delta} \psi(\mathbf{Y^n}_x) + \epsilon ) \leq \exp \left\{ -2 \epsilon^2 \cdot \delta_Y^4 \sum_{i \in [n]} w_i^4(x)
		\right\}
	\end{eqnarray*}
	The number of possible sets $\mathbf{Y}_x^n$ (over $x \in \mathcal{X}$) is at most the $n$-shattering number of balls in $\R^d$, using Sauer's lemma we get the number is bounded by $(2n)^{d+2}$. By union bound, with probability at least $1 - \delta$, for all $x \in \mathcal{X}$ satisfying $B(x,h/2) \bigcap \mathbf{X}^n \neq \emptyset$, 
	\begin{eqnarray*}
		\psi(\mathbf{Y}^n_x) &\leq& 2 \e_{\mathbf{Y}^n | \mathbf{X}^n} \psi(\mathbf{Y^n}_x) + \sqrt{ 0.25 \cdot (d+2) \cdot \log(n/\delta)  \cdot \sum_{i \in [n]} w_i^4(x)} \\
		&\leq& 2 \sqrt{\e_{\mathbf{Y}^n | \mathbf{X}^n} \psi^2(\mathbf{Y^n}_x)} + \sqrt{ 0.25 \cdot (d+2) \cdot \log(n/\delta)  \cdot \delta_Y^4 \max_{i \in [n]} w_i^2(x)} \\
		&\leq& \sqrt{Cd \cdot \log(n/\delta) \cdot 0.25/n^2 \mu_n^2(B(x,h/2))}
	\end{eqnarray*}
	Take a union bound over $k \in [c]$, and take empirical expectation, we get $\forall k \in [c]$
	\begin{eqnarray*}
		\e_n |\tilde{f}_{n,h,k}(X + s e_i) - f_{n,h,k}(X + s e_i)|^2 \leq \frac{0.25 \cdot \sqrt{C d \cdot \ln(cn/\delta)}}{n} \sum_{i \in [n]} \frac{1}{n(x_i,h/2)}
	\end{eqnarray*}
	where $n(x_i,h/2) = n\mu_n(B(x_i,h/2))$ is the number of points in Ball $B(x_i,h/2)$. 
	
	\vspace{2mm}
	\noindent
	Let $\mathcal{Z}$ denote the minimum $h/4$ cover of $\{x_1,...,x_n\}$, which means for any $x_i$, there is a $z \in \mathcal{Z}$, such that $x_i$ is contained in the ball $B(z,h/4)$. Since $x_i \in B(z,h/4)$, we have $B(z,h/4) \in B(x_i,h/2)$.
	We also assume every $x_i$ is assigned to the closest $z \in \mathcal{Z}$, and write $x_i \rightarrow z$ to denote such $x_i$.
	Then we have:
	\begin{eqnarray*}
		\sum_{i \in [n]} \frac{1}{n(x_i,h/2)} &=& \sum_{z \in \mathcal{Z}} \sum_{x_i \rightarrow z} \frac{1}{n(x_i,h/2)} \\
		&\leq& \sum_{z \in \mathcal{Z}} \sum_{x_i \rightarrow z} \frac{1}{n(z,h/4)} \\
		&\leq& \sum_{z \in \mathcal{Z}} \frac{n(z,h/4)}{n(z,h/4)} \\
		&=& |\mathcal{Z}| \leq C_\mu (h/4)^{-d}
	\end{eqnarray*}
	Combining above analysis finishes the proof.
\end{proof}

\begin{lemma}
	There exists a constant $C = C(\mu,K(\cdot))$, such that the following holds with probability at least $1 - 2 \delta$. Define $A(n) = 0.25 \cdot \sqrt{C d \cdot \ln(kn/\delta)} $, $\forall k \in [c]$:
	\begin{eqnarray*}
		\e_n \|\hat{\nabla} f_k(X) - \hat{\nabla} f_{n,h,k}(X)\|_2 \leq \frac{\sqrt{d}}{t} \sqrt{\frac{A(n)}{nh^d} + h^2 R^2}
	\end{eqnarray*}
	\label{lemma7}
\end{lemma}
\begin{proof}
	First we can write the following bound for the l.h.s:
	\begin{eqnarray*}
		\e_n \|\hat{\nabla} f_k(X) - \hat{\nabla} f_{n,h,k}(X)\|_2 &\leq& \e_n \sqrt{\sum_{i \in [d]} |\Delta_{t,i}f_{n,h,k}(X) - \Delta_{t,i} f_k(X)|^2 \cdot  \mathbb{1}_{A_{n,i}(X)}} \\
		&\leq& \sqrt{\sum_{i \in [d]} \e_n |\Delta_{t,i}f_{n,h,k}(X) - \Delta_{t,i} f_k(X)|^2 \cdot  \mathbb{1}_{A_{n,i}(X)}} \\
		&\leq& \sqrt{\sum_{i \in [d]} \frac{1}{t^2} \max_{s \in \{-t,t\}} \e_n |f_{n,h,k}(X + s e_i) - f_k(X + s e_i)|^2 \cdot  \mathbb{1}_{A_{n,i}(X)}}
	\end{eqnarray*}
	First observe that:
	\begin{eqnarray*}
	\e_n |f_{n,h,k}(X + s e_i) - f_k(X + s e_i)|^2 \cdot  \mathbb{1}_{A_{n,i}(X)} 
	&\leq& \e_n |\tilde{f}_{n,h,k}(X + s e_i) - f_k(X + s e_i)|^2 \cdot  \mathbb{1}_{A_{n,i}(X)} \\
	&+& \e_n |\tilde{f}_{n,h,k}(X + s e_i) - f_{n,h,k}(X + s e_i)|^2 \cdot  \mathbb{1}_{A_{n,i}(X)}
	\end{eqnarray*}
	Also notice that: $\e_n |\tilde{f}_{n,h,k}(X + s e_i) - f_k(X + s e_i)|^2 \cdot  \mathbb{1}_{A_{n,i}(X)}$ and $\e_n |\tilde{f}_{n,h,k}(X + s e_i) - f_{n,h,k}(X + s e_i)|^2 \cdot  \mathbb{1}_{A_{n,i}(X)}$ can be respectively bounded by two lemmas from above, thus we get with probability at least $1 - 2\delta$
	\begin{eqnarray*}
		\e_n |f_{n,h,k}(X + s e_i) - f_k(X + s e_i)|^2 \leq h^2 R^2 + \sqrt{\frac{A(n)}{nh^d}}
	\end{eqnarray*}
	Combining above we get with probability at least $1 - 2\delta$, $\forall k \in [c]$
	\begin{eqnarray*}
		\|\hat{\nabla} f_k(X) - \hat{\nabla} f_{n,h,k}(X)\|_2 \leq \frac{\sqrt{d}}{t} \sqrt{\frac{A(n)}{nh^d} + h^2 R^2}
	\end{eqnarray*}
\end{proof}

\noindent
The following theorem provides a bound on $\e_n \|\nabla f(X) - \hat{\nabla} f_{n,h}(X)\|_2$:
\begin{theorem}
	With probability at least $1 - 2 \delta$ over the choice of $X$, we have $\forall k \in [c]$:
	\begin{eqnarray*}
		\e_n \|\nabla f_k(X) - \hat{\nabla} f_{n,h,k}(X)\|_2 \leq \frac{\sqrt{d}}{t} \sqrt{\frac{A(n)}{nh^d} + h^2 R^2} &+& R \left( \sqrt{\frac{d \ln \frac{d}{\delta}}{2n}} + \sqrt{\sum_{i \in [d]} \mu^2(\partial_{t,i}(\mathcal{X}))} \right) \\ 
		&+& \sqrt{\sum_{i \in [d]} \epsilon^2_{t,i}}
	\end{eqnarray*}
	\label{thm:f}
\end{theorem}
\begin{proof}
	We start with the now familiar decomposition:
	\begin{align*}
	\e_n \|\nabla f_k(X) - \hat{\nabla} f_{n,h,k}(X)\|_2 \leq&
	 \e_n \|\hat{\nabla} f_k(X) - \hat{\nabla} f_{n,h,k}(X)\|_2 \\
	&+ \e_n \|\nabla f_k(X) \circ \mathbb{I}_n(X) - \hat{\nabla} f_k(X)\|_2 + \e_n \|\nabla f_k(X) \circ \overline{\mathbb{I}_n(X)}\|_2
	\end{align*}
	By Lemma \ref{lemma1} we bound $\e_n \|\nabla f(X) \circ \overline{\mathbb{I}_n(X)}\|_2$; by Lemma \ref{lemma2} we bound $\e_n \|\nabla f(X) \circ \mathbb{I}_n(X) - \hat{\nabla} f(X)\|_2$; by Lemma \ref{lemma7} we bound $\e_n \|\hat{\nabla} f(X) - \hat{\nabla} f_{n,h}(X)\|_2$. Combining these results concludes the proof.
\end{proof}

\section{Bounds on Eigenvalues and Eigenspace variations}\label{sec:eigenspace}

In the above section, we established that  $\e_n \hat{G}(X) $ is a consistent estimator of $\e_X G(X) $. In this section, we also establish consistency of its eigenvalues and eigenspaces,. The analysis here is based upon results from matrix perturbation theory \cite{RPT1,RPT2}.

\subsection{Eigenvalues variation}
We begin by considering the following lemma for eigenvalues variation from matrix perturbation theory:

\begin{lemma}
	\cite{RPT1} Suppose both $G$ and $\hat{G}$ are Hermitian matrices of size $d \times d$, and admit the following eigen-decompositions:
	\begin{eqnarray*}
		G = X \Lambda X^{-1}  \quad \text{and} \quad \hat{G} = \hat{X} \hat{\Lambda} \hat{X}^{-1}
	\end{eqnarray*}
	where $X$ and $\hat{X}$ are nonsingular and
	\begin{eqnarray*}
		\Lambda = \text{diag}(\lambda_1,\lambda_2,...\lambda_d) \quad \text{and} \quad
		\hat{\Lambda} = \text{diag}(\hat{\lambda}_1,\hat{\lambda}_2,...\hat{\lambda}_d)
	\end{eqnarray*}
	and $\lambda_1 \geq \lambda_2 \geq ... \geq \lambda_d$, $\hat{\lambda}_1 \geq \hat{\lambda}_2 \geq ... \geq \hat{\lambda}_d$. Thus for any unitary invariant norm $\| \cdot \|$, we have
	\begin{eqnarray*}
		\|\diag(\lambda_1 - \hat{\lambda}_1,\lambda_2-  \hat{\lambda}_2, ..., \lambda_d - \hat{\lambda}_d) \| \leq \| G - \hat{G} \|
	\end{eqnarray*}
	More specifically, when considering the spectral norm, we have
	\begin{eqnarray*}
		\max_{i \in [d]} |\lambda_i - \hat{\lambda}_i| \leq \| G - \hat{G} \|_2
	\end{eqnarray*}
	and when considering the Frobenius norm, we have
	\begin{eqnarray*}
		\sqrt{\sum_{i \in [d]} |\lambda_i - \hat{\lambda}_i|^2} \leq \| G - \hat{G} \|_F
	\end{eqnarray*}
	\label{lemma:eigvalue}
\end{lemma}

\vspace{2mm}
\noindent
Using the above lemma, we obtain the following theorem that bounds the eigenvalue variation:
\begin{aside}{Eigenvalue Variation Bound}
\begin{theorem}
	Let $\lambda_1 \geq \lambda_2 \geq ... \geq \lambda_d$ be the eigen-values of $\e_X G(X)$, let $\hat{\lambda}_1 \geq \hat{\lambda}_2 \geq ... \geq \hat{\lambda}_d$ be the eigen-values of $\e_n \hat{G}(X)$. There exist $C = C(\mu,K(\cdot))$ and $N = N(\mu)$ such that the following holds with probability at least $1 - 2\delta$. Define $A(n) = \sqrt{Cd \cdot \log(n/\delta)} \cdot C_Y^2(\delta/2n) \cdot \sigma_Y^2/\log^2(n/\delta)$. Let $n \geq N$, we have:
	\begin{eqnarray*}
		&&\max_{i \in [d]} |\lambda_i - \hat{\lambda}_i| \leq \frac{6 R^2}{\sqrt{n}}(\sqrt{\ln d} + \sqrt{\ln \frac{1}{\delta}}) + \left(3R + \sqrt{\sum_{i \in [d]} \epsilon^2_{t,i}} + \sqrt{d}(\frac{hR + C_Y(\delta)}{t})\right)\\
		&&\left[\frac{\sqrt{d}}{t} \sqrt{\frac{A(n)}{nh^d} + h^2 R^2} + R \left( \sqrt{\frac{d \ln \frac{d}{\delta}}{2n}} + \sqrt{\sum_{i \in [d]} \mu^2(\partial_{t,i}(\mathcal{X}))} \right) + \sqrt{\sum_{i \in [d]} \epsilon^2_{t,i}} \right]
	\end{eqnarray*}
\end{theorem}
\end{aside}
\begin{proof}
	By Lemma \ref{lemma:eigvalue}, we bound $\max_{i \in [d]} |\lambda_i - \hat{\lambda}_i|$ with respect to $\| \e_n \hat{G}(X) - \e_X G(X) \|_2$; by Theorem \ref{thm:main} we bound $\| \e_n \hat{G}(X) - \e_X G(X) \|_2$.
\end{proof}
\subsection{Eigenspace variation}
First we introduce the following definition:
\begin{definition}
	\textbf{(Angles between two subspaces)} Let $X, \hat{X} \in \R^{d \times k}$ have full column rank $k$. The angle matrix $\Theta(X, \hat{X})$ between $X$ and $\hat{X}$ is defined as:
	\begin{eqnarray*}
		\Theta(X, \hat{X}) = \arccos ((X^TX)^{-\frac{1}{2}} X^T \hat{X} (\hat{X}^T \hat{X} )^{-1} \hat{X}^T X (X^TX)^{-\frac{1}{2}})^{\frac{1}{2}}
	\end{eqnarray*}
	More specifically, when $k = 1$, it reduces to the angle between two vectors:
	\begin{eqnarray*}
		\Theta(\x, \hat{\x}) = \arccos \frac{ | \x^T \hat{\x} | }{\|\x\|_2 \|\hat{\x}\|_2}
	\end{eqnarray*}
\end{definition}

\vspace{2mm}
\noindent
Armed with this definition, we consider the following lemma on eigenspace variation:

\begin{lemma}
	\cite{RPT2} Suppose both $G$ and $\hat{G}$ are Hermitian matrices of size $d \times d$, and admit the following eigen-decompositions:
	\begin{eqnarray*}
		G = \begin{bmatrix}
			X_1 & X_2
		\end{bmatrix}
		\begin{bmatrix}
			\Lambda_1 & 0 \\
			0 & \Lambda_2
		\end{bmatrix}
		\begin{bmatrix}
			X_1^{-1} \\
			X_2^{-1}
		\end{bmatrix}
		\quad \text{and} \quad \hat{G} = \begin{bmatrix}
			\hat{X}_1 & \hat{X}_2
		\end{bmatrix}
		\begin{bmatrix}
			\hat{\Lambda}_1 & 0 \\
			0 & \hat{\Lambda}_2
		\end{bmatrix}
		\begin{bmatrix}
			\hat{X}_1^{-1} \\
			\hat{X}_2^{-1}
		\end{bmatrix}
	\end{eqnarray*}
	where $X = \begin{bmatrix}
	X_1 & X_2
	\end{bmatrix}$ and $\hat{X} = \begin{bmatrix}
	\hat{X}_1 & \hat{X}_2
	\end{bmatrix}$ are unitary. We have
	\begin{eqnarray*}
		\| \sin  \Theta(X_1, \hat{X}_1) \|_2 \leq \frac{\|(\hat{G} - G) X_1\|_2}{\min_{\lambda \in \lambda(\Lambda_1), \hat{\lambda} \in \lambda(\Lambda_2)} |\lambda - \hat{\lambda} | }
	\end{eqnarray*}
	\label{lemma:eig}
\end{lemma}

\vspace{2mm}
\noindent
Using the above lemma, we get the following theorem for eigenspaces variant:
\begin{aside}{Eigenspace Variation}
\begin{theorem}
	Write the eigen-decompositions of $\e_X G(X)$ and $\e_n \hat{G}(X)$ as $$
	\e_X G(X) = \begin{bmatrix}
	X_1 & X_2
	\end{bmatrix}
	\begin{bmatrix}
	\Lambda_1 & 0 \\
	0 & \Lambda_2
	\end{bmatrix}
	\begin{bmatrix}
	X_1^{-1} \\
	X_2^{-1}
	\end{bmatrix}, \e_n \hat{G}(X) = \begin{bmatrix}
	\hat{X}_1 & \hat{X}_2
	\end{bmatrix}
	\begin{bmatrix}
	\hat{\Lambda}_1 & 0 \\
	0 & \hat{\Lambda}_2
	\end{bmatrix}
	\begin{bmatrix}
	\hat{X}_1^{-1} \\
	\hat{X}_2^{-1}
	\end{bmatrix}$$ There exist constants $C = C(\mu,K(\cdot))$ and $N = N(\mu)$ such that the following holds with probability at least $1 - 2\delta$. Define $A(n) = \sqrt{Cd \cdot \log(n/\delta)} \cdot C_Y^2(\delta/2n) \cdot \sigma_Y^2/\log^2(n/\delta)$. Let $n \geq N$:
	\begin{small}
		\begin{eqnarray*}
			\| \sin  \Theta(X_1, \hat{X}_1) \|_2 \leq \frac{\|X_1\|_2}{\min_{\lambda \in \lambda(\Lambda_1), \hat{\lambda} \in \lambda(\Lambda_2)} |\lambda - \hat{\lambda} | } \Bigg( \frac{6 R^2}{\sqrt{n}}(\sqrt{\ln d} + \sqrt{\ln \frac{1}{\delta}}) + \\ \left(3R + \sqrt{\sum_{i \in [d]} \epsilon^2_{t,i}} + \sqrt{d}(\frac{hR + C_Y(\delta)}{t})\right) \\
			\left[\frac{\sqrt{d}}{t} \sqrt{\frac{A(n)}{nh^d} + h^2 R^2} + R \left( \sqrt{\frac{d \ln \frac{d}{\delta}}{2n}} + \sqrt{\sum_{i \in [d]} \mu^2(\partial_{t,i}(\mathcal{X}))} \right) + \sqrt{\sum_{i \in [d]} \epsilon^2_{t,i}} \right] \Bigg)
		\end{eqnarray*}
	\end{small}
	\label{eigenspace}
\end{theorem}
\end{aside}
\begin{proof}
	By Lemma \ref{lemma:eig}, we bound $\| \sin  \Theta(X_1, \hat{X}_1) \|_2$ with respect to $\|X_1 (\e_n \hat{G}(X) - \e_X G(X)) \|_2$, since $\|X_1 (\e_n \hat{G}(X) - \e_X G(X)) \|_2 \leq \|X_1\|_2 \cdot \|\e_n \hat{G}(X) - \e_X G(X)\|_2$, and by Theorem \ref{thm:main} we bound $\| \e_n \hat{G}(X) - \e_X G(X) \|_2$. Combining these concludes the proof.
\end{proof}

\section{Recovery of projected semiparametric regression model}\label{sec:semiparametric}
In this section, the last on the theoretical analysis, we return to the multi-index motivation of the EGOP and EJOP discussed in the introduction to this chapter. For ease of exposition, we restrict our discussion to the EGOP, but the same argument also works for the EJOP. 

\vspace{2mm}
\noindent
Consider the following projected semiparametric regression model:
\begin{eqnarray*}
	f(\x) = g(V^T \x)
\end{eqnarray*}
where $V \in \R^{d \times r}, r \ll d$ is a dimension-reduction projection matrix, and $g$ is a nonparametric function. Without loss of generality, we assume $V = [v_1,v_2,....,v_r]$, where $v_i \in R^d, i \in [r]$ is a set of orthonormal vectors, and the gradient outer product (GOP) matrix of $g: \e_{X} [\nabla g(V^T X) \cdot \nabla g(V^T X)^T]$ is nonsingular. The following proposition gives the eigen-decomposition of gradient outer product (GOP) matrix of $f$: $\e_X G(\x)$
\begin{proposition}
	Suppose the eigen-decomposition of $\e_{X} [\nabla g(V^T X) \cdot \nabla g(V^T X)^T]$ is given by:
	\begin{eqnarray*}
		\e_{X} [\nabla g(V^T X) \cdot \nabla g(V^T X)^T] = Z \Lambda Z^{-1}
	\end{eqnarray*}
	then we have the following eigen-decomposition of $\e_X G(X)$:
	\begin{eqnarray*}
		\e_X G(X) = \begin{bmatrix}
			VZ & U
		\end{bmatrix}
		\begin{bmatrix}
			\Lambda & 0 \\
			0       & 0
		\end{bmatrix}
		\begin{bmatrix}
			Z^{-1} V^T \\
			U^T
		\end{bmatrix}
	\end{eqnarray*}
	where $U = [u_1,u_2,...,u_{d-r}]$, $u_i \in [d-r]$ is a set of orthonormal vectors in $\text{ker}(V^T)$.
	\label{prop}
\end{proposition}
\begin{proof}
	Since $f(\x) = g(V^T \x)$, we have $\nabla f(\x) = V \nabla g(V^T \x)$. Thus we get:
	\begin{eqnarray*}
		\e_X G(X) = V \e_{X} [\nabla g(V^T X) \cdot \nabla g(V^T X)^T] V^T = VZ \Lambda Z^{-1}V^T
	\end{eqnarray*}
	When we check the eigen-decomposition given in the proposition, the above equation is satisfied. Moreover, since $$\begin{bmatrix}
	VZ & U
	\end{bmatrix} \begin{bmatrix}
	Z^{-1} V^T \\
	U^T
	\end{bmatrix} = \begin{bmatrix}
	Z^{-1} V^T \\
	U^T
	\end{bmatrix} \begin{bmatrix}
	VZ & U
	\end{bmatrix}= I$$ concludes the proof.
\end{proof}

\vspace{2mm}
\noindent
Since $Z$ in the above proposition is nonsingular, we get that $\text{im}(V) = \text{im}(VZ)$, which means that the column space of projection matrix $V$ is exactly the subspace spanned by the top-$r$ eigenvectors of the GOP matrix $\e_X G(X)$. This point has also been noticed by \cite{sliced,RSSB:RSSB341,DBLP:journals/jmlr/WuGMM10}.

\vspace{2mm}
\noindent
Lastly, we need to show that the projection matrix $V$ can be recovered using the estimated GOP matrix. This is captured in the following two theorems:

\begin{aside}{Recovery of Semi-parametric model}
\begin{theorem}
	Suppose the function $f$ we want to estimate has the form $f(\x) = g(V^T \x)$, and $\tilde{V} \in \R^{d \times r}$ is the matrix composed by the top-$r$ eigenvectors of $\e_n \hat{G}(X)$,
	then with probability at least $1 - 2 \delta$:
	\begin{eqnarray*}
		\| \sin  \Theta(V, \tilde{V}) \|_2 \leq \frac{1}{\lambda_{\min}}\Bigg(
		\frac{6 R^2}{\sqrt{n}}(\sqrt{\ln d} + \sqrt{\ln \frac{1}{\delta}}) + \left(3R + \sqrt{\sum_{i \in [d]} \epsilon^2_{t,i}} + \sqrt{d}(\frac{hR + C_Y(\delta)}{t})\right) \\
		\left[\frac{\sqrt{d}}{t} \sqrt{\frac{A(n)}{nh^d} + h^2 R^2} + R \left( \sqrt{\frac{d \ln \frac{d}{\delta}}{2n}} + \sqrt{\sum_{i \in [d]} \mu^2(\partial_{t,i}(\mathcal{X}))} \right) + \sqrt{\sum_{i \in [d]} \epsilon^2_{t,i}} \right]\Bigg)
	\end{eqnarray*}
	where $\lambda_{\min}$ is the smallest eigenvalue of $\e_{X} [\nabla g(V^T X) \cdot \nabla g(V^T X)^T]$.
	\vspace{2mm}
	\noindent
	Suppose $\lambda_1, \lambda_2, ..., \lambda_{d-r}$ are the lowest $d - r$ eigenvalues of $\e_n \hat{G}(X)$,
	and with probability at least $1 - 2 \delta$:
	\begin{eqnarray*}
		max_{i \in [d-r]} |\lambda_i| \leq (
		\frac{6 R^2}{\sqrt{n}}(\sqrt{\ln d} + \sqrt{\ln \frac{1}{\delta}}) + \left(3R + \sqrt{\sum_{i \in [d]} \epsilon^2_{t,i}} + \sqrt{d}(\frac{hR + C_Y(\delta)}{t})\right) \\
		\left[\frac{\sqrt{d}}{t} \sqrt{\frac{A(n)}{nh^d} + h^2 R^2} + R \left( \sqrt{\frac{d \ln \frac{d}{\delta}}{2n}} + \sqrt{\sum_{i \in [d]} \mu^2(\partial_{t,i}(\mathcal{X}))} \right) + \sqrt{\sum_{i \in [d]} \epsilon^2_{t,i}} \right])
	\end{eqnarray*}
\end{theorem}
\end{aside}
\begin{proof}
	We only sketch the proof. First of all, notice that $V$ is a semi-orthogonal matrix, therefore $\|V\|_2 = 1$. When this observation is combined with above proposition and Theorem \ref{eigenspace}, we get a proof of the first part of the theorem. For proving the second part of the theorem, first observe that by proposition \ref{prop}, the lowest $d-r$ eigenvalues of $\e_X {G}(X)$ are all zeros. This observation when combined with lemma \ref{lemma:eigvalue} finishes the proof.
\end{proof}

\section{Classification Experiments}
In this section, we give a brief experimental evaluation that examines the utility of the EJOP as a technique for metric estimation, when used in the setting of non-parametric classification. As in chapter \ref{ch:EGOP}, we consider non-parametric classifiers that rely on the notion of distance, parameterized by a matrix $\mathbf{M} \succeq 0 $, with the squared distance computed as $(\mathbf{x} - \mathbf{x}')^T \mathbf{M}(\mathbf{x} - \mathbf{x}')$.

\vspace{2mm}
\noindent
In the experiments reported in this section, we consider three different choices for $\mathbf{M}$:
\begin{enumerate}
	\item[1] $\mathbf{M} = \mathbf{I}$, which corresponds to the Euclidean distance
	\item[2] $\mathbf{M} = \mathbf{D}$, where $\mathbf{D}$ is a diagonal matrix, the notion of distance in this case corresponds to a scaled Euclidean distance. In particular, in the absence of a gradients weights \cite{DBLP:conf/nips/KpotufeB12}, \cite{GWJMLR} like approach for the multiclass case, we instead obtain weights by using the ReliefF procedure~\cite{relieff1992}, which estimates weights for the multiclass case by a series of one versus all binary classifications.
	\item[3] $\mathbf{M} = \e_n {G}_n(X)$, where $\e_n {G}_n(X)$ is the estimated EJOP matrix. 
\end{enumerate}

\vspace{2mm}
\noindent
In particular, letting $VDV^\top$ denote the spectral decomposition of $\mathbf{M}$, we use it to transform the input $\mathbf{x}$ as $D^{1/2}V^\top \mathbf{x}$ for the distance computation. Next, for a fixed choice of $\mathbf{M}$, we can define nearest neighbors of a query point $\mathbf{x}$ in various ways. We consider the following two ways:

\begin{enumerate}
	\item[1] $k$ nearest neighbors (denoted henceforth as $k$NN) for fixed $k$
	\item[2] Neighbors that have distance $\leq h$ for fixed $h$ from the query. We denote this as $h$NN. This corresponds to nonparametric classification using a boxcar kernel.
\end{enumerate}

\subsection{A First Experiment on MNIST}
We first consider the MNIST dataset to test the quality of the EJOP metric, and if it improves upon plain Euclidean distance. In this case, we only test it for the $k$NN case, fixing $k=7$. We set aside 10,000 points as a validation set, which is used to obtain the ReliefF weights, as well as for tuning the parameter $t_i$ for $i = 1, \dots, 784$ in the EJOP estimation. While the $t_i$ can be tuned separately for each class, we ignore that option in this set of experiments. Note that no preprocessing is applied on the images, and the metric estimation, as well as classification is done using the raw images. The results on the test set are illustrated in the following table:

\begin{table}
\begin{center}
	\begin{tabular}{l*{1}{c}}
		Method              & Error \%  \\
		\hline \hline
		Euclidean & 4.93  \\
		ReliefF & 4.11 \\
		EJOP & 2.37 \\
	\end{tabular}
	\caption{Error rates on MNIST using EJOP as the underlying metric, and comparison to Euclidean distance and scaled Euclidean distance}
	\label{tb:MNISTEJOP}
\end{center}
\end{table}
\noindent
While MNIST is a considerably easy task, the improvement given by the use of the EJOP as the distance metric over the plain Euclidean distance is substantial. This could perhaps be improved further by tuning $t_i$ separately for each class. We will take this approach in the experiments described in the next section.

\subsection{Experiments on Datasets in \cite{trivediNIPS14} and \cite{kedem2012non}}

Next, we consider the datasets considered in \cite{trivediNIPS14} and \cite{kedem2012non}, on which experiments are described in Chapters \ref{ch:mlng} and \ref{ch:extensions} as well. First we report experiments using plain Euclidean distance, $h$-NN and $k$-NN when the EJOP is used as the metric. The train/test splits are reported in the table. We split 20 \% of the training portion to tune for $h$, $k$ and $t_i$, the results reported are over 10 random runs as in Chapter \ref{ch:EGOP}.

\begin{table}[!th]
	\centering
	\begin{tabu}{|l|l|r|r|r|r|r|r|r|}
		\hline
		Dataset & d & N & train/test & Euclidean & h-NN & $k$-NN \\
		\hline 
		\hline
		\small{Isolet} & 172 &  7797 & 4000/2000  &  14.17 $\pm$ 0.7 & 10.14 $\pm$ 0.9  & 8.67 $\pm$ 0.6 \\
		\hline
		\small{USPS} & 256 & 9298 & 4000/2000  & 7.87 $\pm$ 0.2  & 7.14 $\pm$ 0.3 & 6.67 $\pm$ 0.4  \\
		\hline
		\small{Letters} & 16 & 20000 &  4000/2000 & 7.65 $\pm$ 0.3 & 5.12 $\pm$ 0.7 & 4.37 $\pm$ 0.4 \\
		\hline
		\small{DSLR} & 800 & 157 & 100/50 & 84.85 $\pm$ 4.8 & 41.13 $\pm$ 2.1 & 35.01 $\pm$ 1.4 \\
		\hline 
		\small{Amazon} & 800  & 958 & 450/450  & 66.17 $\pm$ 2.8  & 41.07 $\pm$ 2.3  & 39.85 $\pm$ 1.5  \\
		\hline 
		\small{Webcam} & 800  & 295 & 145/145  & 61.43 $\pm$ 1.7  & 24.86 $\pm$ 1.2  & 23.71 $\pm$ 2.1  \\
		\hline 
		\small{Caltech} & 800  & 1123 & 550/500  & 85.41 $\pm$ 3.5  & 54.65 $\pm$ 2.6  & 52.86 $\pm$ 3.1  \\
		\hline
		\hline
	\end{tabu}
	\caption{Results comparing classification error rates on the datasets used in \cite{kedem2012non} using plain Euclidean distance, $h$NN and $k$NN while using the EJOP as the metric}
	\label{tb:EJOPsecond}
\end{table}

\vspace{2mm}
\noindent
Next, we consider the same datasets, and report results obtained on the same folds using three popular metric learning methods. In particular, we consider Large Margin Nearest Neighbors (LMNN)~\cite{weinberger2009distance}, Information Theoretic Metric Learning (ITML)~\cite{itml2007davis} and Metric Learning to Rank (MLR)~\cite{mcfee10_mlr}. Since these methods explicitly optimize for the metric over a space of possible metrics, the comparison is manifestly unfair, since in the case of the EJOP, there is only one metric, which is estimated from the training samples. The setup is the same as discussed above, with the following addition for the metric learning methods: We learn the metric for $k=5$, and test is using whatever $k$ that was returned while tuning for the EJOP. We observe that despite its simplicity, EJOP does a decent job as compared to the metric learning methods, in some cases returning error rates comparable to those returned by MLR and ITML. 

\begin{table}[!th]
	\centering
	\begin{tabu}{|l|l|r|r|r|r|r|r|r|}
		\hline
		Dataset & h-NN & $k$-NN & ITML & LMNN & MLR \\
		\hline 
		\hline
		\small{Isolet} & 10.14 $\pm$ 0.9  & 8.67 $\pm$ 0.6  & 8.43 $\pm 0.3$ & 5.3 $\pm$ 0.4 & 6.59 $\pm$ 0.3 \\
		\hline
		\small{USPS} & 7.14 $\pm$ 0.3 & 6.67 $\pm$ 0.4  & 6.57 $\pm$ 0.2 & 6.23 $\pm$ 0.5 &  6.76 $\pm$ 0.3 \\
		\hline
		\small{Letters} & 5.12 $\pm$ 0.7 & 4.37 $\pm$ 0.4 & 5 $\pm$ 0.7 &  4.1 $\pm$ 0.4& 17.81 $\pm$ 5.1\\
		\hline
		\small{DSLR} & 41.13 $\pm$ 2.1 & 35.01 $\pm$ 1.4 & 21.65 $\pm$ 3.1 &  29.65 $\pm$ 3.7 & 41.54 $\pm$ 2.3\\
		\hline 
		\small{Amazon} & 41.07 $\pm$ 2.3  & 39.85 $\pm$ 1.5  & 39.83 $\pm$ 3.5 & 33.08 $\pm$ 4.2& 29.65 $\pm$ 2.6\\
		\hline 
		\small{Webcam} & 24.86 $\pm$ 1.2  & 23.71 $\pm$ 2.1  & 15.31 $\pm$ 4.3 & 19.78 $\pm$ 1.5& 27.54 $\pm$ 3.9\\
		\hline 
		\small{Caltech} & 54.65 $\pm$ 2.6  & 52.86 $\pm$ 3.1   & 52.37 $\pm$ 4.2 & 52.15 $\pm$ 3.2& 51.34 $\pm$ 4.5\\
		\hline
		\hline
	\end{tabu}
	\caption{Results comparing classification error rates given by the EJOP, and three popular metric learning methods}
	\label{tb:EJOPthird}
\end{table}

\section{Summary of Part on Metric Estimation}

We conclude this part of the dissertation with a summary of the work undertaken, and some potential avenues for future work. Chapters \ref{ch:EGOP} and \ref{ch:EJOP} made the following contributions:

\begin{aside}{Summary of Part II}
\begin{enumerate}
	\item[1] We described a simple estimator for the Expected Gradient Outerproduct (EGOP) 
	$$\expectation_{\mathbf{x}} G(\mathbf{x}) \triangleq \expectation_\mathbf{x}\paren{\nabla f(\mathbf{x}) \cdot \nabla f(\mathbf{x})^\top} .$$  and demonstrated that it remains statistically consistent under mild assumptions. The estimated EGOP was then showed to be useful in nonparametric regression tasks when used as the underlying metric. 
	\item[2] We extended the EGOP to the multiclass case, proposing a generalization that we refer to as the Expected Jacobian Outer Product (EJOP)
	$$\mathbb{E}_{\mathbf{x}} G(\mathbf{x}) \triangleq  \expectation_\mathbf{x}\paren{\mathbf{J}_{f}(\mathbf{x}) \mathbf{J}_{f}(\mathbf{x})^T}$$ As in the case of the EGOP, we proposed a rough estimator for the EJOP, and also showed that it remained statistically consistent under similar assumptions. The EJOP was then used and shown to be experimentally useful as a metric in non-parametric classification tasks. 
\end{enumerate}
\end{aside}

\section{Potential Avenues for Future Work}

\subsection{Label Aware Dimensionality Reduction}
As discussed in Chapter \ref{ch:egopintro}, an attractive quality of the EGOP is that it recovers the average variation of $f$ in \emph{all} directions. It is this property that makes it useful for \emph{effective dimension reduction}, that is, finding a $k << d$ dimensional subspace that is most relevant to predicting the output $y$. As discussed in Section \ref{sec:semiparametric}, this multi-index motivation also carries through for the multiclass case by the EJOP. 

\vspace{2mm}
\noindent
Although explored somewhat cursorily by the dissertation author, it would be interesting to leverage the multi-index motivation of both the EGOP and the EJOP for the task of dimensionality reduction of data that takes into account the labels as well. This is contrasted to methods such as PCA, where the covariance matrix construction is completely label oblivious. Some experiments for dimensionality for the case of regression are reported by \cite{Mukherjee_wu2010learning, mukherjee2010learning} and by using metric learning are reported by \cite{MLKR}, however not many applications were explored. The EGOP and EJOP can possibly be used to give a handy method for class aware dimensionality reduction. 

\subsection{Operators that take into account local geometry}

We have the following, somewhat hand-wavy analogy between the EGOP and EJOP when put side by side with PCA. PCA helps recover directions according to how much variance in the data is explained by them, whereas the EGOP and EJOP help us recover directions according to the average variation of $f$. Both methods involve construction of a covariance matrix, and lose local information. We illustrate this with the EGOP 
	$$\expectation_{\mathbf{x}} G(\mathbf{x}) \triangleq \expectation_\mathbf{x}\paren{\nabla f(\mathbf{x}) \cdot \nabla f(\mathbf{x})^\top} .$$
While gradients are local objects, since in the estimation of the EGOP, we take expectation over $\mathbf{x}$, all information about the local geometry is averaged out. We would like to construct operators that don't lose local information, and maybe give a non-linear map to a subspace that is most relevant to predict the output. 

\vspace{2mm} 
\noindent
We can perhaps take inspiration from the literature in non-linear dimensionality reduction to search for an alternative. An attractive method, that unlike PCA does retain local information is exemplified by Laplacian Eigenmaps of Belkin and Niyogi \cite{Belkin2003}. In such methods, dimensionality reduction is achieved by the spectral decomposition of an operator that encodes the local geometry of the data. Usually, such an operator is a diffusion based object, such as the Graph Laplacian, defined as:
$$ \mathbf{L} = \mathbf{I} - \mathbf{D}^{-\frac{1}{2}}\mathbf{W}\mathbf{D}^{-\frac{1}{2}}$$
where $\mathbf{D}$ and $\mathbf{W}$ are the degree and adjacency matrices respectively, of an appropriate nearest neighbor graph constructed on the data points. Taking a cue from this, we could define a {\em diffusion map using gradients} $\mathbf{W}$, for the regression and binary classification case as follows:

$$\mathbf{W}_{i,j} = \mathbf{W}_{f}(\mathbf{x}_i, \mathbf{x}_j) = exp \Bigg( - \frac{\|\mathbf{x}_i - \mathbf{x}_j\|^2}{\sigma_1} - \frac{\|\frac{1}{2}(\nabla f(\mathbf{x}_i) + \nabla f(\mathbf{x}_j) \dot (\mathbf{x}_i - \mathbf{x}_j) \|^2}{\sigma_2} \Bigg)$$

\noindent
Such an operator has infact been discussed by \cite{Mukherjee_wu2010learning, mukherjee2010learning}, but not explored in detail. For the multiclass case, we could consider the following:

$$\mathbf{W}_{i,j} = \mathbf{W}_{f}(\mathbf{x}_i, \mathbf{x}_j) = exp \Bigg( - \frac{\|\mathbf{x}_i - \mathbf{x}_j\|^2}{\sigma_1} - \frac{\|\frac{1}{2}(|\nabla f_c(\mathbf{x}_i)| + |\nabla f_c(\mathbf{x}_j|) \dot (\mathbf{x}_i - \mathbf{x}_j) \|^2}{\sigma_2} \Bigg)$$

\noindent
Where the operation $| \cdot |$ takes a matrix and sums over rows. In the above case $|\nabla f_c(\mathbf{x})|$ would be a $d$ dimensional object, rather than $d \times c$. 

\vspace{2mm}
\noindent
Preliminary experiments on using the above operators for non-linear class-aware dimensionality reduction, as well as metric reweighing has yielded encouraging results. However, a detailed study is left for future work. 

\vspace{2mm}
\noindent
Finally, a somewhat more challenging avenue for future work would be to obtain consistent estimators for such objects, which are also cheap to estimate. Recall that in Eignemaps type methods, proving consistency involves showing that the eigenvectors of the graph Laplacian approach the eigenfunctions of the corresponding Laplace-Beltrami operator in the limit (see for example \cite{LuxSpectral, NiyogiMaps}). It is not clear if such results (akin to those in sections \ref{sec:eigenspace} and \ref{sec:semiparametric}) could be shown for the gradient based operators defined above. However, it could be a fruitful line of work to try and extend the EGOP and EJOP in such a way that the local geometry of the data could be taken into account. 

%*****************************************
%*****************************************
%*****************************************
%*****************************************
%*****************************************

\part{Group Equivariant Representation Learning}\label{pt:GCNN}
\cleardoublepage
\cleardoublepage
\cleardoublepage
%*****************************************
\chapter{Discriminative Representation Learning for Spherical Data}\label{ch:S2CNN}

In the previous chapter we motivated group equivariant representation learning, in particular discriminative learning of such representations. In this chapter, we give a particular example: We describe a $SO(3)$ equivariant spherical CNN, which while learning $SO(3)$ equivariant representations discriminatively, also has the unusual feature that it can operate completely in Fourier space.  Work presented in this chapter has appeared in the following publication \cite{trivediSpheres2018}.

\vspace{2mm}
\noindent
Our starting point is the following theorem:
\begin{theorem}[Kondor and Trivedi \cite{pmlr-v80-kondor18a}]
	A neural network connecting layers of the form $L^2(X_i, \mathbb{C}^{n_i})$ for a sequence of $G$-spaces $X_i$ is G-equivariant if and only if it is a composition of $G$-convolutions on the $X_i$ spaces and nonlinearities applied to $\mathbb{C}^{n_i}$
\end{theorem}

\vspace{2mm}
\noindent
A more general result for \emph{steerable convolution} appears in the recent works of Cohen \emph{et al.} \cite{Cohen2018a}, \cite{Cohen2018b}. However, for our discussion it suffices to only consider the discussion in \cite{pmlr-v80-kondor18a}. One of the main contributions of \cite{pmlr-v80-kondor18a} is to give a spectral account for group equivariant networks, making the above theorem actionable to design neural networks that are equivariant to the action of general compact groups. In particular, \cite{pmlr-v80-kondor18a} demonstrates that if a compact group $G$ acts on the inputs of the neural network, then there is a natural Fourier transformation with respect to the group $G$, which gives a sequence of Fourier matrices at each layer. In particular, the linear operation at a given layer will be equivariant to the action of $G$ if and only if it involves multiplying the Fourier matrices with learnable weight matrices from the right. It is this insight that we will use to present a neural network architecture that operates on spherical data, while being equivariant to rotations of the sphere. 

\vspace{2mm}
\noindent
We follow recent work on Spherical CNNs by Cohen \emph{et al.} \cite{CohenSpherical18} (also see \cite{Esteves}), which presents a $SO(3)$ equivariant spherical neural network architecture using a generalized $SO(3)$ Fourier transform. One of the drawbacks of their approach is that the non-linearity still needs to be applied in real space, which leads to a non-conventional architecture which involves forward and backward Fourier transforms, which while being expensive can also cause numerical errors. In what follows we propose a spherical CNN architecture that is strictly more general, but at the same time operates entirely in Fourier space. It must be noted that our methodology is more general in its import--it can be used to design neural networks that are equivariant to the action of any continuous compact group. 

\vspace{2mm}
\noindent
In the next section, we describe the general set-up and notation to explicate on our approach.

\section{Notation and Basic Definitions}

\subsection{The Unit Sphere}
The sphere $\mathcal{S}^2$ with unit radius can be defined as the set of points $\mathbf{x} \in \mathbb{R}^3$ such that $\|\mathbf{x} - \mathbf{x}_0\| = 1$, where $\mathbf{x}_0$ is the origin. We can represent a sphere conveniently in spherical coordinates: for some $\mathbf{x} = [x_1, x_2, x_3]$ we can write $x_1 = r \cos\theta \sin\phi$, $x_2 = r \sin\theta \sin\phi$ and $x_3 = r \cos\phi$, where $\theta \in [0,2\pi]$ is the azimuthal coordinate i.e. the longitude and $\phi \in [0, \pi]$ is the polar coordinate i.e. the co-latitude.

\subsection{Signals}

We work with spherical images represented by $f(\theta, \phi)$ and corresponding filters $h(\theta, \phi)$, which are taken to be continuous, complex valued functions. That is:

$$f, h : \mathcal{S}^2 \to \mathbb{C}^k$$
\noindent
For most of the discussion in this chapter we simply work with $f, h : \mathcal{S}^2 \to \mathbb{C}$ for ease of exposition.

\subsection{Rotations}\label{sec:rotations}
We denote a rotation $R \in SO(3)$, and parametrize it by the familiar $ZYZ$ Euler angles $\alpha, \beta, \gamma$ and denote it as $R(\alpha, \beta, \gamma)$. Any rotation $R(\alpha, \beta, \gamma)$ can thus be written as the following sequence of rotations along the $z$ and $y$ axes:
$$R(\alpha, \beta, \gamma) = R_z(\gamma) R_y(\beta) R_z(\alpha) \qquad \alpha, \gamma \in [0,2\pi), \beta \in [0, \pi]$$

\noindent
Thus any spherical image $h(\theta, \phi)$ when subject to rotation $R$ could be denoted as:

\begin{equation}
h_R(\theta, \phi) = R_z(\gamma) R_y(\beta) R_z(\alpha)(h)(\theta, \phi)
\end{equation}

\noindent
Alternatively, if $x$ denotes the point at position $(\theta, \phi)$, we denote it as 
\begin{equation}
h_R(x) = h(R^{-1}x) \qquad R \in SO(3)
\end{equation}

\section{Correlation on the Sphere}
In classical convolutional neural networks, given an input feature map $f: \mathbb{Z}^2 \to \mathbb{R}$ and a filter $g: \mathbb{Z}^2 \to \mathbb{R}$, the value of the output feature map at some point $(-x, -y)$ is simply the inner product between the input and the filter translated by $(x, y)$. Thus the process of correlation here can just be seen as pattern matching: the output map would have a stronger activation if it has a high correlation with the filter. 

\vspace{2mm}
\noindent
In order to define a spherical CNN, we would want to first state an appropriate notion for correlation between $f, g \in L^2(\mathcal{S}^2)$, when $g$ is rotated and matched with $f$ in analogy with the planar CNN case. The difference in this case however is that, unlike in the planar case, where the translation group and the input (the plane) that it acts on are isomorphic to each other, in the spherical case, they are no longer the same. This can lead to some consternation regarding the correct notion of spherical correlation. 

\vspace{2mm}
\noindent
However, as beautifully pointed out by Chirikjian and Kyatkin \cite{Chirikjian}, a definition of correlation that does not veer off from the notion of pattern matching discussed above is rather simple:

\begin{equation}\label{eq:SphereicalCorr}
(h \star f)(R) =  \frac{1}{4 \pi}\int_{0}^{2\pi} \int_{-\pi}^{\pi} [h_R(\theta, \phi)]^\ast f(\theta, \phi) \cos \theta d\theta d\phi \qquad R \in SO(3)
\end{equation}

\noindent
* denotes complex conjugation. Thus the spherical correlation is function on the rotation group $SO(3)$ rather than on $\mathcal{S}^2$.

\vspace{2mm}
\noindent
At first blush, the rather foreboding double integral in equation \ref{eq:SphereicalCorr} is what we would want to implement in our neural network. But this is problematic, one reason for which is that no perfectly symmetrical discretizations for spheres exist \cite{Thurston}. 

\section{Filters and Feature Maps in Fourier Space}
Instead of working with $f(\theta, \phi)$ and $h(\theta, \phi)$ in real space, we instead move to the Fourier domain. It is well known that for functions on the sphere $f \in L^2(\mathcal{S}^2)$, in direct analogy for periodic functions on the circle, the eigenfunctions of the spherical Laplacian give a basis. These basis functions are the so-called spherical harmonics. We can thus represent $f(\theta, \phi)$ and $h(\theta, \phi)$ in terms of their spherical harmonics expansions. 

\begin{equation}
f(\theta, \phi) = \sum_{\ell = 0}^{\infty} \sum_{m = -\ell}^{\ell} \hat{f_{m}^{\ell}}(\theta, \phi) Y_{m}^{\ell}(\theta, \phi) 
\end{equation}

\begin{equation}
h(\theta, \phi) = \sum_{\ell = 0}^{\infty} \sum_{m = -\ell}^{\ell} \hat{h_{m}^{\ell}}(\theta, \phi) Y_{m}^{\ell}(\theta, \phi) 
\end{equation}
\noindent
As might be already clear, $Y_{m}^{\ell}(\theta, \phi)$ are the spherical harmonics with $\ell \geq 0$ and $m \in \{-\ell, \dots, \ell\}$, and are written as:

\begin{equation}
Y_{m}^{\ell}(\theta, \phi) = (-1)^m \sqrt{\frac{(2\ell+1)(\ell - m)!}{4 \pi (\ell +m)!}} P_{m}^{\ell}(\cos \theta) e^{i m \phi}, \qquad m = -\ell, \dots, \ell
\end{equation}
here $P_{m}^{\ell}$ denote the associated Legendre functions.

\vspace{2mm}
\noindent
The coefficients of this spherical Fourier transform are found as follows:

\begin{equation}
\hat{f_{m}^{\ell}} = \frac{1}{4 \pi} \int_{(\theta, \phi) \in \mathcal{S}^2} f(\theta, \phi) Y_{m}^{\ell}(\theta, \phi) \cos \theta d\theta d\phi
\end{equation}

\begin{equation}
\hat{h_{m}^{\ell}} = \frac{1}{4 \pi} \int_{(\theta, \phi) \in \mathcal{S}^2} h(\theta, \phi) Y_{m}^{\ell}(\theta, \phi) \cos \theta d\theta d\phi
\end{equation}

\vspace{2mm}
\noindent
Above we have described how to write $f(\theta, \phi)$ and $h(\theta, \phi)$ in Fourier space. However, recall that correlation defined in equation \ref{eq:SphereicalCorr} was a function on $SO(3)$. We thus need to work with a Fourier transform on the rotation group. Thankfully, non-commutative harmonic analysis \cite{Chirikjian} provides us with such a notion. For functions $f \in L^2(SO(3))$. The Fourier transform can be seen as a change of basis for the $L_2$ space of complex valued functions on $SO(3)$ to the irreducible representations. More specifically, for some function $g: SO(3) \to \mathbb{C}$, the SO(3)-Fourier transform is the collection of the following matrices:

\begin{equation}
G_{\ell} = \int_{SO(3)} g(R)\rho_{\ell}(R) d \mu(R)  \qquad \ell = 0, 1, 2, \dots
\end{equation}
Where $\rho_{\ell}(R) \in \mathbb{C}^{2 \ell + 1 \times 2 \ell + 1}$ are the Wigner D-matrices, which are the irreducible representations for the group $SO(3)$. As a corollary of Schur's first lemma, we also know that the spherical harmonics also provide us with a basis for the irreducible representations of $SO(3)$. That is $Y_R^{\ell} = \rho_{\ell}(R) Y^{\ell}(\theta, \phi)$, and the elements of $\rho_{\ell}(R)$ are given as:

\begin{equation}\label{eq:wigner}
\rho_{\ell}^{mn}(R) = e^{-im\gamma}d^{\ell}_{mn}(\cos \beta) e^{-i n \alpha} \qquad m,n = -\ell \dots, \ell
\end{equation}
Where $d^{\ell}_{mn}$ correspond to the Wigner little-d matrices. This is a good point to revisit the choice to keep activations and filters to be complex valued $f, h : \mathcal{S}^2 \to \mathbb{C}$. Note that $\rho_{\ell}$ matrices are complex valued, and thus allowing activations and filters to be also complex valued simplifies implementation, as well will see when we describe our network. 

\vspace{2mm}
\noindent
Coming back, having defined the Fourier transform, we would also need the inverse Fourier transform, which is defined as below:

\begin{equation}
g(R) = \sum_{\ell=0}^{\infty} Tr[G_{\ell} \rho_{\ell}(R^{-1})]
\end{equation}

\noindent
Having described Fourier transforms for $f \in L^2(\mathcal{S}^2)$ and $f' \in L^2(SO(3))$, we now consider our spherical correlation formulation again:

\begin{equation}\label{eq:SphereicalCorr2}
(h \star f)(R) =  \frac{1}{4 \pi}\int_{0}^{2\pi} \int_{-\pi}^{\pi} [h_R(\theta, \phi)]^\ast f(\theta, \phi) \cos \theta d\theta d\phi \qquad R \in SO(3)
\end{equation}

\vspace{2mm}
\noindent
It can be shown (see \cite{Chirikjian} and Appendix of \cite{CohenSpherical18}) that the $SO(3)$ correlation satisfies a Fourier theorem, reducing finding $SO(3)$ Fourier coefficients to simply pointwise multiplications of the spherical Fourier coefficients. That is, in the above equation, each component is simply given as (here $^\dagger$ denotes the hermitian conjugate): 

\begin{equation}\label{eq:Cohen1}
[\widehat{h \star f}]_{\ell} = \hat{f}_{\ell} \hat{h}_{\ell}^{\dagger} \qquad \ell = 0, 1, \dots, L
\end{equation}
In layers $s = 2, \dots, S$ of a spherical CNN, the filters and the activations are no longer a function on the sphere, but rather on $SO(3)$. In that case, rather unsurprisingly (see equation \ref{eq:wigner} and preceding discussion), we have a similar convolution theorem

\begin{equation}\label{eq:Cohen2}
[\widehat{h \star f}]_{\ell} = F_{\ell} H_{\ell}^{\dagger} \qquad \ell = 0, 1, \dots, L
\end{equation}
and since we are working with functions on $SO(3)$, $F_{\ell}$ and $H_{\ell}$ are of course matrices. 

\vspace{2mm}
\noindent
The approach of Cohen \emph{et al.} is essentially based on equations \ref{eq:Cohen1} and \ref{eq:Cohen2}, where instead of working with the continuous function $f$, which as we have already seen might be complicated to work with, we work with the coefficients $\hat{f}_{\ell} \text{ with } \ell = 0, 1, \dots, L$ and regard them as the activations of the neural network. Likewise $\hat{h}_{\ell} \text{ with } \ell = 0, 1, \dots, L$ are regarded as the learn-able filters. 

\section{A Generalized SO(3)-covariant Spherical CNN}
In the previous section we discussed two Fourier theorems, which form the bedrock on which the work of \cite{CohenSpherical18} was based. We now consider $SO(3)$ correlation i.e. equation \ref{eq:SphereicalCorr2} again, but view it from an algebraic point of view. Specifically, we would like to first nail down, how it behaves under rotations. To begin, we consider the fact that when a spherical function $f(\theta, \phi)$ is subject to a rotation as discussed in \ref{sec:rotations}, then the Fourier components are modulated by the Wigner D-matrix corresponding to the rotation $R$ i.e. 

\begin{equation}
f\mapsto f_R  \iff \hat{f}_{\ell} \mapsto \rho_{\ell}(R) \hat{f}_{\ell}
\end{equation}

\noindent
Likewise, for a function $h: SO(3) \to \mathbb{C}$, which is subject to a rotation $R$, we have an analogous effect on the Fourier matrices i.e. they  are modulated by the corresponding Wigner D-matrix\footnote{The usage of modulation is in analogy with classical Fourier analysis on the real line. Where a shift in the time domain causes the frequency to be multiplied by a complex exponential $x(t-t_0) \iff e^{-i\omega t_0} X(\omega)$. In the case of Fourier analysis on compact groups, a \emph{shift} in the time domain, in this case a rotation, corresponds to a modulation by the irreducible representation in the frequency domain (in this case the Wigner D-matrix corresponding to $R$). Note that while $\mathbb{R}$ is not compact $e^{-i\omega t_0}$ is infact an irreducible representation for $t_0$.}. 

\begin{equation}
h(R') \mapsto h(R^{-1}R')  \iff G_{\ell} \mapsto \rho_{\ell}(R) G_{\ell}
\end{equation}

\noindent
Where $G_{\ell}$ are the Fourier matrices of $h$. The following proposition states that matrices output in equation \ref{eq:Cohen1} exhibits similar behavior. 

\begin{proposition}
Suppose $f:\mathcal{S}^2 \to \mathbb{C}$ is an activation function that under a rotation $R$ transforms as $f \mapsto f(R^{-1}x) \quad R\in SO(3)$, and also suppose $h:\mathcal{S}^2 \to \mathbb{C}$ is a filter. Then, each component in the cross-correlation formula \ref{eq:Cohen1} transforms as:
\begin{equation}\label{eq:crosscorr1}
[\widehat{h \star f}]_{\ell} \mapsto \rho_{\ell}(R)[\widehat{h \star f}]_{\ell}
\end{equation}
\end{proposition}

\noindent
An identical claim can be made in the context of equation \ref{eq:Cohen2}, which we state separately for the sake of completeness. 
\begin{proposition}
	Suppose $f:SO(3) \to \mathbb{C}$ is an activation function that under a rotation $R$ transforms as $f \mapsto f_R(R') \quad R\in SO(3)$, and also suppose $h:SO(3)  \to \mathbb{C}$ is a filter. Then, each component in the cross-correlation formula \ref{eq:Cohen2} transforms as:
	\begin{equation}\label{eq:crosscorr2}
	[\widehat{h \star f}]_{\ell} \mapsto \rho_{\ell}(R)[\widehat{h \star f}]_{\ell}
	\end{equation}
\end{proposition}

\noindent
Notice that equation \ref{eq:crosscorr1} describes how spherical harmonic vectors transform under a rotation, while equation \ref{eq:crosscorr2} describes the behaviour of Fourier matrices under a rotation. This similarity is not superficial. Indeed, we could understand the latter to mean that each column of the Fourier matrices will instead transform according to \ref{eq:crosscorr1}. It is this observation that leads us to a general definition of a SO(3) covariant Spherical CNN. 

\begin{aside}{Generalized $SO(3)$-covariant Spherical CNN}
\begin{definition}
	Let $\mathcal{N}$ be a $S+1$ layer feed-forward network which takes as input $f^0: \mathcal{S}^2 \to \mathbb{C}$. We say that $\mathcal{N}$ is a generalized $SO(3)$-covariant Spherical CNN if the output of each layer can be expressed as a collection of vectors:
	\begin{equation}\label{eq:fragments}
	\hat{f^s} = \Big(\hat{f^s}_{0,1}, \hat{f^s}_{0,2}, \dots, \hat{f^s}_{0,\tau_0^s}, \hat{f^s}_{1,1}, \hat{f^s}_{1,2}, \dots, \hat{f^s}_{1,\tau_1^s}, \dots \dots \dots \hat{f^s}_{L,\tau_L^s}  \Big)
	\end{equation}
	where each $\hat{f^s}_{\ell,j} \in \mathbb{C}^{2 \ell + 1}$ is a $\rho_{\ell}$-covariant vector in the sense of \ref{eq:crosscorr1}. We call each individual $\hat{f^s}_{\ell,j}$ vector an irreducible fragment of $\hat{f^s}$. The integer vector $\tau^s = (\tau_0^s, \tau_1^s, \dots, \tau_L^s)$ that counts the number of fragments for each $\ell$, we call as the type of $\hat{f^s}$
\end{definition}
\end{aside}

\noindent
The above gives a concrete definition of a $SO(3)$-covariant spherical CNN, however, to fully specify the neural network, we have to explicate on three things:

\begin{enumerate}
	\item[1] A linear transformation in each layer that involves learnable weights. Given that the output of each layer has the form in equation \ref{eq:fragments}, we need to specify how they can be mixed. Moreover, the linear transformation must be covariant.
	\item[2] A covariant non-linearity on top of the linear transformation.
	\item[3] Final output that is rotation-invariant. 
\end{enumerate}

We consider these points one by one. 

\section{Covariant Linear Transformations}\label{sec:covlintx}

For a neural network to be covariant, the linear transformation applied at each layer must also be covariant. In the case of the network defined above, the prescription for this is encapsulated in the following proposition. Note that this proposition is a special case of the theorem introduced in the introduction of this chapter.

\begin{proposition}
	Suppose $\hat{f^s}$ is a $SO(3)$-covariant activation function that has the form $	\hat{f^s} = \Big(\hat{f^s}_{0,1}, \hat{f^s}_{0,2}, \dots, \hat{f^s}_{0,\tau_0^s}, \hat{f^s}_{1,1}, \hat{f^s}_{1,2}, \dots, \hat{f^s}_{1,\tau_1^s}, \dots \dots \dots \hat{f^s}_{L,\tau_L^s}  \Big)$, and yet another function $\hat{g^s} = \mathcal{L}(\hat{f^s})$, which is a linear function of $\hat{f^s}$ expressed similarly. Then $\hat{g^s}$ is $SO(3)$-covariant iff each $\hat{g^s}_{\ell,j}$ fragment is a linear combination of fragments from $\hat{f^s}$ with the same $\ell$
\end{proposition}

\noindent
Recall that each fragment $\hat{f^s}_{\ell,j}$ is $2 \ell +1 $ dimensional. If we concatenate all the fragments corresponding to a fixed $\ell$ into a matrix denoted $F_{\ell}^{s}$, and likewise do the same for $\hat{g}$. Then the proposition basically says that $G_{\ell}^s = F_{\ell}^s W_{\ell}^s$ for all $\ell$. It is these parameters that are learned in our network. We must also note the generality of this formulation by considering that both equations \ref{eq:crosscorr1}and \ref{eq:crosscorr2} are particular cases, although the $W_{\ell}$ does not yield to a good interpretation in terms of cross-correlation. 

\section{Covariant Non-Linearities}
Next we turn our attention to the design of a non-linearity that is both differentiable as well as covariant. The choice of non-linearity is absolutely crucial to the success of neural networks. Besides, in the case of networks that are equivariant, usually we work with non-linearities in real space. The reason for this to easy to understand. Being pointwise operations, these are automatically equivariant. Designing a non-linearity that is both covariant and differentiable in Fourier space is far more challenging. It is for this reason that other work in group equivariant networks always apply the non-linearity in real space. However, these backward-forward transformations can be expensive, and can be a cause for a number of complications, including partially losing equivariance due to quadrature.

\vspace{2mm}
\noindent
Here we take a rather unusual route to solve this problem: We take tensor products between fragments, but note that since each of the fragments was irreducible, after tensor products they no longer might be so. To maintain covariance, we would want the fragments to be irreducible. This problem can be solved exactly by the so called Clebsch-Gordan decomposition.

\vspace{2mm}
\noindent
In representation theory, the Clebsch-Gordan decomposition arises in the context of decomposing the tensor product of irreducible representations in a direct sum of irreducibles. In particular, for the group $SO(3)$, it takes the form: 

$$\displaystyle \rho_{\ell_1}(R) \otimes \rho_{\ell_2}(R) = C_{\ell_1, \ell_2} \Bigg[\bigoplus_{\ell = |\ell_1 - \ell_2|}^{\ell_1 + \ell_2} \rho_{\ell}(R) \Bigg] C_{\ell_1, \ell_2}^T$$

\noindent
Equivalently, we can write:

$$\rho_{\ell}(R) = C_{\ell_1, \ell_2, \ell}^T \Big[\rho_{\ell_1}(R) \otimes \rho_{\ell_2}(R) \Big] C_{\ell_1, \ell_2, \ell}$$
Where $C_{\ell_1, \ell_2, \ell}$ are appropriate blocks of $C_{\ell_1, \ell_2}$.
\noindent
The utility of the CG-transform for our purpose is encapsulated in the following lemma:

\begin{lemma}
	Let $\hat{f}_{\ell_1}$ and $\hat{f}_{\ell_2}$ denote $\rho_{\ell_1}$ and $\rho_{\ell_2}$ covariant vectors, and let $\ell$ denote any integer between $|\ell_1 - \ell_2|$ and $\ell_1 + \ell + 2$. Then 
	\begin{equation}\label{eq:CG}
	\hat{g}_{\ell} = C_{\ell_1, \ell_2, \ell}^T \Big[ \hat{f}_{\ell_1} \otimes \hat{f}_{\ell_2}\Big]
	\end{equation}
	is a $\rho_{\ell}$ covariant vector.
\end{lemma}

\vspace{2mm}
\noindent
The algorithm then consists of finding \ref{eq:CG} between all pairs of fragments and then stacking them horizontally, resulting in possibly very wide matrices: in our parlance the activations, or number of channels increase substantially. This can be controlled by fixing, for each $\ell$, the maximum number of fragments to be $\bar{\tau}_{\ell}$. Thankfully, this can be done by using the learnable weight matrices (discussed in section \ref{sec:covlintx}).

\section{Final Invariant Layer}

Since we need the network to be rotation invariant, we implement this by considering only the $\hat{f}_{0,j}^S$ fragments in the last layer. This is because the $\ell = 0$ representation is constant, and thus rotation invariant. We can then connect fully connected layers on top of this last Fourier layer.

\vspace{2mm}
\noindent
With all the ingredients in place, we now describe our experiments.

\section{Experiments}\label{sec-S2CNNexpts}

In this section we describe experiments that give a direct comparison with those reported by Cohen \emph{et al.} \cite{CohenSpherical18}. We choose these experiments as the Spherical CNN proposed in \cite{CohenSpherical18} is the only direct competition to our method. Besides, the comparison is also instructive for two different reasons: Firstly, while the procedure used in \cite{CohenSpherical18} is exactly equivariant in the discrete case, for the continuous case they use a discretization which causes their network to partially lose equivariance with changing bandwidth and depth, whereas our method is always equivariant in the exact sense. Secondly, owing to the nature of their architecture and discretization, \cite{CohenSpherical18} use a more traditional non-linearity i.e. the ReLU, which is also quite powerful. In our case, to maintain full covariance and to avoid the quadrature, we use an unconventional quadratic non-linearity in Fourier space. Because of these two differences, the experiments will hopefully demonstrate the advantages of avoiding the quadrature and maintaining full equivariance despite using a purportedly weaker nonlinearity.  

\vspace{2mm}
\noindent
Cohen \emph{et al.} present two sets of experiments: In the first sequence, they study the numerical stability of their algorithm and quantify the equivariance error due to the quadrature. In the second, they present results on three datasets comparing with other methods. Since our method is fully equivariant, we focus on the second set of experiments.

\subsection{Rotated MNIST on the Sphere} \label{sec - MNIST}

We use a version of MNIST in which the images are painted onto a sphere and use two instances as in \cite{CohenSpherical18}: One in which the digits are projected onto the Northern hemisphere and another in which the digits are projected on the sphere and are also randomly rotated. 

\vspace{2mm}
\noindent
The baseline model is a classical CNN with 5 $\times$ 5 filters and 32, 64, 10 channels with a stride of 3 in each layer (roughly 68K parameters). This CNN is trained by mapping the digits from the sphere back onto the plane, resulting in nonlinear distortions. The second model that we compare to is the Spherical CNN proposed in \cite{CohenSpherical18}. For this method, we use the same architecture as reported by the authors i.e. having layers $S^2$ convolution -- ReLU -- $SO(3)$ convolution -- ReLU -- Fully connected layer with bandwidths 30, 10 and 6, and the number of channels being 20, 40 and 10 (resulting in a total of 58K parameters). 

\vspace{2mm}
\noindent
For our method we use the following architecture: We set the bandlimit $L_{max} = 8$, and keep $\tau_l = \ceil{\frac{12}{\sqrt{L+1}}}$, using a total of 5 layers as described in section \ref{sec: summary}, followed by a fully connected layer of size 256 by 10. We use batch normalization \cite{IoffeSzegedy} on the fully connected layer, and a variant of batch normalization that preserves covariance in the Fourier layers. This method takes a moving average of the standard deviation for a particular fragment for all examples seen during training till then and divides by it, the parameter corresponding to the mean in usual batch normalization is kept to be zero as anything else will break covariance. Finally, we concatenate the output of each \sm{F^s_0} in each internal layer, which are $SO(3)$ invariant scalars, along with that of the last layer to construct the fully connected layer. We observed that having these skip connections was crucial to facilitate smooth training. The total number of parameters was 342086, the network was trained by using the ADAM optimization procedure \cite{adam2015} with a batch size of 50 and a learning rate of $5 \times 10^{-4}$. We also used L2 weight decay of 0.00001 on the trainable parameters. 

\vspace{2mm}
\noindent
We report three sets of experiments: For the first set both the training and test sets were not rotated (denoted NR/NR), for the second, the training set was not rotated while the test was randomly rotated (NR/R) and finally when both the training and test sets were rotated (denoted R/R).

\begin{center}
	\begin{tabular}{l*{3}{c}}
		Method              & NR/NR & NR/R & R/R \\
		\hline \hline
		Baseline CNN & 97.67 & 22.18 & 12  \\
		Cohen \emph{et al.} & 95.59 & 94.62 & 93.4  \\
		Ours (FFS2CNN)  & 96 & 95.86 & 95.8  \\
	\end{tabular}
\end{center}

\vspace{2mm}
\noindent
We observe that the baseline model's performance deteriorates in the three cases, effectively reducing to random chance in the R/R case. While our results are better than those reported in \cite{CohenSpherical18}, they also have another characteristic: they remain roughly the same in the three regimes, while those of \cite{CohenSpherical18} slightly worsen. We think this might be a result of the loss of equivariance in their method. 

\subsection{Atomization Energy Prediction}

Next, we apply our framework to the QM7 dataset \cite{QM71, QM72}, where the goal is to regress over atomization energies of molecules given atomic positions ($p_i$) and charges ($z_i$). Each molecule contains up to 23 atoms of 5 types (C, N, O, S, H). We use the Coulomb Matrix (CM) representation proposed by \cite{QM72}, which is rotation and translation invariant but not permutation invariant. The Coulomb matrix $C \in \mathbb{R}^{N \times N}$ is defined such that for a pair of atoms $i \neq j$, $C_{ij} = (z_iz_j)/(|p_i - p_j|)$, which represents the Coulomb repulsion, and for atoms $i=j$, $C_{ii} = 0.5 z_i^{2.4}$, which denotes the atomic energy due to charge. To test our algorithm we use the same set up as in \cite{CohenSpherical18}: We define a sphere $S_i$ around $p_i$ for each atom $i$. Ensuring uniform radius across atoms and molecules and ensuring no intersections amongst spheres during training, we define potential functions $U_z(x) = \sum_{j\neq i, z_j = z} \frac{zi z}{|x - p_i|}$ for every $z$ and for every $x$ on $S_i$. This yields a $T$ channel spherical signal for each atom in a molecule. This signal is then discretized using Driscol-Healy \cite{DriscollHealy} grid using a bandwidth of $b = 10$. This gives a sparse tensor representation of dimension $N \times T \times 2b \times 2b$ for every molecule.

\vspace{2mm}
\noindent
Our spherical CNN architecture has the same parameters and hyperparameters as in the previous subsection except that $\tau_l = 15$ for all layers, increasing the number of parameters to 1.1 M. Following \cite{CohenSpherical18}, we share weights amongst atoms and each molecule is represented as a $N \times F$ tensor where $F$ represents \sm{F^s_0} scalars concatenated together. Finally, we use the approach proposed in \cite{Zaheer2017} to ensure permutation invariance. The feature vector for each atom is projected onto 150 dimensions using a MLP. These embeddings are summed over atoms, and then the regression target is trained using another MLP having 50 hidden units. Both of these MLPs are jointly trained. The final results are presented below, which show that our method outperforms the Spherical CNN of Cohen \emph{et al.}. The only method that delivers better performance is a MLP trained on randomly permuted Coulomb matrices \cite{Montavon2012}, and as \cite{CohenSpherical18} point out, this method is unlikely to scale to large molecules as it needs a large sample of random permutations, which grows rapidly with $N$.

\begin{center}
	\begin{tabular}{l*{2}{c}}
		Method  &  RMSE \\
		\hline \hline
		MLP/Random CM \cite{Montavon2012} & \bf{5.96}\\
		\hline
		LGIKA (RF) \cite{Raj2016} & 10.82\\
		\hline
		RBF Kernels/Random CM \cite{Montavon2012} & 11.42\\
		\hline
		RBF Kernels/Sorted CM \cite{Montavon2012} & 12.59\\
		\hline
		MLP/Sorted CM \cite{Montavon2012} & 16.06\\
		\hline
		Spherical CNN \cite{CohenSpherical18} &  8.47  \\
		\hline
		Ours (FFS2CNN)& \bf{7.91} \\
	\end{tabular}
\end{center}

\subsection{3D Shape Recognition}

Finally, we report results for shape classification using the SHREC17 dataset \cite{SHREC}, which is a subset of the larger ShapeNet dataset \cite{ShapeNet} having roughly 51300 3D models spread over 55 categories. It is divided into a 70/10/20 split for train/validation/test. Two versions of this dataset are available: A regular version in which the objects are consistently aligned and another where the 3D models are perturbed by random rotations. Following \cite{CohenSpherical18} we focus on the latter version, as well as represent each 3D mesh as a spherical signal by using a ray casting scheme. For each point on the sphere, a ray towards the origin is sent which collects the ray length, cosine and sine of the surface angle. In addition to this, ray casting for the convex hull of the mesh gives additional information, resulting in 6 channels. The spherical signal is discretized using the Discroll-Healy grid \cite{DriscollHealy} with a bandwidth of 128. We use the code provided by \cite{CohenSpherical18} for generating this representation. 

\vspace{2mm}
\noindent
We use a ResNet style architecture, but with the difference that the full input is not fed back but rather different frequency parts of it. We consider $L_{max} = 14$, and first train a block only till $L = 8$ using $\tau_l = 10$ using 3 layers. The next block consists of concatenating the fragments obtained from the previous block and training for two layers till $L = 10$, repeating this process till $L_{max}$ is reached. These later blocks use $\tau_l = 8$.  As earlier, we concatenate the \sm{F^s_0} scalars from each block to form the final output layer, which is connected to 55 nodes forming a fully connected layer. We use Batch Normalization in the final layer, and the normalization discussed in \ref{sec - MNIST} in the Fourier layers. The model was trained with ADAM using a batch size of 100 and a learning rate of $5 \times 10^{-4}$, using L2 weight decay of 0.0005 for regularization. The total number of parameters was roughly 2.3M. We compare our results using the SHREC competition evaluation script to some of the top performing models on SHREC (which use architectures specialized to the task) as well as the model of Cohen \emph{et al.}. Our method, like the model of Cohen \emph{et al.} is task agnostic and uses the same representation. Despite this, it is able to consistently come second or third in the competition (while being neck to neck with Cohen \emph{et al.}), showing that it affords an efficient method to learn from spherical signals. 

\begin{center}
	\begin{tabular}{l*{5}{c}}
		Method              & P@N & R@N & F1@N & mAP & NDCG \\
		\hline \hline
		Tatsuma\_ReVGG & 0.705 & 0.769 & 0.719 & 0.696 & 0.783  \\
		Furuya\_DLAN & 0.814 & 0.683 & 0.706 & 0.656 & 0.754  \\
		SHREC16-Bai\_GIFT  & 0.678 & 0.667 & 0.661 & 0.607 & 0.735  \\
		Deng\_CM-VGG5-6DB  & 0.412 & 0.706 & 0.472 & 0.524 & 0.624  \\
		Spherical CNNs \cite{CohenSpherical18}  & 0.701 & 0.711 & 0.699 & 0.676 & 0.756  \\
		\hline
		FFS2CNNs (ours)  & 0.707 & 0.722 & 0.701 & 0.683 & 0.756  \\
	\end{tabular}
\end{center}

\section{Conclusion}
In conclusion, in this chapter, we presented a SO(3)-equivariant neural network architecture for spherical data, that operates entirely in Fourier space, while using tensor products and the Clebsch-Gordan decomposition as the only source of non-linearity. We report strong (and perhaps surprising) experimental results. While we specifically presented a spherical CNN, our approach is more widely applicable in that it also provides a formalism for the design of fully Fourier neural networks that are equivariant to the action of any continuous compact group.

%*****************************************
%*****************************************
%*****************************************
%*****************************************
%*****************************************

\cleardoublepage
%\include{Chapters/Chapter12}
%\cleardoublepage
%*****************************************
\chapter{Conclusions and Future Directions}\label{ch:future}

\section{Part I}
\subsection{Conclusions}
In chapter \ref{ch:discintro} we reviewed some relevant literature on metric learning; following which,  in chapter \ref{ch:mlng}, we proposed a metric learning method that makes a more direct attempt to optimize for $k$-NN accuracy than existing methods. While the approach is more general in its formulation (in that it can handle non-linear metrics as well), in chapter \ref{ch:mlng} we demonstrated its efficacy for learning Mahalanobis metrics while comparing to a number of popular competing methods. In chapter \ref{ch:extensions}, we proposed a number of extensions of this approach, applying it to asymmetric similarity learning, discriminative learning of Hamming distance, and metric learning for improving $k$-NN regression performance. In each case we reported competitive results. Below we underline some straightforward avenues for future work:
\subsection{Future Directions}
\begin{enumerate}
	\item[1] A drawback of the approaches presented in Part \ref{pt:discml}, with the exception of section \ref{sec:mlknnregression}, is poor scalability. For every gradient update, the procedures require exact inference and loss-augmented inference. For small dataset sizes this is desirable, however being expensive operations (see section \ref{app:proof}) they restrict scaling these methods to very large datasets. One future avenue of work is to make these methods more scalable while retaining some of their positive characteristics as constrasted to methods such as Large Margin Nearest Neighbors (LMNN). Some strategies to achieve this could take the route of doing exact inference for $h^\ast$ and $\hat{h}$ for a fixed number of gradient updates $N' << N$ (where $N$ is the number of training examples) in the beginning of the optimization. Once the initial Euclidean metric is improved to a somewhat better performing metric, the sets $h^\ast$ and $\hat{h}$ can be fixed for the next $p$ gradient updates, after which they are updated again by doing exact inference. This process could be repeated to convergence. Another route could be to pick large batches and do exact inference in only a given batch, and not the entire dataset.
	\item[2] Yet another avenue for future work is to propose approximate inference procedures for $h^\ast$ and $\hat{h}$, and combining them with the approaches outlined above. 
	\item[3] Extending the approach using deep neural networks to do the mapping is also an obvious extension. This was explored by the dissertation author, but not extensively. It is arguable that the metric thus learned could be a better proxy for similarity than approaches based on triplet based losses. 
	\item[4] For section \ref{sec:mlknnregression}, unlike other approaches presented in Part  \ref{pt:discml}, the inference procedures were intractable. We thus resorted to a modification of the loss function to make inference tractable. It would be interesting to explore approximation algorithms for the original, intractable formulations for $h^\ast$ and $\hat{h}$. Yet another approach to this problem, explored by the dissertation author to some degree, is to keep the original intractable objective, and devising a Metropolis-Hastings type procedure for sampling sets and then updating the metric. Lastly, it would also be interesting to cast the framework in section \ref{sec:mlknnregression} as a structured prediction energy network in the spirit of Tu and Gimpel \cite{Lifu}.
	
\end{enumerate}

\section{Part II}
\subsection{Conclusions}

In part \ref{pt:EGOP} of the dissertation, we proposed a simple estimator for the Expected Gradient Outerproduct (EGOP) $$\expectation_{\mathbf{x}} G(\mathbf{x}) \triangleq \expectation_\mathbf{x}\paren{\nabla f(\mathbf{x}) \cdot \nabla f(\mathbf{x})^\top}, $$  moreover, we also showed that it remains statistically consistent under mild assumptions. The primary use of the estimated EGOP was as the underlying metric in non-parametric regression, and we showed that it improved performance as compared to the Euclidean distance in several real world datasets. 
We also generalized the EGOP to the multiclass case, proposing a variant called the Expected Jacobian Outer Product (EJOP)
$$\mathbb{E}_{\mathbf{x}} G(\mathbf{x}) \triangleq  \expectation_\mathbf{x}\paren{\mathbf{J}_{f}(\mathbf{x}) \mathbf{J}_{f}(\mathbf{x})^T},$$
for which we also proposed a simple estimator and showed that it remained statistically consistent under similarly mild assumptions. We also showed that the estimated EJOP improved non-parameteric classificaiton when used as a metric.

\subsection{Future Directions}
\begin{enumerate}
	\item[1] One immediate use case for the approaches presented in Part \ref{pt:EGOP}, namely, the estimated Expected Gradient Outer Product (EGOP) and Expected Jacobian Outer Product (EJOP), is for dimensionality reduction, that unlike PCA type methods recover a subpsace most relevant to predicting the output. This has not been explored in detail and could potentially be a useful addition to the standard toolbox for dimensionality reduction.
	\item[2]  The EGOP and the EJOP use gradients, which are local objects, but due to the expectation taken over $\mathbf{x}$, they lose all local information, only giving the average variation of the unknown regression or classification function $f$ in direction $\mathbf{v}$. It would be interesting to explore the utility of diffusion based objects that take into account local geometry as well, first analogous to the EGOP 
	$$\mathbf{W}_{i,j} = \mathbf{W}_{f}(\mathbf{x}_i, \mathbf{x}_j) = exp \Bigg( - \frac{\|\mathbf{x}_i - \mathbf{x}_j\|^2}{\sigma_1} - \frac{\|\frac{1}{2}(\nabla f(\mathbf{x}_i) + \nabla f(\mathbf{x}_j) \dot (\mathbf{x}_i - \mathbf{x}_j) \|^2}{\sigma_2} \Bigg)$$
	and then analogous to the EJOP
	$$\mathbf{W}_{i,j} = \mathbf{W}_{f}(\mathbf{x}_i, \mathbf{x}_j) = exp \Bigg( - \frac{\|\mathbf{x}_i - \mathbf{x}_j\|^2}{\sigma_1} - \frac{\|\frac{1}{2}(|\nabla f_c(\mathbf{x}_i)| + |\nabla f_c(\mathbf{x}_j|) \dot (\mathbf{x}_i - \mathbf{x}_j) \|^2}{\sigma_2} \Bigg)$$
	$| \cdot |$ takes a matrix and sums over rows, and explore them for both recovering a metric, as well as for non-linear label-aware dimensionality reduction. 
	\item[3] For the operators defined above it would also be interesting to obtain consistency results, similar in spirit to those obtained for Laplacian Eigenmaps type methods \cite{LuxSpectral, NiyogiMaps}. For such methods, demonstrating consistency amounts to showing that the eigenvectors of the graph Laplacian converge to the eigenfucntions of the corresponding Laplace-Beltrami operator in the limit. However, this promises to be a rather challenging project, which might also require considerable refinement in definitions of these objects.
\end{enumerate}

\section{Part III}

\subsection{Conclusions}
In chapter \ref{ch:gcnnintro} we briefly reviewed work on discriminative group equivariant representation learning, arguing that equivariance to symmetry transformations affords a strong inductive bias in various tasks. In chapter \ref{ch:S2CNN}, following recent work by Kondor and Trivedi \cite{pmlr-v80-kondor18a}, we proposed a SO(3)-equivariant neural network architecture for spherical data, that operates entirely in Fourier space, while using tensor products and the Clebsch-Gordan decomposition as the only source of non-linearity. We reported strong experimental results, and emphasized the wider applicability of our approach, in that it also provides a formalism for the design of fully Fourier neural networks that are equivariant to the action of any continuous compact group.
\subsection{Future Directions}
\begin{enumerate}
	\item[1] We first outline a future avenue for work that relates directly to the contributions presented in Chapter \ref{ch:S2CNN}. Although the network architecture presented is the most general possible for the problem and mathematically elegant in its conception, while also giving excellent performance, it does lead to networks that are considerably bulkier than networks trained by \cite{CohenSpherical18}. In view of the dissertation author this inefficiency might be a consequence of the non-locality of filters. In vision tasks, it is perhaps much better motivated to use filters that operate on a small, spatially contiguous domain of the input. Thus, the use of more global filters could be the reason that the networks slid toward having more parameters to also pick up more local features on their own. Enforcing locality of filters in order to improve the efficiency of our network further is the most immediate line of future work. This should have relevance to not just $SO(3)-$equivariant networks that operate on data that lives on $\mathcal{S}^2$, but to networks for vision tasks that are required to be equivariant to the action of general compact continuous groups.
	\item[2] More directly related to the more general theme covered in Part \ref{pt:GCNN} of this dissertation, is to design convolutional neural networks that encode more structure from the data and task at hand, by considering different groups and their homogeneous spaces. A simple example would be to design a $SIM(3)-$equivariant architecture, to follow works that present $SE(3)-$equivariant networks (for example see \cite{CohenSE3}). This could perhaps be considered low-hanging fruit, notwithstanding the fact that fast implementations of such architectures may require considerable engineering effort.
	\item[3] There are many applications where exact invariances to symmetry transformations are important, such as in Robotics and motion planning, tomography, camera calibration, molecular dynamics, protein kinematics etc. In these areas there already exists a large literature on using non-commutative harmonic  analysis for functions defined on some homogeneous space of a group of interest. An extensive review of such approaches and half a dozen applications is given in \cite{Chirikjian}. Naturally, designing equivariant architectures that extend older approaches to also avoid feature engineering is an obvious line of work to pursue. 
	\item[4] Most of the work on group equivariant neural networks reviewed in chapter \ref{ch:gcnnintro} relies on the assumption that the functions are defined on a suitable homogeneous space of the symmetry group of interest. Indeed, the architecture presented in chapter \ref{ch:S2CNN} is rooted in the fact that the manifold $\mathcal{S}^2$ is a homogeneous space of the rotation group $SO(3)$. There is recent interest in extending the convolutional neural network formalism to more general manifolds \cite{Masci:2015:GCN:2919341.2920992} that might not come equipped with a clear group action. Work that prescribes construction of theoretically well motivated convolutional networks on such spaces promises to be a very fruitful line for future work. 
	\item[5] In most of the work discussed thus far, we have considered neural networks that are equivariant to explicit (and known) symmetry transforms. A considerably difficult project would be to instead to \emph{learn the symmetry group}, without prior knowledge of symmetries in the data. The only work that we are aware of in this direction is that of Anselmi \emph{et al.} \cite{Anselmi2017}.
	\item[6] The theory and design of covariant neural architectures in the context of recurrent neural networks and general dynamical systems also promises to be an interesting project. A simple situation that illustrates this occurs in Koopman mode analysis \cite{doi:10.1063/1.4772195}, where we want the Koopman invariant subspace to respect some underlying symmetry (for example if the physical system is subject to a rotation). To our knowledge there is no work toward this very reasonable end goal.
	\item[7] It would be interesting to explore connections and usages of equivariant networks in the context of \emph{Pattern Theory} \cite{MumfordPatternTheory}, which is a mathematical formalism to study patterns from the bottom up--with a focus on building compositional vocabularies. While pattern theory has influenced directly, or indirectly many modern machine learning algorithms, it by itself has largely been forgotten in the machine learning mainstream. However, the general philosophy and approach of Pattern Theory remains relevant and can be seen as echoed in many recent works in group equivariant architectures, and could also provide inspiration for future work. While broad in its coverage, it would be useful to consider salient aspects in some notable works within pattern theory. First is the construction of shapes, manifolds and surfaces. Second is the comparison of such objects, and lastly is a methodology to define variability (deformations) and using appropriate probability measures for inference. Usually the space of variability of such objects is an orbit under symmetry transformations. As already hinted, yet another important aspect about Pattern Theory is its focus on compositionality. The case for general covariant compositional architectures has been made in recent work \cite{doi:10.1063/1.5024797}, \cite{trivediCCN1}, and it would also be interesting to extend these works to also be able to define probabilistic models in the spirit of \cite{MumfordPatternTheory}.There is rich mathematical literature on probabilities on algebraic structures (see for example \cite{Grenander}) that Pattern Theory draws upon and could also provide fertile ground for the growth of more general, topologically sane work in neural networks.
\end{enumerate}

%*****************************************
%*****************************************
%*****************************************
%*****************************************
%*****************************************

%*********************************
% ********************************************************************
% Backmatter
%*******************************************************
%%\appendix
%\renewcommand{\thechapter}{\alph{chapter}}
%%\cleardoublepage
%%\part{Appendix}
%%\include{Chapters/Chapter0A}
%********************************************************************
% Other Stuff in the Back
%*******************************************************
\nocite{*}
\cleardoublepage%********************************************************************
% Bibliography
%*******************************************************
% work-around to have small caps also here in the headline
% https://tex.stackexchange.com/questions/188126/wrong-header-in-bibliography-classicthesis
% Thanks to Enrico Gregorio
\defbibheading{bibintoc}[\bibname]{%
  \phantomsection
  \manualmark
  \markboth{\spacedlowsmallcaps{#1}}{\spacedlowsmallcaps{#1}}%
  \addtocontents{toc}{\protect\vspace{\beforebibskip}}%
  \addcontentsline{toc}{chapter}{\tocEntry{#1}}%
  \chapter*{#1}%
}
\printbibliography[heading=bibintoc]

@book{GregThesis,
	author={Shakhnarovich, Gregory},
	title={Learning task-specific similarity},
	year={2005},
	publisher={Ph.D. thesis, Massachutsetts Institute of Technology}
}

@article{MacQueen,
	title={Some Methods for classification and Analysis of Multivariate Observations},
	author={James B. MacQueen},
	journal={Proceedings of 5th Berkeley Symposium on Mathematical Statistics and Probability},
	volume={1},
	pages={281--297},
	year={1967},
	publisher={University of California Press}
}

@article{Luxburg,
	title={A tutorial on spectral clustering},
	author={Von Luxburg, Ulrike},
	journal={Statistics and computing},
	volume={17},
	number={4},
	pages={395--416},
	year={2007},
	publisher={Springer}
}

@book{GPBook,
	author={Rasmussen, Carl Edward and Williams, Christopher K. I.},
	title={Gaussian Processes for Machine Learning (Adaptive Computation and Machine Learning)},
	year={2005},
	isbn = {026218253X },
	publisher = {MIT Press},
	address = {Cambridge, MA, USA},
}

@book{EncycDistances,
	author={Deza, Michel Marie and Deza, Elena},
	title={Encyclopedia of Distances},
	year={2009},
	isbn = { 3642002331},
	publisher = {Springer, Berlin, Heidelberg}
}

@article{NN1967,
	title={Nearest Neighbor Pattern Classification},
	author={Cover, Thomas and Hart, Peter},
	journal={IEEE Transactions on Information Theory},
	volume={13},
	number={1},
	pages={21-27},
	year={1967},
	publisher={IEEE}
}

@book{GregBook,
	author = {Shakhnarovich, Gregory and Darell, Trevor and Indyk, Piotr},
	title = {Nearest-neighbor methods in learning and vision: theory and practice (neural information processing)},
	year = {2006},
	isbn = {026219547X},
	publisher = {MIT Press},
	address = {Cambridge, MA, USA},
}

@article{Stone77,
	title={Consistent nonparametric regression},
	author={Stone, Charles J.},
	journal={The Annals of Statistics},
	pages={595-620},
	year={1977}
}

@article{mahalanobis,
	title={On the generalized distance in statistics},
	author={Mahalanobis, Prasanta Chandra},
	journal={Proceedings National Institute of Science of India},
	volume={49},
	number={2},	
	pages={234-256},
	year={1977}
}

@article{Fukunaga1981,
	title={The optimal distance measure for nearest neighbor classification},
	author={Fukunaga, Keinosuke and Short, R.},
	journal={IEEE transactions on Information Theory},
	volume={27},
	number={5},	
	pages={622-627},
	year={1981}
}

@article{HastieTibshirani1996,
	title={Discriminant Adaptive Nearest Neighbor Classification},
	author={Hastie, Trevor and Tibshirani, Robert.},
	journal={IEEE Transactions on Pattern Analysis and Machine Intelligence},
	volume={18},
	number={6},	
	pages={607-616},
	year={1996}
}

@inproceedings{mcfee10_mlr,
	author  = {McFee, B. and Lanckriet, G. R. G.},
	year    = {2010},
	title   = {Metric Learning to Rank},
	booktitle = {Proceedings of the 27th International Conference on Machine
	Learning (ICML'10)},
}

@inproceedings{yu2009learning,
	title={Learning structural SVMs with latent variables},
	author={Yu, Chun-Nam John and Joachims, Thorsten},
	booktitle={Proceedings of the 26th Annual International Conference on Machine Learning},
	pages={1169--1176},
	year={2009},
	organization={ACM}
}

@article{joachims2009cutting,
	title={Cutting-plane training of structural SVMs},
	author={Joachims, Thorsten and Finley, Thomas and Yu, Chun-Nam John},
	journal={Machine Learning},
	volume={77},
	number={1},
	pages={27--59},
	year={2009},
	publisher={Springer}
}

@inproceedings{weinberger2008fast,
	title={Fast solvers and efficient implementations for distance metric learning},
	author={Weinberger, Kilian Q and Saul, Lawrence K},
	booktitle={Proceedings of the 25th international conference on Machine learning},
	pages={1160--1167},
	year={2008},
	organization={ACM}
}

@article{weinberger2009distance,
	title={Distance metric learning for large margin nearest neighbor classification},
	author={Weinberger, Kilian Q and Saul, Lawrence K},
	journal={The Journal of Machine Learning Research},
	volume={10},
	pages={207--244},
	year={2009},
	publisher={JMLR. org}
}

@inproceedings{kedem2012non,
	title={Non-linear Metric Learning},
	author={Kedem, Dor and Tyree, Stephen and Weinberger, Kilian and Sha, Fei and Lanckriet, Gert},
	booktitle={Advances in Neural Information Processing Systems 25},
	pages={2582--2590},
	year={2012}
}

@article{tsochantaridis2006large,
	title={Large margin methods for structured and interdependent output variables},
	author={Tsochantaridis, Ioannis and Joachims, Thorsten and Hofmann, Thomas and Altun, Yasemin and Singer, Yoram},
	journal={Journal of Machine Learning Research},
	volume={6},
	number={2},
	pages={1453},
	year={2006},
	publisher={THE MIT PRESS}
}

@article{yuille2002concave,
	title={The concave-convex procedure (CCCP)},
	author={Yuille, Alan L and Rangarajan, Anand and Yuille, AL},
	journal={Advances in neural information processing systems},
	volume={2},
	pages={1033--1040},
	year={2002},
	publisher={MIT; 1998}
}

@article{mcfee2010metric,
	title={Metric learning to rank},
	author={McFee, Brian and Lanckriet, Gert},
	year={2010},
	publisher={Citeseer}
}

@inproceedings{goldberger2004neighbourhood,
	title={Neighbourhood components analysis},
	author={Goldberger, Jacob and Roweis, Sam and Hinton, Geoff and Salakhutdinov, Ruslan},
	booktitle={NIPS},
	year={2004}
}

@article{dannytarlowkNCA,
	title ={Stochastic k-Neighborhood Selection for Supervised and Unsupervised Learning},
	author={Tarlow, David and Swersky, Kevin and Sutskever, Ilya and Zemel, Richard S},
	journal={Proceedings of the 30th International Conference on Machine Learning},
	volume={28},
	pages={199--207},
	year={2013},
	publisher={JMLR. org}
}

@inproceedings{globersoncollaps06,
	title = {Metric Learning by Collapsing Classes},
	author = {Globerson, Amir and Roweis, Sam},
	booktitle = {Advances in Neural Information Processing Systems 18},
	editor = {Y. Weiss and B. Sch\"{o}lkopf and J. Platt},
	publisher = {MIT Press},
	address = {Cambridge, MA},
	pages = {451--458},
	year = {2006}
}

@inproceedings{Joachims05asupport,
	author = {Thorsten Joachims},
	title = {A support vector method for multivariate performance measures},
	booktitle = {Proceedings of the 22nd International Conference on Machine Learning},
	year = {2005},
	pages = {377--384},
	publisher = {ACM Press}
}

@inproceedings{Xing02distancemetric,
	author = {Eric P. Xing and Andrew Y. Ng and Michael I. Jordan and Stuart Russell},
	title = {Distance Metric Learning, with Application to Clustering with Side-information},
	booktitle = {Advances in Neural Information Processing Systems 15},
	year = {2002},
	pages = {505--512},
	publisher = {MIT Press}
}

@article{felzenszwalb2010object,
	title={Object detection with discriminatively trained part-based models},
	author={Felzenszwalb, Pedro F and Girshick, Ross B and McAllester, David and Ramanan, Deva},
	journal={Pattern Analysis and Machine Intelligence, IEEE Transactions on},
	volume={32},
	number={9},
	pages={1627--1645},
	year={2010},
	publisher={IEEE}
}

@inproceedings{datar2004locality,
	title={Locality-sensitive hashing scheme based on p-stable distributions},
	author={Datar, Mayur and Immorlica, Nicole and Indyk, Piotr and Mirrokni, Vahab S},
	booktitle={Proceedings of the twentieth annual symposium on Computational geometry},
	pages={253--262},
	year={2004},
	organization={ACM}
}

@inproceedings{itml2007davis,
	title={Information-theoretic metric learning},
	author={Davis, Jason V and Kulis, Brian and Jain, Prateek and Sra, Suvrit and Dhillon, Inderjit S},
	booktitle={Proceedings of the 24th international conference on Machine learning},
	pages={209--216},
	year={2007},
	organization={ACM}
}

@inproceedings{beygelzimer2006cover,
	title={Cover trees for nearest neighbor},
	author={Beygelzimer, Alina and Kakade, Sham and Langford, John},
	booktitle={Proceedings of the 23rd international conference on Machine learning},
	pages={97--104},
	year={2006},
	organization={ACM}
}

@article{arya1998optimal,
	title={An optimal algorithm for approximate nearest neighbor searching fixed dimensions},
	author={Arya, Sunil and Mount, David M and Netanyahu, Nathan S and Silverman, Ruth and Wu, Angela Y},
	journal={Journal of the ACM (JACM)},
	volume={45},
	number={6},
	pages={891--923},
	year={1998},
	publisher={ACM}
}

@article{kulis2009learning,
	title={Learning to hash with binary reconstructive embeddings},
	author={Kulis, Brian and Darrell, Trevor},
	journal={Advances in neural information processing systems},
	volume={22},
	pages={1042--1050},
	year={2009}
}

@inproceedings{wang2010sequential,
	title={Sequential projection learning for hashing with compact codes},
	author={Wang, Jun and Kumar, Sanjiv and Chang, Shih-Fu},
	booktitle={Proceedings of International Conference on Machine Learning},
	year={2010}
}

@article{norouzi2011minimal,
	title={Minimal loss hashing for compact binary codes},
	author={Norouzi, Mohammad and Fleet, David J},
	journal={mij},
	volume={1},
	pages={2},
	year={2011}
}

@inproceedings{norouzi2012hamming,
	title={Hamming Distance Metric Learning},
	author={Norouzi, Mohammad and Fleet, David and Salakhutdinov, Ruslan},
	booktitle={Advances in Neural Information Processing Systems 25},
	pages={1070--1078},
	year={2012}
}

@inproceedings{gong2011iterative,
	title={Iterative quantization: A procrustean approach to learning binary codes},
	author={Gong, Yunchao and Lazebnik, Svetlana},
	booktitle={Computer Vision and Pattern Recognition (CVPR), 2011 IEEE Conference on},
	pages={817--824},
	year={2011},
	organization={IEEE}
}

@inproceedings{boser1992,
	title={A training algorithm for optimal margin classifiers},
	author = {Boser, Bernhard E. and Guyon, Isabelle M. and Vapnik, Vladimir N.},
	booktitle = {Proceedings of the 5th Annual Workshop on Computational Learning Theory (COLT'92)},
	pages = {144--152},
	year={1992},
	publisher = {ACM Press},
}

@article{stange2008efficient,
	title={On the efficient update of the singular value decomposition},
	author={Stange, Peter},
	journal={PAMM},
	volume={8},
	number={1},
	pages={10827--10828},
	year={2008},
	publisher={Wiley Online Library}
}

@inproceedings{neyshabur13,
	title={The power of asymmetry in binary hashing},
	author={Behnam Neyshabur and Nathan Srebro and Ruslan Salakhutdinov and Yury Makarychev and Payman Yadollahpour},
	booktitle={Advances in Neural Information Processing Systems},
	pages={2823--2831},
	year={2013}
}

@inproceedings{neyshabur15,
	title={On Symmetric and Asymmetric LSHs for Inner Product Search},
	author={Behnam Neyshabur and Nathan Srebro},
	booktitle={International Conference on Machine Learning},
	pages={1926-1934},
	year={2015}
}

@inproceedings{weiss08,
	title={Spectral Hashing},
	author={Yair Weiss and Antonio Torralba and Rob Fergus},
	booktitle={Advances in Neural Information Processing Systems},
	year={2008}
}

@inproceedings{IndykMotwani,
	title={Approximate Nearest Neighbors: Towards Removing the Curse of Dimensionality},
	author={Piotr Indyk and Rajeev Motwani},
	booktitle={Proceedings of the 13th annual ACM Symposium on Theory of Computing},
	pages={604--613},
	year={1998}
}

@article{Rabani,
	title={Efficient search for approximate nearest neighbor in high dimensional spaces},
	author={Kushilevitz, Eyal and Ostrovsky, Rafail and Rabani, Yuval},
	journal={SIAM Journal on Computing},
	volume={30},
	number={2},
	pages={457--474},
	year={2000},
	publisher={SIAM}
}

@article{SemanticHashing,
	title={Semantic Hashing},
	author={Salakhutdinov, Ruslan and Hinton, Geoffrey E.},
	journal={International Journal of Approximate Reasoning},
	volume={50},
	number={7},
	pages={969--978},
	year={2009},
	publisher={Elsevier}
}

@article{gordoasym,
	title={Asymmetric distances for binary embeddings},
	author={Gordo, Albert and Perronnin, Florent and Gong, Yunchao and Lazebnik, Svetlana},
	journal={IEEE transactions on pattern analysis and machine intelligence},
	volume={36},
	number={1},
	pages={33--47},
	year={2014},
	publisher={IEEE}
}

@article{charikarasym,
	title={Asymmetric distance estimation with sketches for similarity search in high-dimensional spaces},
	author={Dong, Wei and Charikar, Moses and Li, Kai},
	journal={Proceedings of the 31st annual international ACM SIGIR conference on Research and development in information retrieval},
	pages={123--130},
	year={2008},
	publisher={ACM}
}

@inproceedings{MLKR,
	title={Metric Learning for Kernel Regression},
	author={Kilian Q. Weinberger and Gerald Tesauro},
	booktitle={Artificial Intelligence and Statistics},
	pages={612-619},
	year={2007}
}

@article{SVR,
	title={A tutorial on support vector regression},
	author={Smola, Alex J. and Sch{\"o}lkopf, Bernhard},
	journal={Statistics and Computing},
	volume={14},
	number={3},
	pages={199-222},
	year={2004},
	publisher={Springer}
}

@inproceedings{GPR,
	title={Gaussian Processes for Regression},
	author={Christopher K. I. Williams and Carl E. Rasmussen},
	booktitle={Advances in Neural Processing Sysmtems},
	pages={514-520},
	year={1996}
}

@inproceedings{Mannor,
	title={Automatic basis function construction for approximate dynamic programming and reinforcement learning},
	author={Philipp W. Keller and Shie Mannor and Doina Precup},
	booktitle={Proceedings of the 23rd International Conference on Machine Learning},
	year={2006}
}

@article{delve,
	title={The DELVE Manual},
	author={Rasmussen, Carl E. and Neal, Radford M. and Hinton, Geoffrey E. and Camp, Drew van and Revow, Michael and Ghahramani, Zoubin and Kustra, R. and Tibshirani, Robert},
	year={1996}
}

@article{doi:10.1137/1116025,
	author = {Vapnik, V. and Chervonenkis, A.},
	title = {On the Uniform Convergence of Relative Frequencies of Events to Their Probabilities},
	journal = {Theory of Probability and Its Applications},
	volume = {16},
	number = {2},
	pages = {264-280},
	year = {1971},
}

@article{sliced,
	author = {Ker-Chau Li},
	title = {Sliced Inverse Regression for Dimension Reduction},
	journal = {Journal of the American Statistical Association},
	volume = {86},
	number = {414},
	pages = {316-327},
	year = {1991},
}

@article{RSSB:RSSB341,
	author = {Xia, Yingcun and Tong, Howell and Li, W. K. and Zhu, Li-Xing},
	title = {An adaptive estimation of dimension reduction space},
	journal = {Journal of the Royal Statistical Society: Series B (Statistical Methodology)},
	volume = {64},
	number = {3},
	publisher = {Blackwell Publishers},
	issn = {1467-9868},
	url = {http://dx.doi.org/10.1111/1467-9868.03411},
	doi = {10.1111/1467-9868.03411},
	pages = {363--410},
	year = {2002},
}

@article{DBLP:journals/jmlr/WuGMM10,
	author    = {Qiang Wu and
	Justin Guinney and
	Mauro Maggioni and
	Sayan Mukherjee},
	title     = {Learning Gradients: Predictive Models that Infer Geometry
	and Statistical Dependence},
	journal   = {Journal of Machine Learning Research},
	volume    = {11},
	year      = {2010},
	pages     = {2175-2198},
	ee        = {http://portal.acm.org/citation.cfm?id=1859926},
	bibsource = {DBLP, http://dblp.uni-trier.de}
}

@article{RPT2,
	author    = {Ren-Cang Li},
	title     = {Relative perturbation theory II: eigenspace and singular space variations},
	journal   = {SIAM Journal on Matrix Analysis and Applications},
	volume    = {20},
	year      = {1999},
	pages     = {471-492},
	ee        = {http://dx.doi.org/10.1137/120874795},
	bibsource = {DBLP, http://dblp.uni-trier.de}
}

@article{RPT1,
	author    = {Ren-Cang Li},
	title     = {Relative perturbation theory I: eigenvalue and singular value variations},
	journal   = {SIAM Journal on Matrix Analysis and Applications},
	volume    = {19},
	year      = {1998},
	pages     = {956-982},
	ee        = {http://dx.doi.org/10.1137/120874795},
	bibsource = {DBLP, http://dblp.uni-trier.de}
}

@inproceedings{DBLP:conf/nips/2012,
	author    = {Samory Kpotufe and	Abdeslam Boularias},
	title     = {Gradient Weights help Nonparametric Regressors},
	booktitle = {NIPS},
	year      = {2012},
	pages     = {2870-2878},
	ee        = {http://books.nips.cc/papers/files/nips25/NIPS2012_1297.pdf},
	crossref  = {DBLP:conf/nips/2012},
	bibsource = {DBLP, http://dblp.uni-trier.de}
}

@article{GWJMLR,
	author    = {Samory Kpotufe and	Abdeslam Boularias and Thomas Schultz and Kyoungok Kim},
	title     = {Gradient Weights improve Regression and Classification},
	journal   = {Journal of Machine Learning Research},
	volume    = {17},
	year      = {2016},
	pages     = {1-34}
}

@article{kakadenotes,
	author    = {Sham Kakade},
	title     = {Lecture Notes on Multivariate Analysis, Dimensionality Reduction, and Spectral Methods},
	journal   = {STAT 991, Spring},
	year      = {2010}
}

@article{randommatrix,
	author    = {Joel. A. Tropp},
	title     = {User-Friendly Tools for Random Matrices: An Introduction},
	journal   = {Tutorial at NIPS},
	year      = {2012}
}

@book {bass,
	AUTHOR = {Richard F. Bass},
	TITLE = {Real Analysis for Graduate Students},
	EDITION = {Second},
}

@book {Rudin-raca,
	AUTHOR = {Walter Rudin},
	TITLE = {Real and complex analysis},
	EDITION = {Third},
	PUBLISHER = {McGraw-Hill Book Co.},
	ADDRESS = {New York},
	YEAR = {1987},
	PAGES = {xiv+416},
	ISBN = {0-07-054234-1},
	MRCLASS = {00A05 (26-01 30-01 46-01)},
	MRNUMBER = {MR924157 (88k:00002)},
}

@article{SDCA,
	author    = {Shai Shalev-Shwartz and Tong Zhang},
	title     = {Stochastic Dual Coordinate Ascent Methods for Regularized Loss Minimization},
	journal   = {Journal of Machine Learning Research},
	volume    = {14},
	year      = {2013},
	pages     = {567-599},
	bibsource = {DBLP, http://dblp.uni-trier.de}
}

@inproceedings{DBLP:conf/nips/RouxSB12,
	author    = {Nicolas Le Roux and Mark W. Schmidt and Francis Bach},
	title     = {A Stochastic Gradient Method with an Exponential Convergence
	Rate for Finite Training Sets},
	booktitle = {NIPS},
	year      = {2012},
	pages     = {2672-2680},
	ee        = {http://books.nips.cc/papers/files/nips25/NIPS2012_1246.pdf},
	crossref  = {DBLP:conf/nips/2012},
	bibsource = {DBLP, http://dblp.uni-trier.de}
}

@article{DBLP:journals/siamjo/NemirovskiJLS09,
	author    = {Arkadi Nemirovski and Anatoli Juditsky and	Guanghui Lan and Alexander Shapiro},
	title     = {Robust Stochastic Approximation Approach to Stochastic Programming},
	journal   = {SIAM Journal on Optimization},
	volume    = {19},
	number    = {4},
	year      = {2009},
	pages     = {1574-1609},
	ee        = {http://dx.doi.org/10.1137/070704277},
	bibsource = {DBLP, http://dblp.uni-trier.de}
}

@article{DBLP:journals/siamjo/Nemirovski04,
	author    = {Arkadi Nemirovski},
	title     = {Prox-Method with Rate of Convergence O(1/t) for Variational
	Inequalities with Lipschitz Continuous Monotone Operators
	and Smooth Convex-Concave Saddle Point Problems},
	journal   = {SIAM Journal on Optimization},
	volume    = {15},
	number    = {1},
	year      = {2004},
	pages     = {229-251},
	ee        = {http://dx.doi.org/10.1137/S1052623403425629},
	bibsource = {DBLP, http://dblp.uni-trier.de}
}

@article{DBLP:journals/orl/BeckT03,
	author    = {Amir Beck and	Marc Teboulle},
	title     = {Mirror descent and nonlinear projected subgradient methods
	for convex optimization},
	journal   = {Oper. Res. Lett.},
	volume    = {31},
	number    = {3},
	year      = {2003},
	pages     = {167-175},
	ee        = {http://dx.doi.org/10.1016/S0167-6377(02)00231-6},
	bibsource = {DBLP, http://dblp.uni-trier.de}
}

@inproceedings{Jaggi13,
	author    = {Martin Jaggi},
	title     = {Revisiting Frank-Wolfe: Projection-Free Sparse Convex Optimization},
	booktitle = {ICML},
	year      = {2013}
}

@article{DBLP:journals/corr/abs-1212-2002,
	author    = {Simon Lacoste-Julien and	Mark W. Schmidt and	Francis Bach},
	title     = {A simpler approach to obtaining an O(1/t) convergence rate
	for the projected stochastic subgradient method},
	journal   = {CoRR},
	volume    = {abs/1212.2002},
	year      = {2012},
	ee        = {http://arxiv.org/abs/1212.2002},
	bibsource = {DBLP, http://dblp.uni-trier.de}
}

@article{DBLP:journals/jmlr/LanckrietCBGJ03,
	author    = {Gert R. G. Lanckriet and Nello Cristianini and	Peter L. Bartlett and Laurent El Ghaoui and
	Michael I. Jordan},
	title     = {Learning the Kernel Matrix with Semidefinite Programming},
	journal   = {Journal of Machine Learning Research},
	volume    = {5},
	year      = {2004},
	pages     = {27-72},
	ee        = {http://www.jmlr.org/papers/v5/lanckriet04a.html},
	bibsource = {DBLP, http://dblp.uni-trier.de}
}

@book{citeulike:4123765,
	author = {Nesterov, Yurii},
	edition = {1},
	howpublished = {Gebundene Ausgabe},
	isbn = {1402075537},
	publisher = {Springer Netherlands},
	title = {Introductory Lectures on Convex Optimization: A Basic Course (Applied Optimization)},
	url = {http://www.amazon.com/exec/obidos/redirect?tag=citeulike07-20\&path=ASIN/1402075537}
}

@book{Ben-Tal:2001:LMC:502969,
	author = {Ben-Tal, Aharon and Nemirovskiaei, Arkadiaei Semenovich},
	title = {Lectures on modern convex optimization: analysis, algorithms, and engineering applications},
	year = {2001},
	isbn = {0-89871-491-5},
	publisher = {Society for Industrial and Applied Mathematics},
	address = {Philadelphia, PA, USA},
}

@article{bubeck,
	author    = {Sebastien Bubeck},
	title     = {The complexities of optimization: Lecture Notes},
	journal   = {Lecture Notes of Princeton ORF 523},
	year      = {2013},
}

@inproceedings{DBLP:conf/colt/ScholkopfHS01,
	author    = {Bernhard Sch{\"o}lkopf and
	Ralf Herbrich and
	Alex J. Smola},
	title     = {A Generalized Representer Theorem},
	booktitle = {COLT/EuroCOLT},
	year      = {2001},
	pages     = {416-426},
	ee        = {http://dx.doi.org/10.1007/3-540-44581-1_27},
	crossref  = {DBLP:conf/colt/2001},
	bibsource = {DBLP, http://dblp.uni-trier.de}
}

@article{DBLP:journals/jmlr/BartlettM02,
	author    = {Peter L. Bartlett and
	Shahar Mendelson},
	title     = {Rademacher and Gaussian Complexities: Risk Bounds and Structural
	Results},
	journal   = {Journal of Machine Learning Research},
	volume    = {3},
	year      = {2002},
	pages     = {463-482},
	ee        = {http://www.jmlr.org/papers/v3/bartlett02a.html},
	bibsource = {DBLP, http://dblp.uni-trier.de}
}

@article{Belkin2003,
	author    = {Mikhail Belkin and	Partha Niyogi},
	title     = {Laplacian eigenmaps for dimensionality reduction and data representation},
	journal   = {Neural computation},
	volume    = {15},
	year      = {2003},
	pages     = {1373--1396}
}

@article{Cortes:1995:SN:218919.218929,
	author = {Cortes, Corinna and Vapnik, Vladimir},
	title = {Support-Vector Networks},
	journal = {Mach. Learn.},
	issue_date = {Sept. 1995},
	volume = {20},
	number = {3},
	year = {1995},
	issn = {0885-6125},
	pages = {273--297},
	numpages = {25},
	url = {http://dx.doi.org/10.1023/A:1022627411411},
	doi = {10.1023/A:1022627411411},
	acmid = {218929},
	publisher = {Kluwer Academic Publishers},
	address = {Hingham, MA, USA}
}

@book{Scholkopf:2001:LKS:559923,
	author = {Scholkopf, Bernhard and Smola, Alexander J.},
	title = {Learning with Kernels: Support Vector Machines, Regularization, Optimization, and Beyond},
	year = {2001},
	isbn = {0262194759},
	publisher = {MIT Press},
	address = {Cambridge, MA, USA},
}

@article{concentration,
	author    = {Gabor Lugosi},
	title     = {Concentration-of-measure Inequalities},
	journal   = {Lecture Notes of Machine Learning Summer School 2003, Australian National University, Canberra},
	year      = {2003},
}

@article{DBLP:journals/cacm/Valiant84,
	author    = {Leslie G. Valiant},
	title     = {A Theory of the Learnable},
	journal   = {Commun. ACM},
	volume    = {27},
	number    = {11},
	year      = {1984},
	pages     = {1134-1142},
	ee        = {http://doi.acm.org/10.1145/1968.1972},
	bibsource = {DBLP, http://dblp.uni-trier.de}
}

@INPROCEEDINGS{Bousquet04introductionto,
	author = {Olivier Bousquet and Stephane Boucheron and Gabor Lugosi},
	title = {Introduction to Statistical Learning Theory},
	booktitle = {Advanced Lectures in Machine Learning},
	year = {2004},
	pages = {169--207},
	publisher = {Springer}
}

@article{citeulike:4221073,
	author = {Boucheron, Stephane and Bousquet, Olivier and Lugosi, Gabor},
	journal = {ESAIM: Probability and Statistics},
	pages = {323--375},
	title = {Theory of classification : A survey of some recent advances},
	url = {http://cat.inist.fr/?aModele=afficheN\&\#38;cpsidt=17367966},
	volume = {9},
	year = {2005}
}

@book{foml,
	author    = {Mehryar Mohri and
	Afshin Rostamizadeh and
	Ameet Talwalkar},
	title     = {Foundations of Machine Learning},
	publisher = {MIT Press},
	year      = {2012},
	isbn      = {978-0-262-01825-8},
}

@article{sasanotes,
	author    = {Alexander Rakhlin and
	Karthik Sridharan},
	title     = {Statistical Learning Theory and Sequential Prediction: Lecture Notes},
	journal   = {Lecture Notes of Upenn STAT 928},
	year      = {2012},
}

@article{percynotes,
	author    = {Percy Liang},
	title     = {Statistical Learning Theory: Lecture Notes},
	journal   = {Lecture Notes of Stanford STAT 231},
	year      = {2013},
}

@ARTICLE{Baraniuk07asimple,
	author = {Richard Baraniuk and Mark Davenport and Ronald Devore and Michael Wakin},
	title = {A simple proof of the restricted isometry property for random matrices},
	journal = {Constr. Approx},
	year = {2007}
}

@article{DBLP:journals/jmlr/RaskuttiWY10,
	author    = {Garvesh Raskutti and
	Martin J. Wainwright and
	Bin Yu},
	title     = {Restricted Eigenvalue Properties for Correlated Gaussian
	Designs},
	journal   = {Journal of Machine Learning Research},
	volume    = {11},
	year      = {2010},
	pages     = {2241-2259},
	ee        = {http://portal.acm.org/citation.cfm?id=1859929},
	bibsource = {DBLP, http://dblp.uni-trier.de}
}

@INPROCEEDINGS{Donoho_eladm,
	author = {David L. Donoho and Michael Elad},
	title = {Optimally sparse representation in general (non-orthogonal) dictionaries via ? 1 minimization},
	booktitle = {Proc. Natl Acad. Sci. USA 100 2197¨C202},
	year = {2003}
}

@article{DBLP:journals/tit/CaiWX10,
	author    = {T. Tony Cai and
	Lie Wang and
	Guangwu Xu},
	title     = {Stable recovery of sparse signals and an oracle inequality},
	journal   = {IEEE Transactions on Information Theory},
	volume    = {56},
	number    = {7},
	year      = {2010},
	pages     = {3516-3522},
	ee        = {http://dx.doi.org/10.1109/TIT.2010.2048506},
	bibsource = {DBLP, http://dblp.uni-trier.de}
}

@incollection{introcs,
	author      = {Mark A. Davenport and Marco F. Duarte and Yonina C. Eldar and  Gitta Kutyniok},
	title       = {Introduction to Compressed Sensing},
	editor      = {Yonina C. Eldar and  Gitta Kutyniok},
	booktitle   = {Compressed Sensing: Theory and Applications},
	publisher   = {Cambridge University Press},
	year        = {2012},
}

@article{citeulike:4109966,
	author = {Candes, E.},
	doi = {10.1016/j.crma.2008.03.014},
	issn = {1631073X},
	journal = {Comptes Rendus Mathematique},
	month = may,
	number = {9-10},
	pages = {589--592},
	title = {The restricted isometry property and its implications for compressed sensing},
	url = {http://dx.doi.org/10.1016/j.crma.2008.03.014},
	volume = {346},
	year = {2008}
}

@article{DBLP:journals/tit/CaiXZ09,
	author    = {T. Tony Cai and
	Guangwu Xu and
	Jun Zhang},
	title     = {On recovery of sparse signals via l1 minimization},
	journal   = {IEEE Transactions on Information Theory},
	volume    = {55},
	number    = {7},
	year      = {2009},
	pages     = {3388-3397},
	ee        = {http://dx.doi.org/10.1109/TIT.2009.2021377},
	bibsource = {DBLP, http://dblp.uni-trier.de}
}

@article{DBLP:journals/tit/CaiWX10a,
	author    = {T. Tony Cai and
	Lie Wang and
	Guangwu Xu},
	title     = {New bounds for restricted isometry constants},
	journal   = {IEEE Transactions on Information Theory},
	volume    = {56},
	number    = {9},
	year      = {2010},
	pages     = {4388-4394},
	ee        = {http://dx.doi.org/10.1109/TIT.2010.2054730},
	bibsource = {DBLP, http://dblp.uni-trier.de}
}

@ARTICLE{Meinshausen09lasso-typerecovery,
	author = {Meinshausen and Bin Yu},
	title = {Lasso-type recovery of sparse representations from highdimensional data},
	journal = {Annals of Statistics},
	year = {2009},
	pages = {246--270}
}

@article{DBLP:journals/tit/CandesRT06,
	author    = {Emmanuel J. Candes and
	Justin K. Romberg and
	Terence Tao},
	title     = {Robust uncertainty principles: exact signal reconstruction
	from highly incomplete frequency information},
	journal   = {IEEE Transactions on Information Theory},
	volume    = {52},
	number    = {2},
	year      = {2006},
	pages     = {489-509},
	ee        = {http://dx.doi.org/10.1109/TIT.2005.862083},
	bibsource = {DBLP, http://dblp.uni-trier.de}
}

@article{DBLP:journals/tit/DonohoH01,
	author    = {David L. Donoho and
	Xiaoming Huo},
	title     = {Uncertainty principles and ideal atomic decomposition},
	journal   = {IEEE Transactions on Information Theory},
	volume    = {47},
	number    = {7},
	year      = {2001},
	pages     = {2845-2862},
	ee        = {http://dx.doi.org/10.1109/18.959265},
	bibsource = {DBLP, http://dblp.uni-trier.de}
}

@ARTICLE{greenshtein04,
	author = {Eitan Greenshtein and Ya'acov Ritov},
	title = {Persistence in high-dimensional linear predictor selection and the virtue of overparametrization},
	journal = {Bernoulli},
	year = {2004},
	volume = {10},
	number = {6},
	pages = {971--988}
}

@ARTICLE{Bickel09simultaneousanalysis,
	author = {Peter J. Bickel and Ya'acov Ritov and Alexandre B. Tsybakov},
	title = {Simultaneous analysis of Lasso and Dantzig selector},
	journal = {The Annals of Statistics},
	year = {2009},
	volume = {37},
	number = {4}
}

@article{DBLP:journals/tit/Wainwright09,
	author    = {Martin J. Wainwright},
	title     = {Sharp thresholds for high-dimensional and noisy sparsity
	recovery using l1-constrained quadratic programming (Lasso)},
	journal   = {IEEE Transactions on Information Theory},
	volume    = {55},
	number    = {5},
	year      = {2009},
	pages     = {2183-2202},
	ee        = {http://dx.doi.org/10.1109/TIT.2009.2016018},
	bibsource = {DBLP, http://dblp.uni-trier.de}
}

@article{DBLP:journals/tit/CandesT06,
	author    = {Emmanuel J. Candes and
	Terence Tao},
	title     = {Near-Optimal Signal Recovery From Random Projections: Universal
	Encoding Strategies?},
	journal   = {IEEE Transactions on Information Theory},
	volume    = {52},
	number    = {12},
	year      = {2006},
	pages     = {5406-5425},
	ee        = {http://dx.doi.org/10.1109/TIT.2006.885507},
	bibsource = {DBLP, http://dblp.uni-trier.de}
}

@article{DBLP:journals/tit/CandesT05,
	author    = {Emmanuel J. Candes and
	Terence Tao},
	title     = {Decoding by linear programming},
	journal   = {IEEE Transactions on Information Theory},
	volume    = {51},
	number    = {12},
	year      = {2005},
	pages     = {4203-4215},
	ee        = {http://dx.doi.org/10.1109/TIT.2005.858979},
	bibsource = {DBLP, http://dblp.uni-trier.de}
}

@article{DBLP:journals/jmlr/ZhaoY06,
	author    = {Peng Zhao and
	Bin Yu},
	title     = {On Model Selection Consistency of Lasso},
	journal   = {Journal of Machine Learning Research},
	volume    = {7},
	year      = {2006},
	pages     = {2541-2563},
	ee        = {http://www.jmlr.org/papers/v7/zhao06a.html},
	bibsource = {DBLP, http://dblp.uni-trier.de}
}

@ARTICLE{Geer_onthe,
	author = {Sara A. Van De Geer and Peter Buhlmann},
	title = {On the conditions used to prove oracle results for the Lasso},
	journal = {Electronic Journal of Statistics},
	year = {2009},
	pages = {1360-1392}
}

@book{HDbook,
	author    = {Peter B{\"u}hlmann and
	Sara A. van de Geer},
	title     = {Statistics for High-Dimensional Data: Methods, Theory and Applications},
	publisher = {Springer},
	year      = {2011},
	isbn      = {978-3-642-20192-9},
	bibsource = {DBLP, http://dblp.uni-trier.de}
}

@book{Wasserman:2006:NS:1202956,
	author = {Wasserman, Larry},
	title = {All of Nonparametric Statistics (Springer Texts in Statistics)},
	year = {2006},
	isbn = {0387251456},
	publisher = {Springer-Verlag New York, Inc.},
	address = {Secaucus, NJ, USA},
}

@inproceedings{DBLP:conf/www/LiCLS10,
	author    = {Lihong Li and
	Wei Chu and
	John Langford and
	Robert E. Schapire},
	title     = {A contextual-bandit approach to personalized news article
	recommendation},
	booktitle = {WWW},
	year      = {2010},
	pages     = {661-670},
	ee        = {http://doi.acm.org/10.1145/1772690.1772758},
	crossref  = {DBLP:conf/www/2010},
	bibsource = {DBLP, http://dblp.uni-trier.de}
}

@article{DBLP:journals/siamcomp/AuerCFS02,
	author    = {Peter Auer and
	Nicol{\`o} Cesa-Bianchi and
	Yoav Freund and
	Robert E. Schapire},
	title     = {The Nonstochastic Multiarmed Bandit Problem},
	journal   = {SIAM J. Comput.},
	volume    = {32},
	number    = {1},
	year      = {2002},
	pages     = {48-77},
	ee        = {http://dx.doi.org/10.1137/S0097539701398375},
	bibsource = {DBLP, http://dblp.uni-trier.de}
}

@article{DBLP:journals/ml/AuerCF02,
	author    = {Peter Auer and
	Nicol{\`o} Cesa-Bianchi and
	Paul Fischer},
	title     = {Finite-time Analysis of the Multiarmed Bandit Problem},
	journal   = {Machine Learning},
	volume    = {47},
	number    = {2-3},
	year      = {2002},
	pages     = {235-256},
	ee        = {http://dx.doi.org/10.1023/A:1013689704352},
	bibsource = {DBLP, http://dblp.uni-trier.de}
}

@article{DBLP:journals/ftml/BubeckC12,
	author    = {S{\'e}bastien Bubeck and
	Nicol{\`o} Cesa-Bianchi},
	title     = {Regret Analysis of Stochastic and Nonstochastic Multi-armed
	Bandit Problems},
	journal   = {Foundations and Trends in Machine Learning},
	volume    = {5},
	number    = {1},
	year      = {2012},
	pages     = {1-122},
	ee        = {http://dx.doi.org/10.1561/2200000024},
	bibsource = {DBLP, http://dblp.uni-trier.de}
}

@book{DBLP:books/daglib/0016248,
	author    = {Nicol{\`o} Cesa-Bianchi and
	G{\'a}bor Lugosi},
	title     = {Prediction, learning, and games},
	publisher = {Cambridge University Press},
	year      = {2006},
	isbn      = {978-0-521-84108-5},
	pages     = {I-XII, 1-394},
	bibsource = {DBLP, http://dblp.uni-trier.de}
}

@book{rlbook,
	author    = {Richard S. Sutton and
	Andrew G. Barto},
	title     = {Reinforcement Learning: An Introduction},
	publisher = {MIT Press},
	year      = {1998},
	isbn      = {978-0262193986},
	bibsource = {DBLP, http://dblp.uni-trier.de}
}

@ARTICLE{Efron04leastangle,
	author = {Bradley Efron and Trevor Hastie and Iain Johnstone and Robert Tibshirani},
	title = {Least angle regression},
	journal = {Annals of Statistics},
	year = {2004},
	volume = {32},
	pages = {407--499}
}

@article{DBLP:journals/mp/TsengY09,
	author    = {Paul Tseng and
	Sangwoon Yun},
	title     = {A coordinate gradient descent method for nonsmooth separable
	minimization},
	journal   = {Math. Program.},
	volume    = {117},
	number    = {1-2},
	year      = {2009},
	pages     = {387-423},
	ee        = {http://dx.doi.org/10.1007/s10107-007-0170-0},
	bibsource = {DBLP, http://dblp.uni-trier.de}
}

@article{CDlasso,
	author    = {Wenjiang Fu},
	title     = {Penalized Regressions: The Bridge versus the Lasso},
	journal   = {Journal of Computational and Graphical Statistics},
	volume    = {7},
	number    = {1},
	year      = {1998},
	pages     = {397-416},
	ee        = {http://dx.doi.org/10.1137/080716542},
	bibsource = {DBLP, http://dblp.uni-trier.de}
}

@article{DBLP:journals/siamis/BeckT09,
	author    = {Amir Beck and
	Marc Teboulle},
	title     = {A Fast Iterative Shrinkage-Thresholding Algorithm for Linear
	Inverse Problems},
	journal   = {SIAM J. Imaging Sciences},
	volume    = {2},
	number    = {1},
	year      = {2009},
	pages     = {183-202},
	ee        = {http://dx.doi.org/10.1137/080716542},
	bibsource = {DBLP, http://dblp.uni-trier.de}
}

@article{DBLP:journals/mp/Nesterov05,
	author    = {Yurii Nesterov},
	title     = {Smooth minimization of non-smooth functions},
	journal   = {Math. Program.},
	volume    = {103},
	number    = {1},
	year      = {2005},
	pages     = {127-152},
	ee        = {http://dx.doi.org/10.1007/s10107-004-0552-5},
	bibsource = {DBLP, http://dblp.uni-trier.de}
}

@ARTICLE{1988nesterov,
	author = {Yurii Nesterov},
	title = {On an approach to the construction of
	optimal methods of minimization of smooth convex functions},
	year = {1988}
}

@article{1983nesterov,
	author = {Yurii Nesterov},
	title = {A method for solving a convex programming problem with convergence rate {$O(\frac{1}{k^2})$}},
	year = {1983}
}

@article{Zou05regularizationand,
	author = {Hui Zou and Trevor Hastie},
	title = {Regularization and variable selection via the Elastic Net},
	journal = {Journal of the Royal Statistical Society, Series B},
	year = {2005},
	volume = {67},
	pages = {301--320}
}

@article{DBLP:journals/jmlr/DuchiS09,
	author    = {John C. Duchi and
	Yoram Singer},
	title     = {Efficient Online and Batch Learning Using Forward Backward
	Splitting},
	journal   = {Journal of Machine Learning Research},
	volume    = {10},
	year      = {2009},
	pages     = {2899-2934},
	ee        = {http://doi.acm.org/10.1145/1577069.1755882},
	bibsource = {DBLP, http://dblp.uni-trier.de}
}

@article{DBLP:journals/jmlr/RifkinL07,
	author    = {Ryan M. Rifkin and
	Ross A. Lippert},
	title     = {Value Regularization and Fenchel Duality},
	journal   = {Journal of Machine Learning Research},
	volume    = {8},
	year      = {2007},
	pages     = {441-479},
	ee        = {http://dl.acm.org/citation.cfm?id=1314515},
	bibsource = {DBLP, http://dblp.uni-trier.de}
}

@article{DBLP:journals/ftml/Shalev-Shwartz12,
	author    = {Shai Shalev-Shwartz},
	title     = {Online Learning and Online Convex Optimization},
	journal   = {Foundations and Trends in Machine Learning},
	volume    = {4},
	number    = {2},
	year      = {2012},
	pages     = {107-194},
	ee        = {http://dx.doi.org/10.1561/2200000018},
	bibsource = {DBLP, http://dblp.uni-trier.de}
}

@inproceedings{DBLP:conf/nips/BottouB07,
	author    = {L{\'e}on Bottou and
	Olivier Bousquet},
	title     = {The Tradeoffs of Large Scale Learning},
	booktitle = {NIPS},
	year      = {2007},
	ee        = {http://books.nips.cc/papers/files/nips20/NIPS2007_0726.pdf},
	crossref  = {DBLP:conf/nips/2007},
	bibsource = {DBLP, http://dblp.uni-trier.de}
}

@book{convex,
	author    = {Stephen Boyd and
	Lieven Vandenberghe},
	title     = {Convex Optimization},
	publisher = {Cambridge University Press},
	year      = {2004},
	isbn      = {0521833787},
	bibsource = {DBLP, http://dblp.uni-trier.de}
}

@ARTICLE{grouplasso,
	author = {Ming Yuan and
	Yi Lin},
	title = {Model selection and estimation in regression with
	grouped variables},
	journal = {Journal of the Royal Statistical Society, Series B},
	year = {2006},
	volume = {68},
	pages = {49--67}
}

@article{DBLP:journals/ftml/BachJMO12,
	author    = {Francis R. Bach and
	Rodolphe Jenatton and
	Julien Mairal and
	Guillaume Obozinski},
	title     = {Optimization with Sparsity-Inducing Penalties},
	journal   = {Foundations and Trends in Machine Learning},
	volume    = {4},
	number    = {1},
	year      = {2012},
	pages     = {1-106},
	ee        = {http://dx.doi.org/10.1561/2200000015},
	bibsource = {DBLP, http://dblp.uni-trier.de}
}

@inproceedings{DBLP:conf/nips/ZhuRHT03,
	author    = {Ji Zhu and
	Saharon Rosset and
	Trevor Hastie and
	Robert Tibshirani},
	title     = {1-norm Support Vector Machines},
	booktitle = {NIPS},
	year      = {2003},
	ee        = {http://books.nips.cc/papers/files/nips16/NIPS2003_AA07.pdf},
	crossref  = {DBLP:conf/nips/2003},
	bibsource = {DBLP, http://dblp.uni-trier.de}
}

@book{DBLP:books/daglib/0097035,
	author    = {Vladimir Vapnik},
	title     = {Statistical learning theory},
	publisher = {Wiley},
	year      = {1998},
	isbn      = {978-0-471-03003-4},
	pages     = {I-XXIV, 1-736},
	bibsource = {DBLP, http://dblp.uni-trier.de}
}

@book{DBLP:books/wa/BreimanFOS84,
	author    = {Leo Breiman and
	J. H. Friedman and
	R. A. Olshen and
	C. J. Stone},
	title     = {Classification and Regression Trees},
	publisher = {Wadsworth},
	year      = {1984},
	isbn      = {0-534-98053-8},
	bibsource = {DBLP, http://dblp.uni-trier.de}
}

@book{boosting,
	author    = {Robert E. Schapire and
	Yoav Freund},
	title     = {Boosting: Foundations and Algorithms},
	publisher = {MIT Press},
	year      = {2012},
	isbn      = {0-534-98053-8},
	bibsource = {DBLP, http://dblp.uni-trier.de}
}

@book{ESL09,
	author    = {Trevor Hastie and
	Robert Tibshirani and
	Jerome Friedman},
	title     = {The Elements of Statistical Learning: Data Mining, Inference and Prediction, Second Edition},
	publisher = {Springer},
	year      = {2009},
	isbn      = {0-534-98053-8},
	bibsource = {DBLP, http://dblp.uni-trier.de}
}

@book{MLAPP,
	author    = {Kevin Murphy},
	title     = {Machine Learning: A Probabilistic Perspective},
	publisher = {MIT Press},
	year      = {2012},
	isbn      = {0-534-98053-8},
	bibsource = {DBLP, http://dblp.uni-trier.de}
}

@article{DBLP:journals/ml/Quinlan86,
	author    = {J. Ross Quinlan},
	title     = {Induction of Decision Trees},
	journal   = {Machine Learning},
	volume    = {1},
	number    = {1},
	year      = {1986},
	pages     = {81-106},
	ee        = {http://dx.doi.org/10.1023/A:1022643204877},
	bibsource = {DBLP, http://dblp.uni-trier.de}
}

@book{DBLP:books/mk/Quinlan93,
	author    = {J. Ross Quinlan},
	title     = {C4.5: Programs for Machine Learning},
	publisher = {Morgan Kaufmann},
	year      = {1993},
	isbn      = {1-55860-238-0},
	bibsource = {DBLP, http://dblp.uni-trier.de}
}

@article{DBLP:journals/tsp/MallatZ93,
	author    = {St{\'e}phane Mallat and
	Zhifeng Zhang},
	title     = {Matching pursuits with time-frequency dictionaries},
	journal   = {IEEE Transactions on Signal Processing},
	volume    = {41},
	number    = {12},
	year      = {1993},
	pages     = {3397-3415},
	ee        = {http://dx.doi.org/10.1109/78.258082},
	bibsource = {DBLP, http://dblp.uni-trier.de}
}

@article{DBLP:journals/tit/TroppG07,
	author    = {Joel A. Tropp and
	Anna C. Gilbert},
	title     = {Signal Recovery From Random Measurements Via Orthogonal
	Matching Pursuit},
	journal   = {IEEE Transactions on Information Theory},
	volume    = {53},
	number    = {12},
	year      = {2007},
	pages     = {4655-4666},
	ee        = {http://dx.doi.org/10.1109/TIT.2007.909108},
	bibsource = {DBLP, http://dblp.uni-trier.de}
}

@article{dantzig07,
	author    = {Emmanuel Candes and
	Terence Tao},
	title     = {The Dantzig selector: Statistical estimation when p is much larger than n},
	journal   = {Annals of Statistics},
	volume    = {35},
	number    = {6},
	year      = {2007},
	pages     = {2313-2351},
	ee        = {http://dx.doi.org/10.1109/TIT.2007.909108},
	bibsource = {DBLP, http://dblp.uni-trier.de}
}

@ARTICLE{Tibshirani94regressionshrinkage,
	author = {Robert Tibshirani},
	title = {Regression Shrinkage and Selection Via the Lasso},
	journal = {Journal of the Royal Statistical Society, Series B},
	year = {1996},
	volume = {58},
	pages = {267--288}
}

@article{Mukherjee_wu2010learning,
	title={Learning gradients: predictive models that infer geometry and statistical dependence},
	author={Wu, Qiang and Guinney, Justin and Maggioni, Mauro and Mukherjee, Sayan},
	journal={The Journal of Machine Learning Research},
	volume={11},
	pages={2175--2198},
	year={2010},
	publisher={JMLR. org}
}

@article{mukherjee2010learning,
	title={Learning gradients on manifolds},
	author={Mukherjee, Sayan and Wu, Qiang and Zhou, Ding-Xuan and others},
	journal={Bernoulli},
	volume={16},
	number={1},
	pages={181--207},
	year={2010},
	publisher={Bernoulli Society for Mathematical Statistics and Probability}
}

@article{powell1989semiparametric,
	title={Semiparametric estimation of index coefficients},
	author={Powell, James L and Stock, James H and Stoker, Thomas M},
	journal={Econometrica: Journal of the Econometric Society},
	pages={1403--1430},
	year={1989},
	publisher={JSTOR}
}

@article{li1991sliced,
	title={Sliced inverse regression for dimension reduction},
	author={Li, Ker-Chau},
	journal={Journal of the American Statistical Association},
	volume={86},
	number={414},
	pages={316--327},
	year={1991},
	publisher={Taylor \& Francis}
}

@article{hardle1993optimal,
	title={Optimal smoothing in single-index models},
	author={Hardle, Wolfgang and Hall, Peter and Ichimura, Hidehiko and others},
	journal={The annals of Statistics},
	volume={21},
	number={1},
	pages={157--178},
	year={1993},
	publisher={Institute of Mathematical Statistics}
}

@article{xia2002adaptive,
	title={An adaptive estimation of dimension reduction space},
	author={Xia, Yingcun and Tong, Howell and Li, WK and Zhu, Li-Xing},
	journal={Journal of the Royal Statistical Society: Series B (Statistical Methodology)},
	volume={64},
	number={3},
	pages={363--410},
	year={2002},
	publisher={Wiley Online Library}
}

@inproceedings{relieff1992,
	title = {The feature selection problem: Traditional methods and a new algorithm},
	author = {Kira, Kenji and Rendell, Larry M.},
	booktitle = {Proceedings of AAAI},
	pages = {129--134},
	year={1992},
	organization = {AAAI}
}

@article{cleveland1988locally,
	title={Locally weighted regression: an approach to regression analysis by local fitting},
	author={Cleveland, William S and Devlin, Susan J},
	journal={Journal of the American Statistical Association},
	volume={83},
	number={403},
	pages={596--610},
	year={1988},
	publisher={Taylor \& Francis Group}
}

@inproceedings{conf/aaai/1992,
	author = {Kira, Kenji and Rendell, Larry A.},
	booktitle = {AAAI},
	crossref = {conf/aaai/1992},
	editor = {Swartout, William R.},
	ee = {http://www.aaai.org/Library/AAAI/1992/aaai92-020.php},
	isbn = {0-262-51063-4},
	pages = {129-134},
	publisher = {AAAI Press / The MIT Press},
	title = {The Feature Selection Problem: Traditional Methods and a New Algorithm.},
	url = {http://dblp.uni-trier.de/db/conf/aaai/aaai92.html#KiraR92},
	year = 1992
}

@ARTICLE{Kononenko97overcomingthe,
	author = {Igor Kononenko and Edvard Simec and Marko Robnik- Sikonja},
	title = {Overcoming the myopia of inductive learning algorithms with RELIEFF},
	journal = {Applied Intelligence},
	year = {1997},
	volume = {7},
	pages = {39--55}
}

@inproceedings{trivedi2014UAI,
	title={A Consistent Estimator of the Expected Gradient Outerproduct},
	author={Trivedi, Shubhendu and Wang, Jialei and Samory Kpotufe and Gregory Shakhnarovich},
	booktitle={Proceedings of the 30th International Conference on Uncertainty in Artificial Intelligence},
	pages={819--828},
	year={2014},
	organization={AUAI}
}

@book{Lugosi,
	author = {Devroye, Luc and Gy{\"o}rfi, L{\'a}szlo and Lugosi, G{\'a}bor},
	title = {A probabilistic theory of pattern recognition},
	year = {1997},
	isbn = {0387946187},
	publisher = {Springer},
	address = {Cambridge, MA, USA},
}

@article{LuxSpectral,
	title={Consistency of spectral clustering},
	author={Von Luxburg, Ulrike and Belkin, Mikhail and Bousquet, Olivier},
	journal={Annals of Statistics},
	pages={555--586},
	year={2008},
	publisher={Wiley Online Library}
}

@inproceedings{NiyogiMaps,
	title={Convergence of Laplacian Eigenmaps},
	author={Belkin, Mikhail and Niyogi, Partha},
	booktitle={Advances in Neural Information Processing Systems},
	pages={129--136},
	year={2007}
}

@article{HintonDBN,
	title={A fast learning algorithm for deep belief nets},
	author={Hinton, Geoffrey E. and Osindero, Simon and Teh, Yee-Whye},
	journal={Neural Computation},
	volume={18},
	number={7},
	pages={1527--1554},
	year={2006},
	publisher={MIT Press}
}

@article{HintonScience,
	title={Reducing the dimensionality of data with neural networks},
	author={Hinton, Geoffrey E. and Salakhutdinov, Ruslan R.},
	journal={Science},
	volume={313},
	number={5786},
	pages={504--507},
	year={2006}
}

@article{BengioGreedy,
	title={Greedy layer-wise training of deep networks},
	author={Bengio, Yoshua and Lamblin, Pascal and Popovici, Dan and Larochelle, Hugo},
	journal={Advances in neural information processing systems},
	pages={153--160},
	year={2007}
}

@article{LeCunAI,
	title={Scaling learning algorithms towards AI},
	author={Bengio, Yoshua and LeCun, Yann},
	journal={Large-scale kernel machines},
	pages={1--41},
	year={2007}
}

@article{SchmidhuberNN,
	title={Deep learning in neural networks: An overview.},
	author={Schmidhuber, J{\"u}rgen},
	journal={Neural networks},
	pages={85--117},
	year={2015}
}

@article{BengioFNT,
	title={Learning deep architectures for AI},
	author={Bengio, Yoshua},
	journal={Foundations and trends in Machine Learning},
	pages={1--127},
	year={2009}
}

@article{HintonRumelhart,
	title={Distributed representations},
	author={Hinton, Geoffrey E. and McClelland, James L. and Rumelhart, David E.},
	journal={Carnegie-Mellon University},
	pages={1--127},
	year={1984}
}

@article{MultiColumn,
	archivePrefix = {arXiv},
	arxivId = {arXiv:1202.2745},
	author = {Dan Ciresan and Ueli Meier and J{\"u}rgen Schmidhuber}, 
	eprint = {arXiv:1202.2745},
	journal={arXiv:1202.2745}, 
	year={2012},
	title = {Multi-column deep neural networks for image classification}
}

@article{Shakir,
	archivePrefix = {arXiv},
	arxivId = {arXiv:1610.03483},
	author = {Mohamed, Shakir and Lakshminarayanan, Balaji}, 
	eprint = {arXiv:1610.03483},
	journal={arXiv:1610.03483}, 
	year={2016},
	title = {Learning in implicit generative models}
}

@article{GoodfellowGAN,
	title={Generative adversarial nets},
	author={Goodfellow, Ian and Pouget-Abadie, Jean and Mirza, Mehdi and Xu, Bing and Warde-Farley, David and Ozair, Shrejil and Courville, Aaron and Bengio, Yoshua},
	journal={Advances in neural information processing systems},
	pages={2672--2680},
	year={2014}
}

@article{LencVedaldi,
	title={Understanding image representations by measuring their equivariance and equivalence},
	author={Lenc, Karel and Vedaldi, Andrea},
	journal={Proceedings of the IEEE conference on Computer Vision and Pattern Recognition},
	pages={991--999},
	year={2015}
}

@InProceedings{pmlr-v80-kondor18a,
	title = 	 {On the Generalization of Equivariance and Convolution in Neural Networks to the Action of Compact Groups},
	author = 	 {Kondor, Risi and Trivedi, Shubhendu},
	booktitle = 	 {Proceedings of the 35th International Conference on Machine Learning},
	pages = 	 {2747--2755},
	year = 	 {2018},
	editor = 	 {Dy, Jennifer and Krause, Andreas},
	volume = 	 {80},
	series = 	 {Proceedings of Machine Learning Research},
	address = 	 {Stockholmsmässan, Stockholm Sweden},
	month = 	 {10--15 Jul},
	publisher = 	 {PMLR},
	url = 	 {http://proceedings.mlr.press/v80/kondor18a.html}
}

@inproceedings{Dieleman:2016:ECS:3045390.3045590,
	author = {Dieleman, Sander and De Fauw, Jeffrey and Kavukcuoglu, Koray},
	title = {Exploiting Cyclic Symmetry in Convolutional Neural Networks},
	booktitle = {Proceedings of the 33rd International Conference on International Conference on Machine Learning - Volume 48},
	series = {ICML'16},
	year = {2016},
	location = {New York, NY, USA},
	pages = {1889--1898},
	numpages = {10},
	url = {http://dl.acm.org/citation.cfm?id=3045390.3045590},
	acmid = {3045590},
	publisher = {JMLR.org},
}

@InProceedings{pmlr-v48-niepert16,
	title = 	 {Learning Convolutional Neural Networks for Graphs},
	author = 	 {Mathias Niepert and Mohamed Ahmed and Konstantin Kutzkov},
	booktitle = 	 {Proceedings of The 33rd International Conference on Machine Learning},
	pages = 	 {2014--2023},
	year = 	 {2016},
	editor = 	 {Maria Florina Balcan and Kilian Q. Weinberger},
	volume = 	 {48},
	series = 	 {Proceedings of Machine Learning Research},
	address = 	 {New York, New York, USA},
	month = 	 {20--22 Jun},
	publisher = 	 {PMLR},
	url = 	 {http://proceedings.mlr.press/v48/niepert16.html}
}

@incollection{NIPS2016_6081,
	title = {Convolutional Neural Networks on Graphs with Fast Localized Spectral Filtering},
	author = {Defferrard, Micha{\"e}l and Bresson, Xavier and Vandergheynst, Pierre},
	booktitle = {Advances in Neural Information Processing Systems 29},
	editor = {D. D. Lee and M. Sugiyama and U. V. Luxburg and I. Guyon and R. Garnett},
	pages = {3844--3852},
	year = {2016},
	publisher = {Curran Associates, Inc.},
	url = {http://papers.nips.cc/paper/6081-convolutional-neural-networks-on-graphs-with-fast-localized-spectral-filtering.pdf}
}

@inproceedings{Duvenaud:2015:CNG:2969442.2969488,
	author = {Duvenaud, David and Maclaurin, Dougal and Aguilera-Iparraguirre, Jorge and G{\'o}mez-Bombarelli, Rafael and Hirzel, Timothy and Aspuru-Guzik, Al{\'a}n and Adams, Ryan P.},
	title = {Convolutional Networks on Graphs for Learning Molecular Fingerprints},
	booktitle = {Proceedings of the 28th International Conference on Neural Information Processing Systems - Volume 2},
	series = {NIPS'15},
	year = {2015},
	location = {Montreal, Canada},
	pages = {2224--2232},
	numpages = {9},
	url = {http://dl.acm.org/citation.cfm?id=2969442.2969488},
	acmid = {2969488},
	publisher = {MIT Press},
	address = {Cambridge, MA, USA},
}

@inproceedings{Masci:2015:GCN:2919341.2920992,
	author = {Masci, Jonathan and Boscaini, Davide and Bronstein, Michael M. and Vandergheynst, Pierre},
	title = {Geodesic Convolutional Neural Networks on Riemannian Manifolds},
	booktitle = {Proceedings of the 2015 IEEE International Conference on Computer Vision Workshop (ICCVW)},
	series = {ICCVW '15},
	year = {2015},
	isbn = {978-1-4673-9711-7},
	pages = {832--840},
	numpages = {9},
	url = {http://dx.doi.org/10.1109/ICCVW.2015.112},
	doi = {10.1109/ICCVW.2015.112},
	acmid = {2920992},
	publisher = {IEEE Computer Society},
	address = {Washington, DC, USA},
}

@inproceedings{DBLP:conf/cvpr/MontiBMRSB17,
	author    = {Federico Monti and
	Davide Boscaini and
	Jonathan Masci and
	Emanuele Rodola and
	Jan Svoboda and
	Michael M. Bronstein},
	title     = {Geometric Deep Learning on Graphs and Manifolds Using Mixture Model
	CNNs},
	booktitle = {2017 {IEEE} Conference on Computer Vision and Pattern Recognition,
	{CVPR} 2017, Honolulu, HI, USA, July 21-26, 2017},
	pages     = {5425--5434},
	year      = {2017},
	crossref  = {DBLP:conf/cvpr/2017},
	url       = {https://doi.org/10.1109/CVPR.2017.576},
	doi       = {10.1109/CVPR.2017.576},
	bibsource = {dblp computer science bibliography, https://dblp.org}
}

@article{DBLP:journals/spm/BronsteinBLSV17,
	author    = {Michael M. Bronstein and
	Joan Bruna and
	Yann LeCun and
	Arthur Szlam and
	Pierre Vandergheynst},
	title     = {Geometric Deep Learning: Going beyond Euclidean data},
	journal   = {IEEE Signal Process. Mag.},
	volume    = {34},
	number    = {4},
	pages     = {18--42},
	year      = {2017},
	url       = {https://doi.org/10.1109/MSP.2017.2693418},
	doi       = {10.1109/MSP.2017.2693418},
	bibsource = {dblp computer science bibliography, https://dblp.org}
}

@article{CohenSpherical18, 
	title={Spherical {CNN}s},
	author={T. S. Cohen and M. Geiger and  J. K{\"o}hler and M. Welling},
	journal={International Conference on Learning Representations},
	year={2018},
}

@article{IoffeSzegedy, 
	title={Batch Normalization: Accelerating deep network training by reducing internal covariate shift},
	author={S. Ioffe and C. Szefedy},
	journal={International Conference on Machine Learning},
	year={2015},
}

@article{adam2015, 
	title={ADAM: A method for stochastic optimization},
	author={D. P. Kingma and J. Ba},
	journal={International Conference on Learning Representations},
	year={2015},
}

@article{DriscollHealy, 
	title={Computing Fourier transforms and convolutions on the 2-sphere},
	author={J. R. Driscoll and D. M. Healy},
	journal={Advances in Applied Mathematics},
	year={1994},
}

@article{QM71,
	title={970 million druglike small molecules for virtual screening in the chemical universe database GDB-13},
	author={L. C. Blum and J.-L. Reymond},
	journal={Journal of the American Chemical Society},
	year={2009},
}

@article{QM72,
	title={Fast and accurate modeling of	molecular atomization energies with machine learning},
	author={M. Rupp and A. Tkatchenko and K.-R. M{\"u}ller and O. A. von Lilienfeld},
	journal={Physical Review Letters},
	year={2012},
}

@article{SHREC,
	title={Large-Scale 3D Shape Retrieval from ShapeNet Core55},
	author={M. Savva and F. Yu and H. Su and A. Kanezaki and T. Furuya and R. Ohbuchi and Z. Zhou and R. Yu and S. Bai and X. Bai and M. Aono and A. Tatsuma and S. Thermos and A. Axenopoulos and G. Th. Papadopoulos and P. Daras and X. Deng and Z. Lian and B. Li and H. Johan and Y. Lu and S. Mk.},
	journal={Eurographics Workshop on 3D Object Retrieval},
	year={2017},
}

@article{ShapeNet,
	archivePrefix = {arXiv},
	arxivId = {arXiv:1512.03012},
	author = {A. X. Chang and T. Funkhouser and L. Guibas and P. Hanrahan and Q. Huang and Z. Li and S. Savarese and M. Savva and S. Song and H. Su and J. Xiao and L. Yi and F. Yu}, 
	eprint = {arXiv:1512.03012},
	journal={arXiv:1512.03012}, 
	year={2015},
	title = {ShapeNet: An Information-Rich 3D Model Repository}
}

@article{Zaheer2017,
	archivePrefix = {arXiv},
	arxivId = {arXiv:1703.06114},
	author = {M. Zaheer and S. Kottur and S. Ravanbakhsh and B. Poczos and R. Salakhutdinov and and A. Smola}, 
	eprint = {arXiv:1703.06114},
	journal = {arXiv:1703.06114},
	year={2017},
	title = {Deep Sets}
}

@article{EquivarianceArxiv18,
	archivePrefix = {arXiv},
	arxivId = {arXiv:1802.03690},
	author = {R. Kondor and S. Trivedi}, 
	eprint = {arXiv:1802.03690},
	journal = {arXiv:1802.03690},
	year={2018},
	title = {On the Generalization of Equivariance and Convolution in Neural Networks to the Action of Compact Groups}
}

@article{Raj2016,
	archivePrefix = {arXiv},
	arxivId = {arXiv:1612.01988},
	author = {A. Raj and A. Kumar and Y. Mroueh and P.T. Fletcher et al.}, 
	eprint = {arXiv:1612.01988},
	journal = {arXiv:1612.01988},
	year={2016},
	title = {Local group invariant representations via orbit embeddings}
}

@inproceedings{Poczos2017,
	Author = {S. Ravanbakhsh and J. Schneider and B. Poczos},
	Booktitle = {Proceedings of International Conference on Machine Learning},
	Title = {Equivariance Through Parameter-Sharing},
	Year = {2017}
}

@inproceedings{Montavon2012,
	Author = {G. Montavon and K. Hansen and S. Fazli and M. Rupp and F. Biegler and A. Ziehe and A. Tkatchenko and O.A. von Lilienfeld and K. Muller},
	Booktitle = {NIPS},
	Title = {Learning invariant representations of molecules for atomization energy prediction},
	Year = {2012}
}

@article{Masci2015,
	archivePrefix = {arXiv},
	arxivId = {arXiv:1501.06297},
	author = {J. Masci and D. Boscaini and M. M. Bronstein and P. Vandergheynst},
	eprint = {arXiv:1501.06297},
	journal = {arXiv:1501.06297},
	year={2015},
	title = {Geodesic convolutional neural networks on Riemannian manifolds}
}

@article{Monti2016,
	archivePrefix = {arXiv},
	arxivId = {arXiv:1611.08402},
	author = {F. Monti and D. Boscaini and J. Masci and E. Rodola and J. Svoboda and M. M. Bronstein},
	eprint = {arXiv:1611.08402},
	journal = {arXiv:1611.08402},
	year={2016},
	title = {Geometric deep learning on graphs and manifolds using mixture model CNNs}
}

@inproceedings{Cohen2017,
	author={T. S. Cohen and M. Welling},
	booktitle={ICLR},
	title={Steerable CNNs},
	year={2017}
}

@article{Cohen2016,
	archivePrefix = {arXiv},
	arxivId = {1602.07576},
	author = {Cohen, T. S. and Welling, M.},
	eprint = {1602.07576},
	journal = {Proceedings of The 33rd International Conference on Machine Learning},
	pages = {2990--2999},
	title = {Group equivariant convolutional networks},
	volume = {48},
	year = {2016}
}

@article{He2016,
	archivePrefix = {arXiv},
	arxivId = {1512.03385},
	author = {He, K. and Zhang, X. and Ren, S. and Sun, J.},
	doi = {10.1109/CVPR.2016.90},
	eprint = {1512.03385},
	isbn = {978-1-4673-8851-1},
	issn = {1664-1078},
	journal = {2016 IEEE Conference on Computer Vision and Pattern Recognition (CVPR)},
	pages = {770--778},
	pmid = {23554596},
	title = {Deep Residual Learning for Image Recognition},
	url = {http://ieeexplore.ieee.org/document/7780459/},
	year = {2016}
}

@article{LeCun1989,
	Author = {Y LeCun and B Boser and J. S. Denker and D. Henderson and R. E. Howard and W. Hubbard and L. D. Jackel},
	Journal = {Neural Computation},
	Pages = {541-551},
	Title = {Backpropagation applied to handwritten zip code recognition},
	Volume = {1},
	Year = {1989}
}

@article{Krizhevsky2012,
	archivePrefix = {arXiv},
	arxivId = {1102.0183},
	author = {Krizhevsky, Alex and Sutskever, Ilya and Hinton, Geoffrey E.},
	doi = {http://dx.doi.org/10.1016/j.protcy.2014.09.007},
	eprint = {1102.0183},
	isbn = {9781627480031},
	issn = {10495258},
	journal = {Advances In Neural Information Processing Systems},
	pages = {1--9},
	pmid = {7491034},
	title = {ImageNet Classification with Deep Convolutional Neural Networks},
	year = {2012}
}

@book{KondorThesis,
	author={Kondor, R.},
	title={Group theoretical methods in machine learning},
	year={2008},
	publisher={Ph.D. thesis, Columbia Univserity}
}

@article{NbodyArxiv18,
	author = {Kondor, R.},
	title = {N-body Networks: a Covariant Hierarchical Neural Network Architecture for Learning Atomic Potentials},
	journal = {ArXiv e-prints},
	eprint = {1803.01588},
	year = {2018}
}

@ARTICLE{TensorFieldNetworksArxiv18,
	author = {Thomas, N. and Smidt, T. and Kearnes, S. and Yang, L. and 
	Li, L. and Kohlhoff, K. and Riley, P.},
	title = "{Tensor Field Networks: Rotation- and Translation-Equivariant Neural Networks for 3D Point Clouds}",
	journal = {ArXiv e-prints},
	eprint = {1802.08219},
	year = {2018}
}

@InCollection{MaslenRockmore97,
	author={D. Maslen and D. Rockmore},
	title={Generalized {FFT}s -- a survey of some recent results},
	booktitle={Groups and Computation II},
	series={DIMACS Ser. Discrete Math. Theor. Comput. Sci.},
	volume={28},
	publisher={AMS, Providence, RI},
	year={1997},
	pages={183-287}
}

@techreport{healy96ffts,
	author = {Dennis M. Healy and Daniel N. Rockmore and Sean S. B. Moore},
	title = {FFTs for the 2-Sphere -- Improvements and Variations},
	number = {PCS-TR96-292},
	institution={Department of Computer Science, Dartmouth College}, 
	year = {1996}
}

@INPROCEEDINGS{Healy96, 
	author={D. M. Healey and D. N. Rockmore and S. B. Moore}, 
	booktitle={1996 IEEE International Conference on Acoustics, Speech, and Signal Processing Conference Proceedings}, 
	title={An FFT for the 2-sphere and applications}, 
	year={1996}, 
	volume={3}, 
	number={}, 
	pages={1323-1326 vol. 3}, 
	doi={10.1109/ICASSP.1996.543670}, 
	ISSN={1520-6149}, 
	month={May}
}

@Article{Kostelec2008,
	author={Kostelec, P. J
	and Rockmore, D. N.},
	title={FFTs on the Rotation Group},
	journal={Journal of Fourier Analysis and Applications},
	year={2008},
	volume={14},
	number={2},
	pages={145--179},
	issn={1531-5851},
	url={https://doi.org/10.1007/s00041-008-9013-5}
}

@article{Skibbe09,
	author = {Skibbe, H. and Reisert, M. and Ronneberger, O. and Burkhardt, H.},
	journal = {Pattern Recognition Proc. DAGM},
	pages = {141-150},
	title = {Increasing the Dimension of Creativity in Rotation Invariant Feature Design Using 3D Tensorial Harmonic},
	year = {2009}
}

@InProceedings{KakaralaTriple,
	title={A group theoretic approach to the triple correlation},
	author={R. Kakarala},
	pages={28--32},
	booktitle={IEEE Workshop on higher order statistics},
	year={1993}
}

@article{InvImagesArXiv,
	author    = {R. Kondor},
	title     = {A complete set of rotationally and translationally invariant features
	for images},
	journal   = {CoRR},
	volume    = {abs/cs/0701127},
	year      = {2007},
	url       = {http://arxiv.org/abs/cs/0701127},
	archivePrefix = {arXiv},
	eprint    = {cs/0701127},
}

@article{Bartok2013cs,
	author = {Bart{\'o}k, A. P. and Kondor, R. and Cs{\'a}nyi, G.},
	title = {On representing chemical environments},
	journal = {Phys Rev B},
	year = {2013},
	volume = {87},
	number = {18},
	pages = {184115},
	month = may,
	doi = {10.1103/PhysRevB.87.184115},
	language = {English}
}

@Book{Serre,
	author ={J-P. Serre},
	title = {Linear Representations of Finite Groups},
	year = {1977},
	publisher={Springer-Verlag},
	series = {Graduate Texts in Mathamatics},
	volume = {42}
}

@article{ThomasTensor18, 
	title={Tensor Field Networks: Rotation- and Translation-Equivariant Neural Networks for 3D Point Clouds}, 
	url={http://arxiv.org/abs/1802.08219}, 
	note={arXiv: 1802.08219}, journal={arXiv:1802.08219 [cs]}, 
	author={Thomas, N. and Smidt, T. and Kearnes, S. and Yang, L. and Li, L. and Kohlhoff, K. and Riley, P.}, year={2018}, month={Feb}
}

@article{KondorNbodyArxiv18,
	author    = {R. Kondor},
	title     = {N-body Networks: a Covariant Hierarchical Neural Network Architecture
	for Learning Atomic Potentials},
	journal   = {CoRR},
	volume    = {abs/1803.01588},
	year      = {2018},
	url       = {http://arxiv.org/abs/1803.01588},
	archivePrefix = {arXiv},
	eprint    = {1803.01588},
}

@Book{Terras,
	author={A. Terras},
	title={Fourier analysis on finite groups and applications}, 
	series={London Mathematical Society Student Texts},
	volume={43}, 
	publisher={Cambridge Univ. Press},
	year={1999}
}

@Book{Diaconis,
	author={P. Diaconis},
	title={Group Representation in Probability and Statistics},
	volume={11},
	series={IMS Lecture Series},
	publisher={Institute of Mathematical Statistics},
	year={1988}
}

@article{Gutman2008,
	author = {Gutman, B. and Wang, Y. and Chan, T. and Thompson, P. M. and Toga, A. W.},
	journal = {2nd MICCAI Workshop on Mathematical Foundations of Computational Anatomy},
	pages = {56--67},
	title = {Shape Registration with Spherical Cross Correlation},
	year = {2008}
}

@article{Zhang2014,
	author = {Zhang, Y. and Song, S. and Tan, P. and Xiao, J.},
	journal = {Lecture Notes in Computer Science (including subseries Lecture Notes in Artificial Intelligence and Lecture Notes in Bioinformatics)},
	number = {PART 6},
	pages = {668--686},
	title = {PanoContext: A whole-room 3D context model for panoramic scene understanding},
	volume = {8694 LNCS},
	year = {2014}
}

@article{Zelnik-Manor2005,
	author = {Zelnik-Manor, L. and Peters, G. and Perona, P.},
	isbn = {0-7695-2334-X-02},
	journal = {IEEE ICCV},
	pages = {1292--1299},
	title = {Squaring the Circles in Panoramas},
	year = {2005}
}

@article{Su2017b,
	archivePrefix = {arXiv},
	arxivId = {1703.00495},
	author = {Su, Y-C and Grauman, K.},
	eprint = {1703.00495},
	title = {Making 360 Video Watchable in 2D: Learning Videography for Click Free Viewing},
	year = {2017}
}

@article{Su2017a,
	archivePrefix = {arXiv},
	arxivId = {1612.02335},
	author = {Su, Y-C and Jayaraman, D. and Grauman, K.},
	journal = {Lecture Notes in Computer Science (including subseries Lecture Notes in Artificial Intelligence and Lecture Notes in Bioinformatics)},
	number = {1},
	pages = {154--171},
	title = {Pano2vid: Automatic cinematography for watching 360 videos},
	volume = {10114 LNCS},
	year = {2017}
}

@article{Cruz-Mota2012,
	author = {Cruz-Mota, J. and Bogdanova, I. and Paquier, B. and Bierlaire, M. and Thiran, J-P},
	journal = {International Journal of Computer Vision},
	number = {2},
	pages = {217--241},
	title = {Scale invariant feature transform on the sphere: Theory and applications},
	volume = {98},
	year = {2012}
}

@article{Lai,
	archivePrefix = {arXiv},
	author = {Lai, W-S and Huang, Y. and Joshi, N. and Buehler, C. and Yang, M-H and Kang, S-B},
	eprint = {arXiv:1703.10798v4},
	journal = {arXiv:1703.10798v4},
	pages = {1--12},
	title = {Semantic-driven Generation of Hyperlapse from 360 @BULLET Video},
	year = {2017}
}

@article{Gens2014,
	archivePrefix = {arXiv},
	arxivId = {1207.6083},
	author = {Gens, R. and Domingos, P.},
	journal = {NIPS 2014},
	pages = {1--9},
	title = {Deep Symmetry Networks},
	year = {2014}
}

@article{Khasanova2017,
	archivePrefix = {arXiv},
	arxivId = {1707.08301},
	journal = {arXiv: 1707.08301},
	author = {Khasanova, R. and Frossard, P.},
	title = {Graph-Based Classification of Omnidirectional Images},
	url = {http://arxiv.org/abs/1707.08301},
	year = {2017}
}

@article{Boomsma2017,
	author = {Boomsma, W. and Frellsen, J.},
	number = {NIPS},
	pages = {3436--3446},
	title = {Spherical convolutions and their application in molecular modelling},
	year = {2017}
}

@article{Su2017c,
	archivePrefix = {arXiv},
	arxivId = {1708.00919},
	author = {Su, Y-C and Grauman, K.},
	eprint = {1708.00919},
	number = {NIPS},
	title = {Flat2Sphere: Learning Spherical Convolution for Fast Features from 360 Imagery},
	year = {2017}
}

@article{EstevesPo,
	archivePrefix = {arXiv},
	arxivId = {1709.01889},
	journal = {arXiv:1709.01889},
	author = {Esteves, C. and Allen-Blanchette, C. and Zhou, X. and Daniilidis, K.},
	title = {Polar Transformer Networks},
	year = {2017}
}

@article{EstevesSph,
	archivePrefix = {arXiv},
	arxivId = {1711.06721},
	journal = {arXiv:1711.06721},
	author = {Esteves, C. and Allen-Blanchette, C. and Makadia, A. and Daniilidis, K.},
	title = {Learning SO(3) Equivariant Representations with Spherical CNNs},
	year = {2017}
}

@article{Worrall2016,
	archivePrefix = {arXiv},
	arxivId = {1612.04642},
	author = {Worrall, D. E. and Garbin, S. J. and Turmukhambetov, D. and Brostow, G. J.},
	eprint = {1612.04642},
	journal = {arXiv:1612.04642},
	title = {Harmonic Networks: Deep Translation and Rotation Equivariance},
	year = {2016}
}

@article{Suyu2012,
	archivePrefix = {arXiv},
	arxivId = {arXiv:1208.6010},
	author = { S. H. Suyu and M. W. Auger and S. Hilbert and P. J. Marshall and M. Tewes and T. Treu and C. D. Fassnacht and L. V. E. Koopmans and D. Sluse and R. D. Blandford and F. Courbin and G. Meylan},
	eprint = {arXiv:1208.6010},
	title = {Two accurate time-delay distances from strong lensing: Implications for cosmology},
	year={2012}
}

@article{Collett2014,
	archivePrefix = {arXiv},
	arxivId = {arXiv:1403.5278},
	author = {Thomas E. Collett and Matthew W. Auger},
	eprint = {arXiv:1403.5278},
	title = {Cosmological Constraints from the double source plane lens SDSSJ0946+1006},
	year={2014}
}

@article{Diehl2017,
	author = {The DES Survey},
	booktitle={The Astrophysical Journal},
	title = {The DES Bright Arcs Survey: Hundreds of Candidate Strongly Lensed Galaxy Systems from the Dark Energy Survey Science Verification and Year 1 Observations},
	year={2017}
}

@article{Nord2016,
	archivePrefix = {arXiv},
	arxivId = {arXiv:1512.03062},
	author = {B. Nord et al.},
	eprint = {arXiv:1512.03062},
	title = {Observation and Confirmation of Six Strong Lensing Systems in The Dark Energy Survey Science Verification Data},
	year={2016}
}

@article{Oguri2010,
	archivePrefix = {arXiv},
	arxivId = {arXiv:1001.2037},
	author = {Masamune Oguri and Philip J. Marshall},
	eprint = {arXiv:1001.2037},
	title = {Gravitationally lensed quasars and supernovae in future wide-field optical imaging surveys},
	year={2010}
}

@article{Collett2015,
	author = {Thomas E. Collett},
	booktitle={The Astrophysical Journal},
	title = {The population of galaxy-galaxy strong lenses in forthcoming optical imaging surveys},
	year={2015}
}

@article{Goldstein2016,
	author = { Daniel A. Goldstein and Peter E. Nugent},
	booktitle={The Astrophysical Journal},
	title = { How to Find Gravitationally Lensed Type Ia Supernovae},
	year={2016}
}

@inproceedings{JebKon03,
	Address = {Heidelberg, Germany},
	Author = {Tony Jebara and Risi Kondor},
	Booktitle = {Proceedings of the Annual Conference on Computational Learning Theory and Kernels Workshop ({COLT/KW})},
	Editor = {B. Sch{\"o}lkopf and M. Warmuth},
	Number = {2777},
	Pages = {57-71},
	Publisher = {Springer-Verlag},
	Series = {Lecture Notes in Computer Science},
	Title = {Bhattacharyya and Expected Likelihood Kernels},
	Year = {2003}
}

@article{Lowe04,
	Author = {David G. Lowe},
	Journal = {International Journal of Computer Vision},
	Month = {November},
	Pages = {91-110},
	Title = {Distinctive Image Features from Scale-Invariant Keypoints},
	Volume = {60},
	Year = {2004}
}

@inproceedings{Hinton1981,
	author={Geoffrey E. Hinton},
	booktitle={ijcai},
	title={A parallel computation that assigns canonical object-based frames of reference},
	year={1981}
}

@article{FreemanAdelson1991,
	Author = {William T. Freeman and Edward H. Adelson},
	Journal = {IEEE Transactions on Pattern Analysis and Machine Intelligence},
	Month = {September},
	Pages = {891-906},
	Title = {The Design and Use of Steerable Filters},
	Volume = {13},
	Year = {1991}
}

@article{Perona1995,
	Author = {Pietro Perona},
	Journal = {IEEE Transactions on Pattern Analysis and Machine Intelligence},
	Month = {May},
	Pages = {488-499},
	Title = {Deformable Kernels for Early Vision},
	Volume = {17},
	Year = {1995}
}

@article{TeoHelOr1999,
	Author = {P. C. Teo and Y. Hel-Or},
	Journal = {IEEE Transactions on Pattern Analysis and Machine Intelligence},
	Month = {June},
	Pages = {552-556},
	Title = {Design of multiparameter steerable functions using cascade basis reduction},
	Volume = {21},
	Year = {1999}
}

@article{Simocelli1992,
	Author = {Eero P. Simoncelli and William T. Freeman and Edward H. Adelson and David J. Heeger},
	Journal = {IEEE Transactions on Information Theory},
	Month = {March},
	Pages = {587-607},
	Title = {Shiftable Multiscale Transforms},
	Volume = {38},
	Year = {1992}
}

@article{Portilla2003,
	Author = {Javier Portilla and Vasily Strela and Martin J. Wainwright and Eero P. Simoncelli},
	Journal = {IEEE Transactions on Image Processing},
	Month = {November},
	Pages = {1338-1351},
	Title = {Image Denoising Using Scale Mixtures of Gaussians in the Wavelet Domain},
	Volume = {12},
	Year = {2003}
}

@article{Do2002,
	Author = {M. N. Do and M. Vetterli},
	Journal = {IEEE Transactions on Multimedia},
	Month = {November},
	Pages = {146-158},
	Title = {Rotation invariant texture characterization and retrieval using steerable wavelet-domain hidden markov models},
	Volume = {4},
	Year = {2002}
}

@article{Michaelis1995,
	Author = {Markus Michaelis and Gerald Sommer},
	Journal = {Pattern Recognition Letters},
	Month = {November},
	Pages = {1165-1174},
	Title = {A lie group approach to steerable filters},
	Volume = {16},
	Year = {1995}
}

@article{Teo1998,
	Author = {Patrick C. Teo and Yacov Hel-Or},
	Journal = {Pattern Recognition Letters},
	Month = {October},
	Pages = {7-17},
	Title = {Lie generators for computing steerable functions},
	Volume = {16},
	Year = {1998}
}

@article{Manduchi1998,
	Author = {Roberto Manduchi and Pietro Perona and Doug Shy},
	Journal = {IEEE Transactions on Signal Processing},
	Month = {April},
	Pages = {1168-1173},
	Title = {Efficient Deformable Filter Banks},
	Volume = {46},
	Year = {1998}
}

@article{Lenz1989,
	Author = {Reiner Lenz},
	Journal = {Journal of the Optical Society of America},
	Month = {June},
	Pages = {827-834},
	Title = {Group Theoretical Model of Feature Extraction},
	Volume = {6},
	Year = {1989}
}

@techreport{Kondor2007,
	archivePrefix = {arXiv},
	arxivId = {arXiv:0701127v3},
	author = {Risi Kondor},
	eprint = {arXiv:0701127v3},
	title = {A novel set of rotationally and translationally invariant features for images based on the non-commutative bispectrum},
	year = {2007}
}

@article{Kondor2008,
	Author = {Risi Kondor},
	Journal = {PhD Thesis, Columbia University},
	Title = {Group theoretical models in machine learning},
	Year = {2008}
}

@article{Casey1998,
	Author = {Michael A. Casey},
	Journal = {PhD Thesis, MIT},
	Title = {Auditory group theory with applications to statistical basis methods for structured audio},
	Year = {1998}
}

@article{Wood1996,
	Author = {Jeffrey Wood},
	Journal = {Pattern Recognition},
	Month = {January},
	Pages = {1-17},
	Title = {Invariant Pattern Recognition: A Review},
	Volume = {29},
	Year = {1996}
}

@article{Kearns2016,
	Author = {Steven Kearns and Kevin McCloskey and Marc Brendl and Vijay Pande and Patrick Riley},
	Journal = {Journal of Computer Aided Molecular Design},
	Month = {August},
	Pages = {595-608},
	Title = {Molecular graph convolutions: Moving beyond fingerprints},
	Volume = {30},
	Year = {2016}
}

@article{Bruna2013,
	Author = {Joan Bruna and Stephane Mallat},
	Journal = {IEEE Transactions on Pattern Analysis and Machine Intelligence},
	Month = {August},
	Pages = {1872-1886},
	Title = {Invariant scattering convolutional networks},
	Volume = {35},
	Year = {2013}
}

@techreport{Henaff2015,
	archivePrefix = {arXiv},
	arxivId = {arXiv:1506.05163},
	author = {Mikael Henaff and Joan Bruna and Yann LeCun},
	eprint = {arXiv:1506.05163},
	title = {Deep convolutional networks on graph structured data},
	year = {2015}
}

@inproceedings{Bruna2014,
	author={Joan Bruna and Wojciech Zaremba and Arthur Szlam and Yann LeCun},
	booktitle={iclr},
	title={Spectral Networks and Locally connected networks on graphs},
	year={2014}
}

@inproceedings{CohenWell2015,
	author={Taco S. Cohen and Max Welling},
	booktitle={iclr},
	title={Transformation properties of learned visual representations},
	year={2015}
}

@inproceedings{Dieleman2016,
	author={S. Dieleman and J. De Fauw and K. Kavukcouglu},
	booktitle={icml},
	title={Exploiting cyclic symmetry in convolutional neural networks},
	year={2016}
}

@inproceedings{Kanazawa2014,
	author={Angjoo Kanazawa and Abhishek Sharma and David Jacobs},
	booktitle={nips},
	title={Locally scale-invariant convolutional neural networks},
	year={2014}
}

@techreport{Marcos2017,
	archivePrefix = {arXiv},
	arxivId = {arXiv:1612.09346},
	author = {Diego Marcos and Michele Volpi and Nikos Komodakis and Devis Tuia},
	eprint = {arXiv:1612.09346},
	title = {Rotation equivariant vector field networks},
	year = {2017}
}

@techreport{Jacobsen2017,
	archivePrefix = {arXiv},
	arxivId = {arXiv:1706.00598},
	author = {Joern-Henrik Jacobsen and Bert de Brabandere and Arnold W.M. Smeulders},
	eprint = {arXiv:1706.00598},
	title = {Dynamic Steerable Blocks in Deep Residual Networks},
	year = {2017}
}

@techreport{Worrall2017,
	archivePrefix = {arXiv},
	arxivId = {arXiv:1612.04642},
	author = { Daniel E. Worrall and Stephan J. Garbin and Daniyar Turmukhambetov and Gabriel J. Brostow},
	eprint = {arXiv:1612.04642},
	title = {Harmonic Networks: Deep Translation and Rotation Equivariance},
	year = {2017}
}

@inproceedings{Sifre2013,
	author={Laurent Sifre and Stephane Mallat},
	booktitle={cvpr},
	title={Rotation, scaling and deformation invariant scattering for texture discrimination},
	year={2013}
}

@inproceedings{Oyallon2015,
	author={E. Oyallon and S. Mallat},
	booktitle={cvpr},
	title={Deep roto-translation scattering for object classification},
	year={2015}
}

@inproceedings{HelOr1996,
	author={Yacov Hel-Or and Patrick C. Teo},
	booktitle={cvpr},
	title={Canonical Decomposition of Steerable Functions},
	year={1996}
}

@inproceedings{Jaderberg2015,
	author={Max Jaderberg and Karen Simonyan and Andrew Zisserman and Koray Kavukcuoglu},
	booktitle={nips},
	title={Spatial Transformer Networks},
	year={2015}
}

@inproceedings{HintonWang2011,
	author={Geoffrey E. Hinton and Alex Krizhevksy and Sida D. Wang},
	booktitle={icann},
	title={Transforming Auto-Encoders},
	year={2011}
}

@article{Tieleman2014,
	Author = {Tijman Tieleman},
	Journal = {PhD Thesis, University of Toronto},
	Title = {Optimizing Neural Networks that Generate Images},
	Year = {2014}
}

@inproceedings{KivinenWilliams2011,
	author={Jyri J. Kivinen and Christopher K. I. Williams},
	booktitle={icann},
	title={Transformation equivariant restricted Boltzmann machines},
	year={2011}
}

@article{Reisert2008,
	Author = {Marco Reisert},
	Journal = {PhD Thesis, Albert-Ludwigs University},
	Title = {Group integration techniques in pattern analysis: A kernel view},
	Year = {2008}
}

@article{Skibbe2013,
	Author = {H. Skibbe},
	Journal = {PhD Thesis, Albert-Ludwigs University},
	Title = {Spherical tensor algebra for biomedical image analysis},
	Year = {2013}
}

@article{Anselmi2014,
	Author = {F. Anselmi and J. Z. Leibo and J. Mutch and A. Tacchetti and T. Poggio},
	Journal = {Technical Report: MIT Center for Brains, Minds and Machines},
	Title = {Unsupervised learning of invariant representations with low sample complexity},
	Year = {2014}
}

@inproceedings{Greenspan1994,
	author={H. Greenspan and S. Belongie and R. Goodman and P. Perona and S. Rakshit and C. H. Anderson},
	booktitle={cvpr},
	title={Overcomplete steerable pyramid filters and rotation invariance},
	year={1994}
}

@article{KoVanDoorn1990,
	Author = {J. J. Koenderink and A. J. van Doorn},
	Journal = {Biological Cybernetics},
	Month = {April},
	Pages = {291-297},
	Title = {Receptive Field Families},
	Volume = {63},
	Year = {1990}
}

@article{MichSomm1995,
	Author = {Markus Michaelis and Gerald Sommer},
	Journal = {Pattern Recognition Letters},
	Month = {March},
	Pages = {1165-1174},
	Title = {A Lie group approach to steerable filters},
	Volume = {16},
	Year = {1995}
}

@article{Krajsek2007,
	Author = {Kai Krajsek and Rudolf Mester},
	Journal = {Communications in Computer and Information Science},
	Month = {March},
	Pages = {5-13},
	Title = {A unified theory for steerable and quadrature filters},
	Volume = {4},
	Year = {2007}
}

@inproceedings{KonJeb03,
	Author = {Risi Kondor and Tony Jebara},
	Booktitle = {icml},
	Title = {A Kernel between Sets of Vectors},
	Year = {2003}
}

@article{ShervEtal11,
	Author = {Nino Shervashidze and Pascal Schweitzer and Erik Jan van Leeuwen and Kurt Mehlhorn and Karsten M. Borgwardt},
	Journal = {neco},
	Month = {November},
	Pages = {2539-2561},
	Title = {Weisfeiler-Lehman Graph Kernels},
	Volume = {12},
	Year = {2011}
}

@article{FelzRoss2010,
	Author = {Pedro F. Felzenszwalb and Ross B. Girshick and  David McAllester and  Deva Ramanan},
	Journal = {IEEE Transactions on Pattern Analysis and Machine Intelligence},
	Pages = {541-551},
	Title = {Object Detection with Discriminatively Trained Part-Based Models},
	Volume = {32},
	Year = {2010}
}

@article{FelzHutt2005,
	Author = {Pedro F. Felzenszwalb and Daniel P. Huttenlocher},
	Journal = {International Journal of Computer Vision},
	Pages = {55-71},
	Title = {Pictorial Structures for Object Recognition},
	Volume = {61},
	Year = {2005}
}

@article{ZhuMumford,
	Author = {Song-Chun Zhu and David Mumford},
	Journal = {Foundations and Trends in Computer Graphics and Vision},
	Pages = {259-362},
	Title = {A stochastic grammar of images},
	Volume = {2},
	Year = {2006}
}

@article{LeCun1998,
	Author = {Y LeCun and L Bottou and Y Bengio and D Henderson and P Haffner},
	Journal = {Proceedings of the IEEE},
	Pages = {2278-2324},
	Title = {Gradient-based learning applied to document recognition},
	Volume = {86(11)},
	Year = {1998}
}

@inproceedings{KonPan16,
	author={Risi Kondor and Horace Pan},
	booktitle={nips},
	title={The Multiscale Laplacian Graph Kernel},
	year={2016}
}

@inproceedings{AlexHinton2012,
	author={Alex Krizhevsky and Ilya Sutskever and Geoffrey E. Hinton},
	booktitle={nips},
	title={ImageNet Classification with Deep Convolutional Neural Networks},
	year={2012}
}

@inproceedings{Niepert2016,
	author={Mathias Niepert and Mohamed Ahmed and Konstantin Kutzkov},
	booktitle={icml},
	title={Learning Convolutional Neural Networks for Graphs},
	year={2016}
}

@inproceedings{Defferrard2016,
	author={Michaël Defferrard and Xavier Bresson and Pierre Vandergheynst},
	booktitle={nips},
	title={Convolutional Neural Networks on Graphs with Fast Localized Spectral Filtering},
	year={2016}
}

@inproceedings{Duvenaud2015,
	author={David Duvenaud and Dougal Maclaurin and Jorge Aguilera-Iparraguirre and Rafael Gomez-Bombarelli and Timothy Hirzel and Alan Aspuru-Guzik and Ryan P. Adams},
	booktitle={nips},
	title={Convolutional Networks on Graphs for Learning Molecular Fingerprints},
	year={2015}
}

@inproceedings{Li2016,
	author={Yujia Li and Daniel Tarlow and Marc Brockschmidt and Richard Zemel},
	booktitle={iclr},
	title={Gated Graph Sequence Neural Networks},
	year={2016}
}

@techreport{Mallat2012,
	archivePrefix = {arXiv},
	arxivId = {arXiv:1101.2286v3},
	author = {Mallat, S},
	eprint = {arXiv:1101.2286v3},
	title = {Group Invariant Scattering},
	year = {2012}
}

@article{Cohen2018a,
	archivePrefix = {arXiv},
	arxivId = {1803.10743},
	author = {Cohen, Taco S. and Geiger, Mario and Weiler, Maurice},
	eprint = {1803.10743},
	journal = {arXiv},
	title = {Intertwiners between Induced Representations (with Applications to the Theory of Equivariant Neural Networks)},
	url = {https://arxiv.org/abs/1803.10743},
	year = {2018}
}

@article{Cohen2018b,
	author = {Cohen, Taco S. and Geiger, Mario and Weiler, Maurice},
	title = {The Quite General Theory of Equivariant Convolutional Networks},
	journal = {to appear},
	year = {2018}
}

@article{Esteves,
	archivePrefix = {arXiv},
	arxivId = {1711.06721},
	author = {Esteves, Carlos and Allen-Blanchette, Christine and Makadia, Ameesh and Daniilidis, Kostas},
	eprint = {1711.06721},
	journal = {arXiv},
	title = {Learning SO(3) Equivariant Representations with Spherical CNNs},
	url = {https://arxiv.org/abs/1711.06721},
	year = {2017}
}

@article{LeCun2015,
	archivePrefix = {arXiv},
	arxivId = {arXiv:1312.6184v5},
	author = {LeCun, Yann and Bengio, Yoshua and Hinton, Geoffrey and Y., Lecun and Y., Bengio and G., Hinton},
	doi = {10.1038/nature14539},
	eprint = {arXiv:1312.6184v5},
	isbn = {3135786504},
	issn = {0028-0836},
	journal = {Nature},
	number = {7553},
	pages = {436--444},
	pmid = {26017442},
	title = {Deep learning},
	volume = {521},
	year = {2015}
}

@article{Fosdick,
	archivePrefix = {arXiv},
	arxivId = {arXiv:1608.07618v2},
	author = {Fosdick, Bailey K and Mccormick, Tyler H and Murphy, Thomas Brendan and Westling, Ted},
	eprint = {arXiv:1608.07618v2},
	pages = {1--47},
	title = {Multiresolution network models}
}

@Book{Chirikjian,
	author ={Gregory S. Chirikjian and Alexander B. Kyatkin},
	title = {Harmonic Analysis for Engineers and Applied Scientists: Updated and Expanded Edition},
	year = {2016},
	publisher={Courier Dover Publications}
}

@Book{Thurston,
	author ={William P. Thurston},
	title = {Three-Dimensional Geometry and Topology, Volume 1},
	year = {1997},
	publisher={Princeton University Press}
}

@article{CohenSE3,
	archivePrefix = {arXiv},
	arxivId = {1807.02547},
	author = {Weiler, Maurice and Geiger, Marco and Welling, Max and Boomsma, Wouter and Cohen, Taco S.},
	eprint = {1807.02547},
	journal = {arXiv preprint},
	title = {3D Steerable CNNs: Learning Rotationally Equivariant Features in Volumetric Data},
	url = {https://arxiv.org/abs/1807.02547},
	year = {2018}
}

@Book{MumfordPatternTheory,
	author ={David Mumford and Agn{\`e}s Desolneux},
	title = {Pattern theory: the stochastic analysis of real-world signals},
	year = {2010},
	publisher={AK Peters/CRC Press}
}

@Book{Grenander,
	author ={Ulf Grenander},
	title = {Probabilities on algebraic structures},
	year = {2008},
	publisher={Courier Corporation}
}

@article{Lifu,
	archivePrefix = {arXiv},
	arxivId = {1803.03376},
	author = {Tu, Lifu and Gimpel, Kevin},
	eprint = {1803.03376},
	journal = {arXiv preprint},
	title = {Learning approximate inference networks for structured prediction},
	url = {https://arxiv.org/abs/1807.02547},
	year = {2018}
}

@article{Anselmi2017,
	Author = {Anselmi, Fabio and Evangelopoulos, Georgios and Rosasco, Lorenzo and Poggio, Tomaso},
	Journal = {Technical Report: MIT Center for Brains, Minds and Machines},
	Title = {Symmetry Regularization},
	Year = {2017}
}

@article{doi:10.1063/1.4772195,
	author = {Budisic, Marko  and Mohr, Ryan  and Mezic, Igor },
	title = {Applied Koopmanism},
	journal = {Chaos: An Interdisciplinary Journal of Nonlinear Science},
	volume = {22},
	number = {4},
	pages = {047510},
	year = {2012},
	doi = {10.1063/1.4772195}	
}

@inproceedings{trivediNIPS14,
	title={Discriminative Metric Learning by Neighborhood Gerrymandering},
	author={Shubhendu Trivedi and David Mcallester and Gregory Shakhnarovich},
	booktitle={Advances in Neural Information Processing Systems},
	pages={3392--3400},
	year={2014}
}

@inproceedings{trivediAsym15,
	title={Notes on Asymmetric Metric Learning for k-NN Classification},
	author={Shubhendu Trivedi},
	booktitle={Unpublished Notes},
	howpublished = "\url{http://ttic.uchicago.edu/~shubhendu/Papers/Asym.pdf}",
	year={2015}
}

@article{trivediSpheres2018,
	archivePrefix = {arXiv},
	arxivId = {arXiv:1806.09231},
	author = {Kondor, Risi and Lin, Zhen and Trivedi, Shubhendu}, 
	eprint = {arXiv:1806.09231},
	journal={arXiv:1806.09231}, 
	year={2018},
	title = {Clebsch-Gordan Nets: a Fully Fourier Space Spherical Convolutional Neural Network}
}

@article{trivediCCN1,
	archivePrefix = {arXiv},
	arxivId = {arXiv:1801.02144},
	author = {Kondor, Risi and Song, Truong Hy and Pan, Horace and Anderson, Brandon M. and Trivedi, Shubhendu}, 
	eprint = {arXiv:1801.02144},
	journal={arXiv:1801.02144}, 
	year={2018},
	title = {Covariant compositional networks for learning graphs}
}

@article{doi:10.1063/1.5024797,
	author = {Hy,Truong Son  and Trivedi,Shubhendu  and Pan,Horace  and Anderson,Brandon M.  and Kondor,Risi },
	title = {Predicting molecular properties with covariant compositional networks},
	journal = {The Journal of Chemical Physics},
	volume = {148},
	number = {24},
	pages = {241745},
	year = {2018},
	doi = {10.1063/1.5024797}
}

% Old version, will be removed later
% work-around to have small caps also here in the headline
%\manualmark
%\markboth{\spacedlowsmallcaps{\bibname}}{\spacedlowsmallcaps{\bibname}} % work-around to have small caps also
%\phantomsection
%\refstepcounter{dummy}
%\addtocontents{toc}{\protect\vspace{\beforebibskip}} % to have the bib a bit from the rest in the toc
%\addcontentsline{toc}{chapter}{\tocEntry{\bibname}}
%\label{app:bibliography}
%\printbibliography

%%\cleardoublepage\include{FrontBackmatter/Declaration}
%%\cleardoublepage\include{FrontBackmatter/Colophon}
% ********************************************************************
% Game Over: Restore, Restart, or Quit?
%*******************************************************
\end{document}